\documentclass{article}

\usepackage{arxiv}

\usepackage[utf8]{inputenc}
\usepackage[T1]{fontenc}

\usepackage{amsmath, amssymb, amsthm}

\usepackage{graphicx}
\usepackage{booktabs}
\usepackage{caption}
\usepackage[section]{placeins}
\usepackage{footnote}    
\makesavenoteenv{figure}
\usepackage{microtype}

\usepackage{enumitem}

\usepackage{natbib}
\usepackage{hyperref}

\usepackage{orcidlink}

\usepackage{tabularx}
\usepackage{array}    
\newcolumntype{Y}{>{\raggedright\arraybackslash}X}

\usepackage{etoolbox}    

\usepackage[most]{tcolorbox}
\usepackage{tikz}
\usetikzlibrary{positioning, arrows.meta, fit, calc, matrix}

\definecolor{codebg}{gray}{0.965}
\definecolor{coderule}{gray}{0.55}
\tcbset{codeblockstyle/.style={
  enhanced,
  breakable,
  colback=codebg,
  colframe=coderule,
  boxrule=0pt,
  leftrule=2pt,
  arc=0pt,
  outer arc=0pt,
  left=8pt,
  right=4pt,
  top=4pt,
  bottom=4pt,
  before skip=4pt,
  after skip=4pt,
  fontupper=\footnotesize
}}
\newenvironment{codeblock}[1][]{%
  \par\addvspace{2pt}%
  \ifstrempty{#1}{}{\noindent\textit{\small #1}\par\nobreak\addvspace{2pt}}%
  \begin{tcolorbox}[codeblockstyle]%
}{%
  \end{tcolorbox}\par%
}
\newenvironment{asciiblock}[1][]{%
  \par\addvspace{2pt}%
  \ifstrempty{#1}{}{\noindent\textit{\small #1}\par\nobreak\addvspace{2pt}}%
  \begin{tcolorbox}[codeblockstyle]%
}{%
  \end{tcolorbox}\par%
}

\tcbset{
  pipelinephase/.style={
    enhanced, breakable, colback=gray!4, colframe=black!55,
    fonttitle=\bfseries\sffamily\small, fontupper=\small,
    boxrule=0.5pt, arc=2pt,
    left=6pt, right=6pt, top=4pt, bottom=4pt,
    before skip=2pt, after skip=2pt,
    title={#1}
  },
  pipelinesub/.style={
    enhanced, colback=white, colframe=black!35,
    fonttitle=\bfseries\sffamily\footnotesize, fontupper=\footnotesize,
    boxrule=0.4pt, arc=1.5pt,
    left=4pt, right=4pt, top=3pt, bottom=3pt,
    before skip=2pt, after skip=2pt,
    title={#1}
  }
}

\hypersetup{
  colorlinks=true,
  linkcolor=blue,
  citecolor=blue,
  urlcolor=blue,
  breaklinks=true,
}

\usepackage{newunicodechar}
\newunicodechar{−}{$-$}    
\newunicodechar{×}{$\times$}
\newunicodechar{÷}{$\div$}
\newunicodechar{±}{$\pm$}
\newunicodechar{≈}{$\approx$}
\newunicodechar{≠}{$\neq$}
\newunicodechar{≤}{$\leq$}
\newunicodechar{≥}{$\geq$}
\newunicodechar{∈}{$\in$}
\newunicodechar{∉}{$\notin$}
\newunicodechar{∞}{$\infty$}
\newunicodechar{α}{$\alpha$}
\newunicodechar{β}{$\beta$}
\newunicodechar{γ}{$\gamma$}
\newunicodechar{δ}{$\delta$}
\newunicodechar{ε}{$\varepsilon$}
\newunicodechar{ζ}{$\zeta$}
\newunicodechar{η}{$\eta$}
\newunicodechar{θ}{$\theta$}
\newunicodechar{κ}{$\kappa$}
\newunicodechar{λ}{$\lambda$}
\newunicodechar{μ}{$\mu$}
\newunicodechar{π}{$\pi$}
\newunicodechar{ρ}{$\rho$}
\newunicodechar{σ}{$\sigma$}
\newunicodechar{τ}{$\tau$}
\newunicodechar{φ}{$\varphi$}
\newunicodechar{χ}{$\chi$}
\newunicodechar{ψ}{$\psi$}
\newunicodechar{ω}{$\omega$}
\newunicodechar{Δ}{$\Delta$}
\newunicodechar{Σ}{$\Sigma$}
\newunicodechar{Π}{$\Pi$}
\newunicodechar{Ω}{$\Omega$}
\newunicodechar{Ψ}{$\Psi$}
\newunicodechar{ℝ}{$\mathbb{R}$}
\newunicodechar{ℕ}{$\mathbb{N}$}
\newunicodechar{ℤ}{$\mathbb{Z}$}
\newunicodechar{·}{$\cdot$}
\newunicodechar{…}{\ldots}
\newunicodechar{—}{---}     
\newunicodechar{–}{--}      
\newunicodechar{‘}{`}
\newunicodechar{’}{'}
\newunicodechar{“}{``}
\newunicodechar{”}{''}
\newunicodechar{§}{\S}
\newunicodechar{ℓ}{$\ell$}
\newunicodechar{✓}{\checkmark}
\newunicodechar{✗}{$\times$}
\newunicodechar{🟢}{}
\newunicodechar{🟡}{}
\newunicodechar{⏸}{}
\newunicodechar{⇒}{$\Rightarrow$}
\newunicodechar{⇐}{$\Leftarrow$}
\newunicodechar{⇔}{$\Leftrightarrow$}
\newunicodechar{→}{$\to$}
\newunicodechar{←}{$\leftarrow$}
\newunicodechar{↔}{$\leftrightarrow$}
\newunicodechar{↑}{$\uparrow$}
\newunicodechar{↓}{$\downarrow$}
\newunicodechar{∂}{$\partial$}
\newunicodechar{∇}{$\nabla$}
\newunicodechar{∼}{$\sim$}
\newunicodechar{∝}{$\propto$}
\newunicodechar{⊂}{$\subset$}
\newunicodechar{⊃}{$\supset$}
\newunicodechar{⊆}{$\subseteq$}
\newunicodechar{⊇}{$\supseteq$}
\newunicodechar{∪}{$\cup$}
\newunicodechar{∩}{$\cap$}
\newunicodechar{∅}{$\emptyset$}
\newunicodechar{∀}{$\forall$}
\newunicodechar{∃}{$\exists$}
\newunicodechar{∧}{$\wedge$}
\newunicodechar{∨}{$\vee$}
\newunicodechar{¬}{$\neg$}
\newunicodechar{∑}{$\sum$}
\newunicodechar{∏}{$\prod$}
\newunicodechar{∫}{$\int$}
\newunicodechar{√}{$\sqrt{}$}
\newunicodechar{ℓ}{$\ell$}
\newunicodechar{†}{$\dagger$}
\newunicodechar{‡}{$\ddagger$}
\newunicodechar{°}{$^{\circ}$}
\newunicodechar{²}{$^2$}
\newunicodechar{³}{$^3$}
\newunicodechar{¹}{$^1$}
\newunicodechar{½}{$\frac{1}{2}$}
\newunicodechar{¼}{$\frac{1}{4}$}
\newunicodechar{¾}{$\frac{3}{4}$}
\newunicodechar{⌈}{$\lceil$}
\newunicodechar{⌉}{$\rceil$}
\newunicodechar{⌊}{$\lfloor$}
\newunicodechar{⌋}{$\rfloor$}
\newunicodechar{⟨}{$\langle$}
\newunicodechar{⟩}{$\rangle$}
\newunicodechar{‖}{$\|$}
\newunicodechar{∥}{$\|$}
\newunicodechar{‐}{-}
\newunicodechar{‑}{-}
\newunicodechar{ᵀ}{$^{\top}$}
\newunicodechar{·}{$\cdot$}
\newunicodechar{‹}{$<$}
\newunicodechar{›}{$>$}
\newunicodechar{«}{``}
\newunicodechar{»}{''}
\newunicodechar{ι}{$\iota$}
\newunicodechar{ν}{$\nu$}
\newunicodechar{ξ}{$\xi$}
\newunicodechar{ο}{o}
\newunicodechar{Φ}{$\Phi$}
\newunicodechar{Λ}{$\Lambda$}
\newunicodechar{Θ}{$\Theta$}
\newunicodechar{Γ}{$\Gamma$}
\newunicodechar{Ξ}{$\Xi$}
\newunicodechar{Υ}{$\Upsilon$}
\newunicodechar{Β}{B}
\newunicodechar{ⁿ}{$^{n}$}
\newunicodechar{₀}{$_0$}
\newunicodechar{₁}{$_1$}
\newunicodechar{₂}{$_2$}
\newunicodechar{₃}{$_3$}
\newunicodechar{₄}{$_4$}
\newunicodechar{₅}{$_5$}
\newunicodechar{ᵢ}{$_i$}
\newunicodechar{ⱼ}{$_j$}
\newunicodechar{ₜ}{$_t$}
\newunicodechar{₊}{$_+$}
\newunicodechar{₋}{$_-$}
\newunicodechar{∘}{$\circ$}
\newunicodechar{⊗}{$\otimes$}
\newunicodechar{⊕}{$\oplus$}
\newunicodechar{≡}{$\equiv$}

\theoremstyle{plain}

\theoremstyle{definition}

\title{Perturbation Dose Responses in Recursive LLM Loops\\[2pt]
       \large\itshape Raw switching, stochastic floors, and persistent escape\\
       \large\itshape under append, replace, and dialog updates}

\renewcommand{\shorttitle}{Perturbation dose responses in recursive LLM loops}

\author{Pawel Kaplanski~\orcidlink{0000-0003-2223-0870}\\{}Kaplanski Ai Lab\\\texttt{pawel@kaplanski.ai}}
\date{April 30, 2026}

\hypersetup{
  pdftitle={Perturbation Dose Responses in Recursive LLM Loops},
  pdfsubject={cs.AI, cs.LG, cs.CL},
  pdfauthor={Pawel Kaplanski},
  pdfkeywords={recursive LLM loops, perturbation dose response,
    attractor-like regimes, context-update rules, basin switching,
    dialog dynamics},
}

\begin{document}
\maketitle


\begin{abstract}
Recursive language-model loops often settle into recognizable attractor-like patterns, but the practical question is how much injected text is needed to move a settled loop somewhere else, and whether that move lasts. We study this in 30-step recursive loops by separating the model from the context-update rule: append, replace, and dialog updates expose different histories to the same generator. The main result is that persistent redirection in append-mode recursive loops is memory-policy-conditioned. Under the canonical bounded-memory loop with a 12,000-character tail clip, destination-coherent persistence (kicked and still in the same post-injection cluster at the terminal step) plateaus at about 16\% across doses 5-400, and retained source-basin escape (kicked and outside the original cluster at the terminal step) reaches about 36\% at dose 400; neither crosses 50\%. Under a full-history protocol with no artificial truncation within the 30-step horizon, retained source-basin escape crosses 50\% near 400 tokens and saturates near 75-80\% by 1,500 tokens, while destination-coherent persistence first reaches a 50\% point estimate near 1,500 tokens, with Wilson 95\% CI for the proportion [0.41, 0.61], so the half-effect region is only localized to roughly 1,000-2,000 tokens.

The persistent-escape family therefore splits into two endpoints that answer different questions. Retained source-basin escape asks whether the loop was durably pushed out of its starting basin. Destination-coherent persistence asks the stricter question of whether it stayed in the specific post-injection basin. At high full-history doses, destination-coherent persistence becomes non-monotonic at the canonical 30-step horizon even though the kicked rate stays near 0.96. A four-step falsification battery (heterogeneity control, cluster-granularity sweep with hierarchical macro-merge, transition-entropy diagnostic, and long-horizon trajectory continuation) recasts the dip as a finite-horizon, endpoint-definition-sensitive feature rather than a stable structural attractor asymmetry. A homogeneous-perturbation control reproduced the high-dose dip almost exactly (0.44, 0.32, 0.40 at doses 1500, 2000, 3000 versus 0.51, 0.32, 0.41 under heterogeneous perturbations), refuting perturbation heterogeneity as the cause. Roughly half the canonical magnitude is attributable to endpoint timing (the destination cluster sampled at the injection step is dominated by raw-injection text, not yet by model-generated continuation), and the residual is substantially reduced when the loop is run another 50 steps: under the frozen canonical basis (PCA + K-means fit on the original 30-step experiment, applied to extended trajectories), the dip drops by 73\% from -0.143 (95\% family-cluster bootstrap CI $[-0.269,-0.034]$) at step 29 to -0.039 (CI $[-0.158,+0.068]$) at step 79, with the step-79 interval now straddling zero. The same control showed lower retained source-basin escape under homogeneous repetition (0.54-0.58) than under heterogeneous concatenation (0.74-0.79), suggesting that repeated content gives the model a clearer signal from which to recover.

For the simpler raw endpoint, append-mode continuation has a bounded in-distribution dose response. Adversarial continuations yield $\mathrm{ED50}_{\mathrm{raw}}\approx 40$ tokens, with convergent estimates from pooled logistic fitting, mixed-effects modeling, and family-cluster bootstrap. Raw switching plateaus near 67\%, paired controls already diverge at about 35\%, and net switching never reaches +50 percentage points within the tested 5-400 token range; the largest observed net effect is about +32 percentage points at dose 400. Thus raw terminal disagreement is not the same as durable redirection.

Replace-mode loops show near-saturated raw switching under the default perturbation, but this mostly reflects state-reset overwrite: the injected text becomes \(X_{t+1}\), so switching is tautological at the state-write step. Insert-mode probes, which expose the model to the perturbation without writing it as the next state, reduce replace-mode switching to 12-32\%. The F3 cross-loop check shows a smaller append-mode overwrite-insert gap (14-34 percentage points) than the replace-mode gap (60-80 percentage points), with insert rates ordered $O1 \ge O2 \ge O3$.

Practically, this means recursive-loop evaluations should
distinguish transient movement from durable escape, always subtract
stochastic floors, and treat context-update rules as first-class
safety-relevant design choices rather than implementation details.
We report 37 experiments on \texttt{gpt-4o-mini}, with within-vendor
replication on \texttt{gpt-4.1-nano} and public code, configurations,
trajectories, and reports
(<https://github.com/kaplan196883/llmattr>).

---
\end{abstract}

\section*{Plain-language summary}

\subsection*{A quick glossary}

This summary uses paper-internal labels and AI-research terms.
Translated up front so the rest reads cleanly. Skip ahead to "Why it
matters" if you already know the vocabulary.

\textbf{LLM basics.}

\begin{itemize}
  \item \textbf{LLM (large language model).} An AI system trained on large
  amounts of text to generate text - the engine behind chatbots like
  ChatGPT. Our main model is OpenAI's \texttt{gpt-4o-mini}.
  \item \textbf{Prompt.} The text given to an LLM as input: instructions,
  questions, prior conversation, retrieved documents, anything else
  the model is meant to use when producing its next answer.
  \item \textbf{System prompt.} Higher-priority developer instructions that
  tell the LLM how to behave (for example: "Continue the text
  naturally."). The user typically does not see it.
  \item \textbf{Token.} A small chunk of text the LLM works with - often a word
  or part of a word. Rough scale: 1 token ≈ 0.75 English words, so
  the headline result of ≈ 40 tokens is about two sentences.
  \item \textbf{Context.} All text the LLM can see at one step: its system
  prompt, prior conversation, retrieved documents, tool outputs, and
  anything else fed into that API call.
\end{itemize}

\textbf{The recursive loop.}

\begin{itemize}
  \item \textbf{Recursive loop.} We feed an LLM its own output back as the next
  input, step after step. A \textit{trajectory} is one such run: the
  sequence of model outputs over those steps.
  \item \textbf{Context-update rule.} The rule for how the loop builds the
  next input from the previous step. The two we study:
  \item \textit{Append-mode} keeps everything and adds each new turn at the
    end, like a transcript.
  \item \textit{Replace-mode} discards prior context and replaces it with the
    latest output, like overwriting a draft.
  Important caveat for replace-mode: under the default ("overwrite")
  perturbation protocol, the injected text literally becomes the
  next state. High switching under that protocol is therefore a
  property of the update rule, not direct evidence that the model
  was redirected. The paper's \textit{insert-mode} counterfactual (§5.2)
  shows the perturbation to the model without writing it as the
  next state, isolating model-mediated behavior from this
  state-write artifact.
  \item \textbf{Memory policy.} The rule for what an LLM agent keeps, drops,
  summarizes, retrieves, or rewrites between steps. Append-mode
  and replace-mode are the two simplest memory policies; production
  agents often use rolling windows, summary buffers, pinned
  instructions, or tool-output stores.
  \item \textbf{Bounded-memory loop / full-history loop.} Two protocols we
  compare. \textit{Bounded-memory} loops keep only the most recent
  ~12,000 characters of conversation, dropping older content (this
  is called \textit{tail-clipping}). \textit{Full-history} loops keep everything
  the model can see - up to its native limit of ~128,000 tokens
  for our model - so the perturbation stays visible throughout the
  experiment. Real production agents are usually bounded; our
  full-history protocol is a controlled comparison.
  \item \textbf{Observable.} The text view we embed at each step to measure
  where the trajectory is. Three observables in the paper: the
  model's \textit{single-generation output} (just one reply, ~600 chars),
  \textit{rolling\_k3} (last 3 outputs concatenated), and \textit{context\_tail}
  (last 4,000 chars of the running conversation - the canonical
  observable for our main analysis). Different observables can give
  slightly different numbers because they integrate over different
  amounts of history.
\end{itemize}

\textbf{Dynamical-systems vocabulary.}

\begin{itemize}
  \item \textbf{Attractor.} A pattern or end-state the loop tends to settle
  into and return to after small disturbances. Folk-talk version:
  the loop "gets stuck" near it.
  \item \textbf{Basin / basin boundary / barrier.} A \textit{basin} is the set of
  nearby trajectory states that all settle into the same attractor.
  A \textit{basin boundary} is the dividing line between basins. A
  \textit{barrier} (or \textit{basin barrier}, \textit{injected-token barrier}) is how
  hard it is - in injected tokens or other pressure - to push the
  loop across that boundary.
  \item \textbf{Regime.} A recurring mode of loop behavior - a coarse cluster
  of similar trajectories or end-states. The paper names five regimes
  by short codes (O1, O2, O3, D1, D2), formally defined in §3.1.
  \item \textbf{End-state (cluster).} A coarse label for where a trajectory
  finishes. We group similar final outputs into a handful of bins
  and ask which bin a run lands in.
\end{itemize}

\textbf{Experimental terms.}

\begin{itemize}
  \item \textbf{Paired control.} The same prompt run twice with only random
  sampling differing - the model's built-in randomness in choosing
  its next word. Used to ask "is the difference we observe bigger
  than what randomness alone produces?"
  \item \textbf{Injection (perturbation, dose).} A chunk of text we splice
  into the loop at a chosen step. Its size is measured in tokens;
  we vary the dose from 5 to 400 tokens.
  \item \textbf{In-distribution adversarial text.} Injected text hand-crafted
  to look like it naturally belongs in the conversation, rather
  than obviously off-topic gibberish.
  \item \textbf{The three endpoints.} "Did the injection work?" has three
  useful versions:
  \item \textit{Raw switching} - the run ended in a different end-state than
    its paired no-injection twin.
  \item \textit{Net switching} - raw switching minus the \textit{stochastic floor}
    (see below).
  \item \textit{Persistent escape} - the run was knocked into a new end-state
    right after the injection AND was still in that end-state at
    the end of the run.
  \item \textbf{Stochastic floor.} The rate at which two paired no-injection
  runs of the same prompt already end up in different end-states,
  purely from the model's random sampling. In our main append-mode
  setting this baseline is about 35\%.
  \item \textbf{ED50.} Half-effect dose, borrowed from pharmacology: the dose
  at which the effect crosses 50\%. $\mathrm{ED50}_{\mathrm{raw}}$ is
  the dose at which raw switching crosses 50\%. The paper reports
  \textit{two} persistence endpoints with their own half-effect doses (see
  §5.1.3): $\mathrm{ED50}_{\mathrm{persist}}^{\mathrm{src}}$ for
  \textit{retained source-basin escape} (kicked AND outside the original
  cluster at terminal step) and $\mathrm{ED50}_{\mathrm{persist}}^{\mathrm{dst}}$
  for the stricter \textit{destination-coherent persistence} (kicked AND
  in same post-injection cluster at terminal step). They answer
  different questions; both are reported.
  \item \textbf{Embedding space.} The numeric vector representation of text
  that LLMs work with internally. We measure where each output
  lands in this space and group nearby outputs into the end-state
  bins above. "Embedding-space stability" means trajectories cluster
  in similar regions across runs; it is \textit{not} a claim about whether
  the answers are factually correct.
\end{itemize}

\textbf{Agent / safety vocabulary.}

\begin{itemize}
  \item \textbf{Agent (recursive agent).} An LLM-based system that acts over
  multiple steps, often using memory, retrieved documents, or tools
  rather than producing a single one-shot answer.
  \item \textbf{Scaffold (agent scaffold, agent-loop architecture).} The
  software wrapper around the LLM - how it repeats steps, stores
  memory, updates context, calls tools, decides what to do next.
  Many production systems are LLM-plus-scaffold rather than a bare
  LLM.
  \item \textbf{Tool output.} Text returned by an external service the LLM can
  call: a search engine, a browser, a calculator, a database, an
  API. Modern agents read tool outputs into context.
  \item \textbf{Indirect prompt injection (IPI).} An attack in which malicious
  or misleading instructions are hidden in material the LLM reads
  (a webpage, document, email, tool output), rather than typed
  directly by the user.
  \item \textbf{Jailbreak.} An attempt to bypass the model's safety
  instructions or redirect it toward disallowed behavior.
  \item \textbf{Redirection.} Steering the loop or agent away from its
  original behavior toward a different one. \textit{Durable redirection}
  means the changed behavior persists for many later steps.
\end{itemize}

\subsection*{Why it matters}

A common folk-belief about recursive LLM loops is that they either
"get stuck in attractors" or are "easy to hijack with a small
prompt." Both claims are too vague to be useful. A loop can visibly
move without staying moved; two runs of the same prompt can diverge
without any attack; and some apparent "fragility" is caused by the
system's context-update rule rather than by the model crossing a
real basin boundary. The right question is no longer whether
recursive loops have regimes, but how moveable those regimes are
under controlled perturbation.

\subsection*{What we did and what we found}

We repeatedly fed an LLM its own outputs, injected text at a chosen
step, and measured whether the run ended in a different end-state
than its paired no-injection twin. The key result is sharp: in
\textbf{append-mode} (the loop keeps adding to a growing transcript),
text crafted to look like it belongs in the conversation produces
a measurable dose response. About \textbf{40 tokens} of injection are
enough to flip the end-state in roughly half of runs - the
half-effect dose, written $\mathrm{ED50}_{\mathrm{raw}} \approx 40$.
In our canonical bounded-memory loop (the loop keeps only the most
recent ~12,000 characters), durable change is not achieved even at
400 tokens. Raw end-state disagreement rises and plateaus near 67\%,
while two \textit{identical} unperturbed runs already disagree about 35\%
of the time from random sampling alone - the \textbf{stochastic floor} -
so the real effect of the injection saturates around +32 percentage
points above noise. The strict endpoint - knocked into a new
end-state and still there at the end of the run - never crosses
50\% in the tested 5-400 token range \textit{under bounded memory}. When
we instead let the loop keep the full conversation history, durable
change does climb past 50\% at higher doses, with two crossings
depending on how strict we are (~400 tokens loose, ~1,500 tokens
strict; see "Five numbers" below). \textbf{Replace-mode} loops (each step rewrites the
context) look much more fragile, but a direct overwrite-vs-insert
probe shows that most of that apparent switching is the update rule
itself overwriting the state, not the model genuinely flipping.

\subsection*{Five numbers to remember}

\begin{itemize}
  \item \textbf{About 40 tokens of injected text flip the final end-state in
  roughly half of append-mode runs} under the raw endpoint,
  $\mathrm{ED50}_{\mathrm{raw}} \approx 40$. This is the simple "did
  the final cluster differ from the paired control?" measure. It
  is not a durable-redirection threshold. Raw switching plateaus
  near \textbf{67\%}, while two no-injection controls already disagree
  about \textbf{35\%} of the time, so the largest measured net effect is
  only about \textbf{+32 percentage points} at dose 400.
  \item \textbf{In the bounded-memory loop, durable change does not reach
  50\%.} With the canonical 12,000-character tail clip,
  destination-coherent persistence stays near \textbf{16\%} on the main
  \texttt{context\_tail} observable, and retained source-basin escape
  reaches only about \textbf{36\%} at dose 400. The perturbation is
  usually clipped out before the final measurement, so bigger
  injections do not automatically produce durable escape if the
  loop forgets them.
  \item \textbf{In the full-history loop, durable departure appears at higher
  doses.} Retained source-basin escape crosses 50\% at about \textbf{400
  tokens} and saturates near \textbf{75-80\%} by 1,500 tokens. The
  stricter destination-coherent persistence endpoint first reaches
  a 50\% point estimate near \textbf{1,500 tokens}, but the Wilson 95\%
  interval at that dose is \textbf{[0.41, 0.61]}, so the crossing is
  statistically coarse; the half-effect region is best described
  as roughly \textbf{1,000-2,000 tokens}.
  \item \textbf{The high-dose destination-coherent dip is a finite-horizon,
  endpoint-definition-sensitive feature, not a stable structural
  attractor split.} At doses 1500 / 2000 / 3000 destination-coherent
  persistence was \textbf{0.51 / 0.32 / 0.41} with heterogeneous
  concatenated perturbations and \textbf{0.44 / 0.32 / 0.40} with
  homogeneous repeated perturbations, refuting the heterogeneity
  hypothesis. A four-step falsification battery further weakens the
  structural reading: about half the dip comes from comparing a
  destination cluster sampled while the perturbation is still raw
  to a terminal cluster dominated by model-generated continuation,
  and under the frozen canonical cluster basis the dip drops by
  73\% when trajectories are extended from step 29 to step 79
  (\textbf{-0.143 → -0.039}, with the step-79 95\% bootstrap interval
  straddling zero). The simpler retained source-basin escape
  endpoint stays high and monotone across all these checks. A
  useful secondary result from the heterogeneity control: retained
  source-basin escape was lower under homogeneous repetition
  (\textbf{0.54-0.58}) than under heterogeneous concatenation
  (\textbf{0.74-0.79}), suggesting that repeated content is easier for
  the loop to recover from.
  \item \textbf{Replace-mode "fragility" is mostly the update rule.} Under
  the default overwrite protocol, replace-mode switching looks
  near-saturated because the injected text becomes the next state.
  Under insert mode, where the perturbation is visible for one
  generation but is not written as the next state, replace-mode
  switching falls to \textbf{12-32\%}. The overwrite-insert gap is
  \textbf{60-80 percentage points} in replace-mode but only \textbf{14-34
  percentage points} in append-mode, and insert-only rates follow
  $O1 \ge O2 \ge O3$.
\end{itemize}

\subsection*{Practical implications}

\begin{enumerate}
  \item \textbf{Stress-test recursive agents with persistence, not motion.} The
   right robustness question is not "did the trajectory move after
   the perturbation?" but "did it escape and stay escaped after
   several more recursive steps?" An evaluation that counts only
   immediate movement (or just whether the final cluster disagrees
   with the paired control) will over-report fragility, because
   much of the movement is transient - the trajectory wanders away
   and then returns.

  \item \textbf{Always measure the stochastic floor.} In the main append-mode
   setting, two paired control runs already end in different
   clusters about 35\% of the time with no perturbation. That means
   a reported 50\% switching rate is not automatically evidence of
   successful redirection; much of it may be ordinary sampling
   divergence. Recursive-loop evaluations should include
   control-vs-control baselines and report raw, net, and
   persistent-escape rates separately.

  \item \textbf{Treat context-update rules as safety-critical design choices.}
   Append-style updates preserve prior context, creating a real but
   bounded barrier: about 40 tokens for raw switching, with
   persistent escape not reached in the tested range. Replace-style
   updates do not show the same kind of injected-token barrier,
   because the rule itself discards the old state. Claims that
   "replace-mode loops are easy to redirect" should be read as
   claims about overwrite mechanics, not necessarily about weak
   model attractors.

  \item \textbf{Do not assume small adversarial nudges durably redirect
   settled loops.} Even adversarial continuations (text crafted to
   look like a natural next part of the trajectory) selected from
   the same regime top out near 67\% raw switching. Under the
   canonical bounded-memory loop, neither persistence endpoint
   reaches 50\% in the tested 5-400 token range. Under the
   full-history protocol the picture is dose-dependent: retained
   source-basin escape crosses 50\% near 400 tokens; destination-
   coherent persistence first crosses 50\% near 1,500 tokens. For
   jailbreak or agent-redirection evaluations, the meaningful
   target is sustained post-attack behavior measured against an
   appropriate persistence endpoint, not a one-step perturbation
   success - and the threshold depends on what the deployed loop's
   memory policy retains.

  \item \textbf{Evaluating indirect prompt injection (IPI).} Many
   prompt-injection benchmarks (standardized tests for comparing
   defenses) mainly test immediate reaction: does the model follow
   the injected instruction right away? That is closest to our
   \textit{raw-switching} endpoint, but it is not the same as durable
   takeover. A defense
   that pushes raw-switching $\mathrm{ED50}$ from 40 to 200 tokens
   is a quantitative defense improvement; a defense that prevents
   \textit{persistent escape} is a qualitatively stronger claim - the
   model may briefly follow the injection but recover within a few
   turns. Serious IPI evaluations should report all three endpoints
   AND the stochastic floor (so that ordinary sampling randomness
   isn't counted as defense failure), AND distinguish injections
   that land in the system prompt or replaced context (replace-
   style, overwhelms by overwriting) from injections that land in
   tool output or appended document text (append-style, has a real
   but measurable token barrier). A defense that only blocks
   immediate following of the injection is leaving durable-
   redirection attacks on the table.

  \item \textbf{Adapting the protocol to human-LLM influence studies.} The
   \textit{lorem} (placeholder gibberish, like "lorem ipsum") /
   \textit{neutral} (off-topic but coherent) / \textit{adversarial} (on-topic
   text designed to steer the model) content contrast is the most
   directly transferable piece of the framework for measuring how
   LLMs affect users. A study aiming to detect content-specific LLM
   influence on humans (rather than generic "the LLM was in the
   loop" effects) should run three matched conditions: (i) targeted
   LLM intervention, (ii) generic on-topic LLM chat without the
   targeted move, (iii) off-topic LLM small-talk, against a no-LLM
   control. If the targeted condition shows persistent change
   relative to the generic and off-topic conditions, the influence
   is content-specific. If all three look similar, what was
   measured is \textit{generic-LLM-presence} - what psychology calls a
   Hawthorne or engagement effect, where the \textit{presence} of the
   intervention matters more than its specific content - not the
   LLM's content per se. The three-endpoint decomposition
   then applies on the human side: did the user's stated goal /
   sentiment change at the LLM turn (raw), above the natural drift
   two humans show without an LLM (net), and was the change still
   in place several turns or sessions later (persistent)?

  \item \textbf{Hallucination-detection design.} A \textit{hallucination} is a
   false or unsupported claim the model presents as if it were
   true. This framework does not detect hallucinations directly:
   it measures embedding-space stability, not whether the answer
   is factually correct, and a locked-in hallucination is a stable
   hallucination. But three pieces transfer to hallucination
   evaluations once an external grounding signal - an outside
   source of truth like retrieved documents, a fact-checker, or a
   trusted reference answer - is available. \textit{(i)} Self-consistency
   detectors (methods that ask the model the same question
   multiple times and flag disagreement as possible hallucination)
   assume that two runs of the same prompt would agree if the
   model were not hallucinating; on our setting paired runs already
   diverge ~35\% of the time from sampling alone. Such detectors
   should subtract a measured stochastic floor before scoring.
   \textit{(ii)} Transient vs persistent hallucination is a real
   distinction: a hallucinated claim that the model self-corrects
   within a turn is qualitatively less dangerous than one that
   propagates. The persistent-escape endpoint maps directly:
   measure whether the hallucinated content stays in the
   trajectory several turns later. \textit{(iii)} Hallucination
   \textit{robustness} (how hard the model is to push off a hallucinated
   answer) under counter-evidence has a clean dose-response
   framing: how many tokens of contradicting evidence make the
   model retract? Same protocol, but the axis is factual
   correctness instead of cluster identity.

  \item \textbf{Use the protocol for model and architecture comparisons.}
   The raw / net / persistent-escape decomposition (splitting the
   total switching rate into its three component parts) gives a
   portable measurement unit for comparing models, prompts, memory
   policies, and agent scaffolds. Instead of saying one system is
   "more stable", an evaluator can report the dose-response curve,
   stochastic floor, plateau, and persistent-escape threshold - 
   numbers that are directly comparable across vendors (different
   model providers).
\end{enumerate}

\subsection*{Caveats}

The main text-generating model is OpenAI's \texttt{gpt-4o-mini}, with a
within-vendor replication (rerun on another model from the same
provider) on \texttt{gpt-4.1-nano} but no full cross-vendor replication yet
(rerun on models from other companies). The "end-state" measurement uses a sliding window of the recent
conversation, roughly the last 6-7 steps, so it dilutes
one-off changes confined to the final reply. When we instead judge
only the final reply, persistence in the bounded-memory loop rises
to 27\% at the largest 400-token dose (§5.1.1). The crucial scope
note is \textit{which loop is being measured}. We deliberately study two
memory policies: a \textit{bounded-memory} loop (the running conversation
is tail-clipped at ~12,000 characters - smaller than the model's
native ~128,000-token capacity, mirroring real production scaffolds
that use rolling windows, summaries, or retrieved-document memory)
and a \textit{full-history} loop (no artificial truncation within the
30-step horizon; only the model's native context is used). The two
give different persistent-escape thresholds: under bounded memory,
destination-coherent persistence plateaus around 16\% (canonical
context\_tail observable) to 27\% (single-generation output observable)
across all tested doses to 400 tokens, because the perturbation gets
clipped out within ~10-15 post-injection steps. Under full-history,
retained source-basin escape (kicked AND outside the starting basin
at terminal step) crosses 50\% near 400 tokens; the stricter
destination-coherent endpoint first crosses 50\% near 1,500 tokens
(§5.1.3, Fig 5). These two pictures are not contradictory: bounded
memory describes what happens when the loop forgets the
perturbation; full-history describes what happens when it doesn't.
Real production systems sit somewhere in between, depending on
their memory policy. A bounded-memory production loop is harder to
durably redirect with a single injection - but, by symmetry, also
harder to course-correct once it does drift. Under full history,
the stricter destination-coherent endpoint becomes non-monotonic at
very high doses (drops at dose 2000, partial recovery at dose
3000) when measured at the canonical 30-step horizon. A
four-step falsification battery (heterogeneity control, cluster-
granularity sweep, transition-entropy diagnostic, and long-horizon
trajectory continuation) recasts this dip as a finite-horizon,
endpoint-definition-sensitive feature rather than a stable
structural asymmetry: roughly half is attributable to endpoint
timing, and the residual drops by 73\% (with bootstrap interval
straddling zero) when trajectories are extended another 50 steps
under the frozen canonical basis (§5.1.3). The regimes are
\textit{operational embedding-space measurements} - patterns defined by
where outputs land in the model's vector representation - and not
mechanistic proofs that the LLM has classical attractors in any
deeper internal sense. The human-LLM,
IPI, and hallucination-detection implications above describe how the
\textit{measurement protocol} transfers; the paper itself does not collect
data on human users, production-system attacks, or factuality
grading, and "trajectories cluster in similar regions" is not the
same as "the answers are correct."

\section{Introduction}

\subsection{Phenomenon}

Recursive LLM systems increasingly feed model outputs back into future prompts: agents revise plans, assistants summarize tool results, and dialog systems carry state forward. Such loops often appear to settle into attractor-like regimes, but an operational question remains unresolved: how many injected tokens are required to move a settled loop, and does that movement persist?

We answer this by separating the generator from the context-update rule. In append-mode continuation, adversarial in-distribution perturbations produce a real raw dose response, with $\mathrm{ED50}_{\mathrm{raw}}\approx 40$ tokens, but paired controls already diverge about 35\% of the time. Persistent effects are conditioned on both memory policy and endpoint definition: under the canonical bounded-memory loop they stay near 16\% across the 5-400 token range. Under a full-history protocol, retained source-basin escape (kicked AND outside the original cluster at the terminal step) reaches 50\% near 400 tokens, while the stricter destination-coherent endpoint (terminal cluster equals the immediate post-injection cluster) first reaches 50\% near 1,500 tokens. In replace-mode, apparent fragility is largely an overwrite effect of the update rule. Thus the stability of recursive loops is not a property of the model alone; it is jointly determined by model, memory policy, perturbation content, perturbation dose, and persistence criterion.

\begin{savenotes}
\begin{figure}[h!]
\centering
\includegraphics[width=0.95\linewidth]{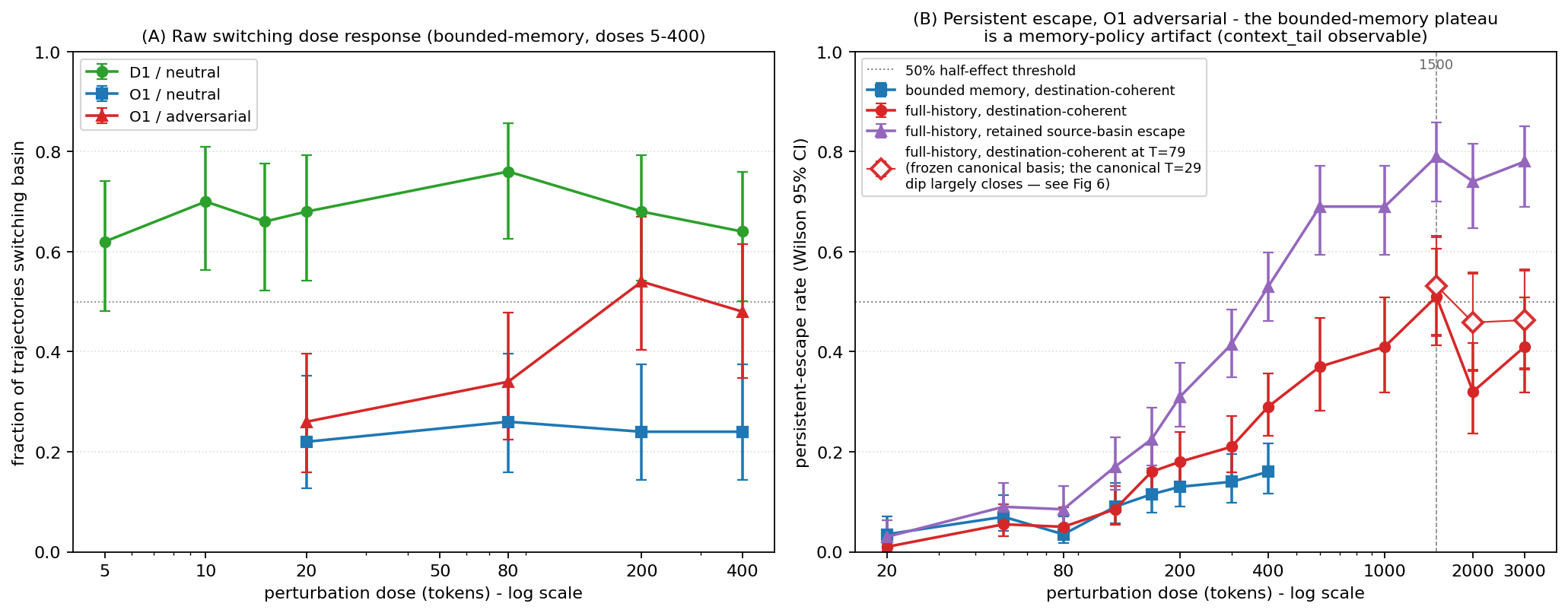}
\caption{\textbf{Headline dose response: raw switching and memory-policy-conditioned persistent escape.} Two panels. (A) Raw-switching dose response in the canonical bounded-memory loop, comparing regime × content (D1/neutral, O1/neutral, O1/adversarial) at doses 5-400 tokens. Replace-mode regimes O2/O3 are excluded because the default state-reset overwrite protocol makes their switching tautological at the state-write step (\(X_{t+1}=\operatorname{clip}(\text{perturbation\_text})\); see §5.2 and Fig 7). (B) Persistent-escape dose response for O1 adversarial under four definitions. Bounded memory, destination-coherent persistence (12K tail-clip; trajectory stays in its post-injection cluster at the terminal step; blue filled squares) plateaus at ~16\% across doses 5-400. Full-history, destination-coherent persistence at the canonical T=29 horizon (red filled circles, n=200 at doses 20-400 merged with n=100 at 600-3000) first reaches 0.51 at dose 1500 then becomes non-monotonic at 2000-3000. Full-history, destination-coherent persistence at the long horizon T=79 under the frozen canonical PCA + K-means $k=12$ basis (red open diamonds, doses 1500/2000/3000 only) sit at 0.53 / 0.46 / 0.46—the canonical T=29 dip is largely closed under longer continuation, see Fig 6 for the full T=29..79 decay table. Full-history, retained source-basin escape (kicked AND outside the pre-injection cluster at the terminal step; purple triangles) is monotone and saturating: it crosses 50\% between doses 300-400 and saturates near 75-80\% by dose 1500. The destination-coherent T=29 dip at high doses is not loss of displacement (the kicked rate stays ~0.96 at doses ≥1500); §5.1.3 traces it to a finite-horizon, endpoint-definition-sensitive feature, formalized via a four-step falsification battery. The bounded-memory plateau holds for both endpoints (full breakdown in §5.1.3 / Fig 5). Source: \texttt{data/aggregated/perturbation\_dose\_response/dose\_response\_2panel.png}.\\[2pt]{\footnotesize\itshape Fig 1 summarizes two dose-response findings. Panel A: raw-switching orientation in the canonical bounded-memory loop. D1/neutral saturates at the smallest dose (~62\% at 5 tokens), O1/neutral stays at the noise floor (22-26\%), O1/adversarial rises with dose to ~54\% at 200 tokens - same loop, different sensitivity to in-distribution vs off-topic text. Panel B: persistent-escape dose response under four endpoints (bounded destination-coherent, full-history destination-coherent at T=29, full-history destination-coherent at T=79 under the frozen canonical basis, full-history source-basin escape). Bounded-memory plateau ~16\% across doses 5-400; full-history source-basin escape crosses 50\% near 400 tokens and saturates ~75-80\%; full-history destination-coherent at T=29 first reaches 50\% near 1,500 tokens but is non-monotonic (0.51/0.32/0.41 at doses 1500/2000/3000). The non-monotonicity at the canonical 30-step horizon is not loss of displacement (kicked rate stays ~0.96); §5.1.3's four-step falsification battery recasts it as a finite-horizon, endpoint-definition-sensitive feature. The T=79 overlay (red open diamonds) measured under the same frozen canonical PCA + K-means k=12 basis as T=29 sits at 0.53/0.46/0.46 — the canonical dip largely closes by T=79, with the dip contrast dropping 73\% from -0.143 at T=29 to -0.039 at T=79 (Fig 6). Falsifying outcomes: D1 not saturating at small doses, O1/adversarial not exceeding O1/neutral at any dose, bounded and full-history curves overlapping throughout in panel B, or the T=79 open diamonds tracing the same dip pattern as the T=29 filled circles. Full breakdown and decomposition table in §5.1.3.}}
\end{figure}
\end{savenotes}

The same distinction appears in coding agents, where the loop state may include tool logs, patches, summaries, pinned requirements, and recent files; the memory policy determines whether new information accumulates, overwrites, or is role-structured.

The state-generator-update view makes this distinction explicit: the state is the prompt-visible memory at the current step, the generator is the stochastic language model, and the update rule maps state and output into the next state. If the term context-update nudge is used, it denotes this update operation, not a separate intervention inside the model.

Recent work has shown that recursive LLM trajectories can exhibit contractive, oscillatory, exploratory, degenerate, or convergent self-referential regimes. These studies establish that attractor-like structure is empirically visible, but they do not measure the perturbation dose required to move a trajectory between regimes, nor do they separate that dose from ordinary stochastic divergence or from update-rule overwrite mechanics.

The resulting measurement problem has three parts. For a settled recursive trajectory, how many injected tokens are required to produce final-cluster switching, how much of that switching exceeds the stochastic control floor, and how often does the perturbation produce a visible basin jump that persists to the terminal step? A single final disagreement cannot answer this, because it may reflect real redirection, unperturbed sampling divergence, or a transient displacement followed by recovery.

\subsection{Question}

We study recursive LLM loops as bounded stochastic systems under a state-generator-update decomposition. The same generator can behave differently under append, replace, and dialog updates because those updates expose different histories, role structures, and memory contents to the next sampling step.

\begin{enumerate}
  \item \textbf{Architecture:} How do append, replace, and dialog update rules alter the accessible loop state?
  \item \textbf{Dose response:} For a settled trajectory, how does switching probability vary with injected-token dose and perturbation type?
  \item \textbf{Persistence:} Which apparent switches exceed the stochastic control floor and remain through the terminal step?
\end{enumerate}

These questions convert "the loop gets stuck" into measurable quantities: raw switching, net switching, and persistent escape. Raw switching is terminal cluster disagreement between a perturbed trajectory and its paired control. Net switching subtracts the measured control compared with control stochastic floor. Persistent escape requires a visible basin jump after injection that remains present at the terminal step.

\subsection{Contributions}

The paper's contributions are best stated as five claim-level takeaways.

\textbf{Claim 1: recursive-loop stability is jointly determined by generator and update rule (§3).} Append, replace, and dialog loops differ because their context-update operators expose different histories to the model. The state-generator-update formalism treats the update operator as a first-class component of the loop, rather than an implementation detail. It also yields a finite-time access result for replace-mode loops and motivates the append-mode prediction that accumulated prior context changes perturbation response.

\textbf{Claim 2: perturbation response decomposes into raw switching, net switching, and persistent escape (§3, §5).} These endpoints separate true redirection from sampling divergence and transient displacement. Raw switching measures final-cluster disagreement with a paired control. Net switching subtracts the natural control compared with control floor. Persistent escape is stricter: the injected text must produce a visible basin jump that remains through the terminal step. The strict stability question is therefore not whether a run moves immediately after injection, but whether it stays moved after subsequent recursive updates.

\textbf{Claim 3: append-mode continuation has a finite raw dose response, and a persistent-escape threshold that is memory-policy-conditioned (§5.1, §5.1.3).} In the dense adversarial in-distribution append-mode rerun, $\mathrm{ED50}_{\mathrm{raw}}$ estimates are 36, 41, and 52 tokens under pooled four-parameter logistic fitting, mixed-effects logistic modeling, and family-cluster bootstrap, respectively. Raw switching rises with dose but plateaus near 67\%, while paired controls already diverge about 35\% of the time. The largest observed net effect is about +32 percentage points at 400 tokens. The persistent-escape endpoint plateaus at ~16\% across doses 5-400 tokens \textit{under the canonical bounded-memory loop} (12K tail-clip). Under a full-history protocol (no artificial truncation within the 30-step horizon), the two persistence endpoints diverge with dose: \textit{retained source-basin escape} (kicked at injection AND outside the pre-injection cluster at terminal step) reaches 50\% near 400 tokens and saturates at ~75-80\%; the stricter \textit{destination-coherent persistence} endpoint (terminal cluster equals the immediate post-injection cluster) first reaches 50\% near 1,500 tokens but becomes non-monotonic at larger doses. The non-monotonicity is a finite-horizon, endpoint-definition-sensitive feature rather than a structural attractor split: a four-step falsification battery in §5.1.3 (heterogeneity control, cluster-granularity sweep, transition-entropy diagnostic, and long-horizon trajectory continuation) attributes roughly half the canonical magnitude to endpoint timing and the residual to finite-horizon scoring; under a 50-step extension (\texttt{exp\_perturb\_O1\_ed50\_higher\_noclip\_extended}) the dip drops from -0.055 at step 29 to 0.000 at step 79, with all three high doses converging on a common destination-coherent floor of ~0.15. The kicked rate is ~0.96 at doses ≥1500 throughout, so the high-dose loss of destination coherence is not loss of displacement. The bounded-memory plateau is a real measurement of the 12K-clipped loop, but not a memory-policy-invariant property of the model's response (§5.1.3, Fig 5). Out-of-distribution neutral and lorem perturbations remain close to a drift floor rather than matching the adversarial continuation response.

\textbf{Claim 4: replace-mode apparent fragility mostly reflects state-reset overwrite (§5.2), not necessarily weak attractors.} Replace-mode paraphrase and summarize-and-negate loops show near-saturated raw switching across tested doses. But under the default intervention, the perturbation replaces \(Y_t\), and replace-mode update then sets \(X_{t+1}=\operatorname{clip}(\text{perturbation\_text})\): the next state simply \textit{is} the perturbation, so switching at the state-write step is tautological. Insert-mode probes, which expose the model to the perturbation but preserve the model's own output as the state, reduce replace-mode switching to 12 to 32\%. The F3 cross-loop check (§5.2.1) shows that the overwrite-insert gap is 14-34 pp in append-mode versus 60-80 pp in replace-mode, and insert switching orders \(O1 \ge O2 \ge O3\), so the insert estimand is regime-conditional rather than a regime-invariant model-behavior constant. These results separate state-write access from model-mediated redirection and make memory policy a safety-relevant design choice.

\textbf{Claim 5: perturbation response resolves regimes that bulk geometry alone cannot distinguish (§5.10).} Drift, dispersion, cluster persistence, and low-dimensional visualisations describe trajectory shape, but they can leave stylistic multi-basin dialog and oscillatory patterns ambiguous. Perturbation dose response adds the missing load test: two regimes with similar bulk geometry can differ in raw switching, stochastic-floor-adjusted switching, and persistent escape. The empirical potential landscape $V(x) = -\log \rho(x)$ is therefore used as a descriptive view of basin organization, not as an independent substitute for the behavioral endpoints.

> \textbf{Originality at a glance.} Prior work on recursive LLM loops classifies regimes (contractive, oscillatory, exploratory). This paper turns regime classification into a quantitative perturbation pharmacology. We separate the model from the context-update rule, treating append, replace, and dialog memory policies as part of the dynamics. Three endpoints (raw, net, persistent-escape) decompose final-cluster disagreement into sampling divergence, transient displacement, and durable redirection. The persistent-escape family further splits into source-basin escape and destination-coherent persistence. Two empirical results that did not exist before: the overwrite-vs-insert counterfactual isolates state-write mechanics from model behavior in replace-mode loops, and a four-step falsification battery on the high-dose destination-coherent dip (heterogeneity control, cluster-granularity sweep, transition-entropy diagnostic, and 50-step trajectory continuation) localizes the dip to a finite-horizon, endpoint-definition-sensitive feature rather than a structural attractor split. Memory policy is a safety-relevant first-class variable, not an implementation detail.

All trajectories, configurations, analysis scripts, and replication artifacts are publicly released, with automated checks linking the reported numerical claims to the underlying result tables; within-vendor replication on \texttt{gpt-4.1-nano} is also provided.

\section{Background and related work}

Existing work has begun to describe recursive LLM inference as a dynamical system, identifying convergence, cycling, divergence, and dimensional collapse in semantic space. What remains under-specified is the \textit{mechanism of intervention}: which part of the loop is the model, which part is the context-update rule, and how much external text is required to move a trajectory between empirically supported basins. This section reviews the relevant attractor, degeneration, dimensionality, and dialog literatures, with emphasis on that gap: attractor-like behavior has been observed, but perturbation dose-response and token-scale barrier measurement have not been systematically measured.

\subsection{Attractors in neural dynamics}

Attractor analysis has long been central to the study of recurrent neural systems. In classical recurrent networks, one studies fixed points, cycles, and low-dimensional manifolds through tools such as Jacobian linearization, Lyapunov spectra, stability analysis, recurrence, and effective dimensionality \citep{hopfield1982,sussillo2013,maheswaranathan2019}. These methods ask whether repeated application of a state-update rule drives a system toward stable regions, cycles, or higher-dimensional wandering. The resulting vocabulary of attractors, basins, recurrence, contraction, and instability remains useful even when the underlying system is not a smooth deterministic map.

Recursive language-model loops differ from classical recurrent neural networks in two important ways. First, the visible state is text, not a continuous hidden vector whose update is directly available for differentiation. Second, the loop usually contains a sampling step and a context-update rule chosen by the experimenter. A model may generate text, but the experimenter decides whether that text replaces the previous state, is appended to it, is summarized, is inserted into a dialog transcript, or is mixed with external material. The effective recurrence is therefore not the language model alone. It is the composition of a generator with a context-update mechanism.

For this reason, exact local linearization is generally unavailable at the level of the visible text loop. Empirical analogs are needed. In this paper, recurrence, dwell structure, ensemble spread, finite-time Lyapunov-like growth, effective dimension, clustering, and density-derived landscapes are used as operational diagnostics rather than as proofs of classical attractors. The goal is not to claim that a recursive LLM loop is a smooth dynamical system in the strict mathematical sense. It is to ask whether repeated text transformations show stable, recurrent, or collapsive structure in embedding space, and whether that structure can be perturbed in a measurable way.

\subsection{Attractor observations in language models}

The dynamical-systems framing of LLM inference loops has emerged rapidly. Recent work formalizes recursive LLM transformations as discrete dynamical systems in semantic space and classifies trajectories into regimes such as contractive convergence, oscillation, and exploratory divergence. The closest regime-classification study identifies contractive, oscillatory, and exploratory behavior using drift, dispersion, and cluster-persistence measures. Our use of recurrence, dimensionality, density landscapes, and perturbation response is complementary: these diagnostics are introduced not to relabel the same regimes, but to estimate how robust basin membership is under controlled textual intervention.

That line of work shows that prompt and loop design can select qualitatively different regimes. Iterative paraphrasing may drive semantic contraction, while iterative negation or adversarial transformations may drive dispersion. Such results establish that recursive LLM systems are not merely independent samples from a generator. The text produced at one step shapes the next state, and the chosen update rule can create repeatable geometric structure in the trajectory. This motivates treating the whole loop as the object of analysis.

A related literature studies degeneration in autoregressive generation. Earlier work characterized repetitive collapse, blandness, and looping as sampling and decoding failures, motivating alternatives such as nucleus sampling and more cautious truncation strategies \citep{holtzman2020,carlini2021}. These studies were not primarily framed in terms of attractors, but they documented phenomena that are naturally read through a dynamical lens: once a text process enters a repetitive template, later outputs may remain trapped in a narrow region of form or content. The attractor vocabulary does not replace the degeneration literature. It provides a way to connect repetitive symptoms to recurrence, dimensionality, and basin stability.

Dimensionality measures are also relevant. \citet{arxiv251021258} use correlation dimension to quantify geometric complexity in generated text and report dimensionality drops during training transitions and degeneration. Their observable differs from ours, but their result motivates treating dimensional collapse as a relevant diagnostic when recursive loops become repetitive or template-bound. We therefore use dimensionality measures as one component of the attractor-like diagnostic battery, without equating correlation dimension with our basin definitions.

Other work on neural dynamics and representation geometry provides useful measurement precedents. Effective dimension, participation ratio, cluster persistence, and trajectory geometry have been used to characterize how neural and model states occupy representational spaces. In LLM recursion, however, the observable trajectory is mediated by text and sampling. This makes basin claims more fragile than in systems where the latent state transition is directly specified. Robustness tests therefore become central. A cluster that appears stable in an unperturbed trajectory may be a sampling artifact, a prompt-template artifact, or a genuinely supported basin candidate. Perturbation response helps distinguish these possibilities.

\subsection{What this paper adds}

Relative to recent regime-classification studies of recursive LLM loops, this paper shifts the unit of analysis from trajectory shape alone to \textit{intervention cost}. Prior work asks whether trajectories contract, cycle, or disperse. We ask how those regimes respond to controlled perturbations, whether switching probability scales with injected-token dose, and whether dialog-state updates create basin structure not captured by single-stream operators. The comparison is therefore not primarily one of model scale or dataset size, but of observable: perturbation dose-response rather than unperturbed trajectory geometry alone.

This paper makes three conceptual moves. First, it separates the recursive loop into the visible state, the generator, and the context-update nudge. Second, it treats perturbations as calibrated interventions rather than informal prompt changes, allowing barrier height to be operationalized as a dose-response quantity. Third, it treats dialog as a distinct update architecture, not merely as another prompt template, and tests whether role-structured state creates basin structure different from single-stream recursion.

Our labels are not a one-to-one replacement for the contractive/oscillatory/exploratory taxonomy. We retain contractive and oscillatory as comparable high-level behaviors, treat absorbing collapse as a distinct low-diversity endpoint observed under specific recursive operators, and introduce D1/D2 as dialog-specific regimes whose definition depends on perturbation response and state-update structure rather than dispersion alone.

\citet{arxiv251210350} is the closest recent precedent for treating recursive LLM text transformations as discrete dynamical systems in semantic space. That work demonstrates regime classification on a small set of proof-of-concept loops; the present paper extends the framing to perturbation dose-response, explicit generator/nudge separation, dialog-state updates, and larger paired-control trajectory batteries. The emphasis here is not that prior work lacked dynamical concepts, but that the intervention mechanism had not been made the central observable.

A separate literature studies recursive degradation under \textit{training}-time recursion, where models are repeatedly retrained on their own generated outputs \citep{arxiv230517493,arxiv230701850}. These works document model collapse and "model autophagy disorder" when synthetic outputs replace fresh data. Our setting is \textit{inference}-time recursion of a frozen generator: the model parameters do not change, only the visible state and its perturbation. The two settings share the question of whether recursive self-feeding produces convergence, but answer it through different mechanisms; we focus on regime stability and intervention cost within a single inference loop, not parameter drift across training rounds.

Indirect prompt injection (IPI) benchmarks (\citet{arxiv230212173}; \citet{arxiv231214197} / BIPIA; \citet{arxiv240613352} / AgentDojo) are also adjacent. They evaluate whether an agent follows malicious instructions hidden in retrieved documents or tool output, typically by measuring immediate compliance at the next step. The endpoint decomposition we develop - raw switching, net switching, persistent escape - translates directly to IPI evaluation, where it provides the missing distinction between transient compliance and durable agent redirection. Our overwrite-vs-insert counterfactual (§5.2) further isolates how much apparent IPI vulnerability is model behavior versus scaffold state-write mechanics, a control IPI benchmarks could readily adopt.

Self-refinement and iterative correction loops \citep{arxiv230317651,arxiv221100053,arxiv230803188,arxiv231001798} are a third adjacent literature: a model is asked to revise its own output across multiple rounds. These works document that refinement loops can degrade rather than improve, which we read as evidence for absorbing or oscillatory regimes under specific operators. We do not contribute to refinement-quality measurement; we contribute the dose-response and persistence machinery that lets such loops be characterized as dynamical systems with measurable barriers.

The state, generator, and nudge separation is important because recursive LLM behavior is not determined by the model alone. The same generator can be placed in a replacement loop, an append loop, a rolling-window loop, a summarization loop, or a role-structured dialog loop. Each choice changes what information persists, what is forgotten, and how external text enters the next step. A dynamical description that treats the prompt as the whole state can miss this architecture. By explicitly separating these components, one can ask whether an apparent attractor belongs to the generator's tendencies, to the context-update rule, or to their interaction.

The perturbation focus also changes the meaning of a basin. In unperturbed geometry, a basin candidate may be identified by repeated returns, dense occupancy, contraction, or cluster persistence. Under intervention, the question becomes how much injected text is required to alter the trajectory's terminal cluster or produce a more durable escape. This yields an operational barrier height in tokens under a specified perturbation protocol. The measure is not a thermodynamic barrier and not a universal property of the model. It is a reproducible dose-response quantity tied to a generator, update rule, observable, clustering convention, injection time, and perturbation family.

This intervention view also clarifies why dialog deserves separate treatment. Dialog updates are not just longer prompts. Role labels, speaker alternation, local adjacency, and conversational memory can determine which parts of the state are salient at the next step. A dialog system may therefore show basin structure tied to conversational role, topic commitment, or local conversational obligations rather than to semantic drift alone. Treating dialog as a nudge architecture allows these possibilities to be tested rather than assumed.

Finally, this paper borrows selectively from related theoretical work while keeping the claims operational. \citet{arxiv260419740} study stochastic gradient descent dynamics on neural-network loss landscapes using random dynamical systems and introduce a sharpness-dimension generalization bound near the edge of stability. Their setting is parameter space, Hessian-anchored, and training-time. Ours is embedding space, gradient-free, and inference-time for a frozen generator. We use comparable dimensionality language only as a diagnostic analogy, not as a transfer of their training-dynamics theory.

\subsection{Effective potential and geometric barriers}

Empirical density landscapes are often summarized by an effective potential \(V(x)=-\log \rho(x)\), where \(\rho\) is an estimated stationary density in a chosen coordinate system. In physical systems this construction is tied to free-energy analysis; here it is used only as a descriptive transformation of trajectory density in embedding space. The resulting landscape can suggest wells, ridges, and low-density separations between frequently visited regions, but its numerical scale depends on the projection, density estimator, and smoothing choices. We therefore treat potential-derived barriers as descriptive geometric summaries, not as thermodynamic quantities or direct estimates of model-internal energy.

This distinction matters because an embedding-space density is not the model's internal probability distribution. It is a distribution over observed trajectory embeddings after choices about text observable, embedding model, dimensionality reduction, and sampling. A high-density region may correspond to repeated semantic content, repeated rhetorical form, or repeated role-state position. A low-density gap may indicate a meaningful separation between basin candidates, but it may also reflect projection artifacts or sparse sampling. Potential landscapes are therefore used as one part of a triangulation strategy, alongside recurrence, dimensionality, clustering, dwell, and perturbation response.

\subsection{Dialog as a distinct dynamical setting}

Most recursive-LLM dynamics studies focus on single-stream operators such as continuation, paraphrase, or negation. Multi-turn dialog differs because generated text is written into a role-structured conversational state rather than merely appended or replaced as a homogeneous string. This makes dialog a natural test case for whether the context-update rule itself can shape basin structure. Although multi-agent and generative-agent work has studied dialog for capability, memory, and alignment, embedding-space attractor analysis of role-structured recursive dialog remains comparatively undeveloped.

Dialog introduces structure that is invisible in single-stream recursion. Speaker roles determine which text is interpreted as user instruction, assistant response, system constraint, memory, or prior conversational context. Turn order creates local obligations, such as answering the immediately preceding message, while the longer transcript can preserve topic commitments or interactional style. These features make the nudge more than a storage policy. It becomes part of the mechanism that selects what the generator treats as relevant state.

For this reason, dialog is not merely a larger prompt class in this study. It is a separate family of recursive update rules whose stability can be probed by the same perturbation logic used for single-stream operators. If role-structured state creates distinct basin candidates, those candidates should be visible not only in trajectory geometry but also in how they respond to injected text, neutral insertions, adversarial turns, or changes in injection time. This motivates analyzing dialog regimes through both geometry and intervention cost.

\subsection{Agent scaffolds, memory architectures, and adversarial attacks}

The paper's claims about memory policy and persistent redirection touch on three additional adjacent literatures that we engage briefly. Scaffolded reasoning systems wrap an LLM in iteration, search, or self-reflection: ReAct \citep{arxiv221003629} interleaves reasoning steps with tool actions; Tree of Thoughts \citep{arxiv230510601} performs test-time search over generated thoughts; Reflexion \citep{arxiv230311366} feeds an explicit self-reflection trace back into the next attempt. These are recursive scaffolds in which past observations and reflections become future state, so the visibility/persistence questions we develop apply to them with adjusted observables. Memory architectures for LLMs include MemGPT \citep{arxiv231008560}, which manages tiered context buffers explicitly, and retrieval-augmented generation \citep{arxiv200511401}, which retrieves external documents at each step. Both correspond to specific memory policies in our state-generator-nudge framework, and our overwrite-vs-insert decomposition (§5.2) provides a probe for whether their apparent robustness or fragility reflects model behavior or scaffold mechanics.

A separate adjacent literature studies prompt-level adversarial attacks: \citet{arxiv190807125} on universal adversarial triggers, \citet{arxiv221109527} on prompt injection, and \citet{arxiv231001405} on universal and transferable adversarial attacks via gradient-based suffix optimization (GCG). These works operate on token-sequence optimization against a single response, not recursive trajectories. Our setting differs in that we use in-distribution adversarial continuations drawn from the loop's own distribution, not optimized adversarial suffixes; and we measure persistence over a 30-step horizon rather than immediate compliance. The two methodologies are complementary: GCG-style attacks find low-token-budget triggers for one-step compliance, while our framework asks whether such compliance survives recursive evolution.

\subsection{Terminology}

We use "attractor-like" operationally rather than as a claim about a smooth deterministic system. The analysis proceeds from visible text state \(X_t\), to an observable text view \(\mathcal{O}(X_t)\), to an embedding \(z_t\), to clusters, basin candidates, and finally regime labels. A "nudge" is the context-update rule that writes text into the next state; a "perturbation" is externally injected text used to test stability. A "switch" denotes final-cluster disagreement under the stated perturbation protocol, whereas "persistent escape" requires a durable post-injection basin change.

\section{Formal framework and hypotheses}

This section defines the recursive-loop object studied in the paper, the perturbation estimands used to measure basin switching, and the operational criteria by which we call a regime attractor-like. Numerical ED50 estimates, pass/fail audits, and regime-specific measurements are results and are reported later. The only substantive structural claim retained here is the replace-vs-append distinction: replace mode admits a generation-budget access bound, whereas append mode motivates an injected-token accumulation conjecture.

\subsection{State, generator, nudge}

Every recursive loop studied here has two distinct moving parts. The \textbf{generator} is the model itself, which samples the next piece of text. The \textbf{nudge} is the memory policy that decides how that text is written back into the next prompt. Prior work often treats these as one object, but they play different roles: the generator determines what can be produced from a state, while the nudge determines what persists as state.

Let \(X_t \in \mathcal{C}\) denote the bounded visible context at step \(t\), where \(\mathcal{C}\) is the space of valid clipped contexts. In this paper, \(\mathcal{C}\) is instantiated as finite-length character strings produced by tail-clipping at 12,000 characters. Let \(f\) denote the content instruction and \(\eta\) the memory policy. The recurrence is

\[Y_t \sim P_\theta(\cdot \mid X_t; f),
\qquad
X_{t+1} = \mathcal{N}_\eta(X_t, Y_t).\]

Here \(P_\theta(\cdot \mid X_t; f)\) is the language-model generator under instruction \(f\), for example continuation, paraphrase, summarize-and-negate, or role-alternating dialog. The map \(\mathcal{N}_\eta : \mathcal{C} \times \mathcal{Y} \to \mathcal{C}\) is the nudge, or context-update operator. The pair \((f,\eta)\), not either component alone, defines the loop dynamics.

The canonical nudges used in the experiments are:

\[\mathcal{N}_{\mathrm{append}}(X_t,Y_t)
=
\operatorname{clip}(X_t \Vert Y_t),\]

\[\mathcal{N}_{\mathrm{replace}}(X_t,Y_t)
=
\operatorname{clip}(Y_t),\]

\[\mathcal{N}_{\mathrm{dialog}}(X_t,Y_t)
=
\operatorname{clip}\!\left(X_t \Vert \operatorname{format\_turn}(r_t,Y_t)\right),\]

where \(\Vert\) denotes string concatenation, \(r_t\) is the role label assigned to the turn (alternating according to the dialog protocol), and \(\operatorname{format\_turn}(r_t,Y_t)\) is the role-labeled rendering of the model output that is appended to the prior context.

\begin{table}[h!]
\centering
\small
\begin{tabularx}{\textwidth}{lYY}
\toprule
Formal nudge & Engineering analogue & Typical risk or behavior \\
\midrule
Append: \(\operatorname{clip}(X_t \Vert Y_t)\) & Full transcript, rolling recent context, accumulated tool logs & Prior evidence remains as ballast; perturbations compete with accumulated state \\
Replace: \(\operatorname{clip}(Y_t)\) & Summarize-and-continue, scratchpad replacement, single-state whiteboard & Old state is discarded; whatever enters the replacement becomes privileged \\
Dialog: \(\operatorname{clip}(X_t \Vert \operatorname{format\_turn}(r_t,Y_t))\) & Role-structured chat state, multi-role agents, user/assistant/tool buffers & Recent role-local turns may dominate despite longer accumulated context \\
Hybrid pinned plus summary & Pinned issue, tests, policy, plus compressed older history & Robustness depends on which facts can be overwritten and which remain invariant \\
\bottomrule
\end{tabularx}
\end{table}

In engineering terms, the nudge is the loop memory policy. It is therefore part of the system's robustness and security boundary, not merely an implementation detail. The hypotheses below use this factorization directly: changing the content instruction \(f\) can change the generator, while changing the nudge \(\eta\) can change which parts of the generated text become future state.

Throughout the paper we refer to five concrete instantiations of the $(f, \eta)$ framework by short code:

\begin{itemize}
  \item \textbf{O1 (contractive append):} continuation operator $f=\text{continue}$ under append nudge.
  \item \textbf{O2 (oscillatory replace):} paraphrase operator under replace nudge.
  \item \textbf{O3 (absorbing replace):} summarize-and-negate operator under replace nudge.
  \item \textbf{D1 (role-structured dialog):} curious-user / helpful-agent dialog under the dialog nudge.
  \item \textbf{D2 (drill-down dialog):} explorer / expert dialog under the dialog nudge, with successive subtopic refinement.
\end{itemize}

The dialog regimes are append-style at the trajectory level but role-structured at the turn level. Wherever "O1", "O2", "O3", "D1", or "D2" appears below, it refers to the corresponding generator-nudge pair listed here.

\subsubsection{Barrier height as a persistent-escape estimand}

Let \(B_1,B_2 \subset \mathcal{C}\) be basin sets in the late-window basin partition defined operationally in the Methods. Let \(t_{\mathrm{inj}}\) be the injection step and let \(X_{t_{\mathrm{inj}}^-}\) denote the state immediately before injection. For an injected dose of \(\tau\) tokens, define the persistent-escape barrier from \(B_1\) to \(B_2\) as

\[\mathrm{B}_{\mathrm{persist}}(B_1 \to B_2)
=
\inf\Bigl\{\tau:
\Pr\bigl[J_\tau(B_1\to B_2)=1,\ X_T\in B_2
\mid X_{t_{\mathrm{inj}}^-}\in B_1,\ \mathrm{inject}_\tau\bigr]
\ge \tfrac12
\Bigr\}.\]

Here \(J_\tau(B_1\to B_2)\) is the operational injection-step jump indicator defined by the late-window basin partition. Without \(J_\tau\), the same formula defines a terminal raw-switching barrier, not a persistent-escape barrier.

The unit of \(\mathrm{B}_{\mathrm{persist}}\) is injected tokens. This unit is intentionally operational: it asks how much text must be inserted into the loop, under a declared injection protocol, for an injection-step basin jump that persists to the terminal measurement to occur with probability at least one half. Because the nudge determines how injected text is retained, overwritten, or role-labeled, this token-valued barrier is expected to vary with \(\eta\).

The definition separates two events that are often conflated. A trajectory may terminate in \(B_2\) without an injection-step jump, because it would have drifted there under sampling noise. Conversely, a trajectory may jump at injection and later return to its reference basin. Persistent escape requires both the injection-step jump and the terminal outcome.

\subsubsection{Operational endpoints for switching and dose response}

A perturbed trajectory can finish in a different cluster than its reference trajectory for several reasons: the injection genuinely redirected it, stochastic sampling would have produced a different terminal cluster anyway, or the injection caused a transient jump that later recovered. The following endpoints separate these cases.

Let \(C_{\mathrm{ref}}\) denote the declared reference basin for a perturbed trajectory: in this paper, the pre-injection basin of that same trajectory. Let \(C_T^{(D)}\) be the terminal basin after injected dose \(D\). Let \(p_0=\Pr(C_T^{(0,1)}\neq C_T^{(0,2)})\) be the control-vs-control natural switching floor under the same sampling protocol.

Define

\[S_{\mathrm{raw}}(D)
=
\Pr(C_T^{(D)}\neq C_{\mathrm{ref}}),\]

\[S_{\mathrm{net}}(D)
=
S_{\mathrm{raw}}(D)-p_0,\]

The persistent-escape family splits into two endpoints, defined in §5.1.3 / Fig 5:

\[S_{\mathrm{persist}}^{\mathrm{src}}(D)
=
\Pr(J_D=1,\ C_T^{(D)}\neq C_{\mathrm{ref}}),\]

\[S_{\mathrm{persist}}^{\mathrm{dst}}(D)
=
\Pr(J_D=1,\ C_T^{(D)}=C_{t_{\mathrm{inj}}+1}^{(D)}),\]

where \(C_{t_{\mathrm{inj}}+1}^{(D)}\) is the immediate post-injection cluster. \(S_{\mathrm{persist}}^{\mathrm{src}}\) (retained source-basin escape) asks whether the trajectory was kicked AND remains outside its starting basin at terminal step. \(S_{\mathrm{persist}}^{\mathrm{dst}}\) (destination-coherent persistence) asks the stricter question: was the trajectory kicked AND still in the \textit{specific} post-injection cluster at terminal step. The destination-coherent event is a subset of the source-basin escape event.

The corresponding ED50 endpoints are

\[\mathrm{ED50}_{\mathrm{raw}}
=
\inf\{D:S_{\mathrm{raw}}(D)\ge 0.5\},
\quad
\mathrm{ED50}_{\mathrm{net}}
=
\inf\{D:S_{\mathrm{net}}(D)\ge 0.5\},\]

\[\mathrm{ED50}_{\mathrm{persist}}^{\mathrm{src}}
=
\inf\{D:S_{\mathrm{persist}}^{\mathrm{src}}(D)\ge 0.5\},
\quad
\mathrm{ED50}_{\mathrm{persist}}^{\mathrm{dst}}
=
\inf\{D:S_{\mathrm{persist}}^{\mathrm{dst}}(D)\ge 0.5\}.\]

If the relevant set is empty over the tested dose range, the corresponding endpoint is not reached in that range.

These endpoints are different estimands. \(S_{\mathrm{raw}}\) measures terminal disagreement from the declared reference basin. It includes natural stochastic divergence. \(S_{\mathrm{net}}\) subtracts the control-vs-control floor \(p_0\), so it estimates excess terminal switching above natural drift, but it is not itself a probability. \(S_{\mathrm{persist}}^{\mathrm{src}}\) measures durable escape from the original basin (regardless of destination); \(S_{\mathrm{persist}}^{\mathrm{dst}}\) measures durable commitment to the specific post-injection destination.

This distinction is central to interpreting perturbation experiments. A raw-switching ED50 can be small if the natural switching floor is large. A persistent-escape ED50 can be unreached even when raw switching is substantial, if most injection-step jumps recover before terminal measurement or if terminal disagreement is dominated by sampling drift. The Results report these endpoints separately.

\subsubsection{Operational criteria for attractor-like regimes}

The term \textbf{attractor-like} is used operationally. It does not assert the existence of a smooth deterministic attractor. A regime earns the label only by passing prespecified diagnostics on a measured trajectory ensemble.

> \textbf{Definition. Operational attractor-like regime.} Fix a trajectory ensemble, an observable map, an embedding map, a late-window basin partition, and a declared null ensemble. A regime \(r\) is evaluated by four binary criteria:
> C1 basin predictability; C2 recurrence or dwell above null; C3 embedder-robust recurrence class; C4 re-entry, contraction, or absorbing collapse.
> A strong attractor passes 4/4. An attractor-like regime passes at least 3/4. A regime passing fewer than 3/4 is not an operational attractor under this definition. Missing measurements count as failure unless the criterion is structurally inapplicable and this inapplicability is declared before evaluation.

The thresholds are:

\textbf{C1. Basin predictability.} Cross-validated accuracy for predicting the late-window basin from the final pre-terminal state must satisfy

\[A_r^{\mathrm{final}} \ge \tau_{\mathrm{acc}},
\qquad
\tau_{\mathrm{acc}}=0.70.\]

The basin partition and predictor protocol are fixed in the Methods. This criterion tests whether late-window basin identity is legible from the measured state rather than being arbitrary cluster noise.

\textbf{C2. Recurrence or dwell above null.} Let \(R_r\) be recurrence and \(D_r\) be dwell for regime \(r\). Let the declared null ensemble provide \((\mu_R^{\mathrm{null}},\sigma_R^{\mathrm{null}})\) and \((\mu_D^{\mathrm{null}},\sigma_D^{\mathrm{null}})\). The regime passes if

\[\max\left\{
\frac{R_r-\mu_R^{\mathrm{null}}}{\sigma_R^{\mathrm{null}}},
\frac{D_r-\mu_D^{\mathrm{null}}}{\sigma_D^{\mathrm{null}}}
\right\}
\ge 2
\quad
\text{and}
\quad
\text{Cohen's } d\ge 0.5.\]

When both time-shuffled and no-feedback nulls are available, the stronger null gate is used. For dialog runs, the no-feedback null is structurally unavailable; the time-shuffled null is required.

\textbf{C3. Embedder-robust recurrence class.} Recurrence is assigned to a coarse bin

\[b_e(r)\in
\{\text{high}\ge 0.70,\ \text{low}\le 0.40,\ \text{mid otherwise}\}\]

for each embedding map \(e\). The criterion passes if the recurrence bin agrees in at least two of three embedders: the canonical embedder plus two alternatives. This criterion tests survival of the regime class under measurement change. It does not require scalar invariance of all diagnostics.

\textbf{C4. Re-entry, contraction, or absorbing collapse.} Let \(\lambda_{1,r}^{\mathrm{late}}\) be the top finite-time ensemble-spread exponent for regime \(r\) computed in the late window (defined in §4.5.2), let \(R_r\) and \(SD_r\) denote regime-mean recurrence and sharpness dimension, and let \(\texttt{best\_period}\) and \(\texttt{period\_2\_score}\) denote the periodicity diagnostics from §4.5.1. The regime passes if at least one of the following holds:

\[\lambda_{1,r}^{\mathrm{late}}\le 0.015\]

for late-window contraction;

\[\texttt{best\_period}=2
\quad\text{and}\quad
\texttt{period\_2\_score}>0\]

for period-two re-entry;

\[R_r\ge 0.90
\quad\text{and}\quad
SD_r\le 1.50\]

for absorbing collapse; or exit-return exceeds the C2 null gate. These alternatives allow different geometric signatures to qualify while still requiring a measured return, contraction, or collapse property.

The criteria above are applied to each experimental regime in the Results. The framework itself does not assume that any regime passes; the pass/fail audit is an empirical outcome.

\subsubsection{Replace-mode access bound}

Replace mode discards the previous context after each generation step. In this regime, the next state depends on the newly generated replacement rather than on an accumulated transcript. This makes the injected-token dose \(\tau\) distinct from the post-injection model-generation budget. It also makes the default ("state-reset overwrite") perturbation protocol tautological at the access step: overriding \(Y_{t_{\mathrm{inj}}}\) with the injection produces \(X_{t_{\mathrm{inj}}+1}=\operatorname{clip}(\text{perturbation\_text})\) directly, so the proposition below should be read as a \textit{post-injection generation-budget bound from the perturbation as a fresh initial condition}, not as an injected-token barrier in the model-resistance sense.

For \(m\) post-injection replace steps, define

\[G_m
=
\sum_{s=t_{\mathrm{inj}}}^{t_{\mathrm{inj}}+m-1} |Y_s|,
\qquad
\bar G_m=\mathbb{E}G_m.\]

The quantity \(G_m\) is not injected text. It is the number of tokens generated by the model after the injection.

\textbf{Proposition. Replace-mode access bound.} In replace mode, suppose every reachable non-target state has one-step probability at least \(q_0\) of generating a target-basin replacement, target entry persists to terminal measurement with probability at least \(r_0\), and expected generation length is at most \(\kappa\) per step. Then after \(m\) post-injection replace steps,

\[\Pr(X_T\in B_2)\ge r_0[1-(1-q_0)^m],
\qquad
\mathbb{E}G_m\le \kappa m .\]

This bounds post-injection generated tokens, not injected-token dose. Full lemma, corollaries, and proof in §12.3.

The proposition captures the structural distinction between replace and append memory. In replace mode, once the loop reaches a state from which target-basin replacement is accessible, repeated replacement attempts accumulate probability through generation steps. The relevant finite quantity in the proposition is \(\mathbb{E}G_m\), the expected post-injection generation budget. It is not \(\tau\), the length of the injected string.

Consequently, the proposition should not be read as an injected-token barrier bound. If the access assumptions hold only after a particular injected string of length \(\tau\), then the injection protocol may certify an injected-dose upper bound of at most that \(\tau\) under those conditions. The factor \(\kappa m\) certifies only the generated-token budget after injection. If comparable access holds without injected text, then replace-mode switching is better understood as endogenous reachability of the replacement dynamics rather than as evidence of a positive injected-token barrier.

\subsubsection{Append-mode accumulation conjecture}

Append mode retains prior context:

\[X_{t+1}
=
\operatorname{clip}(X_t \Vert Y_t),
\qquad
Y_t\sim P_\theta(\cdot\mid X_t;f).\]

A perturbation injected into an append-mode loop therefore does not overwrite the current state in one step. Instead, it is incorporated into an already formed bounded context and must compete with incumbent basin evidence that remains visible until displaced by clipping. The injected text, subsequent model outputs, and pre-existing context all coexist inside the same clipped state.

This motivates an injected-token accumulation conjecture rather than a generation-budget theorem. Let \(B_1,B_2\subset\mathcal{C}\) be basin sets for an append-mode loop. There is no reason to expect a one-step overwrite bound analogous to replace mode. Instead, switching should depend on how much basin-relevant counter-context survives inside the effective context window.

\textbf{Conjecture. Append-mode accumulation barrier.} For append-mode loops, basin switching from \(B_1\) to \(B_2\) is governed by the accumulated share of semantically legible counter-context inside the clipped state. In particular, there exists a basin-dependent threshold such that subthreshold injections remain near the stochastic floor, while injections that supply sufficient basin-relevant counter-context produce a rising switching curve.

A testable response-curve formulation is as follows. Let \(a_\tau\) be the semantically weighted share of the clipped context occupied by basin-relevant injected text. Append-mode accumulation predicts a nondecreasing switching curve

\[S_{B_1\to B_2}(\tau)
=
\Pr(X_T\in B_2\mid X_{t_{\mathrm{inj}}}\in B_1,\tau)
\approx
p_{\mathrm{floor}}+
(p_{\max}-p_{\mathrm{floor}})
F(a_\tau-\alpha^\star_{B_1\to B_2}),\]

where \(F\) is increasing and the threshold \(\alpha^\star\) is lower for in-distribution counter-context than for out-of-distribution text.

Here \(p_{\mathrm{floor}}\) is the relevant stochastic floor, \(p_{\max}\) is the attainable upper switching level under the protocol, and \(\alpha^\star_{B_1\to B_2}\) is a basin-pair threshold in effective context share. The word "semantically" is important: two injections with the same token count can have different \(a_\tau\) if one is in-distribution counter-context for \(B_2\) and the other is irrelevant or out-of-distribution text. This makes the conjecture falsifiable. It predicts dose dependence for append-mode in-distribution perturbations, weaker or absent dose response for semantically irrelevant perturbations at the same token count, and a threshold that shifts when the perturbation family changes.

A geometric refinement is possible. Let \(V(x)=-\log\rho(x)\) be an empirical potential over the representation space, and let \(V^\star(B_1,B_2)\) denote the saddle height along a minimum-cost path between basins. One may hypothesize that injected-token barriers in append mode scale with this saddle height up to representation and perturbation-family factors. This geometric relation is not used as a theorem in the framework. It is a secondary hypothesis to be evaluated empirically.

All barrier quantities in this section are operational token counts unless explicitly stated otherwise. Token-valued barriers are tokenizer-dependent and model-specific. A model-normalized alternative would measure injected information in nats using per-token conditional log probabilities under \(P_\theta\). That requires log-probability capture during generation. The present framework therefore treats nat-valued barriers as a future measurement target rather than as reported quantities.

\subsection{Observable maps and embedding}

Attractor-like structure is not measured directly from raw recurrence definitions alone. We introduce observable maps that select text views of the trajectory, and embedding maps that lift those views into vector space. Let

\[O_t=h(X_t,Y_t,\mathrm{metadata})\]

be an observable string selected from the state, generation, and recorded metadata. Let

\[z_t=\phi(O_t)\in\mathbb{R}^m\]

be its vector representation. For the canonical embedding used in the main battery, \(m=1536\) for \texttt{text-embedding-3-small}. A trajectory therefore produces one or more embedded polylines \(\{z_t^{(O)}\}_{t=0}^{T-1}\), depending on the observable family.

Observable maps are post-hoc measurement functions; they do not feed back into the recurrence unless explicitly used as part of a nudge. Thus \(X_t\) is the dynamical state, \(Y_t\) is the sampled continuation, \(O_t=h(X_t,Y_t,\mathrm{metadata})\) is the text view selected for measurement, and \(z_t=\phi(O_t)\) is the vector representation used for clustering, recurrence, dwell, and basin-predictability metrics. The Methods section fixes the observable family, embedding normalization, clustering procedures, and null ensembles before applying the C1-C4 criteria.

This separation matters for interpretation. The loop dynamics are defined by \((P_\theta,f,\mathcal{N}_\eta)\). The observables and embeddings define the measurement apparatus. A claimed attractor-like regime must therefore survive not only within one projection, but also under the robustness checks in C3. Conversely, failure under one observable does not by itself imply absence of structure in the text dynamics; it implies that the declared measurement battery did not certify the operational definition.

\subsection{Hypotheses}

\textbf{H1.} At least one recursive-loop regime passes the operational attractor-like definition under the declared observable, embedding, basin partition, and null ensemble.

\textbf{H2.} The pass/fail pattern and attractor subtype depend on both content operator \(f\) and nudge \(\eta\); changing \(\eta\) while holding content approximately fixed changes the measured regime class.

\textbf{H3.} Perturbation sensitivity differs by nudge: append mode exhibits a dose-dependent in-distribution raw-switching curve above the stochastic floor; replace mode reaches high switching at the smallest tested dose or without dose; dialog lies between these extremes and depends on role/state structure.

\textbf{H4.} The qualitative regime labels and endpoint ordering observed in small-N exploration remain unchanged under the large-N battery within declared uncertainty.

\section{Methods}

Before the per-component details below, the experimental skeleton is short: we run paired recursive trajectories from the same seed and prompt; in the treatment run we inject text at a fixed step; we embed every step's observable output; we project to a low-dimensional space; and we cluster final states jointly across treatment and control runs. The perturbation is summarized by the perturbed run's final cluster relative to its paired control's. We separately estimate the control-vs-control divergence rate, namely how often two paired control runs already disagree from sampling alone, and the persistent-escape rate, namely how often a perturbed run jumps clusters at injection and remains in the new cluster at the terminal step. The remainder of §4 details each component.

\textbf{Scope of the experimental object: bounded-memory loops.} The recursive loop studied here has an explicit memory policy: each step's running context is tail-clipped at \texttt{max\_context\_chars = 12000}, well below the generator's native ~128K-token context window. The clip is part of the experimental object, not an implementation accident. Real recursive systems use rolling windows, summary buffers, retrieval-augmented memory, and other bounded-memory policies, and the framework's whole point (§3.1's state-generator-nudge separation) is to study how such policies shape loop dynamics. Headline append-mode claims (ED50\_raw, persistence at the terminal step, the 4PL saturation extrapolation in §5.1.2) describe the canonical bounded-memory loop and should be read as memory-policy-conditioned, not as claims about loops with unbounded memory. §5.1.3 reports a buffer-size sensitivity check that quantifies how much of the saturation pattern depends on this specific memory policy.

\subsection{Primary endpoints and inferential contract}

Before listing implementation details, we pre-specify five primary endpoint families used for claims: operational attractor score, group-aware basin predictability, perturbation switching signature, behavioral ED50, and persistent escape (with two complementary definitions in the third family, see §3.1.2 and §5.1.3). All other quantities in §4.5 are diagnostic or visualization-support metrics unless explicitly mapped to one of these endpoints.

This contract separates exploratory dynamics measurements from claim-bearing endpoints. Recurrence, dwell, basin occupancy, periodicity, dispersion, finite-time ensemble-spread spectra, sharpness dimension, flow fields, and density landscapes are used to characterize a regime and to support interpretation. They are not, by themselves, sufficient for a headline claim unless they enter one of the five endpoint families defined operationally in §4.13.

The five primary endpoint families are:

\begin{enumerate}
  \item \textbf{Operational attractor score}, based on the C1-C4 attractor criteria: late-window basin persistence, recurrence or dwell above null, embedder robustness, and contraction, re-entry, or collapse.
  \item \textbf{Group-aware basin predictability}, measured by predicting the late basin from early 10-component PCA projection (PCA-10) state while holding out prompt families.
  \item \textbf{Perturbation switching signature}, measured by paired treatment-control final-cluster disagreement, interpreted alongside the paired control-control stochastic floor.
  \item \textbf{Behavioral ED50}, the token dose at which the O1 adversarial perturbation reaches 50\% raw switching, with uncertainty estimated by family-cluster bootstrap and model-based fits.
  \item \textbf{Persistent escape} (two complementary definitions): \textit{destination-coherent persistence}, where the perturbation induces an injection-time cluster jump and the trajectory remains in that \textit{specific} post-injection cluster at terminal step; and \textit{retained source-basin escape}, where the trajectory is kicked AND outside its pre-injection cluster at terminal step (regardless of destination). Destination-coherent is a strict subset of source-basin escape; the two answer different safety-relevant questions (§5.1.3).
\end{enumerate}

All thresholds, pass rules, and current endpoint status are consolidated in §4.13.

\subsection{The recurrence}

We instantiate the formal recurrence from §3.1 with three context-update rules:

\[\text{Append:}\quad X_{t+1}=\mathcal{N}_{\text{append}}(X_t,Y_t)=\operatorname{clip}(X_t\Vert Y_t)\]

\[\text{Replace:}\quad X_{t+1}=\mathcal{N}_{\text{replace}}(X_t,Y_t)=\operatorname{clip}(Y_t)\]

\[\text{Dialog:}\quad X_{t+1}=\mathcal{N}_{\text{dialog}}(X_t,Y_t)=\operatorname{clip}\!\left(X_t\Vert \operatorname{format\_turn}(r_t,Y_t)\right)\]

where

\[Y_t \sim P_\theta(\cdot \mid X_t; f)\]

and \(P_\theta\) is the language-model distribution parameterized by \(\theta\), here \texttt{gpt-4o-mini}. The clipping operator \(\operatorname{clip}(\cdot)\) truncates context from the head, namely the oldest text, once the running string exceeds 12,000 characters, preserving the most recent state. The content operator \(f\) enters through the system prompt fed to \(P_\theta\), for example "Continue the text" for \(f=\text{continue}\) and "Paraphrase the following" for \(f=\text{paraphrase}\).

Append mode creates a growing memory trace until the context cap is reached. Replace mode overwrites the state with the latest output, making the loop maximally sensitive to the current sample. Dialog mode alternates formatted turns between two roles and retains the conversation history subject to the same context-management principles. These three update rules define the operator families used throughout the experiments.

\subsection{Sampling}

Each experiment runs

\[N_{\text{traj}}=N_{\text{families}}\times N_{\text{ICs}}\times N_{\text{runs}}\]

trajectories, where \(N_{\text{families}}\) is the number of distinct prompt families, \(N_{\text{ICs}}\) is the number of initial conditions per family, and \(N_{\text{runs}}\) is the number of independent runs per (family, IC) cell. Publication-scale defaults differ by experiment family:

\begin{table}[h!]
\centering
\small
\begin{tabularx}{\textwidth}{lYYY}
\toprule
experiment family & design & steps & point count \\
\midrule
Operator runs O1, O2, O3 & 15 prompt families x 30 initial conditions x 3 runs = 1,350 trajectories per regime & 40 & 54,000 points per experiment \\
Dialog run D1 & 5 dialog-suitable families x 30 initial conditions x 3 runs = 450 trajectories & 40 & 18,000 points per role, 36,000 effective role-specific points \\
D2 drill-down dialog & 5 families x 5 initial conditions x 1 run = 25 trajectories & 50 & exploratory scale only \\
\bottomrule
\end{tabularx}
\end{table}

D2 is below the \(N\geq 2\)-runs minimum required for ensemble-spread diagnostics and is therefore treated as exploratory rather than publication-scale evidence.

Initial conditions are short seed texts grouped into families. Families include philosophical prompts, practical-advice prompts, creative-writing prompts, reflective prompts, and emotional prompts. Across families we obtain variation in topic and tone; within families we obtain variation across seeds. Unless a temperature sweep is explicitly being run, sampling uses temperature \(T=0.8\). The Phase 2b temperature sweep varies temperature and, in some cells, \texttt{steps\_per\_run}.

\subsection{Embedding}

All trajectories are embedded with \texttt{text-embedding-3-small}, producing 1536-dimensional vectors. We embed multiple observables per step. Each observable is a different text view of the same recursive state, and all analyses are repeated per observable to expose representation-dependent findings.

\begin{table}[h!]
\centering
\small
\begin{tabularx}{\textwidth}{lYY}
\toprule
observable & source location & what it captures \\
\midrule
\texttt{output} & \texttt{core/observables.py} & the model's \(Y_t\) text alone, with no context \\
\texttt{rolling\_k3} & \texttt{core/observables.py} & concatenation of the last 3 outputs \\
\texttt{context\_tail} & \texttt{core/observables.py} & the last 4000 characters of the running context \\
\texttt{context\_full} & \texttt{core/observables.py} & fixed-window 8000-character tail, used in longer-memory pilot configurations \\
\texttt{last\_user\_turn} & \texttt{experiments/dialog/observables.py} & dialog-only: most recent user or role-A utterance \\
\texttt{last\_agent\_turn} & \texttt{experiments/dialog/observables.py} & dialog-only: most recent agent or role-B utterance \\
\texttt{rolling\_user\_k3} & \texttt{experiments/dialog/observables.py} & dialog-only: rolling window of last 3 user turns \\
\texttt{rolling\_agent\_k3} & \texttt{experiments/dialog/observables.py} & dialog-only: rolling window of last 3 agent turns \\
\texttt{turn\_pair} & \texttt{experiments/dialog/observables.py} & dialog-only: most recent user-agent exchange concatenated \\
\bottomrule
\end{tabularx}
\end{table}

The role-specific observables are essential for dialog analysis because basin scores, recurrence, and predictability can differ strongly when computed on the user's questions, the agent's answers, or the concatenated exchange. D1 uses user and agent labels. D2 uses explorer and expert labels. Role names are read from \texttt{cfg.dialog.role\_a.name} and \texttt{cfg.dialog.role\_b.name} at embed time, so the observable wiring accepts any configured role-name pair.

For one observable string at one trajectory step, we obtain exactly one 1536-dimensional vector. There is no user-managed chunking, no per-token output, and no sliding window internal to the analysis pipeline. After the embedding API returns, each row is L2-normalized so downstream cosine similarities reduce to dot products. Thus one operator publication trajectory yields 3 vectors per step for \texttt{output}, \texttt{rolling\_k3}, and \texttt{context\_tail}, or 4 if \texttt{context\_full} is enabled. One dialog publication trajectory yields 8 vectors per step, or 9 if \texttt{context\_full} is enabled.

The token budget is held below the 8,191-token embedding limit by construction. Per-step generations are bounded by \texttt{max\_output\_tokens} of 120 for operator runs and 160 for dialog runs. The running context is capped at 12,000 characters, approximately 3,000 English tokens, and every context-based observable slices only the recent tail before embedding. Typical upper bounds are approximately 160 tokens for \texttt{output}, 480 for \texttt{rolling\_k3}, 1,000 for \texttt{context\_tail}, 2,000 for \texttt{context\_full}, 320 for \texttt{turn\_pair}, and 480 for each role-specific rolling window. Exact constructor calls, caching details, adjacent-step similarity checks, cost estimates, and verification snippets are provided in Supplementary §12.4.

\subsection{Perturbation materials and paired-run protocol}

Perturbation experiments use paired recursive runs from the same seed, prompt family, initial condition, and generation settings. The treatment run receives injected text at a pre-specified injection step \(t_{\mathrm{inj}}\), recorded in the experiment configuration and held fixed within a condition unless the experiment is explicitly an injection-step sweep. Terminal steps are 29 for 30-step perturbation pilots and 49 for 50-step perturbation runs.

\subsubsection*{Perturbation corpora}

\begin{table}[h!]
\centering
\small
\begin{tabularx}{\textwidth}{lYYY}
\toprule
corpus or condition & source & intended semantic role & dose handling \\
\midrule
\texttt{control} & no injected text & estimates the unperturbed recursive endpoint & zero dose \\
\texttt{neutral} & off-topic Wikipedia-style paragraph drawn from the hand-written \texttt{corpora.NEUTRAL\_WIKI} pool & coherent but task-irrelevant text & full paragraph in pilot runs; token-resized variants use \texttt{neutral\_<N>} \\
\texttt{lorem} & 70 random English words from the curated neutral \texttt{corpora.\_WORD\_POOL} & incoherent or weakly coherent lexical mass with low affective load & fixed 70-word pilot form or token-resized variants \\
\texttt{adversarial} & late-step output from a different trajectory of the same regime & on-manifold text likely to point toward another basin & token-resized variants for dose-response \\
\bottomrule
\end{tabularx}
\end{table}

The neutral word pool was curated to avoid emotional, introspective, or obviously directive vocabulary. The adversarial condition is not an instruction to attack the model. It is adversarial in the dynamical-systems sense: it is a plausible state fragment taken from another trajectory that may redirect the recurrence.

\subsubsection*{Dose definitions}

A perturbation dose \(\tau\) is measured in tokens. For dose-parametrized variants, the perturbation text is resized to the requested token length before injection. Sparse O1 adversarial dose sweeps use \(\tau \in \{20,80,200,400\}\). Dense reruns use more dose levels and larger per-cell sample sizes. Pilot conditions that do not carry an explicit dose label use their default corpus length, for example a full neutral paragraph or a 70-word lorem sample.

The injection is applied to the recursive state at \(t_{\mathrm{inj}}\) in a condition-specific but pre-specified manner. In append and dialog settings, the injected text is inserted into the context stream as an additional state fragment. In replace settings, the default ("state-reset overwrite") intervention is \textbf{tautological at the state-write step}: by overriding \(Y_{t_{\mathrm{inj}}}\) with the injection, replace-mode update yields \(X_{t_{\mathrm{inj}}+1}=\operatorname{clip}(\text{perturbation\_text})\). Any subsequent observation reflects only the convergence of the operator \(f\) starting from the perturbation as a fresh initial condition; it is \textit{empirical only in the post-injection convergence}. This is why replace-mode default-protocol perturbation results are interpreted as a capitulation signature rather than as evidence of a subtle barrier-crossing mechanism, and why the insert-mode counterfactual in §5.2 is the appropriate measurement of model behavior under replace-mode loops.

\subsubsection*{Adversarial-source rule}

For adversarial perturbations, the source text must come from a different trajectory than the target. The source is drawn from a different family-initial-condition trajectory of the same regime, using late-step output so that the perturbation resembles an endpoint state rather than an early transient. We exclude self-source cases, missing-source cases, and cases where the source text is empty after tokenization or cannot be resized to the requested dose.

\subsubsection*{Paired-control design and exclusion rules}

For each seed-level unit we run two unperturbed controls, denoted \(A\) and \(B\), and one matched treatment, denoted \(Z\). Control \(A\) is the treatment comparator. Control \(B\) estimates the stochastic floor, namely the rate at which two unperturbed recursive samples disagree at the terminal step from ordinary sampling alone.

A trajectory unit is excluded from perturbation endpoint aggregation if any required member of the tuple \((A,B,Z)\) is missing; if the terminal step is missing for any member; if embeddings or cluster labels are unavailable for the selected observable; if the adversarial source violates the source rule; if the perturbation text is empty after tokenization; or if the run does not match the configured injection step and terminal horizon. Exclusions are applied before endpoint computation and are counted in audit tables.

\subsubsection*{Algorithm 1: paired perturbation evaluation}

Input: seed or task \(x\), generator \(P_\theta\), context-update rule \(\mathcal{N}\), injection condition \(c\), injection step \(t_{\mathrm{inj}}\), terminal step \(T\), observable map \(O\), and equivalence rule \(C\). In this paper \(C\) is a K-means cluster in joint PCA-10 space, but the algorithm only requires a pre-specified equivalence rule.

\begin{enumerate}
  \item Run two unperturbed controls:
   \[A=\operatorname{RunLoop}(x,P_\theta,\mathcal{N},\text{no injection})\]
   \[B=\operatorname{RunLoop}(x,P_\theta,\mathcal{N},\text{no injection})\]

  \item Estimate the stochastic-floor event:
   \[\operatorname{floor}=\left[C(O(A_T))\neq C(O(B_T))\right]\]

  \item Run the matched treatment:
   \[Z=\operatorname{RunLoop}(x,P_\theta,\mathcal{N},\text{inject }c\text{ at }t_{\mathrm{inj}})\]

  \item Define raw switching:
   \[\operatorname{raw}=\left[C(O(Z_T))\neq C(O(A_T))\right]\]

  \item Define the injection-time jump:
   \[\operatorname{jump}=\left[C(O(Z_{t_{\mathrm{inj}}+1}))\neq C(O(Z_{t_{\mathrm{inj}}-1}))\right]\]

  \item Define persistent escape (the strict variant: trajectory remains in the \textit{post-injection} cluster, not merely in any cluster other than the reference):
   \[\operatorname{persist}=\operatorname{jump}\wedge
   \left[C(O(Z_T))=C(O(Z_{t_{\mathrm{inj}}+1}))\right]\]

  \item Aggregate over seeds, tasks, and families:
   \[\operatorname{raw\_rate}=\mathbb{E}[\operatorname{raw}],\quad
   \operatorname{floor}=\mathbb{E}[\operatorname{floor}],\quad
   \operatorname{net\_rate}=\operatorname{raw\_rate}-\operatorname{floor}\]
   \[\operatorname{persistent\_escape\_rate}=\mathbb{E}[\operatorname{persist}]\]
\end{enumerate}

This same endpoint decomposition applies to non-embedding observables in engineering settings, including final patch family, files touched, test pass/fail sets, tool-call sequences, plan categories, security-policy violations, or embeddings of full traces.

\subsection{Representation spaces and metric battery}

For each observable's embedding matrix \(Z\in\mathbb{R}^{N\times1536}\), we compute joint projections across the full point cloud. Projections are never fit per-run or per-family, because coordinates must be comparable across trajectories, conditions, and roles.

PCA-2 is used for density estimation, potential landscapes, and two-dimensional plots. For short-output observables it typically carries 10\% to 15\% of total variance; for longer-context observables it can carry approximately 25\%. PCA-10 is used for K-means clustering, basin classification, basin predictability, recurrence, dwell, and most endpoint-level metrics. PCA-50 is used only as a pre-reduction stage before t-SNE.

t-SNE is fit after PCA-50 pre-reduction using cosine distance, perplexity 30 capped at \((N-1)/4\) for small \(N\), PCA initialization, automatic learning rate, and random seed 42. t-SNE is computed once per experiment and observable. It is used only for visualization because local neighborhoods are informative but global distances are not physically interpretable. Quantitative metrics are computed in PCA-10 unless explicitly stated otherwise. Exact constructor calls and verification snippets are provided in Supplementary §12.4.

In perturbation experiments, PCA and clustering are fit jointly across all conditions. In dialog experiments, the relevant observable-specific point cloud is projected jointly before role-specific subsets are inspected. This joint-fit rule is required for comparing basins, geodesics, switching events, and role-conditioned trajectories.

\subsubsection{Regime-structure diagnostics}

\textbf{Definition.} Regime-structure diagnostics summarize the temporal organization of a single trajectory or a set of trajectories in PCA-10 cluster space.

Recurrence for trajectory points \(z_0,\ldots,z_{T-1}\) is

\[\operatorname{recurrence}(\epsilon,\tau)=
\frac{\#\{(t,s):\lVert z_t-z_s\rVert_2<\epsilon\ \wedge\ |t-s|>\tau\}}
{T(T-1)/2}\]

with \(\epsilon=0.15\) and \(\tau=3\). The lag exclusion suppresses trivially nearby adjacent steps.

Dwell is computed after K-means clustering with \(k=12\) in PCA-10. It is the run-length distribution within a cluster. Long dwell times indicate basin persistence; short dwell times indicate transience or rapid cycling.

A target region is defined as the K-means cluster containing the trajectory's final 30\% of points. Basin score is the fraction of post-\(t^*\) points in that cluster, with \(t^*=0.7T\). Basin entry is the first step at which the trajectory's cluster matches its late-window target.

Late recurrence restricts the recurrence statistic to the second half of the trajectory. Exit-return asks whether a trajectory that has visited its target basin subsequently leaves and re-enters it. This separates tight basins from metastable basins.

Periodicity is measured by lag-distance autocorrelation. For lag \(k\),

\[\operatorname{mean\_dist}(k)=
\mathbb{E}_t\left[\lVert z_t-z_{t+k}\rVert_{\cos}\right].\]

The period-2 score is \(\operatorname{mean\_dist}(1)-\operatorname{mean\_dist}(2)\). Positive values indicate that lag-2 points are closer than lag-1 points. Analogous scores are computed for period 3, and the best period is the lag \(k\in[1,T/2]\) minimizing the mean distance.

Dispersion compares ensemble spread early and late in the trajectory. Initial dispersion is the mean pairwise distance over \(t\in[0,T/4]\). Final dispersion is the mean pairwise distance over \(t\in[3T/4,T]\). Dispersion growth is \((\operatorname{final}-\operatorname{initial})/\operatorname{initial}\). Global drift is the distance between the centroid at the terminal step and the centroid at the initial step. Drift monotonicity is the correlation between centroid distance and step.

\textbf{Use in claims.} These diagnostics feed the operational attractor score and the three-axis classifier. Recurrence, dwell, basin score, basin entry, late recurrence, and exit-return support the convergence and persistence components of the attractor score. Periodicity is the primary diagnostic for O2-style oscillatory regimes. Dispersion and drift support the distinction between contractive, oscillatory, absorbing, and divergent behavior.

\textbf{Rationale or limitation.} These metrics test whether the trajectory has temporal structure beyond its marginal point cloud. A high recurrence or dwell statistic is not sufficient on its own because the same point cloud can appear recurrent after time shuffling. Conversely, a low recurrence statistic on \texttt{output} can coexist with strong basin persistence on \texttt{context\_tail} if the individual generations fluctuate while the integrated context is stable. For this reason, regime claims require baseline comparison and cross-observable agreement.

\subsubsection{Ensemble-spread diagnostics}

\textbf{Definition.} We compute a finite-time ensemble-spread spectrum from multiple runs sharing the same family and initial condition. We use the term finite-time Lyapunov-like exponent for an ensemble-spread growth rate, not for a Jacobian-derived exponent of a smooth map.

For each family-initial-condition cell with \(N\) runs, the embeddings at step \(t\) define a covariance matrix

\[\Sigma_t=\frac{1}{N-1}\sum_i (z_i^t-\bar z^t)(z_i^t-\bar z^t)^\top .\]

Let \(\mu_k(t)\) be the \(k\)-th eigenvalue of \(\Sigma_t\). The \(k\)-th finite-time ensemble-spread exponent over a window from \(t_{\mathrm{baseline}}\) to \(T-1\) is

\[\lambda_k=
\frac{1}{2(T-1-t_{\mathrm{baseline}})}
\log\left(\frac{\mu_k(T-1)}{\mu_k(t_{\mathrm{baseline}})}\right).\]

The factor \(1/2\) converts variance growth to amplitude growth. We compute early and late windows. The early window captures transient divergence as runs move away from the seed. The late window captures on-attractor or post-transient behavior.

We also compute sharpness dimension and effective rank as secondary diagnostics. Sharpness dimension uses the ordered spectrum and the Tuci-style fractional form

\[j^*=\max\left\{i:\sum_{k\leq i}\lambda_k\geq0\right\}\]

with \(j^*=0\) if \(\lambda_1<0\), and

\[\operatorname{SD}=j^*+
\frac{\sum_{k\leq j^*}\lambda_k}{|\lambda_{j^*+1}|}.\]

If the spectrum is everywhere negative, SD is 0. If the cumulative sum remains non-negative through the full spectrum, SD is \(d\). Effective rank counts exponents above \(-0.01\).

\textbf{Use in claims.} The late ensemble-spread spectrum contributes to the contraction component of the operational attractor score. A negative or shrinking late spread supports a contractive or absorbing interpretation. A positive early exponent combined with late contraction indicates transient divergence followed by settling. Sharpness dimension and effective rank are reported as secondary diagnostics, not as primary endpoints.

\textbf{Rationale or limitation.} Text generation is discrete and sampling-based, so there is no smooth one-step Jacobian from which to derive a classical Lyapunov spectrum. The spectrum here measures growth or contraction of stochastic ensemble spread under matched prompts and seeds. With \(N=3\) runs per initial condition, the covariance rank is at most 2, so the spectrum length is at most 2 and sharpness dimension can saturate at 2.0. These diagnostics are comparative across regimes, not absolute estimates of a dynamical invariant.

\subsubsection{Predictability endpoint}

\textbf{Definition.} Basin predictability asks whether the late-window basin can be predicted from early state. For each trajectory, the target label \(y\) is the K-means cluster occupied in the late window. For each early step \(k\), we train a multinomial logistic regression from the PCA-10 state at step \(k\) to \(y\).

Two cross-validation schemes are reported. The first is stratified \(k\)-fold cross-validation after dropping singleton classes that cannot be split into train and test folds. The number of folds is adaptive: \(n_{\mathrm{splits}}=\min(5,\text{smallest class size})\). If fewer than two non-singleton classes remain, the cell is recorded as missing. The second is group-aware cross-validation with prompt family held out as the grouping variable. This GroupKFold analysis tests whether predictability generalizes across prompt families rather than exploiting family identity.

Audit columns record the number of dropped singleton classes and trajectories. Publication-scale runs usually reach 5 stratified folds. Reduced-scope sweeps and pilots can fall back to 2 to 4 folds.

\textbf{Use in claims.} Group-aware basin predictability is one of the five primary endpoints. To claim cross-family basin predictability, accuracy at step \(k=10\) must satisfy

\[\operatorname{acc}_{\mathrm{group}}(k=10)\geq0.70.\]

To claim that the original stratified estimate is leakage-free, the difference

\[\Delta=\operatorname{acc}_{\mathrm{stratified}}-\operatorname{acc}_{\mathrm{group}}\]

must be below 0.10. Stratified accuracy is reported as a diagnostic, but it is not sufficient for the leakage-free endpoint.

\textbf{Rationale or limitation.} A high stratified accuracy can arise because prompt families occupy different regions of embedding space. Group-aware splitting is therefore the stricter test: the model must predict the late basin for held-out families. This endpoint is intentionally conservative. It can fail even when within-family predictability is high, because the claim is cross-family generalization.

\subsubsection{Perturbation-response endpoints}

\textbf{Definition.} Perturbation-response endpoints are computed from the paired-control design in §4.4. Raw switching is the fraction of treatment runs whose terminal cluster differs from paired control \(A\). The stochastic floor is the fraction of paired controls \(A\) and \(B\) whose terminal clusters differ without perturbation. Net switching is raw switching minus the stochastic floor. Persistent escape is the stricter event in which the treatment changes cluster immediately after injection and remains in that post-injection cluster at the terminal step.

The behavioral ED50 is estimated from dose-response data. It is the token dose \(\tau\) at which the fitted O1 adversarial dose-response reaches 50\% raw switching. Fits include a four-parameter logistic curve, a generalized linear mixed model, and family-cluster bootstrap summaries. The endpoint is reported with uncertainty intervals and with a separate assessment of whether net switching or persistent escape reaches the same threshold.

\textbf{Use in claims.} The perturbation switching signature is a primary endpoint. For O1 selective sensitivity, adversarial switching at 200 tokens must be at least 0.50, must be at least twice the larger of neutral or lorem switching at the same dose, and the maximum out-of-distribution neutral or lorem switching must be at most 0.30. For O2 and O3 replace-mode perturbations, capitulation is supported if all non-control perturbation conditions produce switching of at least 0.85, with the interpretive caveat that replace mode overwrites state.

Behavioral ED50 is a primary endpoint only for raw O1 adversarial switching. A localized token barrier requires a finite point estimate and a 95\% family-cluster bootstrap interval contained within the probed dose interval. Net ED50 and persistent ED50 are reported separately. If the net effect or persistent escape does not reach 50\% in the tested range, those stricter ED50 values are undefined.

\textbf{Rationale or limitation.} Raw switching alone is deliberately sensitive to ordinary stochastic divergence. It is meaningful only when read alongside the paired control-control floor and persistent-escape rate. Persistent escape is stricter but can undercount cases where a perturbation changes the long-term basin through intermediate clusters. ED50 is a behavioral dose summary, not a physical energy barrier. The term barrier is therefore operational: it refers to a token-valued perturbation scale under the specified equivalence rule and sampling design.

\subsubsection{Three-axis classifier summary}

\textbf{Definition.} The three-axis classifier maps diagnostic metrics to three hypotheses. H1a, convergence to a basin, uses basin-positive and dwell-above-null signals. H1b, recurrence or oscillation, uses late recurrence above null, period-2 score above threshold, and best-period majority greater than 1. H1c, divergence or no attractor, uses dispersion growth, monotonically outward drift, and absence of a stable basin. Signal counts are converted to qualitative strengths: not supported, weak, moderate, or strong.

\textbf{Use in claims.} The classifier provides a structured summary that justifies regime labels. O1 contractive behavior is expected to show strong H1a and weak H1b. O2 oscillation is expected to show strong H1b driven by period-2 structure. O3 absorbing behavior is expected to show strong H1a plus reduced spread. H1c supports a divergent or unsupported label. The classifier is not a replacement for the five primary endpoints in §4.0 and §4.13.

\textbf{Rationale or limitation.} The classifier reduces many correlated diagnostics to a small set of pre-registered signal counts. Its strength is transparency: each verdict can be traced to thresholded signals. Its limitation is that it remains a summary of diagnostics rather than a direct endpoint. A regime can receive a plausible classifier label while failing a stricter decision-grade endpoint such as group-aware predictability or persistent escape.

\subsubsection*{Metric, baseline, and pass-rule mapping}

\begin{table}[h!]
\centering
\small
\begin{tabularx}{\textwidth}{lYYY}
\toprule
Metric family & Null/baseline & Claim use & Pass rule \\
\midrule
Recurrence/dwell & time-shuffled, no-feedback, independent regeneration & attractor score & at least baseline plus 2 standard deviations and Cohen's \(d\geq0.5\) \\
Basin score/entry and exit-return & time-shuffled and recursive baselines by observable & attractor score and regime label support & consistent late-window persistence across canonical observables \\
Periodicity & time-shuffled trajectory order and non-oscillatory regimes & O2 recurrence or oscillation support & positive period-2 score with best-period evidence above threshold \\
Dispersion and finite-time ensemble spread & matched-run ensemble spread and no-feedback baselines where applicable & contraction component of attractor score & late spread contraction or collapse relative to baseline \\
Switching & paired control-control floor & perturbation signature & raw, net, and persistent reported separately \\
Predictability & GroupKFold by family & basin predictability & \(\operatorname{acc}_{\mathrm{group}}(k=10)\geq0.70\) \\
Behavioral ED50 & family-cluster bootstrap and model-based dose fit & token barrier endpoint & finite estimate with 95\% interval contained within the probed dose range \\
Persistent escape & injection-time cluster jump plus terminal retention & persistent basin-escape endpoint & persistent escape rate at least 0.50 at the claimed barrier dose \\
\bottomrule
\end{tabularx}
\end{table}

\subsection{Baselines}

We use three baselines to separate endogenous recursive structure from marginal embedding geometry or seed-conditioned sampling.

\begin{table}[h!]
\centering
\small
\begin{tabularx}{\textwidth}{lYY}
\toprule
baseline & implementation role & interpretation \\
\midrule
\texttt{time\_shuffled} & reshuffles step labels within each trajectory and recomputes dynamics metrics & tests whether a statistic depends on temporal order rather than the point cloud alone \\
\texttt{no\_feedback} & samples each step from the seed only, ignoring accumulated context & removes recurrence while preserving seed conditioning \\
\texttt{independent\_regeneration} & regenerates each step from the system prompt and seed with no carryover & removes history dependence completely \\
\bottomrule
\end{tabularx}
\end{table}

A regime is treated as endogenous only if its diagnostic statistic differs from all applicable baselines beyond the statistical gate. Effect size relative to each baseline is computed as Cohen's \(d\), namely the recursive mean minus the baseline mean divided by pooled standard deviation. Baseline applicability differs by experiment: no-feedback and independent-regeneration baselines are operator-regime baselines, while time shuffling applies broadly to trajectory metrics.

\subsection{Statistical procedures}

Confidence intervals for trajectory-level quantities are computed by 1000-iteration bootstrap. Between-condition differences, including perturbation-condition comparisons, are tested by permutation tests on the relevant mean difference. Switching-rate proportions in small-denominator dose-response cells use Wilson-style confidence intervals where ordinary bootstrap intervals would be unstable.

Effect sizes for recursive-vs-baseline comparisons use Cohen's \(d\). A signal counts only if it passes both parts of the significance gate: the diagnostic statistic must be at least 2 standard deviations above the baseline mean, and Cohen's \(d\) must be at least 0.5. This prevents very large samples from turning negligible effects into claim-bearing signals.

Basin-predictability cross-validation uses the adaptive stratified protocol and the group-aware protocol described in §4.5.3. Singleton classes are dropped before cross-validation and recorded. Cells that cannot be split into at least two non-singleton classes are recorded as missing rather than forced into an unstable estimate.

All endpoint-level analyses use the pre-specified observable and projection choices described above. Sensitivity to other observables is reported as diagnostic support, not as a replacement for the primary endpoint unless the endpoint itself specifies cross-observable agreement.

\subsection{Static visualization battery}

Every experiment generates a standardized set of static plots. These figures are used to inspect trajectories, detect failures, and communicate dynamics, but they are not primary endpoints unless explicitly tied to the criteria in §4.13.

The visualization battery includes joint t-SNE plots colored by regime, family, and step; per-family trajectory grids in shared coordinates; ensemble-spread timelines; PCA-2 quiver fields; sampled t-SNE trajectories with temporal ordering; streamlines with density; speed-colored streamlines; divergence fields; even/odd step parity plots for oscillatory regimes; family-by-initial-condition final-cluster maps; and distributional plots for basin entry, basin score, cluster occupancy, and dwell.

Plots are rendered at 200 DPI to PNG. Each experiment's \texttt{reports/plots/} directory contains approximately 50 to 150 figures depending on the number of observables and optional visualization modules. The plotting code is deterministic given the stored embeddings, projections, and metric CSVs.

\subsection{Flow-field computation}

Flow-field visualizations share a bin-and-aggregate kernel that converts a trajectory ensemble into a spatially resolved displacement field on a two-dimensional projection. For each trajectory group, points are sorted by step. Each adjacent pair contributes a start location and a displacement vector. The collection of starts and displacements is binned over the projection plane.

For each grid bin, the average displacement vector is computed from all transitions whose start point falls inside the bin. Empty bins are left undefined, so streamline integration does not pass through unsupported regions. This produces an empirical vector field \(v(x)=(u(x),w(x))\) that represents the average one-step projected motion observed in that region.

Density fields use a higher-resolution two-dimensional histogram followed by Gaussian smoothing. The smoothed density \(\hat\rho(x)\) is used as a background heatmap and, in perturbation analyses, as the basis for the effective potential

\[V(x)=-\log(\hat\rho(x)+\epsilon),\]

where \(\epsilon>0\) is a small numerical stabilizer preventing log of zero in low-density grid cells.

Streamlines are integral curves of the empirical vector field. Divergence is computed as

\[\nabla\cdot v(x)=\frac{\partial u}{\partial x}+\frac{\partial w}{\partial y}.\]

Negative divergence indicates sink-like behavior, while positive divergence indicates source-like behavior. For recursive LLM loops, we expect weakly negative average divergence in contractive regimes, strong local minima at absorbing sinks, and saddle-like structure in oscillatory regimes. Implementation details and grid parameters are provided in Supplementary §12.8.

\subsection{Perturbation visualization toolkit}

Perturbation experiments additionally generate an empirical-potential and barrier-visualization toolkit. PCA-2 coordinates are converted into a smoothed density estimate \(\hat\rho(x)\), then into an effective potential \(V(x)=-\log \hat\rho(x)\). Basin centers are detected as local minima on the potential grid. Geodesic barriers between basin pairs are computed by shortest-path search on the grid, with the barrier height defined by the maximum potential encountered along the path.

The toolkit also produces streamlines over potential contours, geodesic overlays, condition-wise flow-field panels, and three-dimensional density-shell animations. The 3D animations use PCA-3 coordinates, iso-density shells, sampled trajectory walks, and visual markers for perturbation events. These visualizations are interpretive aids for the perturbation endpoint analyses. Full grid parameters, smoothing constants, local-minimum detection, Dijkstra settings, marching-cubes details, transparency schedules, and parallel rendering settings are provided in Supplementary §12.8.

\subsection{End-to-end pipeline schematic}

The complete pipeline is organized into seven deterministic phases:

\begin{table}[h!]
\centering
\small
\begin{tabularx}{\textwidth}{lYY}
\toprule
phase & operation & persistent output \\
\midrule
Phase 1 Generation & run recursive trajectories with \texttt{gpt-4o-mini}, applying append, replace, or dialog updates & \texttt{raw/steps.jsonl} \\
Phase 2 Observables & derive text views such as \texttt{output}, \texttt{rolling\_k3}, \texttt{context\_tail}, role-specific turns, and \texttt{turn\_pair} & observable text streams \\
Phase 3 Embeddings & embed each observable with \texttt{text-embedding-3-small} and L2-normalize rows & \texttt{embeddings/<obs>/embeddings.npy} and \texttt{metadata.parquet} \\
Phase 4 Joint projections & fit PCA-2, PCA-10, PCA-50, and t-SNE jointly per observable & projected-coordinate files \\
Phase 5 Metrics and endpoints & compute clusters, recurrence, dwell, basin metrics, ensemble-spread diagnostics, predictability, and perturbation endpoints & metric CSVs \\
Phase 6 Baselines and statistics & compute baselines, bootstrap intervals, permutation tests, effect sizes, and pass-rule summaries & bootstrap and effect-size summaries \\
Phase 7 Reports & render plots, perturbation visualizations, endpoint tables, and narrative reports & per-experiment \texttt{reports/} directories \\
\bottomrule
\end{tabularx}
\end{table}

Each phase writes a deterministic intermediate, allowing downstream analyses to be rerun without regenerating trajectories. The full TikZ source, shape annotations, persistence-boundary table, and rerun semantics live in Supplementary §12.9.

The source of truth is \texttt{steps.jsonl}. Embeddings, projections, metrics, figures, and reports are regenerable from that file plus the code and configuration. In routine development, metric and plotting changes can be rerun without additional model-generation calls.

\subsection{Hardware and software}

All experiments run locally on a single workstation with API calls to OpenAI for generation and embeddings; no GPU is required. The host used to build the released artefacts is an HP ProLiant DL360 Gen9 with two Intel Xeon E5-2687W v3 processors (2 x 10 physical cores at 3.10 GHz base, 40 logical threads total) and 256 GB of RAM, running Windows 10 Pro 64-bit. Embedding ingestion, dimensionality reduction, clustering, density-and-geodesic-barrier computation, and animation rendering are all CPU-only.

The Python environment is Python 3.14 with numpy 2.3, scipy 1.16, scikit-learn 1.8, scikit-image 0.26, pandas 2.3, matplotlib 3.10, and imageio-ffmpeg 0.6 (resolved versions used to produce the released artefacts; the code itself targets Python 3.10+). The full dependency lock is in \texttt{requirements.txt}. Animations are stitched via imageio-ffmpeg using the libx264 codec. The pytest suite of 99 tests is green end-to-end and runs in roughly 13 seconds in this environment.

Parallel rendering of trajectory animations and basin diagnostics uses \texttt{concurrent.futures.ProcessPoolExecutor} with up to 40 workers, matching the number of logical threads on the host. The framework makes no other hardware assumptions; the analysis pipeline runs on any Linux, macOS, or Windows machine with the dependency stack above and enough RAM to hold a single experiment's trajectories and PCA-10 embeddings in memory (a few GB per experiment).

\subsection{Decision-grade endpoints}

The five primary endpoints introduced in §4.0 are now defined operationally. The metric battery in §4.5 is intentionally broad: it is used to diagnose, visualize, and stress-test recursive dynamics from several angles. The paper's headline claims, however, should not depend on dozens of partially redundant quantities. For decision purposes, we treat the following five endpoints as load-bearing. Each endpoint has a fixed numerical pass rule; results that do not clear the rule are reported as diagnostic, exploratory, or in-flight rather than as supported regime claims.

\begin{table}[h!]
\centering
\small
\begin{tabularx}{\textwidth}{lYYYY}
\toprule
endpoint & definition & measured at & threshold for "regime claim is supported" & defined in \\
\midrule
\textbf{Operational attractor score C1-C4} & Count of the four attractor criteria passed: late-window basin persistence, recurrence or dwell above null, embedder robustness, and contraction, re-entry, or collapse. & Publication-scale O1, O2, O3, and D1 on canonical observables. D2 exploratory status is checked separately. & \textbf{Strong attractor:} 4/4 criteria pass. \textbf{Attractor-like:} at least 3/4 pass. \textbf{Not attractor:} fewer than 3/4 pass. Missing publication-scale measurements count as fail unless structurally inapplicable. & §4.5.1, §4.5.2, §4.6, Supplementary §12.2 \\
\textbf{Leakage-free basin predictability acc\_group(k=10)} & GroupKFold-by-prompt-family accuracy of predicting the late-window K-means basin from the PCA-10 state at step \(k=10\). & Publication-scale O1, O2, O3, and D1; \texttt{context\_tail}; K-means \(k=12\). & To claim \textbf{cross-family basin predictability:} \(\operatorname{acc}_{\mathrm{group}}(k=10)\geq0.70\). To claim the original stratified number is \textbf{leakage-free:} \(\Delta=\operatorname{acc}_{\mathrm{stratified}}-\operatorname{acc}_{\mathrm{group}}<0.10\). & §4.5.3, Supplementary §12.3 \\
\textbf{Perturbation switching signature} & Final-step switching rate: fraction of perturbed trajectories whose final K-means cluster differs from the paired control trajectory. & O1 dose-response at matched 200-token dose; O2, O3, and D1 perturbation pilots. & \textbf{O1 selective sensitivity:} \(S_{\mathrm{adv}}(200)\geq0.50\) and \(S_{\mathrm{adv}}(200)/\max(S_{\mathrm{neutral}}(200),S_{\mathrm{lorem}}(200))\geq2.0\), with maximum out-of-distribution switching at most 0.30. \textbf{Replace-mode capitulation:} minimum non-control switching across O2/O3 neutral, lorem, and adversarial conditions at least 0.85. & §4.4, §4.5.4, Supplementary §12.5 \\
\textbf{Behavioral ED50 token barrier} & The perturbation dose \(\tau\) at which a four-parameter logistic fit to the O1 adversarial dose-response reaches 50\% switching, with prompt-family-cluster bootstrap uncertainty. & O1 adversarial dose sweep. Sparse run: \(\tau\in\{20,80,200,400\}\), \(n=50\) per cell. Dense rerun: \(n=200\) per cell across 8 doses. & To claim a \textbf{localized token barrier:} ED50 point estimate finite and the 95\% family-cluster bootstrap interval lies wholly inside the probed interval \([20,400]\) tokens. If the point estimate is inside but the interval crosses the boundary, report only "finite but unlocalized" or "in flight." & §4.5.4, Supplementary §12.1 \\
\textbf{Destination-coherent persistence} \(S_{\mathrm{persist}}^{\mathrm{dst}}\) & Fraction of trajectories that visibly change cluster at injection AND remain in that \textit{specific} post-injection cluster at the terminal step. & O1 adversarial dose sweeps; joint PCA-10 K-means clusters; persistence summary aggregated under §5.1.3. & To claim \textbf{destination-coherent persistent escape}: rate at least 0.50 at the claimed barrier dose. & §4.4, §4.5.4, §5.1.3, Supplementary §12.6 \\
\textbf{Retained source-basin escape} \(S_{\mathrm{persist}}^{\mathrm{src}}\) & Fraction of trajectories kicked at injection AND outside the pre-injection cluster at the terminal step (regardless of destination). The strict endpoint above is a subset of this one. & Same data sources as above; full-history extension under the no-clip protocol of §5.1.3. & To claim \textbf{durable departure from the source basin}: rate at least 0.50 at the claimed barrier dose. & §4.4, §5.1.3 \\
\bottomrule
\end{tabularx}
\end{table}

On the current data, after the dense-dose rerun and the endpoint-decomposition analysis:

\begin{itemize}
  \item \textbf{Operational attractor score C1-C4:} O1, O2, O3 are \textit{attractor-like} under the original criteria (≥ 3/4 pass) and remain attractor-like under the stricter group-aware C1 + z-tested C2 reading (3/4 z-tested PASS, borderline; see §12.10 honest reading). D1 is borderline: it passes the attractor-like threshold under the original criteria but fails group-aware C1 (acc(k=10)=0.34) under the stricter reading. D2 does not pass at any threshold. The scorecard is therefore best read as \textit{operational attractor-like} status, not as a claim about strong attractors in any deeper sense.
  \item \textbf{Leakage-free basin predictability:} only O1 passes the stricter \(\operatorname{acc}_{\mathrm{group}}(k=10)\geq0.70\) and \(\Delta<0.10\) rule. O2, O3, and D1 fail under group-aware cross-validation.
  \item \textbf{Perturbation switching signature:} O1 selective sensitivity passes, with \(S_{\mathrm{adv}}(200)=0.620\) in the dense rerun and a ratio to neutral or lorem switching of approximately 2.8. Replace-mode O2/O3 capitulation passes by point estimate but is \textbf{tautological at the state-write step} (replace-mode intervention sets \(X_{t+1}=\operatorname{clip}(\text{perturbation\_text})\)); only the post-injection convergence under \(f\) is empirical, and the insert-mode counterfactual (§5.2) is the appropriate model-behavior measurement.
  \item \textbf{Behavioral \(\mathrm{ED50}_{\mathrm{raw}}\) token barrier:} passes at approximately 40 tokens, with estimates of 36 from the four-parameter logistic fit, 41 from the generalized linear mixed model, and 52 as the bootstrap median. The 95\% interval \([8.5,242]\) is wide because of the 5-family-cluster heavy tail.
  \item \textbf{\(\mathrm{ED50}_{\mathrm{net}}\) above natural floor:} does not pass. The net effect saturates at +32 percentage points at dose 400, below the +50 percentage-point threshold.
  \item \textbf{Destination-coherent persistence:} under the canonical bounded-memory loop, does not pass - at dose 400, only 16\% of dense-data trajectories are kicked AND remain in the post-injection cluster, well below the 50\% threshold; \(\mathrm{ED50}_{\mathrm{persist}}^{\mathrm{dst}}\) is undefined in the tested 5-400 token range. Under the full-history protocol (§5.1.3), the destination-coherent first crossing is approximately 1,500 tokens (non-monotonic at higher doses).
  \item \textbf{Retained source-basin escape:} under the canonical bounded-memory loop, plateaus around 36\% at dose 400 (kicked AND outside the pre-injection cluster), still below threshold. Under the full-history protocol, \(\mathrm{ED50}_{\mathrm{persist}}^{\mathrm{src}}\approx 400\) tokens (saturating at ~75-80\% by dose 1,500); see §5.1.3 / Fig 5.
\end{itemize}

In the present experiments, the equivalence rule \(C(O(X_T))\) is a K-means cluster of an embedding-space observable. In tool-using coding agents, the same endpoint structure can be instantiated with engineering observables: final patch family, files touched, the failing or passing test set, the selected plan category, the tool-call sequence, a security-policy violation, or an embedding of the full trajectory trace. Algorithm 1 requires only a consistent, pre-specified equivalence rule and paired controls; it does not require that "cluster" literally mean an embedding cluster.

\section{Results}

Adversarial append-mode perturbations produce a clear raw-switching dose response in the O1 continuation loop, with $\mathrm{ED50}_{\mathrm{raw}} \approx 40$ tokens. The dense rerun under the canonical bounded-memory loop shows a raw plateau near 67\%, a natural stochastic-divergence floor near 35\%, a maximum net adversarial effect of +32 percentage points at 400 tokens, and a maximum destination-coherent persistence rate of 16\% under K-means $k=12$. Persistent escape is memory-policy-conditioned: under a full-history protocol (no artificial truncation within the 30-step horizon), retained source-basin escape reaches 50\% near 400 tokens and saturates at ~75-80\%, while destination-coherent persistence first reaches 50\% near 1,500 tokens (full breakdown in §5.1.3 / Fig 5). Replace-mode loops initially appear almost fully perturbation-transparent, but the overwrite-versus-insert probe shows that most of this effect comes from the memory policy discarding prior state, not from a low injected-token barrier. Phase-0 and Phase-1 pilots validated the measurement pipeline and identified candidate regimes; full pilot history is in §12.7, the aggregation and per-experiment catalog is in §12.9, and row-level endpoint audit tables are in Extended Data Tables 1 and 2 (§12.1, §12.2).

\textbf{How to read this section.} §5 is organized in two phases. \textit{Phase A - headline endpoints} contains §5.1 (append-mode raw, net, and persistent endpoints; observable robustness; memory-policy sensitivity, Fig 5) and §5.2 (replace-mode tautology and the F3 cross-loop check). These two subsections carry the load-bearing numerical claims. \textit{Phase B - regime characterization and robustness} spans §5.3 through §5.10 and supports the regime taxonomy: §5.3 publication-scale ordering; §5.4 perturbation-content separation; §5.5 dialog drill-down; §5.6 injection-timing basin hardening; §5.7 cluster-granularity stability; §5.8 within-regime structure (cluster semantics and family heterogeneity); §5.9 robustness (embedder ablation and within-vendor cross-model verification); §5.10 bulk geometry is descriptive. Readers tracking only the headline finding can stop after §5.2; readers evaluating the regime taxonomy and robustness story should continue.

\subsection*{Phase A, headline endpoint}

\subsection{Adversarial append perturbations: raw switching, persistent escape, and memory-policy sensitivity}

The central perturbation endpoint is O1 append-mode continuation under in-distribution adversarial text. In the sparse pilot, O1 adversarial perturbations showed a graded response, while O1 neutral perturbations remained flat near the out-of-distribution drift floor and D1 neutral perturbations saturated even at very small doses. The dense rerun then localized the O1 adversarial raw-switching curve at $n=200$ trajectories per dose and separated three quantities that must not be conflated:

\begin{enumerate}
  \item \textbf{Raw switching:} final K-means cluster differs from the paired control trajectory.
  \item \textbf{Net switching:} raw switching minus the control-control stochastic-divergence floor.
  \item \textbf{Persistent escape:} the trajectory visibly changes cluster at injection and remains in the post-injection cluster to the terminal step.
\end{enumerate}

The dense rerun was pre-registered before execution: $n=200$ per cell, equal to 5 families × 10 ICs × 4 runs, with 8 adversarial dose conditions plus one control condition, for 1,800 trajectories total. The configuration was \texttt{configs/perturbation/O1\_ed50\_dense.yaml}; the analysis script was \texttt{scripts/fit\_ed50\_hierarchical.py}.

\textbf{Dense O1 adversarial dose response, separating raw, net, and persistent endpoints}

\begin{table}[h!]
\centering
\small
\begin{tabularx}{\textwidth}{rYYYY}
\toprule
dose (tokens) & raw switch rate & Wilson 95\% CI & net over natural floor & persistent escape, K-means $k=12$ \\
\midrule
20 & 0.415 & [0.349, 0.484] & +0.068 & 0.035 \\
50 & 0.510 & [0.441, 0.578] & +0.163 & 0.070 \\
80 & 0.575 & [0.506, 0.641] & +0.228 & 0.035 \\
120 & 0.630 & [0.561, 0.694] & +0.283 & 0.090 \\
160 & 0.605 & [0.536, 0.670] & +0.258 & 0.115 \\
200 & 0.620 & [0.551, 0.684] & +0.273 & 0.130 \\
300 & 0.655 & [0.587, 0.717] & +0.308 & 0.140 \\
400 & 0.670 & [0.602, 0.731] & +0.323 & 0.160 \\
\bottomrule
\end{tabularx}
\end{table}

The control-control natural floor is 34.7\% [31.0\%, 38.6\%] across $n=600$ ordered control-control pairs. Thus two trajectories with the same family and IC seed but different generation RNG end in different K-means clusters 35\% of the time without any perturbation. The raw 50\% crossing occurs between 20 and 50 tokens, but much of that apparent switching is baseline stochastic divergence. Under the stricter net endpoint, the curve does not reach +50 percentage points within the tested range.

Three independent ED50 estimates agree on the raw-switching scale:

\begin{table}[h!]
\centering
\small
\begin{tabular}{lrl}
\toprule
method & ED50 (tokens) & uncertainty \\
\midrule
4PL pooled fit & 36 & point estimate \\
Mixed-effects logistic GLMM & 41 & point estimate, log10-dose slope \\
Family-cluster bootstrap median & 52 & 95\% CI [8.5, 242] \\
\bottomrule
\end{tabular}
\end{table}

\begin{savenotes}
\begin{figure}[h!]
\centering
\includegraphics[width=0.95\linewidth]{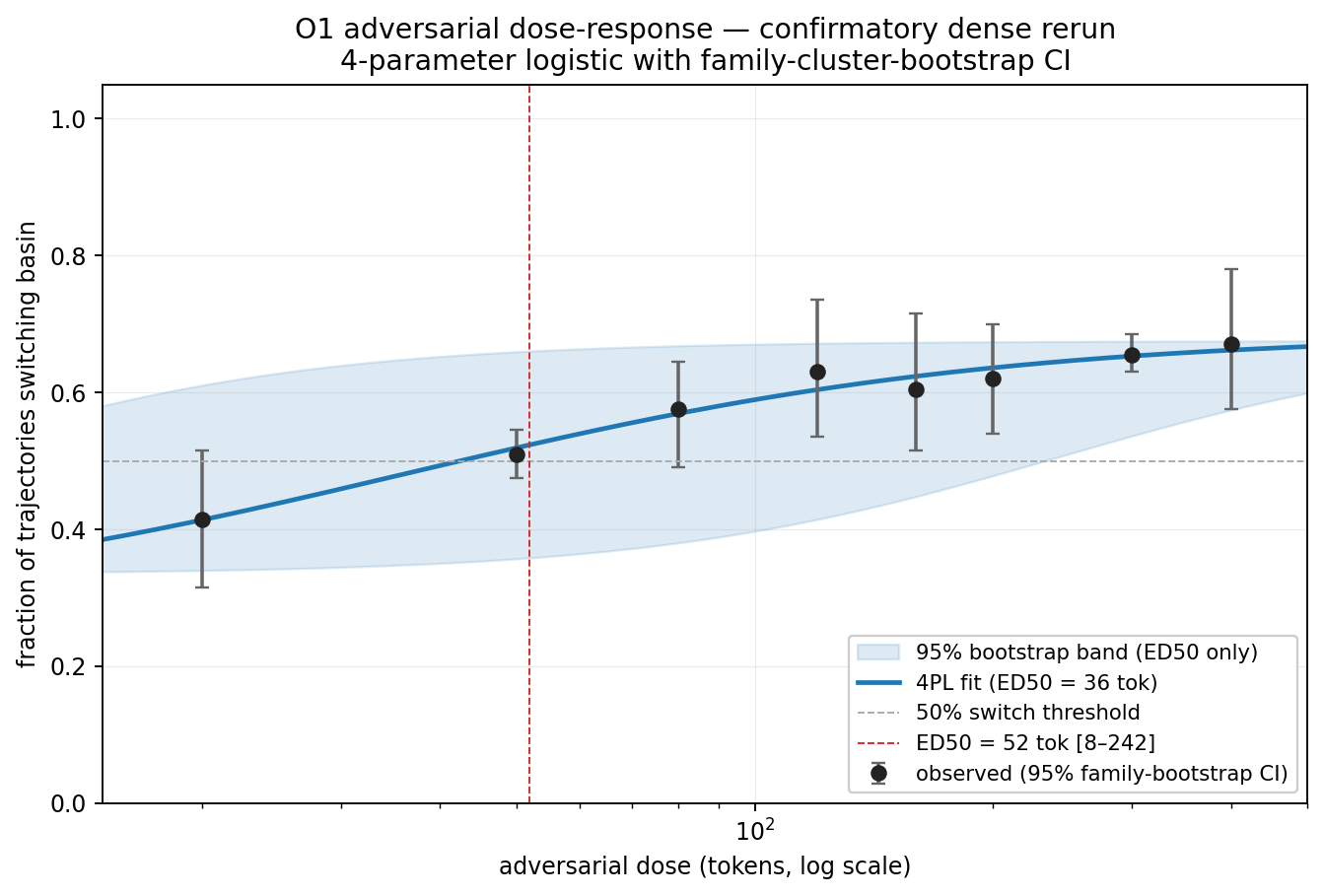}
\caption{\textbf{Dense O1 adversarial ED50 fit.} O1 append-mode adversarial dose response from the dense confirmatory rerun, with 8 doses x $n=200$ per cell. Black points are observed switching rates with family-cluster bootstrap 95\% CIs; the blue curve is a 4-parameter logistic fit (\texttt{a=0.69, d=0.28, b=1.16, ED50=36 tok}); the dashed red line marks the bootstrap-median ED50 = 52 tokens [CI 8.5, 242]. Source: \texttt{data/exp\_perturb\_O1\_ed50\_dense/reports/perturbation/ed50\_curve.png}.\\[2pt]{\footnotesize\itshape Fig 2 shows the dense O1 confirmatory adversarial append dose-response: the x-axis, adversarial dose (tokens, log scale), gives appended perturbation length, and the y-axis, fraction of trajectories switching basin, gives raw basin-switching probability. The observed switching rates (black points showing mean switching rates per dose, with gray vertical intervals showing family-cluster bootstrap 95 percent confidence intervals) rise from the control-vs-control natural floor of 0.347 [0.310, 0.386] toward a plateau near 67 percent. The 4PL fit (solid blue 4-parameter logistic curve estimating the dose-response, with a=0.69, d=0.28, b=1.16, ED50=36 tokens) crosses the 50 percent switch threshold (gray horizontal reference line marking half of trajectories switching basin) at broadly similar doses to the other estimators. The 95 percent bootstrap band (pale blue ribbon showing bootstrap uncertainty in the fitted ED50 while holding other structure fixed) is wide, so the threshold location is imprecise. Three convergent estimates appear: 4PL=36, mixed-effects GLMM=41, and the family-cluster bootstrap median ED50 = 52 tokens with CI 8.5 to 242 (dashed red vertical line). Adversarial dose clearly increases raw switching, with maximum net effect +0.323 at dose 400, but does not imply persistent escape: at dose 400, persistent escape is only 0.16 for K-means k=12, 0.10 for K-means k=4, and 0.395 under HDBSCAN. The claim would have been falsified by near-complete high-dose switching with persistent escape rates near the raw plateau, robust across clustering choices and clearly separated from the natural floor.}}
\end{figure}
\end{savenotes}

The point estimates cluster in the 36-52 token range, substantially below the earlier sparse-grid estimate near 150 tokens. The family-cluster bootstrap interval remains wide because only five prompt families are available for resampling.

Two structural findings matter more than the exact ED50 point estimate. First, the raw curve plateaus near 67\%, not near 100\%. The 4-parameter logistic upper asymptote is $a = 0.69$, implying a substantial non-switching subpopulation under the present protocol. Second, the persistent-escape endpoint is much smaller than raw switching. At dose 400, raw switching is 67\%, but persistent escape is 16\% under K-means $k=12$. Most raw switching is therefore not clean barrier crossing. It is final-step divergence from the paired control, often without a durable at-injection jump into a new basin.

The persistence decomposition on the dense rerun confirms this interpretation. At dose 400, 69 of 200 trajectories visibly changed cluster at injection. Of those, 32 persisted in the post-injection cluster, 13 drifted back to the pre-injection cluster, and 24 drifted elsewhere. Even among trajectories that visibly jump at injection, roughly half do not remain in the post-injection basin.

\begin{savenotes}
\begin{figure}[h!]
\centering
\includegraphics[width=0.95\linewidth]{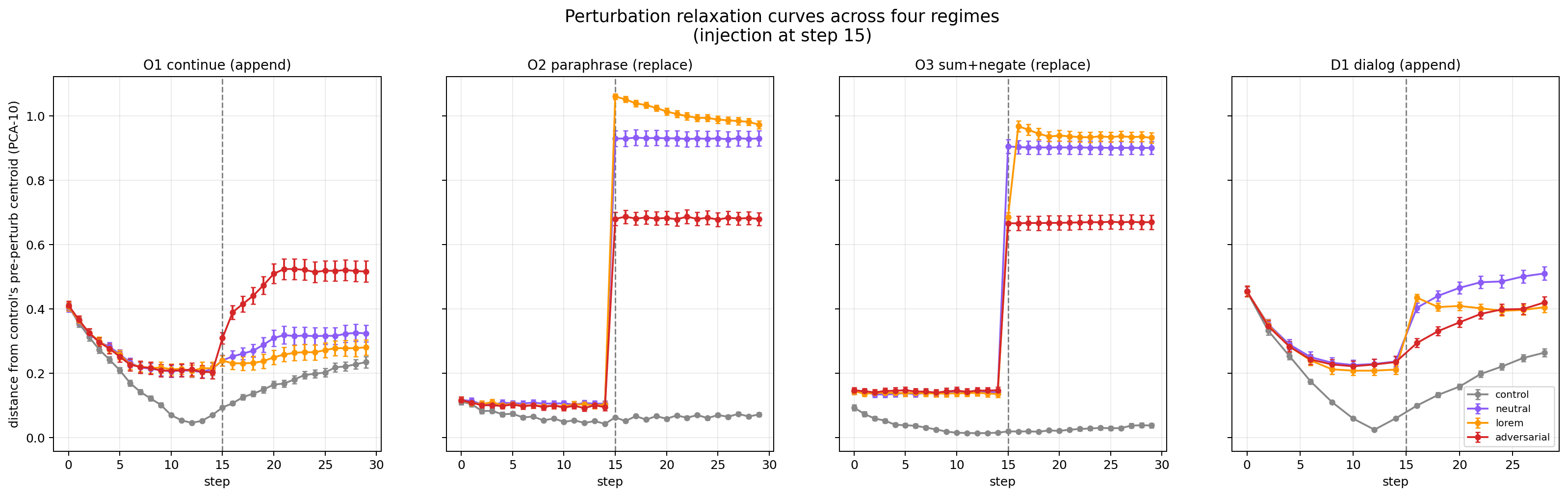}
\caption{\textbf{Post-perturbation relaxation and recovery.} Relaxation curves after perturbation show that many trajectories move transiently but do not remain in the injected post-jump basin. The curves support the distinction between raw switching and durable escape. Source: \texttt{data/aggregated/perturbation\_cross\_regime/cross\_relaxation\_curves.png}.\\[2pt]{\footnotesize\itshape Fig 3 shows post-perturbation relaxation curves across O1 (O1, the append-mode continuation loop), O2 (O2, the paraphrase replace-mode loop where each output replaces context), O3 (O3, the summarize-and-negate replace-mode loop), and D1 (D1, the dialog-style multi-basin recursive loop). The y-axis is distance from the pre-perturbation centroid in PCA-10 space, the x-axis is recursive step, and the vertical dashed line marks the injection at step 15. Curves compare control (control, the unperturbed trajectory), neutral (neutral, off-topic filler text), lorem (lorem, lorem-ipsum-style out-of-distribution gibberish), and adversarial (adversarial, on-distribution text drawn from another trajectory of the same regime) perturbations. The main pattern is that replace regimes, O2 and O3, respond almost discontinuously: neutral, lorem, and adversarial jump at injection and then partially relax or plateau, indicating that replacement directly relocates the loop state. Append regimes, O1 and D1, instead show slower cumulative drift, with perturbation effects accumulating after injection rather than appearing as a single state reset. D1 also preserves multi-basin structure, with different perturbations relaxing toward distinct offsets. The O1 dense rerun clarifies that an apparent jump is not automatically persistent basin transfer: at dose 400, 69 of 200 trajectories visibly jumped at injection, but only 32 persisted, 13 returned to the pre-injection cluster, and 24 drifted elsewhere, so 32/69, or 46 percent, of jumped trajectories remained in the post-injection basin. This interpretation would have been falsified if replace loops changed smoothly without injection discontinuities, append loops jumped and stabilized immediately, or all perturbation labels produced indistinguishable relaxation traces.}}
\end{figure}
\end{savenotes}

Because persistence is cluster-defined, we also recomputed it under three granularities: K-means $k=12$, K-means $k=4$, and HDBSCAN. The formal persistent-escape ED50, the dose at which persistent escape reaches 50\%, is not reached under any of the three definitions.

\begin{table}[h!]
\centering
\small
\begin{tabularx}{\textwidth}{rYYYY}
\toprule
dose & persistent escape, $k=12$ & persistent escape, $k=4$ & persistent escape, HDBSCAN & kicked at injection, HDBSCAN \\
\midrule
20 & 3.5\% & 1.5\% & 7.0\% & 12.0\% \\
50 & 7.0\% & 3.0\% & 16.5\% & 28.5\% \\
80 & 3.5\% & 5.0\% & 28.5\% & 48.0\% \\
120 & 9.0\% & 4.5\% & 35.5\% & 58.0\% \\
160 & 11.5\% & 9.5\% & 41.0\% & 64.5\% \\
200 & 13.0\% & 13.5\% & 40.5\% & 60.0\% \\
300 & 14.0\% & 8.5\% & 40.0\% & 66.5\% \\
400 & 16.0\% & 10.0\% & 39.5\% & 68.5\% \\
\bottomrule
\end{tabularx}
\end{table}

\begin{savenotes}
\begin{figure}[h!]
\centering
\includegraphics[width=0.95\linewidth]{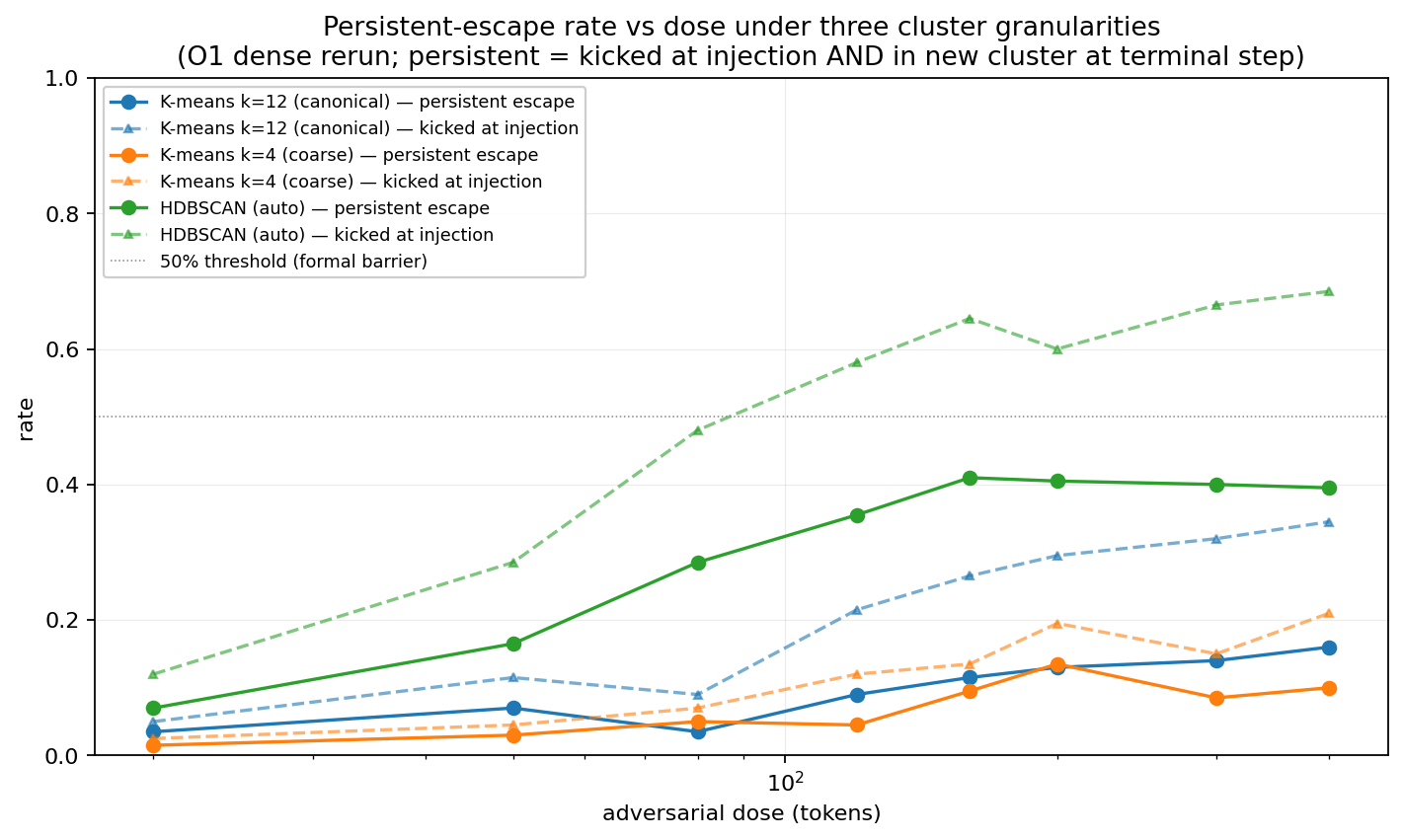}
\caption{\textbf{Persistent escape under cluster granularity (canonical bounded-memory loop).} Persistent-escape rates recomputed under K-means $k=12$, K-means $k=4$, and HDBSCAN, all on the canonical \texttt{context\_tail} observable. No clustering convention reaches the 50\% destination-coherent persistence threshold up to 400 injected tokens \textit{in the canonical 12K bounded-memory loop}. Observable-robustness of the same conclusion is checked in §5.1.1 (\texttt{output} single-step rises to 27\% at dose 400, still below 50\%); the bounded-memory plateau breaks under the full-history protocol (§5.1.3 / Fig 5), where source-basin escape crosses 50\% near 400 tokens. Source: \texttt{data/aggregated/multi\_granularity\_persistence.png}.\\[2pt]{\footnotesize\itshape Fig 4 plots dose response curves for two event definitions under three clustering granularities: persistent escape (the trajectory was kicked at injection AND remained in the post-injection cluster at the terminal step) and kicked at injection (any visible cluster jump immediately after perturbation regardless of final state). K-means k=12 (the canonical 12-cluster partition), K-means k=4 (the coarser 4-cluster partition), and HDBSCAN (the density-based variable-k partition) give the same qualitative conclusion. The dotted horizontal line marks the 50 percent formal barrier threshold, and no persistent escape curve crosses it. At dose 400, persistent escape is only 16 percent for k=12, 10 percent for k=4, and 39.5 percent for HDBSCAN. Thus even the most permissive density-based partition stops below majority persistence, while the stricter coarse partition yields only 10 percent. The dashed kicked-at-injection curves are substantially higher, showing that the perturbation often produces an immediate visible displacement: HDBSCAN reaches 68.5 percent kicked at injection at dose 400, and already 60 percent at dose 200, where its persistent escape rate is only 40.5 percent. This separation between immediate displacement and terminal persistence is the central point of the panel: many samples are knocked into a different visible region, but most do not remain there. The result is robust to cluster granularity and rules out the alternative that persistent escape is merely hidden by the choice of clustering. The claim would have been falsified if a reasonable clustering choice, especially permissive HDBSCAN, had driven persistent escape above the 50 percent barrier or revealed majority terminal escape where k=12 or k=4 did not.}}
\end{figure}
\end{savenotes}

HDBSCAN is the most permissive definition and gives the largest persistent-escape values, but even there the maximum is 39.5\%, below the 50\% threshold. The conclusion is robust to cluster granularity within the canonical bounded-memory loop: O1 adversarial append perturbations create a finite raw-switching dose response, but persistent basin escape is not demonstrated up to 400 injected tokens \textit{under the canonical 12,000-character tail clip}. The scope qualifier matters: §5.1.3 demonstrates that this saturation is largely a memory-policy artifact, and that under unbounded memory persistent escape \textit{does} reach 50\% at higher doses.

\subsubsection{Observable-robustness of the persistence finding}

The Fig 4 numbers use the canonical observable \texttt{context\_tail} (last 4000 chars of running context per step; §4.4). At each step we embed only the text slice defined by the observable: the current output, the last 3 outputs, or the last 4000 characters of running context. \textbf{We do not average embeddings over the entire trajectory.} Persistence is computed from these step-level embeddings (cluster at pre-injection step 14, at-injection step 15, and the terminal step), not from any pooled or averaged trajectory representation.

Because \texttt{context\_tail} integrates over multiple recent generations - in O1 append-mode the running context crosses the 4000-char threshold at step 6, so from step 7 onward \texttt{context\_tail} is a sliding window over roughly the last 6-7 generations - single-step shifts can be diluted by surrounding content still in the tail. We checked observable-robustness by recomputing the kicked / persisted decomposition under two alternatives: \texttt{output} (model's single-step generation alone, ~600 chars) and \texttt{rolling\_k3} (concatenation of the last 3 outputs, ~1500-2000 chars). Joint PCA-10 + K-means $k=12$ as in §5.1; n=200 trajectories per dose-cell from the dense rerun. Persistence rate is the unconditional fraction of all n=200 trajectories per cell satisfying "kicked at injection AND still in the post-injection cluster at the terminal step" - not conditional on having been kicked. Cluster labels are observable-specific: each observable reruns the same unsupervised PCA-10 + K-means pipeline, so cluster IDs are not comparable across observables - only persistence rates are.

\begin{table}[h!]
\centering
\small
\begin{tabular}{lrrr}
\toprule
observable & approximate window & dose 200 persisted & dose 400 persisted \\
\midrule
\texttt{output} & 1 generation (~600 chars) & 0.310 [0.250, 0.377] & 0.270 [0.213, 0.335] \\
\texttt{rolling\_k3} & 3 generations (~1500-2000 chars) & 0.160 [0.116, 0.217] & 0.155 [0.111, 0.212] \\
\texttt{context\_tail} (canonical) & ~6-7 generations (~4000 chars) & 0.130 [0.090, 0.184] & 0.160 [0.116, 0.217] \\
\bottomrule
\end{tabular}
\end{table}

Full 8-dose grid for all three observables: \texttt{data/aggregated/multi\_observable\_persistence/\{long.csv, summary.png\}}.

\textbf{The qualitative claim is robust.} Under every tested observable, persistent escape stays well below the 50\% threshold at every tested dose from 5 to 400 tokens. The "no $\mathrm{ED50}_{\mathrm{persist}}$" conclusion is independent of observable choice.

\textbf{The quantitative magnitude is observable-conditional.} At dose 400, persistence ranges from 16\% under \texttt{context\_tail} (the conservative recent-context estimate) to 27\% under \texttt{output} (the more permissive final-generation estimate). The largest point estimate observed at any dose is 31\% under \texttt{output} at dose 200. These should be read as observable-conditional sensitivity checks rather than estimates of a single latent rate: there is no observable-independent "true" persistence rate. The single-step observable necessarily registers any generation-level shift, whereas \texttt{context\_tail} blends each recent generation against ~5-6 surrounding ones, so a shift that affects only the latest generation is partially absorbed. Reports of persistence rates should specify the observable; the §5.1 / Fig 4 numbers use \texttt{context\_tail}.

\textbf{The same observable-robustness check on D1.} D1 dialog runs the same canonical pipeline (\texttt{context\_tail}, agent-role-filtered) and reports high raw switching already at the smallest tested dose. We re-ran the D1 neutral dose sweep (\texttt{exp\_perturb\_D1\_dose}, n=50 trajectories per cell) under four observables: \texttt{output}, \texttt{last\_agent\_turn} (the cleanest single-turn dialog observable, which embeds only the responder's most recent turn), \texttt{rolling\_agent\_k3} (last 3 agent turns), and \texttt{context\_tail}. Terminal-cluster raw switching (matching the protocol of §4.0):

\begin{table}[h!]
\centering
\small
\begin{tabularx}{\textwidth}{rYYYY}
\toprule
dose & \texttt{output} & \texttt{last\_agent\_turn} & \texttt{rolling\_agent\_k3} & \texttt{context\_tail} (canonical) \\
\midrule
20 & 0.84 & 0.84 & 0.76 & 0.70 \\
80 & 0.82 & 0.82 & 0.76 & 0.76 \\
200 & 0.78 & 0.78 & 0.62 & 0.66 \\
400 & 0.72 & 0.74 & 0.66 & 0.58 \\
\bottomrule
\end{tabularx}
\end{table}

Across observables, D1 shows high raw terminal switching already at dose 20 and stays high across the sweep, but the point estimates are flat-to-decreasing rather than monotone increasing - D1 should not be read as exhibiting a conventional dose-response. The low-dose fragility claim is robust to observable choice: at dose 20, raw switching is at least 70\% under every observable; at dose 400 it is at least 58\%. The single-turn observables (\texttt{output}, \texttt{last\_agent\_turn}) have higher point estimates than \texttt{context\_tail} at every tested dose, though with n=50 per cell these differences should be treated as descriptive rather than inferential. The Fig 1 / §5.4 D1 numbers use \texttt{context\_tail}; the qualitative conclusion - high terminal switching even at very small doses - holds across all four observables. Full data: \texttt{data/aggregated/multi\_observable\_d1\_raw/\{long.csv, summary.png\}}.

\subsubsection{Apparent saturation under the canonical bounded-memory policy}

Within the canonical loop (max\_context\_chars = 12,000, tail-clip), the persistence dose-response is well-fit by a saturating 4PL with ceiling well below 50\%:

\begin{table}[h!]
\centering
\small
\begin{tabular}{lrrl}
\toprule
observable & ceiling (apparent plateau) & midpoint dose (inflection) & apparent $\mathrm{ED50}_{\mathrm{persist}}$ \\
\midrule
\texttt{output} (single-step) & 0.31 ± 0.03 & ~52 tokens & not reached \\
\texttt{rolling\_k3} (3-step window) & 0.18 ± 0.02 & ~107 tokens & not reached \\
\texttt{context\_tail} (canonical) & 0.15 ± 0.02 & ~139 tokens & not reached \\
\bottomrule
\end{tabular}
\end{table}

Under this fit all three observables predict an asymptotic plateau below the 50\% threshold. The midpoints sit inside the tested 5-400 range, so the curves appear to have reached their plateaus by ~200-300 tokens. Plot: \texttt{data/aggregated/multi\_observable\_persistence/extrapolation.png}.

\textbf{This apparent saturation is a real measurement of the bounded loop, but not a memory-policy-invariant property of the model.} The 12K buffer clips the perturbation out of the model's running context after ~10-15 post-injection steps. Once the perturbation is no longer visible to the model, the trajectory drifts back toward its natural basin and persistence drops. §5.1.3 reports a re-run under a full-history protocol (max\_context\_chars = 200,000, large enough that no artificial truncation occurs during the 30-step experiment) and shows that the first tested dose whose point estimate reaches the half-effect level is 1,500 tokens. The clipped saturation should therefore be read as a \textit{property of the bounded-memory loop's memory policy}, not as a memory-policy-invariant property of the model's response to perturbation.

\subsubsection{Memory-policy sensitivity: persistent-escape ED50 under unbounded memory}

We re-ran the O1 adversarial dose response with \texttt{max\_context\_chars = 200000} (effectively unbounded for 30-step trajectories) at the same n=200 design across doses 20 to 400 (\texttt{exp\_perturb\_O1\_ed50\_dense\_noclip}), then extended at n=100 to higher doses 600, 1000, 1500, 2000, 3000 (\texttt{exp\_perturb\_O1\_ed50\_higher\_noclip}). All other parameters are identical to the canonical dense rerun: gpt-4o-mini, 5 prompt families × 10 ICs × 4 runs at low doses (×2 at high doses), injection at step 15, K-means $k=12$ on joint PCA-10. The full no-clip dose-response, alongside the canonical clipped curve, is shown in Fig 5.

\begin{savenotes}
\begin{figure}[h!]
\centering
\includegraphics[width=0.95\linewidth]{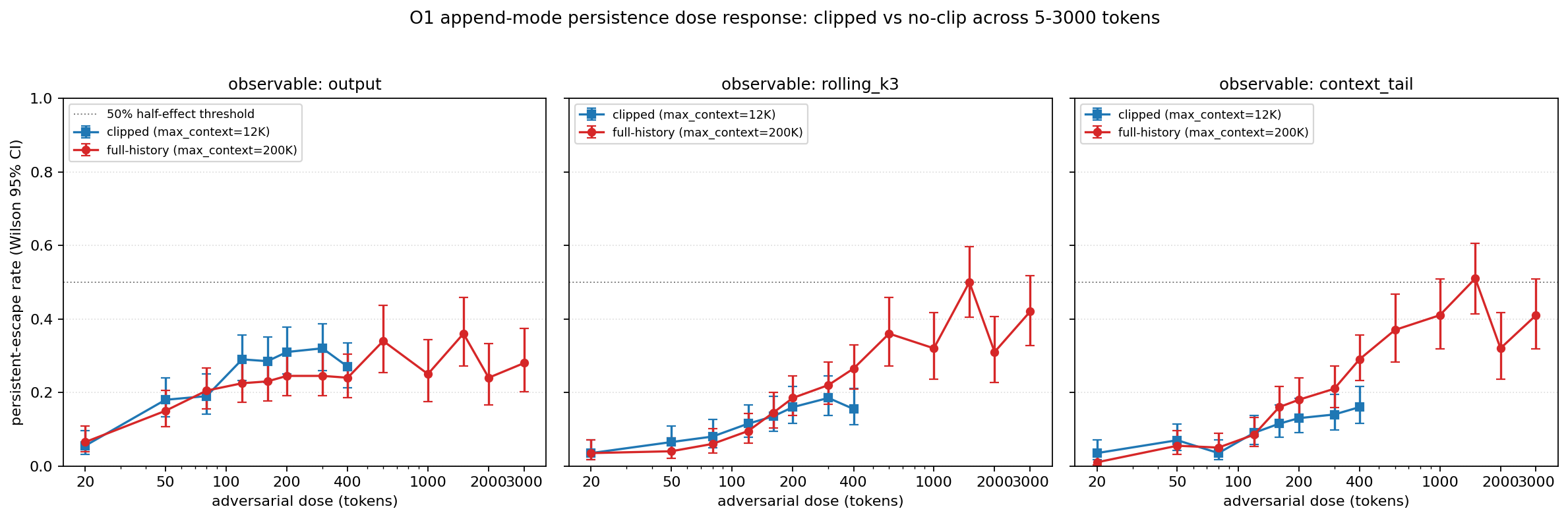}
\caption{\textbf{O1 append-mode destination-coherent persistence: clipped vs full-history memory, across observables.} All curves show \textit{destination-coherent persistence} (kicked AND in same post-injection cluster at terminal step); see Fig 1B for the source-basin escape curve under full-history. Three panels, one per observable (\texttt{output}, \texttt{rolling\_k3}, \texttt{context\_tail}). Blue squares: canonical bounded-memory loop (\texttt{max\_context\_chars = 12000}, n=200). Red circles: full-history loop (\texttt{max\_context\_chars = 200000}), plotted as a single continuous series across the merged dose grid (n=200 per cell at doses 20-400, n=100 per cell at doses 600-3000; the underlying protocol is otherwise identical, so the two N regimes are presented as one curve, the only visible difference being slightly wider Wilson 95\% CIs above dose 400). Dotted gray line marks the 50\% half-effect threshold. The clipped and full-history curves track closely at low doses (≤200 tokens) but diverge at moderate doses, with the full-history rate rising substantially higher under \texttt{rolling\_k3} and \texttt{context\_tail}. At dose 1500, both \texttt{rolling\_k3} (0.50, 95\% Wilson CI for the proportion [0.40, 0.60]) and \texttt{context\_tail} (0.51, [0.41, 0.61]) reach the 50\% point estimate, though the lower CI bound sits below 0.5. The upper tail is non-monotonic at the canonical 30-step horizon: persistence drops to 0.32 at dose 2000 and recovers to 0.41 at dose 3000 under \texttt{context\_tail}. A four-step falsification battery in §5.1.3 (heterogeneity control, cluster-granularity sweep, transition-entropy diagnostic, and 50-step trajectory continuation) localizes this dip to a finite-horizon, endpoint-definition-sensitive feature; it closes to zero by step 79. The single-step \texttt{output} observable stays below 50\% across all doses (peak 0.36 at dose 1500). The Fig 1B source-basin escape curve, under the same full-history protocol, is monotone and crosses 50\% near dose 400. Source: \texttt{data/aggregated/multi\_observable\_persistence\_full\_noclip/plot.png}.\\[2pt]{\footnotesize\itshape Fig 5 compares O1 persistent-escape dose response under bounded (max\_context\_chars=12000) and unbounded (max\_context\_chars=200000) memory across three observables. Clipped and unclipped curves overlap at low doses (≤120 tokens), diverge at moderate doses (160-400 tokens, +5-13 pp under context\_tail), and at high doses (600-3000) the unclipped curve crosses 50\% at dose 1500 under rolling\_k3 and context\_tail. The upper tail is non-monotonic (0.32 at dose 2000, 0.41 at dose 3000) but CIs at n=100 are wide. Single-step output stays below 50\%, peaking at 0.36 at dose 1500. Falsifying outcomes: matched curves throughout (would mean clip irrelevant) or no crossing under unbounded memory (would confirm intrinsic plateau).}}
\end{figure}
\end{savenotes}

\textbf{Headline finding.} Under the full-history protocol, \textit{retained source-basin escape} (the loose endpoint) crosses 50\% near 400 tokens (monotone, saturating ~75-80\% by 1,500 tokens). The stricter \textit{destination-coherent persistence} endpoint first reaches 0.50 at \textbf{1,500 tokens} (≈ 6,000 characters of in-distribution adversarial text injected at step 15 in a 30-step append-mode loop with gpt-4o-mini); \texttt{context\_tail} gives 0.51 and \texttt{rolling\_k3} gives 0.50 at this dose. \textbf{Caveat on the destination-coherent threshold.} The Wilson 95\% CIs for the \textit{proportion} at dose 1500 are [0.41, 0.61] under context\_tail and [0.40, 0.60] under rolling\_k3 - the lower bound sits below 0.5, so the threshold-crossing is at the boundary of statistical resolution at n=100. We therefore localize the destination-coherent half-effect region to roughly the \textbf{1,000-2,000 token neighborhood} rather than treating 1,500 as a resolved ED50, and we describe its 50\%-crossing rather than calling it a clean ED50 - a fitted ED50-dose CI would require a monotone dose-response model, which is precluded by the upper-tail non-monotonicity (see below).

\textbf{Why the bounded-memory loop hides this finding.} Under the 12,000-character clip, a 1,500-token perturbation (≈ 6,000 characters) plus the pre-injection trajectory (~10,000 characters at step 15) overflows the buffer immediately at injection: $X_{16} = \operatorname{clip}(X_{15} \,\Vert\, \text{perturbation})$ retains only the last 12,000 characters. By step 25-30, both the seed prompt and the perturbation have been clipped out of the model's context. The trajectory then continues from a context that no longer carries the perturbation, and the model's natural-basin generation tendency reasserts. The "saturation at ~15\%" reported in §5.1.2 is therefore not a property of the LLM's intrinsic response to perturbation; it is a property of the bounded-memory loop discarding the perturbation within ~10-15 post-injection steps.

\textbf{Two finite-horizon persistence endpoints.} Let \(C_0\) denote the pre-injection cluster (step 14), \(C_+\) the immediate post-injection cluster (step 15), and \(C_T\) the terminal cluster (step 29). Define a kick by \(K=[C_+\neq C_0]\). We report two endpoints, both under K-means \(k=12\) on joint PCA-10 of the canonical \texttt{context\_tail} observable:

\begin{itemize}
  \item \textbf{Retained source-basin escape}: \(K\,\wedge\,[C_T\neq C_0]\). The trajectory was kicked at injection and lies outside its pre-injection cluster at the terminal step. This is the more inclusive endpoint, asking only whether the trajectory has \textit{left} its starting basin and remained outside through the 30-step horizon.
  \item \textbf{Destination-coherent persistence}: \(K\,\wedge\,[C_T=C_+]\). The trajectory was kicked AND remains in the same non-source cluster into which it was immediately displaced. This is the stricter endpoint, asking whether the trajectory committed to a \textit{specific} new basin.
\end{itemize}

The second is a subset of the first; their difference quantifies destination heterogeneity among escaped trajectories.

\textbf{Why the strict endpoint is non-monotonic at high dose.} Under destination-coherent persistence, the dose response shows an apparent non-monotonic dip in the upper tail: 0.51 at dose 1500, 0.32 at dose 2000, 0.41 at dose 3000. The dip is \textit{not} explained by failed displacement: the kicked rate stays ~0.85-0.97 across doses 600-3000 and source-basin escape stays correspondingly high.

We tested four candidate mechanisms with a four-step falsification battery: a perturbation-heterogeneity control, a cluster-granularity sweep with hierarchical macro-merge, a transition-entropy diagnostic, and a long-horizon trajectory continuation. The conclusion is that \textbf{the dip is a finite-horizon, endpoint-definition-sensitive feature that resolves under a 50-step trajectory continuation and should not be interpreted as a stable structural attractor asymmetry}. It does not contradict the loose endpoint (retained source-basin escape), which stays high and monotone throughout.

\textit{(i) Perturbation-heterogeneity hypothesis} (cause: multi-source concatenation). When the requested dose exceeds a single source-trajectory output (~1,000 chars), our sampler concatenates multiple late-step outputs from different cross-family source trajectories. At dose 1,500 the perturbation is ~6 concatenated outputs; at dose 3,000 it is ~12. If the dip were caused by this construction, then a homogeneous control (one source output repeated to the target length) should produce \textit{higher} destination-coherent persistence at matched dose by reducing destination scatter. We ran the control (\texttt{exp\_perturb\_O1\_homogeneous\_control}, n=50 per cell at doses 1500/2000/3000):

\begin{table}[h!]
\centering
\small
\begin{tabular}{rrrr}
\toprule
dose & heterogeneous (concatenated) & homogeneous (repeated) & delta \\
\midrule
1500 & 0.51 [0.41, 0.61] & 0.44 [0.31, 0.58] & -0.07 \\
2000 & 0.32 [0.24, 0.42] & 0.32 [0.21, 0.46] & +0.00 \\
3000 & 0.41 [0.32, 0.51] & 0.40 [0.28, 0.54] & -0.01 \\
\bottomrule
\end{tabular}
\end{table}

The heterogeneity hypothesis is not supported by the control. Destination-coherent persistence is essentially identical between heterogeneous and homogeneous perturbations at all three high doses; the dip at dose 2000 reproduces under homogeneous repetition. A separate finding from the same control: source-basin escape is meaningfully \textit{lower} under homogeneous perturbation (0.54-0.58 across doses 1500-3000) than heterogeneous (0.74-0.79). Repetition of the same content gives the model a clearer single signal to recover from, so a larger fraction of trajectories return to the original basin; heterogeneous concatenation pushes more trajectories permanently away.

\textit{(ii) Endpoint timing and cluster granularity.} We swept the K-means partition $k\in\{3,6,8,12,18,24,32\}$ and the step at which $C_+$ is recorded (step 15 vs steps 16-18, giving the model 1-3 generation steps after injection before the destination cluster is fixed). The dip magnitude $\Delta = S^{\mathrm{dst}}(2000) - \tfrac{1}{2}[S^{\mathrm{dst}}(1500)+S^{\mathrm{dst}}(3000)]$ is non-monotonic in $k$: it peaks at the canonical $k=12$ ($\Delta=-0.140$, family-cluster bootstrap 95\% CI $[-0.263,-0.030]$) and is smaller at coarser ($k=3$: $\Delta\approx 0$) and finer ($k=32$: $\Delta=-0.040$) partitions. We additionally hierarchically merged the canonical $k=12$ centroids via Ward agglomerative clustering to $k_{\mathrm{macro}}\in\{3,4,6,8\}$; the dip is preserved at $-0.085$ to $-0.090$ across all macro levels. We emphasize that this is robustness \textit{to coarsening of the canonical $k=12$ partition}, not robustness \textit{to coarse partitioning in absolute}: a direct K-means refit at $k=3$ recovers different cluster boundaries that obscure the dip ($\Delta\approx 0$). The macro-merge result therefore says the dip is not purely a $k=12$ microcluster-label artifact, but it remains conditional on the canonical microcluster geometry. Delaying $C_+$ from step 15 to step 18 cuts the dip at $k=12$ from $-0.140$ to $-0.065$: roughly half the canonical dip is attributable to comparing a \texttt{context\_tail} window dominated by raw injection text against a terminal window dominated by model-generated continuation. Transition entropy $H(C_T \mid C_+, \text{kicked})$ at dose 2000 (2.215 bits) is 0.181 bits above the neighbor average. We pre-registered a 0.25-bit support threshold for the perturbation-induced exploration hypothesis and the diagnostic \textit{fails} that threshold; the sign is directionally consistent with a small destination-scatter excess at dose 2000 but does not contribute affirmative support.

\textit{(iii) Long-horizon trajectory continuation} (\texttt{exp\_perturb\_O1\_ed50\_higher\_noclip\_extended}, n=100 per cell at doses 1500/2000/3000, append-mode continuation from step 30 to step 79 under identical generator parameters). We score the dip under two cluster bases. The \textbf{frozen-canonical basis} fits PCA-10 + K-means $k=12$ on the original 30-step experiment alone—the same partition that produced the published $\Delta=-0.140$ at step 29—and projects the extended embeddings onto those fixed centroids; this is a direct longer-horizon test of the canonical metric. The \textbf{joint basis} refits PCA-10 + K-means $k=12$ on the union of original and extended embeddings; this is a robustness check that allows the cluster geometry to adapt to the longer horizon. Family-cluster 95\% bootstrap CIs are reported on the dip contrast $\Delta = S^{\mathrm{dst}}_{2000} - \tfrac{1}{2}[S^{\mathrm{dst}}_{1500}+S^{\mathrm{dst}}_{3000}]$.

\textbf{Frozen-canonical basis} (canonical metric, longer horizon):

\begin{table}[h!]
\centering
\small
\begin{tabularx}{\textwidth}{rYYYYY}
\toprule
terminal step T & $S^{\mathrm{dst}}(1500)$ & $S^{\mathrm{dst}}(2000)$ & $S^{\mathrm{dst}}(3000)$ & dip $\Delta$ & 95\% CI on $\Delta$ \\
\midrule
29 (original) & 0.511 & 0.312 & 0.400 & $-0.143$ & $[-0.269,-0.034]$ \\
40 & 0.511 & 0.354 & 0.442 & $-0.122$ & $[-0.221,-0.026]$ \\
50 & 0.489 & 0.385 & 0.442 & $-0.080$ & $[-0.199,+0.003]$ \\
60 & 0.511 & 0.438 & 0.453 & $-0.044$ & $[-0.155,+0.043]$ \\
70 & 0.532 & 0.406 & 0.453 & $-0.086$ & $[-0.231,+0.020]$ \\
79 & 0.532 & 0.458 & 0.463 & $-0.039$ & $[-0.158,+0.068]$ \\
\bottomrule
\end{tabularx}
\end{table}

Under the frozen canonical basis, the dip drops by 73\% from step 29 to step 79 ($-0.143 \to -0.039$). At step 29 the bootstrap interval excludes zero; at step 79 it does not. The point estimate is monotone non-increasing in $T$ to step 60 and roughly stable thereafter. Importantly, the absolute persistence values stay close to the canonical 0.31-0.51 range, so this is the same metric, just measured later.

\textbf{Joint basis} (allow the partition to adapt to the longer horizon):

\begin{table}[h!]
\centering
\small
\begin{tabularx}{\textwidth}{rYYYYY}
\toprule
terminal step T & $S^{\mathrm{dst}}(1500)$ & $S^{\mathrm{dst}}(2000)$ & $S^{\mathrm{dst}}(3000)$ & dip $\Delta$ & 95\% CI on $\Delta$ \\
\midrule
29 & 0.191 & 0.152 & 0.222 & $-0.055$ & $[-0.179,+0.055]$ \\
40 & 0.202 & 0.130 & 0.200 & $-0.071$ & $[-0.101,-0.035]$ \\
50 & 0.191 & 0.141 & 0.211 & $-0.060$ & $[-0.109,-0.011]$ \\
60 & 0.223 & 0.163 & 0.211 & $-0.054$ & $[-0.087,-0.017]$ \\
70 & 0.160 & 0.152 & 0.167 & $-0.011$ & $[-0.056,+0.029]$ \\
79 & 0.149 & 0.152 & 0.156 & $0.000$ & $[-0.066,+0.044]$ \\
\bottomrule
\end{tabularx}
\end{table}

The joint basis dilutes absolute persistence to ~0.15-0.22 because the partition adapts to the broader cluster geometry of original-plus-extended embeddings. This is a different estimand from the canonical metric—useful as a sensitivity check, not as a direct continuation of the published $\Delta=-0.140$ number. The qualitative time evolution under both bases is the same: monotone closure of the dip toward zero, with the step-79 contrast statistically indistinguishable from zero.

\begin{savenotes}
\begin{figure}[h!]
\centering
\includegraphics[width=0.95\linewidth]{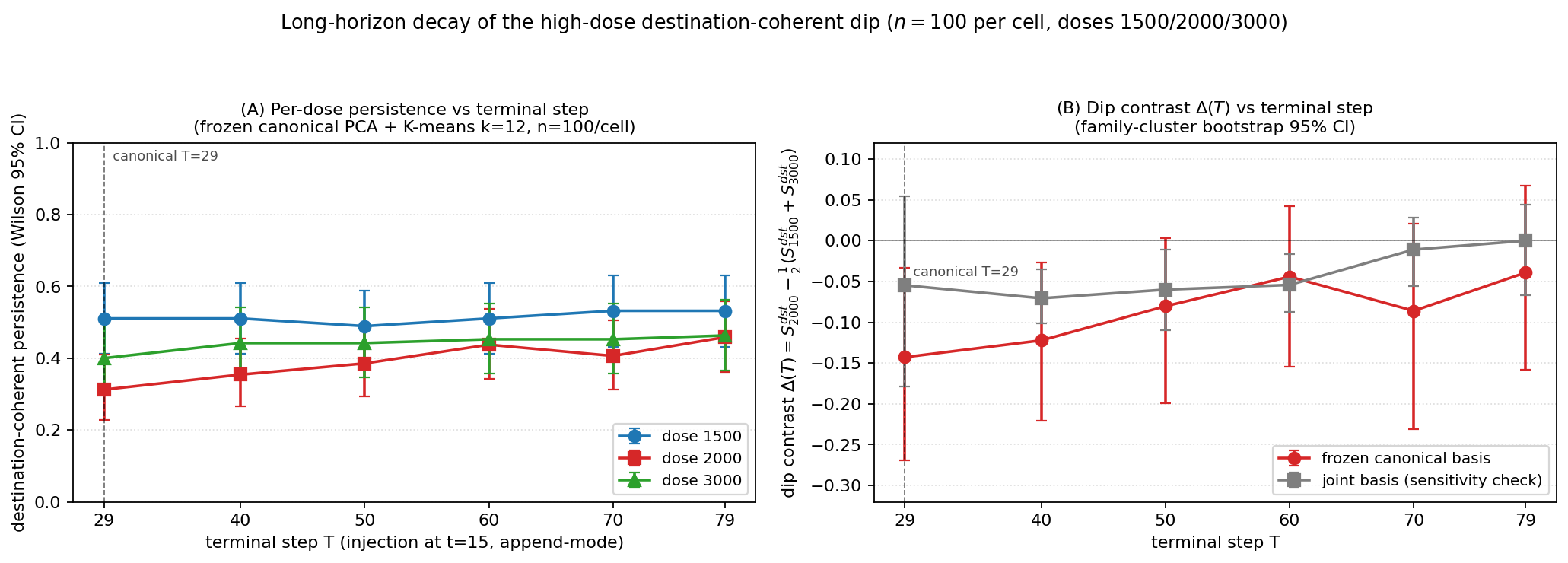}
\caption{\textbf{Long-horizon decay of the high-dose destination-coherent dip.} Two panels, $n=100$ per cell at doses 1500/2000/3000. (A) Per-dose destination-coherent persistence vs terminal step T under the frozen canonical PCA + K-means $k=12$ basis (fit on the original 30-step experiment, applied to extended trajectories). Wilson 95\% CIs at each terminal step. Dose 2000 (red squares) climbs from 0.31 at the canonical $T=29$ to 0.46 at $T=79$; doses 1500 (blue circles) and 3000 (green triangles) stay essentially flat. (B) Dip contrast $\Delta(T) = S^{\mathrm{dst}}_{2000}(T) - \tfrac{1}{2}[S^{\mathrm{dst}}_{1500}(T)+S^{\mathrm{dst}}_{3000}(T)]$ vs T under both the frozen canonical basis (red) and the joint basis (gray, sensitivity check), with family-cluster bootstrap 95\% CIs. The frozen-basis $\Delta$ drops 73\% from $-0.143$ at $T=29$ to $-0.039$ at $T=79$, with the step-79 interval straddling zero. Reading: the high-dose destination-coherent dip at the canonical $T=29$ horizon is a finite-horizon, endpoint-definition-sensitive feature; under longer trajectory continuation with the same canonical cluster definition, it substantially weakens and becomes statistically indistinguishable from zero. Source: \texttt{data/aggregated/dip\_mechanism\_B/dip\_vs\_terminal\_step.png}.\\[2pt]{\footnotesize\itshape Fig 6 shows the long-horizon decay of the high-dose destination-coherent dip. The two panels share data from \texttt{exp\textbackslash\{\}\_perturb\textbackslash\{\}\_O1\textbackslash\{\}\_ed50\textbackslash\{\}\_higher\textbackslash\{\}\_noclip\textbackslash\{\}\_extended}, which extended the original n=100 trajectories at doses 1500/2000/3000 from step 30 to step 79 in append mode under identical generator parameters. Panel A reports per-dose destination-coherent persistence at terminal steps T ∈ \{29, 40, 50, 60, 70, 79\} under the \textit{frozen canonical basis}: PCA-10 + K-means k=12 fit on the original 30-step experiment alone, with extended embeddings projected onto those fixed centroids. This is a direct longer-horizon test of the published metric; the absolute persistence values at T=29 reproduce the canonical 0.31-0.51 range. Dose 2000 (red squares) climbs from 0.312 [0.227, 0.408] at T=29 to 0.458 [0.362, 0.557] at T=79; doses 1500 and 3000 stay roughly flat. Panel B reports the dip contrast $\Delta(T) = S^{\mathrm{dst}}_{2000}(T) - \tfrac{1}{2}[S^{\mathrm{dst}}_{1500}(T) + S^{\mathrm{dst}}_{3000}(T)]$ under both the frozen canonical basis (red) and the joint basis (gray, which refits PCA + K-means across original and extended embeddings; sensitivity check, different estimand). The frozen-basis $\Delta$ moves from -0.143 [-0.269, -0.034] at T=29 (excludes zero) to -0.039 [-0.158, +0.068] at T=79 (straddles zero), a 73\% reduction. The joint-basis $\Delta$ shows the same qualitative closure on a smaller absolute scale. Falsifying outcomes: dose 2000 staying low while 1500 and 3000 stayed high through T=79 (would have supported a stable-asymmetry reading), or $\Delta$ widening rather than closing (would have supported a structural attractor split). Caveats: n=100 yields a CI half-width near 0.11 on $\Delta$, so the population-level dip at T=79 could be a small remaining negative effect or a small positive effect; the analysis was post-hoc, designed after observing the canonical T=29 dip.}}
\end{figure}
\end{savenotes}

\textbf{The mechanism is not "dose 2000 catches up."} Under both bases, doses 1500 and 3000 either decay (joint basis) or oscillate (frozen basis) toward dose 2000's level rather than dose 2000 rising to theirs.

\textbf{Combined verdict.} The dip is real \textit{as a feature of the canonical 30-step measurement window with K-means $k=12$ and $C_+$ taken from raw-injection-dominated \texttt{context\_tail}}, with a family-cluster 95\% bootstrap CI of $[-0.263,-0.030]$ at step 29 that excludes zero. Three pieces of evidence push back against interpreting it as a stable structural attractor asymmetry. (a) Approximately half the canonical magnitude comes from endpoint-timing: delaying $C_+$ by three generation steps cuts $|\Delta|$ from $0.140$ to $0.065$. (b) Under the frozen canonical basis, extending trajectories from step 29 to step 79 reduces the dip by 73\% (to $-0.039$, CI $[-0.158,+0.068]$) without changing any cluster definition or analysis parameter. (c) The retained source-basin escape endpoint is monotone in dose, unaffected by all of these analysis choices, and saturates at 75-80\% by 1,500 tokens. We therefore retain the dose-1500 destination-coherent half-effect \textit{crossing} as a property of the canonical 30-step measurement protocol while relinquishing any mechanism-level claim about a structural high-dose attractor split. The simpler retained source-basin escape endpoint is the safer load-bearing claim for downstream safety uses.

\textbf{Caveats on the long-horizon analysis.} (a) n=100 per cell still yields wide CIs on individual contrasts; the step-79 dip estimate of $-0.039$ has a CI half-width of $\sim 0.11$, so the population-level dip could be anywhere from a small remaining negative effect to a small positive effect. We claim only that the canonical $-0.143$ asymmetry has been substantially reduced under longer continuation, not that the population dip is exactly zero. (b) The frozen-canonical basis projects extended embeddings into a PCA-10 space fit on the original 30-step trajectories; the late-window extended trajectories may drift toward regions of embedding space underrepresented in the original fit. (c) The continuation analysis was not pre-registered; it was designed and run after observing the canonical dip and reading reviewer feedback. We treat all four mechanism analyses as \textit{post-hoc decision rules locked for future replication}, not as confirmatory tests.

\textbf{Joint table.}

\begin{table}[h!]
\centering
\small
\begin{tabular}{rrrr}
\toprule
dose & kicked & destination-coherent & source-basin escape \\
\midrule
200 & 0.41 & 0.18 & 0.31 \\
400 & 0.68 & 0.29 & 0.53 \\
600 & 0.85 & 0.37 & 0.69 \\
1000 & 0.90 & 0.41 & 0.69 \\
1500 & 0.94 & 0.51 & 0.79 \\
2000 & 0.96 & 0.32 & 0.74 \\
3000 & 0.96 & 0.41 & 0.78 \\
\bottomrule
\end{tabular}
\end{table}

Under retained source-basin escape the curve is monotone and saturating: it crosses 50\% between dose 300 (0.42) and dose 400 (0.53), and saturates at ~75-80\% by dose 1500. The empirical half-effect dose is therefore approximately \textbf{400 tokens} for source-basin escape. Under destination-coherent persistence, the curve first reaches 50\% near \textbf{1,500 tokens} but is non-monotonic at higher doses, so we describe its 50\%-crossing rather than calling it a clean ED50.

\textbf{Which endpoint is more relevant.} The two endpoints answer different questions. For safety analyses where the concern is \textit{any durable departure from the original behavior} (agent hijacking, content drift, persistent adversarial steering), retained source-basin escape is the more directly relevant measure. For \textit{targeted steering or basin capture} (where the goal is committing to a specific alternative basin), destination-coherent persistence is the appropriate stricter criterion. We therefore lead with source-basin escape as the primary safety endpoint and report destination-coherent persistence as the stricter complement.

\textbf{Disclosure on endpoint-promotion order.} Destination-coherent persistence was the pre-specified definition (§3.1.2 / §4.13). The source-basin escape endpoint was added after we saw the upper-tail non-monotonicity and observed that destination-coherent persistence does not capture all forms of durable redirection. We report both, not one in place of the other. The 1,500-token destination-coherent crossing is the stricter number under the original definition; the 400-token source-basin escape number is the looser, safety-oriented complement. Destination-coherent persistence is a strict subset of source-basin escape, so the two numbers describe the same data — not a re-fit.

Under bounded memory both endpoints plateau well below 50\% across the tested 5-400 token range (destination-coherent ~16\%, source-basin escape ~36\% at dose 400), confirming that the memory-policy effect operates on both endpoints.

\textbf{Scope and caveats.} The unbounded-memory protocol requires that the perturbation fit within the model's native context window (~128K tokens for gpt-4o-mini); doses up to 3,000 tokens are well within this limit. The non-monotonic upper tail should be tested with larger N before being treated as a load-bearing finding. The result is observable-conditional: under the strict single-step \texttt{output} observable, persistence stays below 50\% across all doses tested (peak 0.36 at dose 1500); the 50\%-crossing is observed only under the multi-step window observables (\texttt{context\_tail}, \texttt{rolling\_k3}).

The sparse dose-response pilot remains useful for the qualitative contrast among perturbation contents. In that pilot, D1 neutral switching was already high at 5 tokens and stayed high across the grid, O1 neutral switching stayed near 22-26\% from 20 to 400 tokens, and O1 adversarial switching rose from 26\% to roughly 50\%. The defensible headline numbers for the canonical bounded-memory loop remain \textbf{raw $\mathrm{ED50} \approx 40$ tokens} and \textbf{bounded-memory destination-coherent persistence plateau ≈ 16\% (\texttt{context\_tail}, $k=12$)}. The full-history headline numbers are \textbf{retained source-basin escape half-effect dose ≈ 400 tokens} (saturating ~75-80\% by dose 1,500; the safety-relevant displacement endpoint) and \textbf{destination-coherent persistence point-estimate first crossing ≈ 1,500 tokens} (the stricter endpoint; ED50-dose CI not estimated due to upper-tail non-monotonicity).

\subsection{Replace-mode "fragility" is primarily a memory-policy effect}

\textbf{Tautology warning before the headline numbers.} The default replace-mode perturbation is a state reset, not a clean test of model resistance to redirection. In replace-mode, the context-update rule is \(X_{t+1}=\operatorname{clip}(Y_t)\): only the latest output becomes the next state. The original ("overwrite") perturbation protocol replaces \(Y_t\) with the injected text, so the replace update yields \(X_{t+1}=\operatorname{clip}(\text{perturbation\_text})\). Switching at this step is \textbf{tautological at the state-write level}: the next state simply \textit{is} the perturbation. After that, the trajectory is just the operator \(f\) iterated from the perturbation as a fresh initial condition. We report the naive overwrite numbers below to expose this trap, then correct it with the insert-mode counterfactual.

The original perturbation pilots made O2 and O3 appear nearly maximally fragile. Under this state-reset overwrite protocol, all non-control perturbations produce 94-100\% final-cluster disagreement with the paired control trajectory:

\begin{table}[h!]
\centering
\small
\begin{tabular}{lrrr}
\toprule
regime & neutral & lorem & adversarial \\
\midrule
O2, paraphrase replace & 100\% [93-100] & 100\% [93-100] & 94\% [84-98] \\
O3, summarize-negate replace & 100\% [93-100] & 100\% [93-100] & 96\% [86-99] \\
\bottomrule
\end{tabular}
\end{table}

Read alone, these numbers suggest that replace-mode regimes have almost zero injected-token barriers. The overwrite-versus-insert probe shows that this is mostly a memory-policy effect; the high overwrite rates are the state-write tautology in action, not evidence of weak attractor structure. We re-ran O2 and O3 with the same adversarial doses under two intervention modes that decouple state-write from model exposure:

\begin{itemize}
  \item \textbf{State-reset overwrite (= state replacement):} the original protocol. The injection replaces step 15's output entirely; under replace-mode update this means \(X_{16}=\operatorname{clip}(\text{perturbation\_text})\). The post-injection trajectory is operator iteration from a fresh initial condition.
  \item \textbf{Insert (= visible-only):} the injected text is prepended to the context for step 15, but the model's own generated output is preserved as the state. The injected text influences a single API call and then disappears from the state by construction. Only model-mediated effects can persist.
\end{itemize}

The O2 paraphrase-replace results were:

\begin{table}[h!]
\centering
\small
\begin{tabular}{lrl}
\toprule
condition & switch rate & 95\% Wilson CI \\
\midrule
control & 0.00 & [0.00, 0.07] \\
\texttt{adversarial\_dose80}, overwrite & 0.92 & [0.81, 0.97] \\
\texttt{adversarial\_insert\_dose80} & 0.32 & [0.21, 0.46] \\
\texttt{adversarial\_dose200}, overwrite & 0.98 & [0.90, 1.00] \\
\texttt{adversarial\_insert\_dose200} & 0.18 & [0.10, 0.31] \\
\bottomrule
\end{tabular}
\end{table}

The O3 summarize-negate-replace results were:

\begin{table}[h!]
\centering
\small
\begin{tabular}{lrl}
\toprule
condition & switch rate & 95\% Wilson CI \\
\midrule
control & 0.00 & [0.00, 0.07] \\
\texttt{adversarial\_dose80}, overwrite & 0.90 & [0.79, 0.96] \\
\texttt{adversarial\_insert\_dose80} & 0.18 & [0.10, 0.31] \\
\texttt{adversarial\_dose200}, overwrite & 0.92 & [0.81, 0.97] \\
\texttt{adversarial\_insert\_dose200} & 0.12 & [0.06, 0.24] \\
\bottomrule
\end{tabular}
\end{table}

The overwrite-minus-insert gap is 60-80 percentage points across both regimes and both doses:

\begin{table}[h!]
\centering
\small
\begin{tabularx}{\textwidth}{lYYYY}
\toprule
regime & dose & overwrite & insert & overwrite minus insert \\
\midrule
O2 & 80 & 0.92 & 0.32 & +0.60 \\
O2 & 200 & 0.98 & 0.18 & +0.80 \\
O3 & 80 & 0.90 & 0.18 & +0.72 \\
O3 & 200 & 0.92 & 0.12 & +0.80 \\
\bottomrule
\end{tabularx}
\end{table}

Thus most apparent replace-mode perturbation transparency comes from the update rule discarding prior state. Once the perturbation no longer overwrites the loop state, switching falls to 12-32\%, below or near the natural stochastic-divergence floor measured for O1. The original O2/O3 result remains an important systems finding, but it should be described as \textbf{overwrite-induced state replacement}, not as a discovered low behavioral barrier comparable to the O1 dose-response measurement.

This has a direct engineering analogue. Any architecture that periodically replaces accumulated context with a generated summary, scratchpad, task state, or memory record can fail by promoting untrusted text into the replacement state. In that case, the system has not merely been persuaded by injected text; its previous state has been removed by the memory policy.

\subsubsection{F3 cross-loop insert validation: insert is regime-conditional}

F3 tests whether the insert-mode estimand from §5.2 is loop-independent. It is not. In append-mode O1, the overwrite-insert gap is 14-34 pp, much smaller than the 60-80 pp gap in replace-mode O2/O3; and at both tested doses, insert-only switching follows \(O1 \ge O2 \ge O3\). Insert therefore removes the replace-mode state-write tautology, but it does not remove regime dependence: the update rule still controls how much of the one-generation perturbation imprint propagates.

The §5.2 overwrite-vs-insert decomposition had previously been run only in replace-mode O2/O3. We therefore ran the same protocol in append-mode O1: 5 families × 5 ICs × 2 runs × 5 conditions = 250 trajectories, gpt-4o-mini, matched to the O2/O3 design (\texttt{configs/perturbation/O1\_overwrite\_vs\_insert.yaml}). Final-cluster switching against the paired control:

\begin{table}[h!]
\centering
\small
\begin{tabular}{lrl}
\toprule
condition & switch rate & 95\% Wilson CI \\
\midrule
control & 0.00 & [0.00, 0.07] \\
\texttt{adversarial\_dose80}, overwrite & 0.54 & [0.40, 0.67] \\
\texttt{adversarial\_insert\_dose80} & 0.40 & [0.28, 0.54] \\
\texttt{adversarial\_dose200}, overwrite & 0.70 & [0.56, 0.81] \\
\texttt{adversarial\_insert\_dose200} & 0.36 & [0.24, 0.50] \\
\bottomrule
\end{tabular}
\end{table}

\begin{savenotes}
\begin{figure}[h!]
\centering
\includegraphics[width=0.95\linewidth]{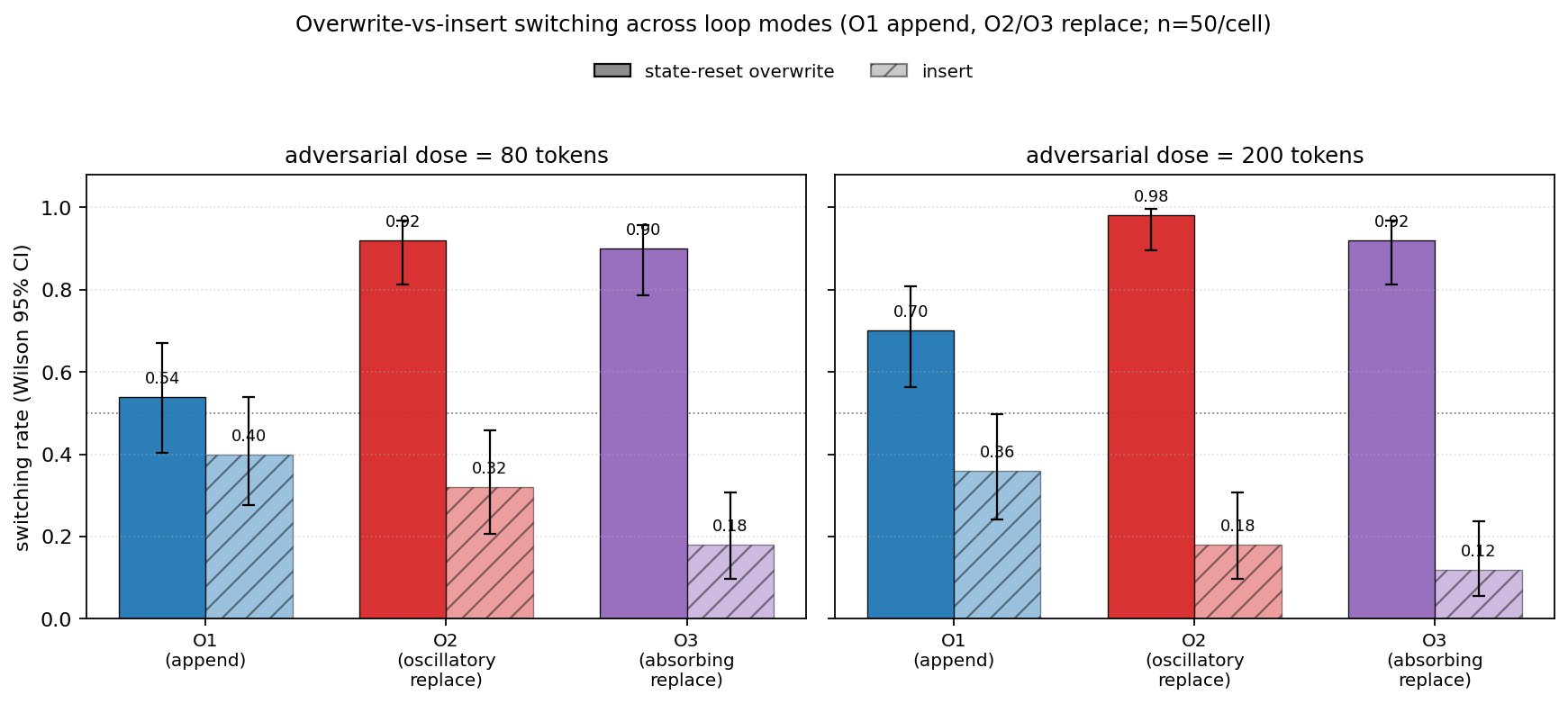}
\caption{\textbf{Cross-loop overwrite-versus-insert switching (round-32, F3).} Matched adversarial doses 80 and 200 across O1 append, O2 oscillatory replace, and O3 absorbing replace; n=50 trajectories per cell. Solid bars show state-reset overwrite; hatched bars show insert; error bars are Wilson 95\% CIs. \textbf{Two takeaways.} (a) The overwrite-insert gap is small in O1 append (14-34 pp) and large in O2/O3 replace (60-80 pp), consistent with the state-write tautology being confined to replace-mode update rules. (b) Insert-mode rates follow \(O1 \ge O2 \ge O3\) at both doses, showing that insert switching is regime-conditional rather than a regime-invariant model-behavior constant. Source: \texttt{data/aggregated/overwrite\_vs\_insert\_cross\_regime/cross\_regime.png}.\\[2pt]{\footnotesize\itshape Fig 7 reports overwrite-versus-insert switching across three loop modes at two adversarial doses, drawn from three matched-design experiments (exp\_perturb\_O1\_overwrite\_vs\_insert, exp\_perturb\_O2\_overwrite\_vs\_insert, exp\_perturb\_O3\_overwrite\_vs\_insert), each with 5 prompt families × 5 ICs × 2 runs = 50 trajectories per condition. The state-reset overwrite protocol overrides the model output at step 15 with the perturbation text; under replace-mode update this sets the entire next state to the perturbation, while under append-mode update it appends the perturbation to the existing transcript (no state-write tautology). The insert protocol prepends the perturbation text to the context for one API call, but preserves the model's natural Y\_15 as the next-state contribution; the perturbation does not persist into the state by construction. At dose 80, switching is 0.54 [0.40, 0.67] (O1 overwrite), 0.40 [0.28, 0.54] (O1 insert), 0.92 [0.81, 0.97] (O2 overwrite), 0.32 [0.21, 0.46] (O2 insert), 0.90 [0.79, 0.96] (O3 overwrite), and 0.18 [0.10, 0.31] (O3 insert). At dose 200: 0.70 [0.56, 0.81] (O1 ow), 0.36 [0.24, 0.50] (O1 ins), 0.98 [0.90, 1.00] (O2 ow), 0.18 [0.10, 0.31] (O2 ins), 0.92 [0.81, 0.97] (O3 ow), 0.12 [0.06, 0.24] (O3 ins). The overwrite-minus-insert gap is +0.14 / +0.34 (O1, by dose), +0.60 / +0.80 (O2), and +0.72 / +0.80 (O3) - small in append, large in replace, consistent with the state-write tautology being a property of the replace-mode update rule (see §3.1, §5.2). The insert-mode ordering O1 > O2 > O3 at matched dose (non-overlapping CIs at both doses for O1 vs O3) reflects the dynamics of the post-injection update: append preserves Y\_15 in a growing transcript so single-shot perturbation imprints carry forward; replace iterates f on its own output and the operator's natural attractor (oscillatory for O2, absorbing for O3) erases the imprint within a few steps, faster for absorbing dynamics. This pattern would have been falsified if the O1 overwrite-insert gap matched the replace-mode 60-80 pp gap (which would have meant the gap is not an update-rule artifact), or if insert rates were equal across regimes (which would have supported a strict regime-invariant model-behavior reading of insert).}}
\end{figure}
\end{savenotes}

\textbf{Observation 1: the overwrite-vs-insert gap is update-rule dependent.}

\begin{table}[h!]
\centering
\small
\begin{tabularx}{\textwidth}{lYYYY}
\toprule
regime & dose & overwrite & insert & overwrite minus insert \\
\midrule
O1 (append) & 80 & 0.54 & 0.40 & +0.14 \\
O1 (append) & 200 & 0.70 & 0.36 & +0.34 \\
O2 (replace) & 80 & 0.92 & 0.32 & +0.60 \\
O2 (replace) & 200 & 0.98 & 0.18 & +0.80 \\
O3 (replace) & 80 & 0.90 & 0.18 & +0.72 \\
O3 (replace) & 200 & 0.92 & 0.12 & +0.80 \\
\bottomrule
\end{tabularx}
\end{table}

In append-mode, the default ("overwrite") protocol is not a state reset. It appends the perturbation tokens to the existing transcript rather than promoting them to the entire next state, so the overwrite rate already reflects model-mediated effects on later generations. The remaining 14-34 pp append-mode gap is plausibly due to persistence: overwrite leaves the perturbation tokens in the appended history, whereas insert exposes the model for one API call and then removes them. By contrast, in replace-mode O2/O3, overwrite makes the perturbation the entire next state; the 60-80 pp replace-mode gap is therefore dominated by the state-write tautology.

\textbf{Observation 2: insert-mode switching is regime-conditional, not a regime-invariant model-behavior constant.}

\begin{table}[h!]
\centering
\small
\begin{tabular}{lrr}
\toprule
regime & insert d=80 & insert d=200 \\
\midrule
O1 (append) & 0.40 [0.28, 0.54] & 0.36 [0.24, 0.50] \\
O2 (oscillatory replace) & 0.32 [0.21, 0.46] & 0.18 [0.10, 0.31] \\
O3 (absorbing replace) & 0.18 [0.10, 0.31] & 0.12 [0.06, 0.24] \\
\bottomrule
\end{tabular}
\end{table}

At both tested doses, insert-only switching orders \(O1 \ge O2 \ge O3\), and the Wilson CIs of O1 append and O3 absorbing replace do not overlap. The interpretation is propagation, not pure exposure: under insert protocol, the perturbation imprints on \(Y_{t_{\mathrm{inj}}}\) for one generation, and the loop update rule determines how much of that imprint survives. Append-mode preserves \(Y_{t_{\mathrm{inj}}}\) in the growing transcript, so any cluster-relevant signal can carry forward. Replace-mode iterates \(f\) on its own output, and its attractor structure erases the imprint within a few post-injection steps; absorbing O3 erases faster than oscillatory O2, matching the observed \(O3 < O2 < O1\) ordering.

\textbf{Scope.} §5.2.1 is a 50-trajectory-per-condition cross-validation cell, not a publication-scale rerun. CIs are wide, and small per-cell differences are not statistically resolved; for example, O1 versus O2 at dose 80 has partly overlapping CIs. The purpose of §5.2.1 is methodological: to validate the insert-mode protocol across loop modes, not to refine the §5.6 dose-response curves. The headline append-mode results are still based on the publication-scale dense rerun (\texttt{exp\_perturb\_O1\_ed50\_dense}) and are unchanged: \(\mathrm{ED50}_{\mathrm{raw}}\approx 40\) tokens, plateau ≈ 0.67, and persistent escape not reached up to 400 tokens.

The cross-loop check refines the §5.2 framing rather than contradicting it. The 12-32\% replace-mode insert numbers are correct for replace-mode, but they should not be read as a regime-invariant measurement of pure model behavior. F1 in §8.6 - the visibility × hard-write factorial across all loop modes - is the principled way to fully disentangle the remaining model-versus-update-rule interaction. F3's 250-trajectory check is the near-term sanity check that motivates F1 as future work.

\subsection*{Phase B, regime establishment with leakage-aware analysis}

\subsection{Publication-scale runs preserve regime ordering}

REPORT5 ran the four diagnostic regimes at publication scale. Operator regimes O1, O2, and O3 used 15 prompt families × 30 ICs × 3 runs, for 1,350 trajectories per regime. Dialog regime D1 used 5 dialog-suitable families × 30 ICs × 3 runs, for 450 trajectories. All four were 40 steps long.

Basin predictability is measured by 5-fold multinomial logistic regression on PCA-10, predicting the trajectory's late-window K-means cluster at $k=12$ from the embedding at step $k$. The late-window cluster is the majority cluster over $t \geq \lceil 0.7T \rceil$. For $T = 40$, this is a 12-step late window. For D1 with role-restricted observables, the latest predictor step is 26, the last agent turn before the late window opens at step 28.

At first exposure we report both the original stratified cross-validation accuracy and the leakage-aware GroupKFold accuracy at $k=10$, where entire prompt families are held out across folds.

\begin{table}[h!]
\centering
\small
\begin{tabularx}{\textwidth}{lYYYYYYY}
\toprule
experiment & regime & acc(k=5) & acc(k=10), stratified & acc(k=10), group-aware & leakage delta & acc(k=20) & acc(final) \\
\midrule
\texttt{exp\_pub\_O1\_continue} & O1, contractive & 0.77 & 0.803 & 0.732 & +0.071 & 0.81 & 0.85 \\
\texttt{exp\_pub\_O2\_paraphrase\_replace} & O2, oscillatory replace & 0.90 & 0.896 & 0.596 & +0.301 & 0.91 & 0.91 \\
\texttt{exp\_pub\_O3\_summarize\_negate\_replace} & O3, absorbing replace & 0.92 & 0.912 & 0.629 & +0.283 & 0.92 & 0.93 \\
\texttt{exp\_pub\_D1\_dialog\_curious\_helpful\_v2} & D1, dialogue-state multi-basin & n/a & 0.604 & 0.336 & +0.269 & 0.69 & 0.77 \\
\bottomrule
\end{tabularx}
\end{table}

\begin{savenotes}
\begin{figure}[h!]
\centering
\includegraphics[width=0.95\linewidth]{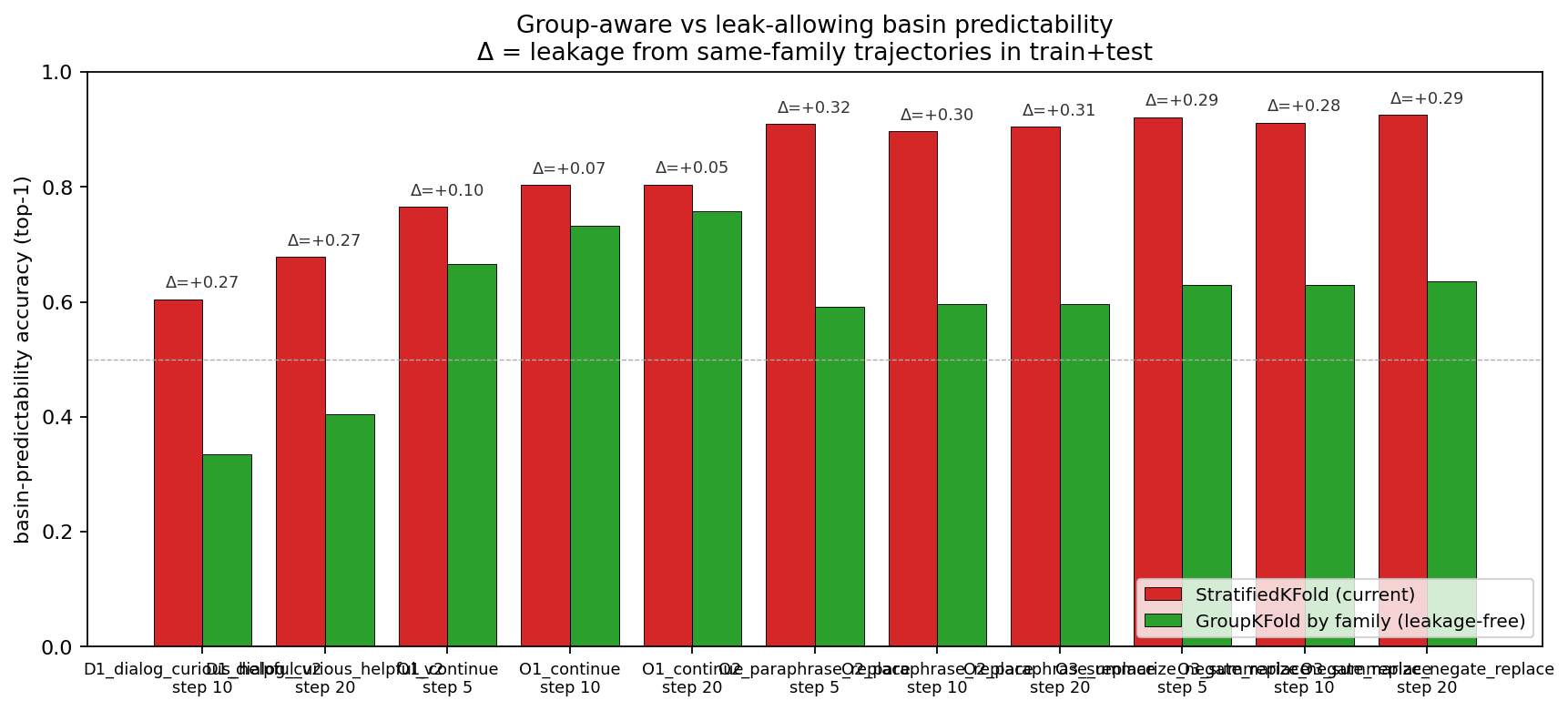}
\caption{\textbf{Leakage-aware basin predictability.} Group-aware basin-predictability with prompt families held out across folds. O1 remains the strongest leakage-free predictability result, while O2, O3, and D1 drop substantially under family-held-out validation. Source: \texttt{data/aggregated/group\_aware\_basin\_pred.png}.\\[2pt]{\footnotesize\itshape Fig 8 shows a leakage-aware basin-predictability audit comparing red bars (stratified k-fold cross-validation accuracy, the old method that allows family-leakage) against green bars (GroupKFold-by-prompt-family accuracy, the leakage-resistant method that holds out entire prompt families); the dashed horizontal line marks chance level. Labels above each red and green pair report the stratified-minus-group-aware gap, so positive values measure how much the old estimate was inflated by same-family trajectory fingerprints rather than genuinely cross-family basin information. At k=10, O1 continue (O1, the append-mode continuation regime) drops only from 0.803 to a post-correction value of 0.732, a +7 percentage-point gap, whereas O2 paraphrase replace (O2, paraphrase-and-replace-state mode) drops from 0.896 to 0.596, a +30 percentage-point gap, O3 summarize-negate (O3, summarize-then-negate replace mode) drops from 0.912 to 0.629, a +28 percentage-point gap, and D1 dialog (D1, the role-structured dialog regime) drops from 0.604 to 0.336, a +27 percentage-point gap. A small gap indicates that stratified accuracy was load-bearing across prompt families, while a large gap indicates that the stratified score mostly reflected family-fingerprint leakage. Under the corrected estimator, the C1 attractor criterion (group-aware accuracy at k=10 at least 0.70) is satisfied only by O1 continue, with 0.732 after correction; the other regimes fall below threshold despite their high leaked red-bar scores. This result would have been falsified if O1 had collapsed below 0.70 under GroupKFold, or if O2, O3, or D1 had retained similarly high group-aware accuracy after holding out prompt families.}}
\end{figure}
\end{savenotes}

The stratified values reproduce the original regime ordering: O2 and O3 lock in very early, O1 is slower but still strongly predictable, and D1 remains the least predictable at early steps. The group-aware analysis changes the interpretation. O1 loses only 7 percentage points when prompt families are held out, while O2, O3, and D1 lose 27-30 percentage points. Thus O1's basin predictability is the most cross-family robust result. For O2, O3, and D1, a substantial part of the original predictability is a family or style fingerprint.

The qualitative regime separation survives. O3 and O2 remain high-recurrence replace-mode regimes, O1 remains a cross-family contractive append regime, and D1 remains a slower, more family-sensitive dialog regime. The main correction is evidential: stratified accuracies should be read as upper bounds, and the leakage-aware columns are the relevant values for cross-family generalization.

\begin{savenotes}
\begin{figure}[h!]
\centering
\includegraphics[width=0.95\linewidth]{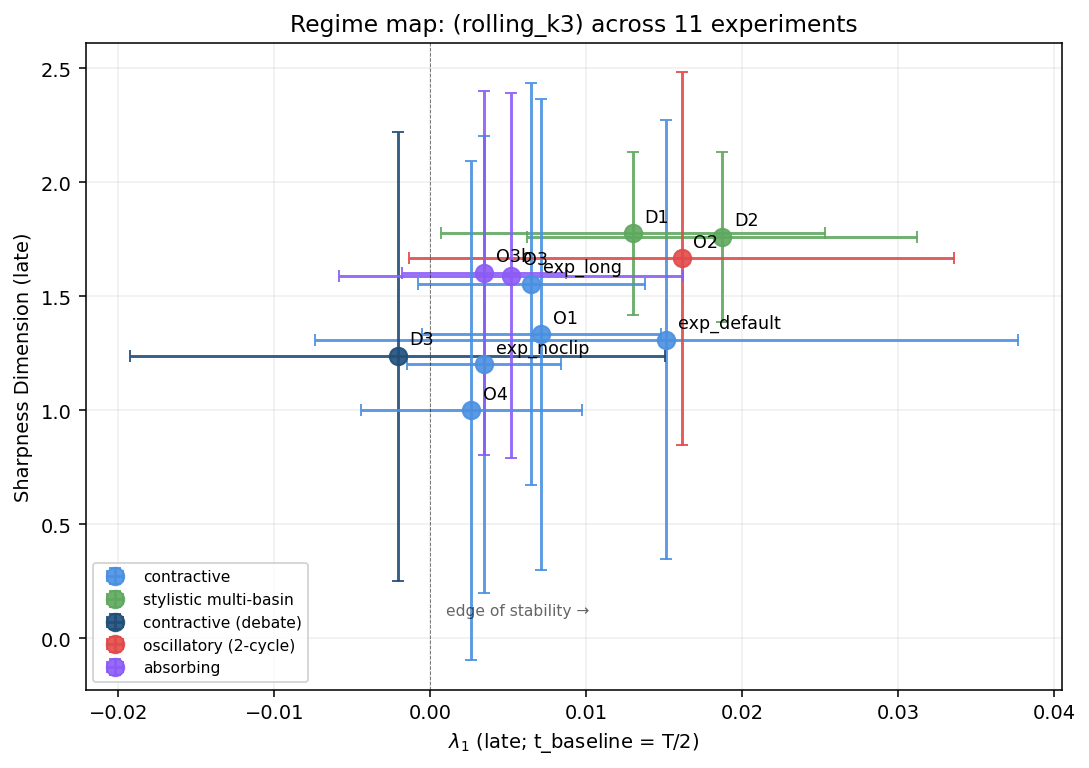}
\caption{\textbf{Cross-experiment dynamics map.} Regime-level map in late-window $\lambda_1$ versus sharpness-dimension space, showing broad separation of replace, append, and dialog regimes. The plot is diagnostic rather than endpoint-defining. Source: \texttt{data/aggregated/dynamics\_plots/regime\_map\_rolling\_k3.png}.\\[2pt]{\footnotesize\itshape Fig 9 plots the late-window Lyapunov-like exponent (an ensemble-spread growth rate, not a true Jacobian Lyapunov exponent) against sharpness dimension (a Tuci-style fractional dimension on the ordered Lyapunov spectrum) for 11 recursive-loop experiments with within-run uncertainty shown by error bars. Legend labels are translated as follows: contractive denotes O1 (append-mode continuation); oscillatory two-cycle denotes O2 (paraphrase replace-mode); absorbing denotes O3 (summarize-negate replace-mode); stylistic multi-basin denotes D1 (role-structured dialog) and D2 (drill-down dialog); contractive debate denotes a pilot debate-style contractive variant. The map is intended as a qualitative diagnostic plane, not as an endpoint-defining classifier. Its main signal is that the replace regimes, O2 paraphrase replace-mode and O3 summarize-negate replace-mode, cluster near zero growth and comparatively low sharpness, whereas append and dialog regimes, especially O1 append-mode continuation and D1 role-structured dialog, sit higher in sharpness and farther to the positive-growth side. However, O1 append-mode continuation and D1 role-structured dialog overlap substantially in this bulk geometry, so the plane by itself cannot separate continuation from dialog dynamics. This motivates the perturbation-based evidence in sections 5.4 to 5.7, where group-aware accuracy at k=10 is 0.732 for O1, passing criterion C1, compared with 0.629 for O3, 0.596 for O2, and 0.336 for D1. The interpretation would have been falsified if replace-mode runs occupied the high-sharpness, positive-growth region, if append and dialog regimes were cleanly separable here without perturbations, or if error bars erased the coarse low-sharpness versus high-sharpness contrast.}}
\end{figure}
\end{savenotes}

\subsection{Perturbation pilots separate append from replace}

The cross-regime perturbation pilots used 5 families × 5 ICs × 2 runs × 30 steps, for $n=50$ trajectories per condition, except D2 where $n=25$. Switching is final-step K-means cluster disagreement with the paired control trajectory.

\begin{table}[h!]
\centering
\small
\begin{tabularx}{\textwidth}{lYYYY}
\toprule
regime & control & neutral & lorem & adversarial \\
\midrule
O1, contractive append & 0\% [0-7] & 24\% [14-37] & 18\% [10-31] & 54\% [40-67] \\
O2, paraphrase replace & 0\% [0-7] & 100\% [93-100] & 100\% [93-100] & 94\% [84-98] \\
O3, summarize-negate replace & 0\% [0-7] & 100\% [93-100] & 100\% [93-100] & 96\% [86-99] \\
D1, dialogue-state dialog & 0\% [0-7] & 76\% [62-86] & 54\% [40-67] & 60\% [46-73] \\
D2, drill-down dialog & 0\% [0-13] & n/a & n/a & 64\% [44-80] \\
\bottomrule
\end{tabularx}
\end{table}

\begin{savenotes}
\begin{figure}[h!]
\centering
\includegraphics[width=0.95\linewidth]{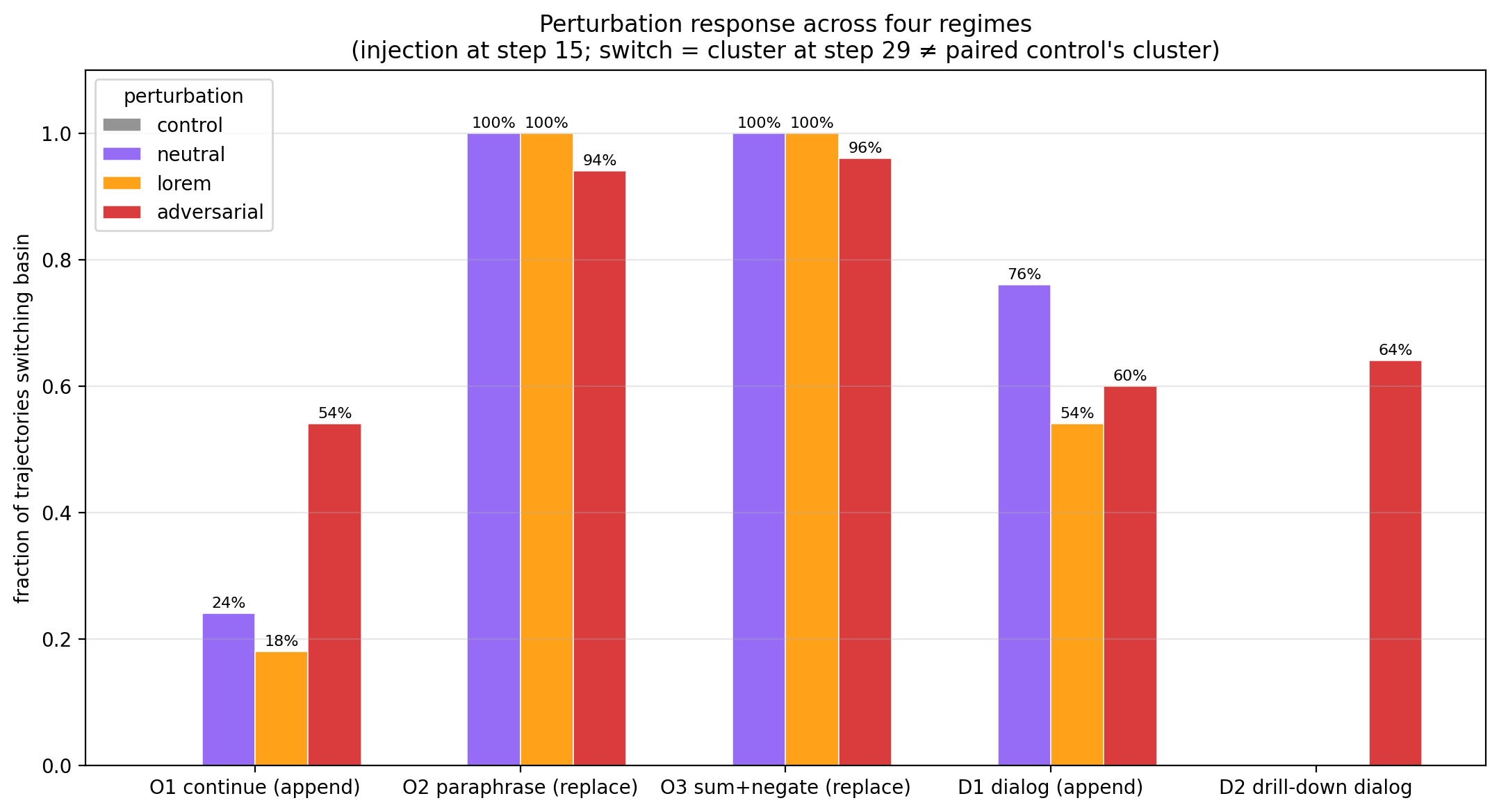}
\caption{\textbf{Cross-regime perturbation switching.} Final-cluster switching rates across append, replace, and dialog perturbation pilots. Replace-mode O2/O3 saturation should be read as overwrite-protocol sensitivity, not as a clean injected-token barrier. Source: \texttt{data/aggregated/perturbation\_cross\_regime/cross\_switching\_rates.png}.\\[2pt]{\footnotesize\itshape Fig 10 reports final-cluster switching rates with n = 50 trajectories per cell, comparing control (no injection, paired baseline), neutral (topic-irrelevant filler, off-topic but coherent text), lorem (lorem-ipsum-style out-of-distribution gibberish), and adversarial (on-distribution text drawn from another trajectory of the same regime) perturbations. Controls switch at 0 percent in every regime. In O1 contractive append (O1, append-mode continuation), switching is modest for neutral at 24 percent and lorem at 18 percent, but rises to 54 percent for adversarial, consistent with append mode requiring a more regime-compatible perturbation. In O2 paraphrase replace (O2, paraphrase replace-mode), switching saturates at 100 percent for neutral and 100 percent for lorem, with adversarial still 94 percent. O3 summarize-negate replace (O3, summarize-then-negate replace mode) shows the same pattern, 100 percent neutral, 100 percent lorem, and 96 percent adversarial. D1 dialog (D1, the dialog regime) is intermediate, with 76 percent neutral, 54 percent lorem, and 60 percent adversarial. D2 drill-down dialog (D2, the drill-down dialog regime) was run for control and adversarial only, yielding 0 percent and 64 percent. The key diagnostic is that O2 and O3 saturate even for lorem, a gibberish out-of-distribution string, and neutral, coherent but off-topic text, which is suspicious if interpreted as semantic susceptibility. Read with the section 5.2 insert-mode probe, where O2 and O3 switching drops to 12 to 32 percent, the roughly 60 to 80 percentage-point overwrite-versus-insert gap indicates that replace-mode fragility mainly reflects the update rule overwriting state rather than a low barrier for injected tokens. This account would have been falsified if controls switched, if only adversarial caused O2/O3 switching, or if insert mode remained near replace-mode rates.}}
\end{figure}
\end{savenotes}

This table is now read through §5.2. The O2/O3 values are real measurements of the original overwrite protocol, but they are not fair injected-token barrier estimates. They mostly measure replacement of prior state. O1 shows the cleanest content-dependent append result: in-distribution adversarial text switches more often than neutral or lorem text. D1 is broadly susceptible across perturbation types, consistent with a dialog-state basin that is easier to redirect than O1 but less mechanically overwritten than O2/O3.

\subsection{Drill-down dialog adds content gravity}

D2 is an Explorer-Expert drill-down dialog. Each user turn asks for a deeper, more specific explanation of one concept from the previous expert turn. The exploratory run used 5 topic families × 5 seed topics, for 25 trajectories at 50 steps each. An adversarial perturbation was injected at step 25, drawing expert text from a different topic family, followed by 25 post-injection steps.

The D2 adversarial switch rate was \textbf{64\%} [44\%, 80\%]. This is not a publication-scale estimate, and it is not perfectly matched to the D1 timing cells because dose, content, and post-injection horizon differ. The qualitative signal is nevertheless useful: 36\% of D2 trajectories did not switch under a late, in-distribution adversarial expert-text injection. The drill-down format imposes content gravity through progressive specialization into a topic tree, which free dialog lacks.

D2 is therefore retained as an exploratory fifth regime. It is distinct from D1 because its dialog state combines conversational style and recent-context capture with an accumulating content path that anchors successive turns.

\subsection{Injection timing reveals basin hardening}

We injected the same perturbation at three times in a 30-step trajectory: D1 neutral at dose 80 and O1 adversarial at dose 200, with $n=50$ per cell.

\begin{table}[h!]
\centering
\small
\begin{tabular}{rrr}
\toprule
inject step & D1, neutral at 80 & O1, adversarial at 200 \\
\midrule
5 & 72\% [58-83] & 60\% [46-73] \\
15 & 78\% [65-87] & 54\% [40-67] \\
25 & 52\% [38-66] & 62\% [48-74] \\
\bottomrule
\end{tabular}
\end{table}

D1 shows partial basin hardening. By step 25, the trajectory has more often committed to a dialog-state basin, and switching falls from 78\% at step 15 to 52\% at step 25. O1 is approximately flat across injection time. The contractive append operator integrates in-domain adversarial text regardless of when it arrives, while D1 becomes harder to redirect late in the trajectory.

\begin{savenotes}
\begin{figure}[h!]
\centering
\includegraphics[width=0.95\linewidth]{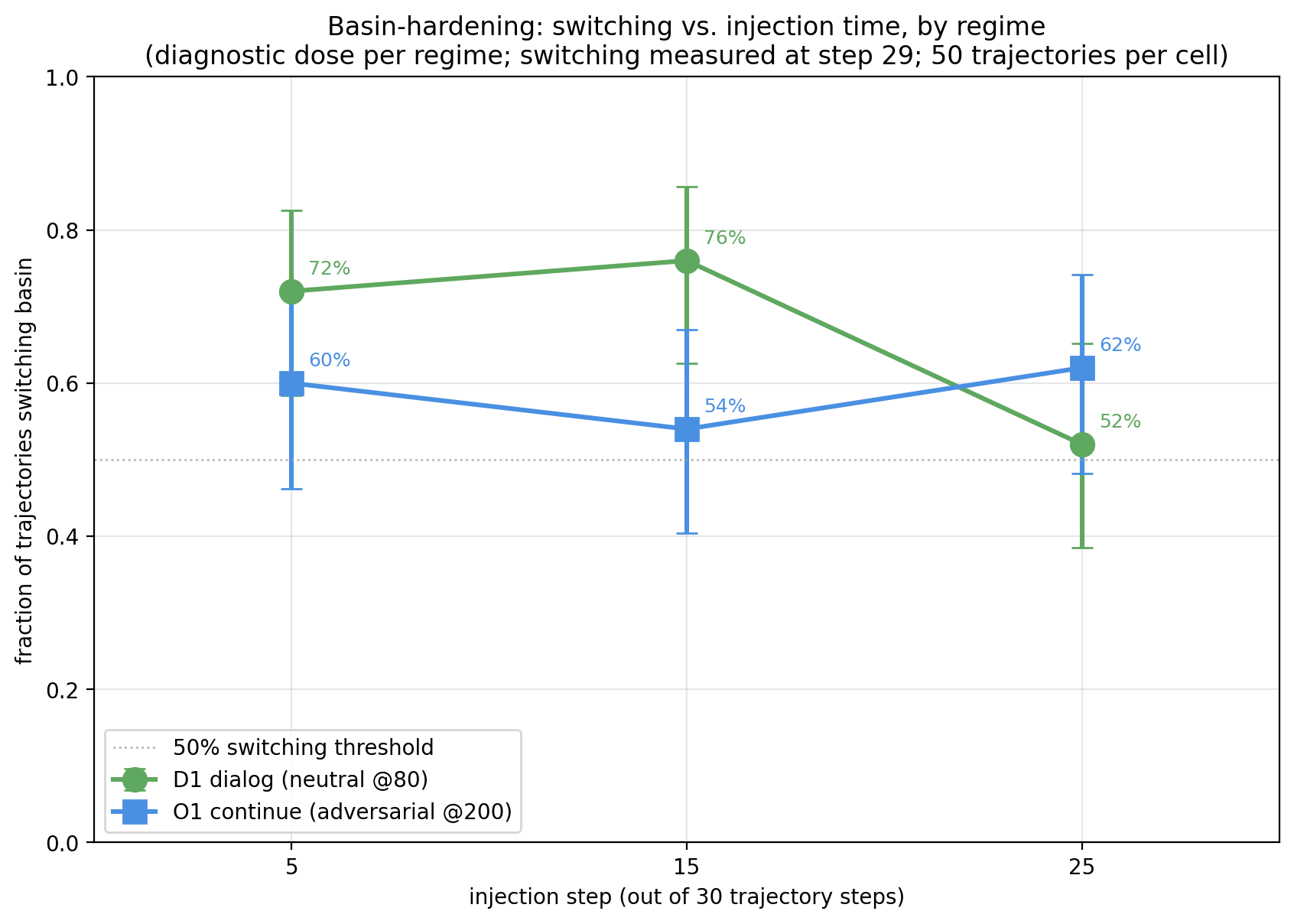}
\caption{\textbf{Basin hardening by injection time.} Switching rates for early, middle, and late injections in O1 and D1, with $n=50$ per cell and 95\% Wilson confidence intervals. D1 shows partial late hardening, whereas O1 adversarial append perturbations remain approximately flat across injection time. Source: \texttt{data/aggregated/perturbation\_basin\_hardening/basin\_hardening.png}.\\[2pt]{\footnotesize\itshape Fig 11 tests whether basin membership becomes harder to perturb as a 30-step trajectory unfolds by injecting a diagnostic dose at steps 5, 15, or 25 and measuring switching at step 29, with 50 trajectories per cell and Wilson 95 percent confidence intervals. The dotted 50 percent switching threshold is the chance-level reference for whether injections move trajectories across basins. In D1 dialog (D1, the role-structured dialog regime) at neutral dose 80 (neutral, off-topic filler text), switching is high when injected early or midway, 72 percent at step 5 and 78 percent at step 15, but falls to 52 percent at step 25, an exact late drop of 26 percentage points from step 15 to step 25. This pattern supports basin hardening: as dialog turns accumulate, the trajectory appears to commit to a basin, so the same neutral perturbation loses leverage late in the run. In O1 continue (O1, the append-mode continuation regime) at adversarial dose 200 (adversarial, on-distribution text from another trajectory), switching is 60 percent at step 5, 54 percent at step 15, and 62 percent at step 25, giving only a 2 percentage point net increase from early to late and no comparable late collapse. This flat timing profile is consistent with a more contractive continuation process that integrates the injected text similarly regardless of when it arrives. The basin-hardening account would be falsified if D1 showed no late decline, especially if step 25 remained near 78 percent, or if O1 showed a similarly large 26 percentage point late drop under the same timing comparison.}}
\end{figure}
\end{savenotes}

\subsection*{Phase C, cluster, granularity, and semantic robustness}

\subsection{Cluster stability and multi-granularity switching}

The canonical basin partition is K-means with $k=12$ on PCA-10. To test whether this partition is an artefact of K-means at one value of $k$, we re-clustered publication-scale late-window clouds using HDBSCAN and spectral clustering, then compared partitions with adjusted Rand index. The result is moderate stability overall, with an important exception for O1.

\begin{table}[h!]
\centering
\small
\begin{tabularx}{\textwidth}{lYYY}
\toprule
regime & median ARI vs K-means $k=12$ & HDBSCAN auto-detected $k$ & interpretation \\
\midrule
O1 contractive & 0.53 & 2 & HDBSCAN sees O1 as one or two large density basins. K-means $k=12$ is a fine sub-partition of a contractive attractor. \\
O2 paraphrase replace & 0.58 & 16 & Replace-mode cluster structure is moderately stable. \\
O3 summarize-negate replace & 0.60 & 16 & Replace-mode cluster structure is moderately stable. \\
D1 dialog & 0.66 & 16 & Highest cross-method stability, but still partly method-dependent. \\
\bottomrule
\end{tabularx}
\end{table}

For O1, this strengthens the attractor interpretation but qualifies the switching metric. A K-means $k=12$ switch can mean movement between sub-regions of a large contractive basin rather than escape from a HDBSCAN density basin. This is why §5.1 separates raw switching from persistent escape and tests persistence under multiple cluster definitions.

We also recomputed perturbation switching in the four diagnostic perturbation pilots under K-means $k=12$, K-means $k=4$, and HDBSCAN.

\begin{table}[h!]
\centering
\small
\begin{tabularx}{\textwidth}{lYYYYY}
\toprule
pilot & condition & $k=12$ & $k=4$ & HDBSCAN & summary \\
\midrule
O1 & adversarial & 0.54 & 0.44 & 0.60 & robustly higher than OOD \\
O1 & neutral & 0.24 & 0.18 & 0.38 & low across all \\
O1 & lorem & 0.18 & 0.18 & 0.30 & low across all \\
O2 & adversarial & 0.94 & 0.72 & 1.00 & saturated except coarse $k=4$ \\
O2 & neutral & 1.00 & 1.00 & 1.00 & saturated \\
O2 & lorem & 1.00 & 1.00 & 1.00 & saturated \\
O3 & adversarial & 0.96 & 0.74 & 0.98 & saturated except coarse $k=4$ \\
O3 & neutral & 1.00 & 0.74 & 1.00 & high across all \\
O3 & lorem & 1.00 & 1.00 & 1.00 & saturated \\
D1 & adversarial & 0.60 & 0.50 & 0.40 & granularity-sensitive \\
D1 & neutral & 0.76 & 0.60 & 0.66 & granularity-sensitive \\
D1 & lorem & 0.56 & 0.46 & 0.44 & granularity-sensitive \\
\bottomrule
\end{tabularx}
\end{table}

The O1 content asymmetry is robust: adversarial switching remains roughly 2-3 times neutral or lorem switching across granularities. O2/O3 overwrite-protocol switching remains high at fine granularities, with some collapse under coarse $k=4$. D1 is the most granularity-sensitive, consistent with its family-leakage and dialog-state dependence.

\begin{savenotes}
\begin{figure}[h!]
\centering
\includegraphics[width=0.95\linewidth]{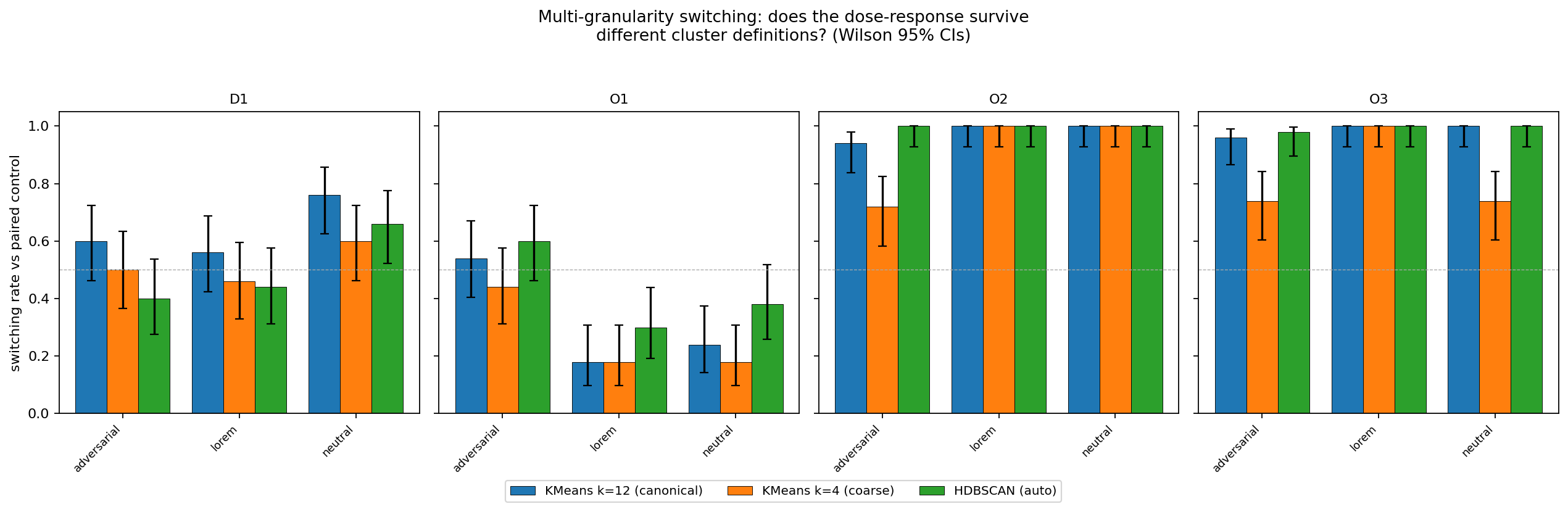}
\caption{\textbf{Switching under alternative basin granularities.} Perturbation switching recomputed under K-means $k=12$, K-means $k=4$, and HDBSCAN. The O1 adversarial-versus-OOD contrast is robust across granularities, while D1 is more granularity-sensitive. Source: \texttt{data/aggregated/multi\_granularity\_switching.png}.\\[2pt]{\footnotesize\itshape Fig 12 evaluates whether the switching dose response is invariant to cluster granularity by comparing K-means k=12 (the canonical 12-cluster partition), K-means k=4 (the coarse 4-cluster partition), and HDBSCAN (the density-based variable-k partition) across O1 (the append-mode continuation regime), O2 (the paraphrase replace-mode regime), O3 (the summarize-negate replace-mode regime), and D1 (the role-structured dialog regime). The falsification test is straightforward: if the apparent dose response were an artifact of one clustering scheme, changing granularity should erase or invert the adversarial-versus-out-of-distribution contrast. In O1, it does not: adversarial switching is 0.54/0.44/0.60 under K-means k=12 / K-means k=4 / HDBSCAN, compared with neutral 0.24/0.18/0.38 and lorem 0.18/0.18/0.30, preserving the roughly 2 to 3 times adversarial-versus-OOD asymmetry across all three definitions. In D1, by contrast, the result is granularity-sensitive: adversarial is 0.60/0.50/0.40, neutral is 0.76/0.60/0.66, and lorem is 0.56/0.46/0.44, so HDBSCAN reverses the adversarial-versus-lorem ordering, with lorem 0.44 exceeding adversarial 0.40. O2 and O3 mainly show ceiling effects: O2 adversarial is 0.94/0.72/1.00 while O2 lorem and neutral are 1.00/1.00/1.00, and O3 adversarial is 0.96/0.74/0.98, neutral is 1.00/0.74/1.00, and lorem is 1.00/1.00/1.00. Thus O1 is robust, whereas D1 requires fixed granularity.}}
\end{figure}
\end{savenotes}

\subsection{Within-regime structure: cluster semantics and family heterogeneity}

Two analyses examine variability within each regime. The first inspects what the K-means clusters represent semantically, on a per-regime basis. The second documents per-family heterogeneity behind the population-level O1 dose response. Together they sharpen the regime taxonomy without changing the headline endpoint claims.

\textbf{Cluster semantics.} We extracted representative trajectory text from each K-means cluster and had a separate held-out reasoning model characterize the clusters blind to the paper's regime labels. The detailed per-cluster tables are available in the released audit artefacts. The main result is that the four canonical regimes are all multi-cluster at $k=12$, but the semantic axis of clustering differs by regime.

\begin{table}[h!]
\centering
\small
\begin{tabularx}{\textwidth}{lYYY}
\toprule
regime & basin axis & mechanism & taxonomy implication \\
\midrule
O1 & register and style & append-mode continuation contracts toward high-probability continuation registers, such as sentimental narrative, policy-discursive exposition, reflective empathic prose, and technical tutorial & label preserved, but specified as register-contractive rather than topic-contractive \\
O2 & seed family and local topic & paraphrase preserves meaning while sanding surface form into conventional paraphrase & period-2 dynamics remain, but basins are family-preserving rather than semantically absorbing \\
O3 & formal template & summarize-then-negate imposes a summary plus antithesis discourse shape while preserving seed-specific content & absorbing means template-absorbing, not content-convergent \\
D1 & dialogue state and recent-context capture & append-mode dialog drifts into conversational acts such as coaching, reassurance, recommendation, journaling, or creative feedback & better described as dialogue-state-driven multi-basin than purely stylistic multi-basin \\
\bottomrule
\end{tabularx}
\end{table}

The semantic inspection preserves the regime taxonomy but sharpens the mechanism. O1 is the strongest true attractor case, with convergence along register rather than content. O2 and O3 are operator-shaped but content-preserving in different ways. D1 is governed by conversational state and recent-context capture rather than by a stable topic basin.

\textbf{Family heterogeneity.} The sparse O1 adversarial dose grid revealed substantial family-level heterogeneity. Each family-dose cell has $n=10$ trajectories, so the table is not a precise family-level ED50 estimate. It is a warning that the population curve mixes different local response profiles.

\begin{table}[h!]
\centering
\small
\begin{tabularx}{\textwidth}{lYYYY}
\toprule
family & dose 20 & dose 80 & dose 200 & dose 400 \\
\midrule
philosophy\_dialog & 0.10 & 0.40 & 0.90 & 0.50 \\
practical\_dialog & 0.40 & 0.20 & 0.70 & 0.80 \\
creative\_dialog & 0.20 & 0.40 & 0.30 & 0.60 \\
reflective & 0.30 & 0.30 & 0.40 & 0.40 \\
emotional & 0.30 & 0.40 & 0.40 & 0.10 \\
\bottomrule
\end{tabularx}
\end{table}

\texttt{philosophy\_dialog} shows a clear threshold-like increase up to dose 200. \texttt{practical\_dialog} increases after dose 80. \texttt{creative\_dialog} increases after dose 200. \texttt{reflective} is nearly flat, and \texttt{emotional} is non-monotone with a low 400-token endpoint. The dense rerun establishes a clean population-level raw dose response, but the wide family-cluster bootstrap interval is consistent with this underlying heterogeneity. Future replications should increase the number of prompt families rather than only the number of ICs per family.

\begin{savenotes}
\begin{figure}[h!]
\centering
\includegraphics[width=0.95\linewidth]{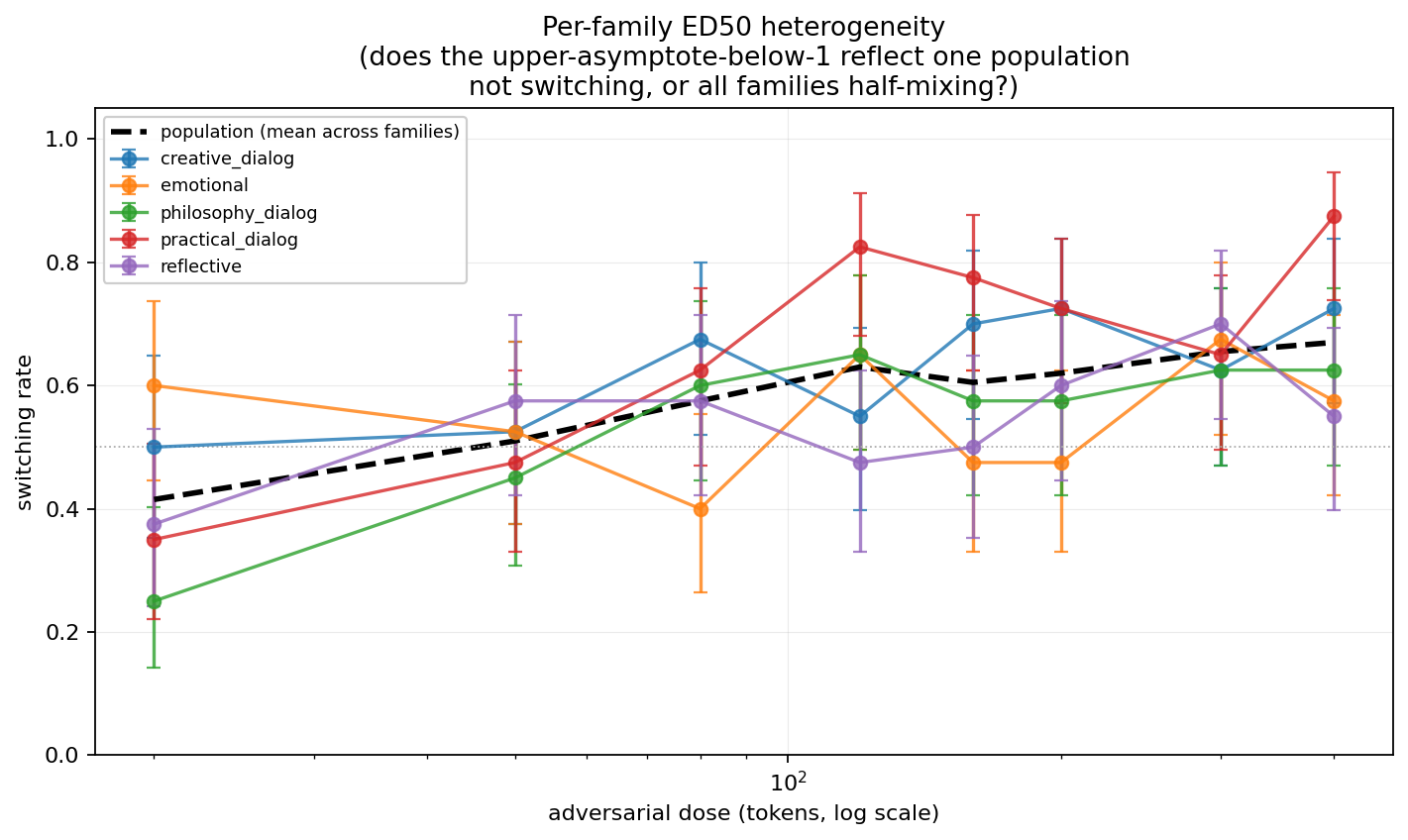}
\caption{\textbf{Per-family O1 adversarial dose response.} Family-level sparse O1 adversarial dose curves with $n=10$ trajectories per family-dose cell. Heterogeneity behind the population-level ED50 explains the wide family-cluster bootstrap interval. Source: \texttt{data/aggregated/per\_family\_ed50.png}.\\[2pt]{\footnotesize\itshape Fig 13 plots per-family O1 (O1, the append-mode continuation regime) adversarial (adversarial, on-distribution text from another trajectory) dose-response curves, where family (a prompt family used as a set of initial conditions) is shown by color on a log dose axis, with population (mean across families) as the thick black dashed curve and the dotted line marking the 50 percent switching threshold. Each family-dose cell has n=10. The exact switching rates at 20/80/200/400 tokens show why a single ED50 is misleading: philosophy\_dialog (philosophical-prompt family) is 10/40/90/50 percent, threshold-sensitive with a peak at 200 then a drop; practical\_dialog (practical-advice family) is 40/20/70/80 percent, a late rise; creative\_dialog (creative-writing family) is 20/40/30/60 percent, a weak upward trend; reflective (reflective-prompt family) is 30/30/40/40 percent, approximately flat; and emotional (emotional-prompt family) is 30/40/40/10 percent, non-monotone and low at 400. Thus the population mean ED50 of about 40 tokens hides five qualitatively different response shapes, not merely five noisy estimates of one curve. The wide bootstrap CI [8.5, 242] should therefore be read as real between-family heterogeneity rather than just sampling noise from the initial conditions within each family. The heterogeneity interpretation would be falsified if future higher-n replications, using the same family definitions and adversarial construction, collapsed these curves to a common monotone dose response with overlapping ED50s and no systematic family-by-dose interaction. Future replications should add more families, not just more initial conditions within the same families.}}
\end{figure}
\end{savenotes}

\subsection*{Phase D, embedder and cross-generator checks}

\subsection{Robustness: embedder ablation and within-vendor cross-model verification}

Two robustness checks address whether the regime taxonomy survives substitution of the measurement embedder and substitution of the generator. The first re-embeds representative experiments with alternative text embedders. The second reruns the regime-level predicates on a within-vendor smaller model. Both checks confirm the qualitative taxonomy without numerically replicating the headline ED50 or persistence endpoints.

\textbf{Embedder ablation.} The main measurements use \texttt{text-embedding-3-small}. We re-embedded 5,000-step subsamples of representative experiments under two alternatives: \texttt{text-embedding-3-large}, a within-vendor scale-up, and \texttt{all-mpnet-base-v2}, a local cross-architecture sentence-transformer. We then recomputed recurrence, late sharpness dimension, and basin predictability.

The rank-order result is selective. Basin predictability is the most embedder-invariant diagnostic, recurrence is partially invariant, and sharpness dimension is not invariant.

\begin{table}[h!]
\centering
\small
\begin{tabularx}{\textwidth}{lYYY}
\toprule
diagnostic & Spearman rank correlation vs \texttt{text-embedding-3-large} & Spearman rank correlation vs \texttt{all-mpnet-base-v2} & interpretation \\
\midrule
recurrence rate & +0.60 & +0.60 & high/low split between replace-mode and append/dialog regimes is preserved \\
basin predictability acc(k=10) & +0.80 & +1.00 & strongest embedder-invariant diagnostic \\
sharpness\_dim\_late & -0.40 & +0.00 & embedding-specific and not load-bearing \\
\bottomrule
\end{tabularx}
\end{table}

The load-bearing taxonomy distinction between replace-mode regimes and append/dialog regimes survives embedder substitution. O2 and O3 remain high-recurrence and high-predictability relative to O1, D1, and D2. The fine-grained sharpness-dimension ordering should be interpreted only within the original \texttt{text-embedding-3-small} measurement pipeline.

\begin{savenotes}
\begin{figure}[h!]
\centering
\includegraphics[width=0.95\linewidth]{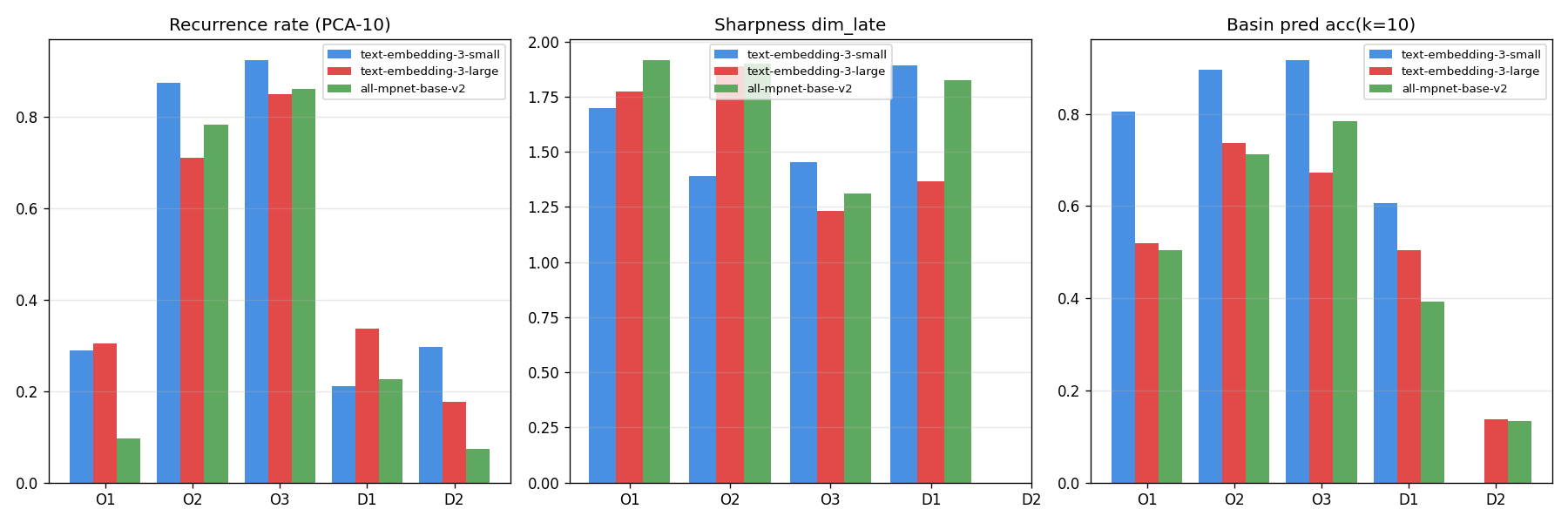}
\caption{\textbf{Embedding-model ablation.} Diagnostics recomputed under \texttt{text-embedding-3-small}, \texttt{text-embedding-3-large}, and \texttt{all-mpnet-base-v2}. Basin predictability and coarse recurrence ordering are more stable than sharpness dimension. Source: \texttt{data/aggregated/embedding\_ablation/comparison.png}.\\[2pt]{\footnotesize\itshape Fig 14 reports an embedding-space invariance check across O1 (append continuation), O2 (paraphrase replace), O3 (summarize-negate replace), D1 (dialog), and D2 (drill-down dialog), comparing text-embedding-3-small (canonical OpenAI 1536-d, used in main analyses), text-embedding-3-large (within-vendor OpenAI scale-up), and all-mpnet-base-v2 (local sentence-transformers cross-architecture). Recurrence rates are only partially invariant by rank against text-embedding-3-small, with Spearman correlations of +0.60 for text-embedding-3-large and +0.60 for all-mpnet-base-v2, but the coarse split is stable: O2 and O3 remain high-recurrence replacement regimes, while O1, D1, and D2 remain lower append or dialog regimes. Exact recurrence rates for text-embedding-3-small, text-embedding-3-large, and all-mpnet-base-v2 are O1 0.289/0.304/0.096, O2 0.875/0.711/0.783, O3 0.924/0.850/0.862, and D1 0.210/0.337/0.226. Basin predictability acc(k=10) shows the strongest invariance, with Spearman correlations of +0.80 and +1.00 versus text-embedding-3-small, supporting the claim that the replace-vs-append/dialog separation is not an artifact of one embedding model. By contrast, sharpness\_dim\_late is not invariant, with Spearman correlations of -0.40 and 0.00, so it should not be interpreted cross-embedder. The falsification test is whether the qualitative replace-vs-append/dialog split disappears under another embedder; it does not, although the fine ordering within replacement regimes is embedder-specific.}}
\end{figure}
\end{savenotes}

The n = 4 cross-metric correlations among recurrence, switching, sharpness, and lock-in are descriptive only and are reported in the released aggregate tables rather than used as inferential evidence here.

\textbf{Cross-model thesis verification.} We also ran a cross-generator audit by substituting \texttt{gpt-4.1-nano} for \texttt{gpt-4o-mini} in the regime-level experiment set and evaluating six machine-checkable thesis predicates. This is a qualitative audit of the regime taxonomy, not a replication of the dense ED50 endpoint, the persistence endpoint, or the overwrite-versus-insert mechanism.

The regime-level predicates all passed on both generators:

\begin{table}[h!]
\centering
\small
\begin{tabularx}{\textwidth}{lYYYY}
\toprule
ID & audited claim & \texttt{gpt-4o-mini} & \texttt{gpt-4.1-nano} & verdict \\
\midrule
T1 & recurrence ordering separates replace from append/dialog & O1 0.272, O2 0.834, O3 0.905, D1 0.146 & O1 0.393, O2 0.840, O3 0.866, D1 0.168 & PASS on both \\
T2 & replace-mode perturbation switching remains high under original overwrite protocol & O2 adv 0.94, O3 adv 0.96 & O2 adv 0.94, O3 adv 0.88 & PASS on both \\
T3 & O1 neutral and lorem remain in drift-floor band & control 0.00, neutral 0.24, lorem 0.18 & control 0.00, neutral 0.22, lorem 0.18 & PASS on both \\
T4 & O1 adversarial switching exceeds O1 lorem switching & adv 0.54 vs lorem 0.18 & adv 0.38 vs lorem 0.18 & PASS on both \\
T5 & D1 neutral switching exceeds 0.30 & neutral 0.76 & neutral 0.80, lorem 0.94 & PASS on both \\
T6 & publication-scale verdict labels match expected H1a/H1b tuples & expected tuples & identical tuples & PASS on both \\
\bottomrule
\end{tabularx}
\end{table}

The structural taxonomy is preserved. Replace-mode regimes remain high-recurrence and high-switching under the original overwrite protocol. O1 remains content-sensitive, with adversarial text exceeding lorem text. D1 remains broadly susceptible to perturbation.

The magnitude shifts are also informative. \texttt{gpt-4.1-nano} has a smaller O1 adversarial-vs-lorem margin, +0.20 compared with +0.36 on \texttt{gpt-4o-mini}, suggesting shallower contractive basins. D1 lorem switching rises to 0.94 on \texttt{gpt-4.1-nano}, suggesting looser dialog-state anchoring.

The audit does \textbf{not} replicate the full headline endpoint. The following claims remain established only for the \texttt{gpt-4o-mini} experiments reported above:

\begin{table}[h!]
\centering
\small
\begin{tabularx}{\textwidth}{lYY}
\toprule
claim & replicated on \texttt{gpt-4.1-nano}? & status \\
\midrule
dense O1 adversarial $\mathrm{ED50}_{\mathrm{raw}} \approx 40$ tokens & no & not rerun in the cross-generator thesis audit \\
O1 natural stochastic-divergence floor of 34.7\% & no & not rerun in the cross-generator thesis audit \\
persistent escape not reaching 50\% up to 400 tokens & no & not rerun in the cross-generator thesis audit \\
overwrite-versus-insert gap of 60-80 percentage points for O2/O3 & no & not rerun in the cross-generator thesis audit \\
V* density-landscape sensitivity and ordinal stability & no & not rerun in the cross-generator thesis audit \\
regime-level qualitative predicates T1-T6 & yes & passed on both generators \\
\bottomrule
\end{tabularx}
\end{table}

The correct interpretation is therefore narrow but useful: the \textbf{regime-level qualitative claims} generalize to \texttt{gpt-4.1-nano}; the \textbf{dense endpoint calibration} remains a \texttt{gpt-4o-mini} result until rerun directly.

\subsection*{Phase E, secondary analyses}

\subsection{Bulk geometry is descriptive, not endpoint-defining}

Two methodological analyses test whether bulk-geometry diagnostics could replace perturbation evidence as the basis for regime classification. The first checks the parameter sensitivity of the empirical potential-barrier summary $V^\star$. The second asks whether unsupervised clustering of bulk diagnostic vectors recovers the five-regime taxonomy. Both analyses conclude that bulk geometry is useful for description but cannot stand in for the perturbation endpoints used in §5.1 and §5.2.

\textbf{Density landscape sensitivity.} The $V(x) = -\log \hat{\rho}(x)$ density-landscape analyses visualize where trajectory clouds spend time in a joint PCA-2 projection. They are useful descriptive summaries of geometry, but they are not calibrated barrier-height estimates and they do not validate the token ED50 values in §5.1.

The main limitation is parameter sensitivity. For the O1 perturbation pilot, we recomputed $V^\star$ across 45 combinations of KDE bandwidth, grid resolution, and basin count. Per-condition coefficients of variation ranged from 14\% to 24\%. A single numerical $V^\star$ value is therefore not stable enough to quote as a calibrated barrier.

\begin{savenotes}
\begin{figure}[h!]
\centering
\includegraphics[width=0.95\linewidth]{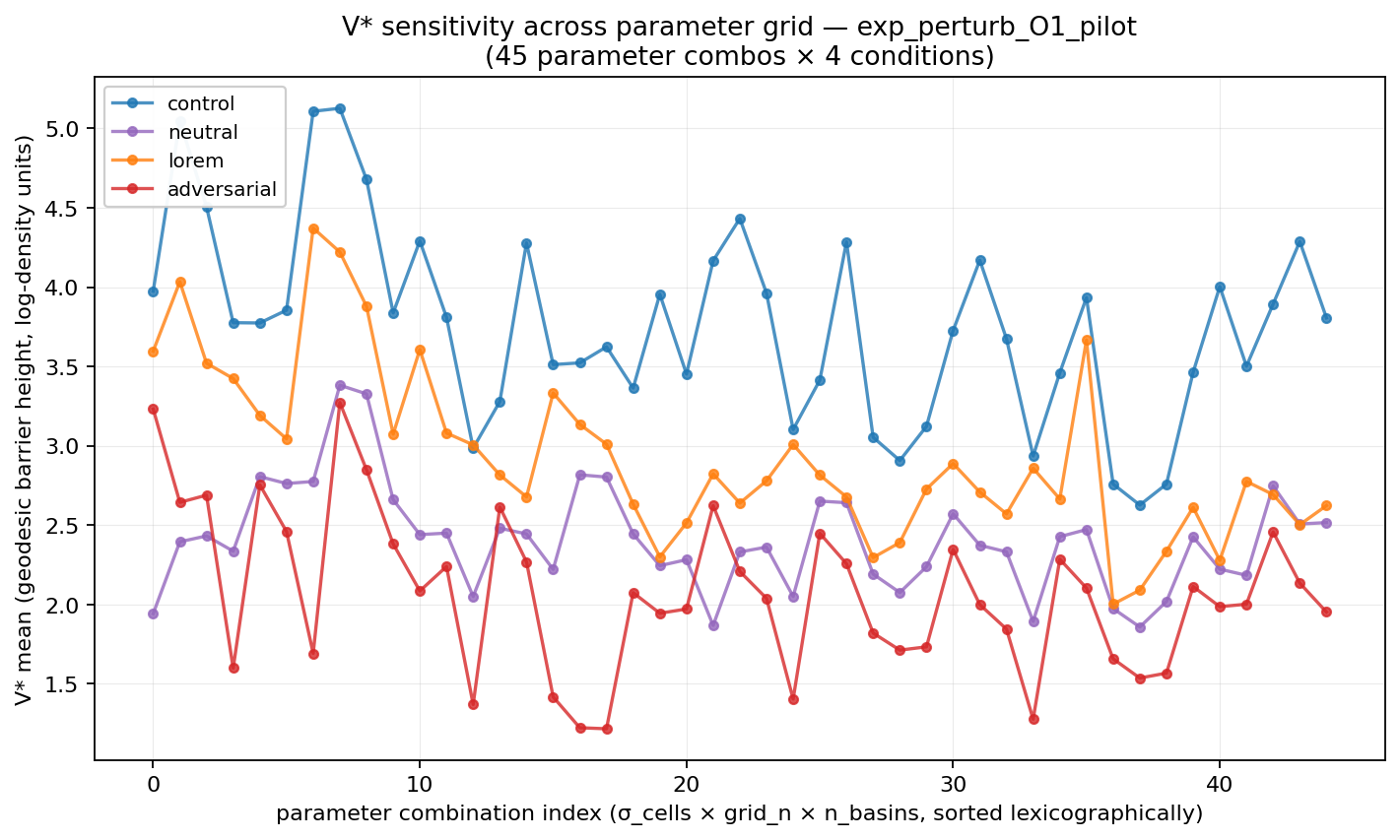}
\caption{\textbf{V* parameter-grid sensitivity.} Sensitivity of empirical potential-barrier summaries across KDE bandwidth, grid resolution, and basin-count settings. The ordinal pattern is more stable than the absolute $V^\star$ values, so density landscapes remain descriptive rather than calibrated. Source: \texttt{data/aggregated/v\_star\_sensitivity.png}.\\[2pt]{\footnotesize\itshape Fig 15 plots V-star (the geodesic barrier height defined as the maximum V along the Dijkstra shortest path between basin centers in the empirical density landscape V(x) = -log rho(x)) across the parameter grid (45 combinations of KDE bandwidth, grid resolution, and basin-count settings). The four lines are control (the unperturbed pilot trajectories), neutral (the off-topic filler-perturbed trajectories), lorem (the lorem-ipsum gibberish-perturbed trajectories), and adversarial (the on-distribution-perturbed trajectories). These density landscapes are descriptive rather than calibrated: absolute V-star magnitudes should not be interpreted as physical barrier heights or as directly comparable effect sizes across parameterizations. Consistent with that limitation, V-star values vary appreciably across the grid, with per-condition coefficients of variation ranging from 14 to 24 percent. The relevant evidence is ordinal. Across the 45 settings, control has the highest V-star in 98 percent of grids, adversarial has the lowest V-star in 89 percent of grids, and the dominant ordering is control greater than neutral and lorem greater than adversarial. Thus the figure supports a robustness claim about rank structure, not a calibrated metric claim about barrier size. The result would be falsified if reasonable bandwidth, grid, or basin-count choices frequently reversed the ordering, especially if adversarial ceased to be the lowest or control ceased to be the highest across a substantial fraction of grids. The token ED50 result at 40 tokens and the V-star ordinal pattern are consistent, but they are not equivalent measurements and should be interpreted as converging ordinal evidence only.}}
\end{figure}
\end{savenotes}

The ordinal pattern was more stable. Across the 45 parameter combinations, control had the highest $V^\star$ in 98\% of combinations, adversarial had the lowest $V^\star$ in 89\%, and neutral and lorem occupied the middle. This supports a weak geometric statement: the density-landscape summaries preserve a robust rank ordering within the O1 perturbation pilot. It does not support a quantitative $V^\star$ to ED50 conversion.

This distinction is especially important for replace-mode regimes. O2 and O3 can show high geometric separation while also showing high overwrite-protocol switching, because the perturbation and memory policy can create or occupy a different basin rather than cross a pre-existing ridge. Full $V^\star$ tables, renormalization-group-style merge-distance tables (RG merge-distance: maximum Ward-linkage merge height across $k=48$ fine cluster centroids, used as a coarse-graining diagnostic of trajectory-cloud expansion), and the parameter-grid sensitivity outputs are moved to §12.11.

\textbf{Why exactly five regimes.} This sub-analysis shows that bulk geometry alone does not recover the five-regime taxonomy, motivating perturbation as the load-bearing test. We assembled five-dimensional diagnostic vectors containing recurrence rate, late sharpness dimension, late $\lambda_1$, basin-predictability accuracy at $k=10$, and adversarial switching rate. Internal validation indices did not select a single cluster count matching the five labels: silhouette favored two clusters, Calinski-Harabasz favored seven, and Davies-Bouldin favored six.

Bulk diagnostics separate O2 and O3 from the append/dialog regimes, but they do not cleanly separate O1 from D1. O1 and D1 have similar recurrence, contraction, and basin-predictability values at this diagnostic resolution.

\begin{savenotes}
\begin{figure}[h!]
\centering
\includegraphics[width=0.95\linewidth]{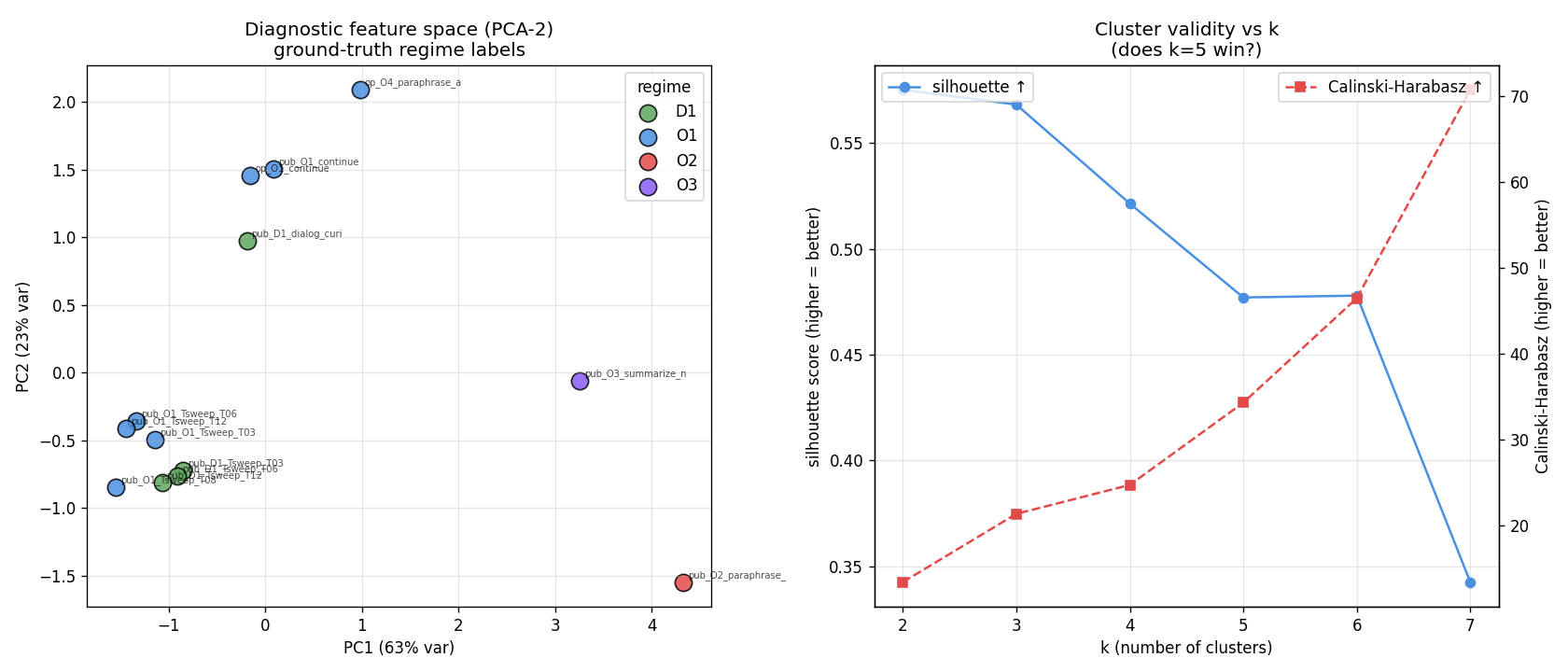}
\caption{\textbf{Regime clustering in diagnostic space.} Scatter view of regime diagnostic vectors used in the unsupervised five-regime check. Bulk geometry separates replace-mode regimes from append/dialog regimes but does not by itself recover the full five-way taxonomy. Source: \texttt{data/aggregated/regime\_cluster\_analysis/cluster\_scatter.png}.\\[2pt]{\footnotesize\itshape Fig 16 is a two-panel diagnostic check on the claim that the proposed five-regime taxonomy is visible in bulk feature geometry alone. The left panel projects each experiment's five-dimensional diagnostic vector into PCA space, with colors showing ground-truth regime labels: O1 (append-mode continuation), O2 (paraphrase replace mode), O3 (summarize-negate replace mode), and D1 (role-structured dialog). The diagnostic vector contains recurrence rate, late sharpness dimension, late lambda\_1, basin-predictability accuracy at k=10, and adversarial switching rate. Visually, the replace regimes O2 and O3 separate strongly from the rest, while O1 and D1 overlap across the dense bulk of the projection, so PCA geometry alone does not cleanly recover all named regimes. The right panel asks the same question with internal clustering criteria: the silhouette index (within-cluster cohesion versus between-cluster separation) peaks at k=2; the Calinski-Harabasz index (ratio of between to within cluster dispersion) peaks at k=7; and the Davies-Bouldin index (average similarity ratio of each cluster to its most similar one) peaks at k=6. None of these validation indices picks k=5. Thus Fig 16 is a falsification of the stronger claim that exactly five regimes are justified by unsupervised bulk diagnostics alone. The five-regime taxonomy therefore requires perturbation evidence, especially the intervention results in Fig 10 and Fig 12, rather than being supported by the diagnostic point cloud by itself.}}
\end{figure}
\end{savenotes}
 The perturbation protocol is what separates them: O1 shows content-dependent adversarial raw switching with out-of-distribution resistance, while D1 is broadly susceptible to dialog-state redirection and hardens with time. D2 is then distinguished by drill-down content gravity.

\begin{savenotes}
\begin{figure}[h!]
\centering
\includegraphics[width=0.95\linewidth]{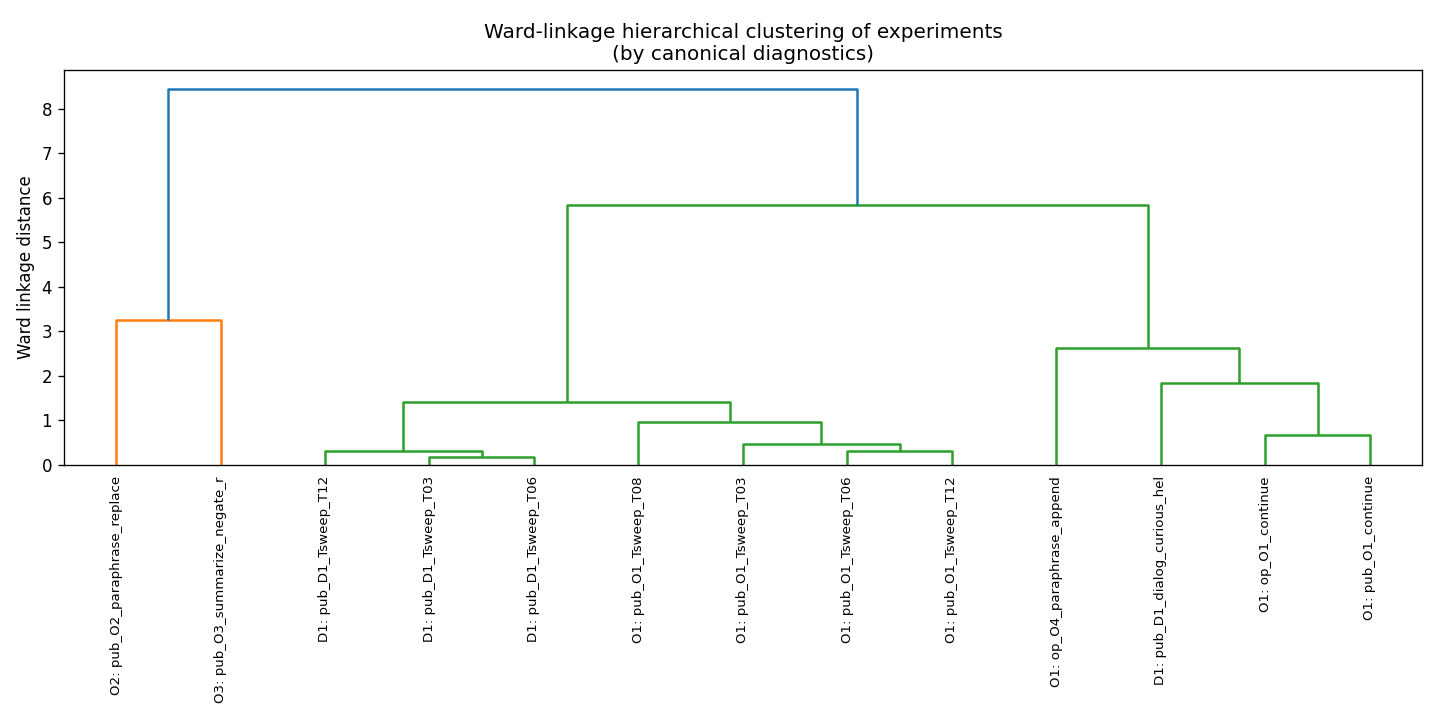}
\caption{\textbf{Regime-clustering dendrogram.} Hierarchical clustering of regime-level diagnostic summaries. The dendrogram reinforces that the five-regime taxonomy is not obtained from bulk diagnostics alone and requires perturbation endpoints for separation. Source: \texttt{data/aggregated/regime\_cluster\_analysis/cluster\_dendrogram.png}.\\[2pt]{\footnotesize\itshape Fig 17 reports a Ward-linkage (Ward-linkage, the minimum-within-cluster-variance criterion) hierarchical clustering dendrogram for the 14 experiments using the same 5-D diagnostic vectors as Fig 16. Leaves are experiment labels, and the vertical merge height is the Ward-linkage distance. The O2: pub\_O2\_paraphrase\_replace (paraphrase replace) and O3: pub\_O3\_summarize\_negate\_r (summarize-negate replace) pair forms a clearly isolated branch: those two leaves first merge at about 3.25, then join all remaining experiments only at the top height, about 8.45. By contrast, the O1: append-mode continuation, and D1: role-structured dialog, experiments mostly occupy nearby low-distance structure. The T-sweep (T-sweep, the temperature-variant of an experiment) leaves verify within-regime consistency: D1 pub\_D1\_Tsweep\_T03 and D1 pub\_D1\_Tsweep\_T06 merge near 0.15, D1 pub\_D1\_Tsweep\_T12 joins near 0.30, O1 pub\_O1\_Tsweep\_T06 and O1 pub\_O1\_Tsweep\_T12 merge near 0.30, O1 pub\_O1\_Tsweep\_T03 joins near 0.48, and O1 pub\_O1\_Tsweep\_T08 joins near 0.95. The broader O1 and D1 T-sweep block merges near 1.40, while the O1 continue pair joins near 0.67 and then links to D1 pub\_D1\_dialog\_curious\_heir near 1.85 and O1 op\_O4\_paraphrase\_append near 2.63. A coarse cut therefore yields roughly 2 to 3 groups, and a fine cut yields about 6 to 7 subgroups, but no natural 5-way partition. This falsifies the hypothesis that the bulk diagnostic geometry recovers the nominal 5-way experimental split, matching Fig 16.}}
\end{figure}
\end{savenotes}

The five-way taxonomy is therefore supported by the union of bulk diagnostics and perturbation endpoints, not by either alone. Full unsupervised-clustering matrices, validation indices, and feature-space plots are available in the released aggregate tables.

At matched reduced scope, D1 showed narrower temperature variation than O1: D1 basin predictability at $k=10$ stayed in a 0.57-0.61 band, while O1 ranged from 0.52 to 0.65. However, the O1 reduced-scope T=0.8 cell was 28 percentage points below the publication-scope T=0.8 anchor, 0.52 versus 0.80, so O1 absolute temperature values are scope-confounded. The full temperature sweep is retained as exploratory secondary material in the released aggregate tables.

\begin{savenotes}
\begin{figure}[h!]
\centering
\includegraphics[width=0.95\linewidth]{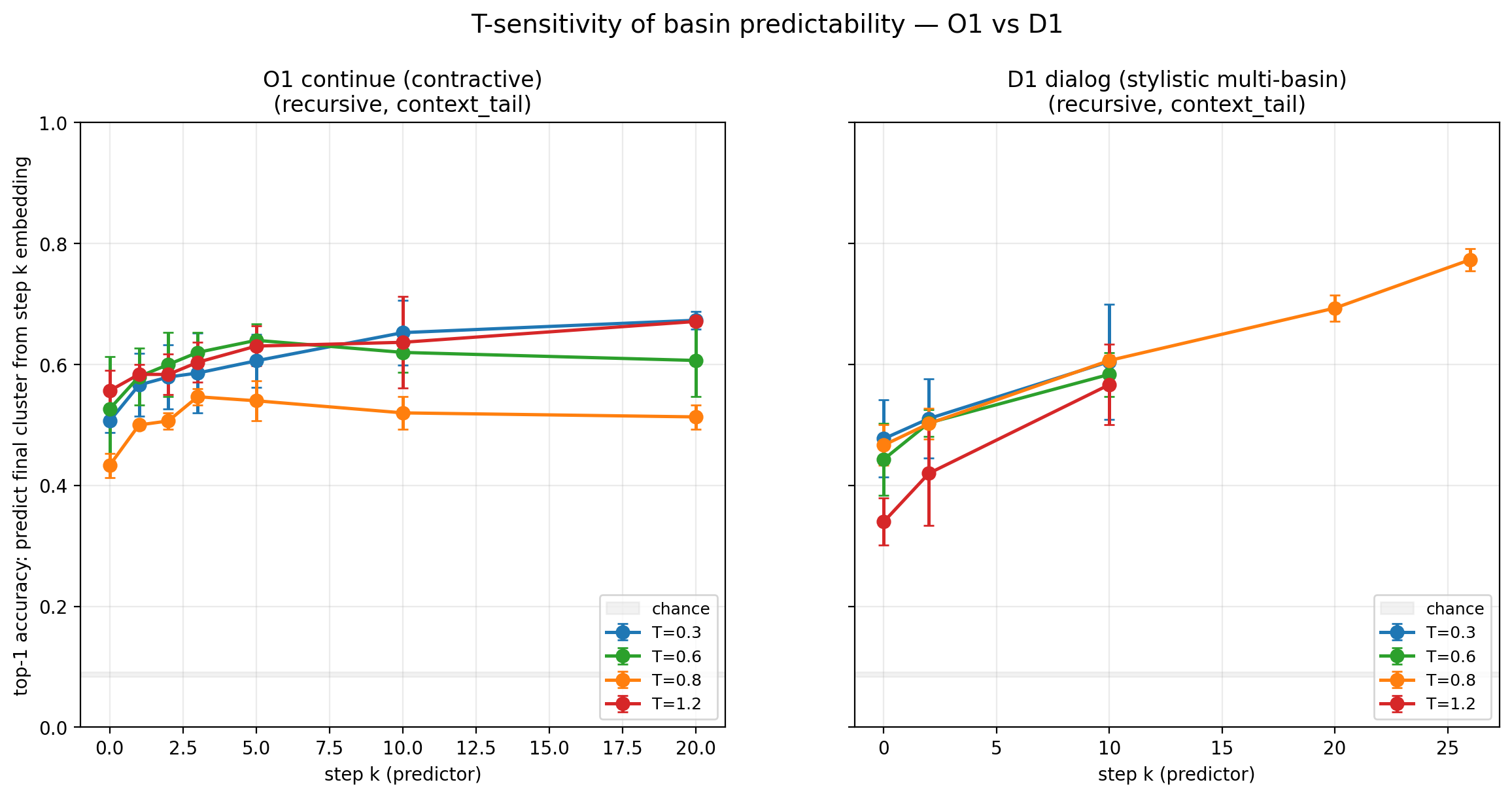}
\caption{\textbf{Cross-experiment temperature sensitivity.} Temperature-sensitivity summary across reduced-scope cells. These results remain explicitly secondary because absolute O1 values are scope-confounded relative to the publication-scale anchor. Source: \texttt{data/aggregated/t\_sensitivity\_cross\_regime/cross\_t\_sensitivity.png}.\\[2pt]{\footnotesize\itshape Fig 18 reports cross-experiment temperature sensitivity as an exploratory secondary check, not a primary result. O1 (append-mode continuation) is shown at left, and D1 (role-structured dialog) at right. In both panels, acc(k) is the top-1 accuracy for predicting the final late-window basin from the embedding at predictor step k (the recursive step at which the embedding is taken). Legend entries are chance (the random baseline); T=0.3 (sampling temperature 0.3); T=0.6 (sampling temperature 0.6); T=0.8 (sampling temperature 0.8); and T=1.2 (sampling temperature 1.2). All cells are reduced-scope cells (lower-N temperature sweep cells with n=150 each), and are not directly comparable to the publication-scope baseline. The key observation is that D1 is temperature-stable: at k=10, D1 remains within a narrow 0.57-0.61 band across all four temperatures. O1 appears more temperature-variable in the reduced-scope sweep, with k=10 accuracy ranging from 0.52 to 0.65. However, that apparent O1 variability is strongly scope-confounded: the reduced-scope O1 T=0.8 value is 0.52, whereas the publication-scope O1 T=0.8 value is 0.80, a 28 percentage-point scope-confounded delta. Thus the falsification is that D1 does not show the large temperature-driven basin-predictability separation that would be expected under a strong temperature-sensitivity account, while O1's apparent temperature effect cannot be cleanly interpreted because scope changes dominate the comparison.}}
\end{figure}
\end{savenotes}

\section{Discussion}

Robustness in recursive LLM systems is not a scalar model property but emergent from how generated text is re-entered, retained, or overwritten. The main result is therefore not simply that recursive LLM loops have attractor-like regimes, but that apparent robustness depends on two separable choices: the memory policy that writes text back into state, and the endpoint used to define a successful perturbation. Append-mode continuation shows a real in-distribution raw dose response, but not a durable escape threshold. Replace-mode loops, meanwhile, look fragile mostly because the update rule overwrites prior state. Thus the practical unit of analysis is not the prompt alone, or the model alone, but the generator-nudge system evaluated under raw, net, and persistent endpoints.

\subsection{Recursive-loop robustness is determined by memory policy and endpoint choice}

The central lesson is that recursive-loop behavior is determined jointly by the generator and the context-update rule. In the formalism of §3.1, the model samples \(Y_t \sim P_\theta(\cdot \mid X_t; f)\), while the nudge \(\mathcal{N}_\eta\) decides how that sampled text becomes the next state. Two loops with the same generator and similar prompt instructions can therefore have different perturbation responses solely because one preserves prior context and the other overwrites it.

The second lesson is that robustness is endpoint-dependent. A loop can move without staying moved; two unperturbed runs can diverge without any perturbation; and an overwrite-style update can make a system appear fragile even when inserted text has little effect. In O1, adversarial append-mode perturbations produce a finite raw-switching dose response, but the stochastic floor and lack of persistent crossing prevent interpreting that response as durable escape. Raw switching is appropriate for detecting sensitivity; net switching estimates perturbation effect above stochastic drift; persistent escape is the relevant endpoint when the safety concern is durable redirection after recovery turns.

\subsection{Append, replace, and dialog implement different memory mechanisms}

Append, replace, and dialog are not implementation details; they are different memory mechanisms. In append mode, the perturbation enters an already accumulated context and must compete with prior trajectory mass. This makes append-mode continuation resistant to irrelevant text: O1 neutral and lorem perturbations remain near the out-of-distribution drift floor, while in-distribution adversarial text produces the graded raw dose response measured in §5.1.

Replace mode implements a different mechanism. The previous state is discarded, so the next state is dominated by the replacement. The overwrite-vs-insert probe makes this visible: overwrite-mode perturbations in O2 and O3 switch far more often than inserted perturbations, indicating that much apparent replace-mode fragility is attributable to overwrite mechanics rather than a low injected-token barrier (§5.2).

Dialog sits between these cases. It preserves a running transcript like append mode, but role-structured turns give recent utterances high local influence. D1 therefore behaves as a dialogue-state-driven multi-basin regime, while D2 suggests topic-anchored content gravity, meaning repeated return to the same semantic cluster despite local perturbation (§5.5, §5.8).

The expected signatures differ: append should show specificity and graded dose response; replace should show a large overwrite-insert gap; dialog should show recency and role sensitivity; summary or pinned-memory systems should be tested for what information survives compression.

\subsection{A raw-switching ED50 is not a persistent-escape barrier}

The most important interpretive correction is that \(\mathrm{ED50}_{\mathrm{raw}}\) is not the same quantity as a persistent-escape barrier. Raw switching asks whether the perturbed trajectory's terminal cluster differs from its paired control. That endpoint is useful, but it conflates true redirection, ordinary stochastic divergence, and transient movement that later recovers. Persistent escape is stricter: the trajectory must visibly jump at injection and remain in the post-injection basin through the terminal step (§3.1.2). Persistent escape is the appropriate strict endpoint for durable basin change, but transient raw switching may still matter in applications where a single action is consequential.

Under the permissive raw endpoint, O1 adversarial append-mode continuation has a clear dose response: \(\mathrm{ED50}_{\mathrm{raw}} \approx 40\) tokens, with estimates 36, 41, and 52 tokens across the 4PL, GLMM, and family-cluster-bootstrap analyses (§5.1). But the raw curve plateaus near 67\%, not 100\%. Paired controls already disagree at about 35\%, so subtracting the stochastic floor leaves a maximum net effect of only +32 pp at 400 tokens, below the +50 pp criterion for \(\mathrm{ED50}_{\mathrm{net}}\).

The persistent endpoint is smaller still under the canonical bounded-memory loop. At the highest tested dose under the 12K tail-clip, destination-coherent persistence reaches only 16\% under K-means \(k=12\), 10\% under K-means \(k=4\), and 39.5\% under HDBSCAN (§5.1). None crosses 50\% in the tested 5 to 400 token range. Under the full-history protocol (§5.1.3), the picture changes: retained source-basin escape crosses 50\% near 400 tokens and destination-coherent persistence first crosses 50\% near 1,500 tokens. The correct conclusion is therefore: O1 has a finite raw-switching dose response (~40 tokens), and persistent-escape thresholds exist but are memory-policy-conditioned - under bounded memory the perturbation is clipped out before the terminal measurement, suppressing all persistence endpoints; under full-history both endpoints cross 50\% at the higher doses reported in §5.1.3.

\subsection{Density landscapes diagnose basin geometry but do not calibrate token barriers}

The empirical potential landscape \(V(x) = -\log \rho(x)\) is useful as a diagnostic picture of trajectory density, but it should not be read as a calibrated token barrier. The landscape is computed after embedding, projection, smoothing, and basin detection, so its absolute values depend on the representation and analysis choices. The parameter-grid sensitivity analysis confirms this: per-condition \(V^\star\) values vary with coefficient of variation 14 to 24\% across KDE bandwidths, grid resolutions, and basin-count settings (§5.10).

Their value is diagnostic: they show whether trajectories occupy one or several basins, whether perturbations move mass toward alternative density regions, and whether basin structure is stable under analysis choices. What survives here is the ordinal claim. In the O1 perturbation pilot, the rank ordering of \(V^\star\) is stable across 89 to 98\% of parameter combinations: control is usually highest, adversarial usually lowest, and neutral or lorem occupy the middle (§5.10). This supports the use of density landscapes as descriptive basin-geometry tools. It does not justify converting \(V^\star\) into tokens, nor does it independently validate \(\mathrm{ED50}_{\mathrm{raw}}\).

\subsection{Robust recursive-loop evaluations must report floors, persistence, and overwrite mechanics}

\begin{enumerate}
  \item \textbf{Report the context-update rule.} Every recursive-loop evaluation should state whether the system appends, replaces, role-structures, summarizes, pins, rolls, or hybridizes memory, because the same generator can show different perturbation responses under different nudges (§3.1, §5.2).

  \item \textbf{Report raw, net, and persistent-escape endpoints rather than raw switching alone.} Raw movement says whether the trajectory changed, net switching subtracts the control-vs-control stochastic floor, and persistent escape asks whether the perturbation produced a durable basin change after recovery turns (§3.1.2, §5.1, §5.1.3).

  \item \textbf{For replace-style or summary-memory systems, run an overwrite-vs-insert control.} If overwrite switching is high but insert switching is low, the measured fragility is partly the memory policy discarding prior state rather than the injected text overcoming a basin barrier (§5.2).
\end{enumerate}

Box 1 in Supplementary §12.12 operationalizes these points as a minimum reporting checklist, including memory policy, perturbation insertion mode, stochastic floor, endpoint definition, recovery horizon, and overwrite-vs-insert controls.

\subsection{The unresolved question is external validity across generators and real agent scaffolds}

The next step is to ask which parts of this decomposition survive in full agent scaffolds: cross-vendor generators, persistent tool state, retrieval memory, code repositories, and policy-constrained actions. The within-vendor replication on \texttt{gpt-4.1-nano} preserves the qualitative thesis predicates (§5.9), but both generators are from the same vendor family. Cross-vendor and open-weight replication remain necessary before the architecture-vs-generator partition can be treated as general (§7.1, §8.1).

The same agenda extends to real agent scaffolds. The present paper studies recursive text loops with controlled update rules and embedding-space observables; it does not directly evaluate coding agents, tool-using assistants, SWE-Bench tasks, jailbreak benchmarks, production indirect prompt injection, or factuality-graded hallucination settings (§7.5). For indirect prompt injection or coding agents, the analogous question is not whether the transcript topic changes, but whether an injected artifact produces action-level divergence above a no-injection floor and whether that divergence persists through subsequent planning or repair steps.

Those domains can adopt the same endpoint decomposition, but their observables must be application-specific: patch family, files touched, tests passed, policy violations, tool-call traces, or grounded factual claims. The paper's position is therefore bounded but actionable: recursive-loop stability is measurable, and the measurement becomes meaningful when memory policy, stochastic floor, and persistence endpoint are made explicit.

\section{Limitations}

The results establish dose-response and regime-separation behavior for bounded, English, text-only recursive loops under the tested generator-nudge families. They do not establish universality across model vendors, memory policies, languages, or deployed tool-using agents.

The experiments therefore support a structured account of recursive LLM dynamics within a defined operating envelope. That envelope is broad enough to distinguish regimes, reproduce perturbation effects, and compare generator-nudge conditions, but it is not broad enough to make representation-free or deployment-general claims. The limitations below specify where the evidence is strongest, where the measurements are operational rather than invariant, and where additional observables would be required.

\subsection{Evidence is concentrated in two OpenAI generators}

The current audit covers two generators within one vendor: \texttt{gpt-4o-mini} for the headline measurements and \texttt{gpt-4.1-nano} for within-vendor replication of the qualitative regime predicates (§5.9). The headline ED50 estimates, persistence endpoints, F3 cross-loop check, and bounded-vs-full-history memory-policy results are all from \texttt{gpt-4o-mini}. The \texttt{gpt-4.1-nano} audit confirms that the regime taxonomy survives a smaller within-vendor model, but it does not replicate the dense ED50 endpoint, the persistence endpoint, or the overwrite-versus-insert mechanism. There is no cross-vendor replication in the present paper.

This matters because the qualitative taxonomy is motivated by the generator-nudge factorization in §3, not by a special property of one model checkpoint. The append, replace, and dialog distinctions plausibly arise from how the update operator rewrites the next context. However, barrier heights, basin geometry, switching thresholds, and even the number of empirically separable regimes can vary with decoding behavior, tokenizer structure, alignment tuning, refusal style, context-management details, and vendor-specific instruction-following behavior.

The within-vendor evidence supports preservation of the append/replace/dialog qualitative pattern under the tested generator class; it does not support equality of barrier heights, switching probabilities, basin counts, or token-dose thresholds across vendors. The hierarchy of evidence is therefore important. The strongest claim is that the headline regimes survive both within-vendor models. A weaker but still useful claim is that the qualitative pattern emerges from the generator-nudge factorization rather than from a single checkpoint. The audit does not establish numerical equivalence across vendors, and it does not license a model-agnostic claim about all current or future LLMs.

The cross-model audit in this paper should consequently be read as evidence of shape preservation within one vendor family. It supports the existence of recurring append-like, replace-like, and dialog-like patterns in the tested systems, but it does not show that these patterns have fixed quantitative parameters. Future work (§8.1) should expand the vendor set to Anthropic, Google, Mistral, and open-weight models, vary decoding policies, context managers, and refusal regimes before treating the taxonomy as a general law of recursive language-model behavior.

\subsection{Basins, barriers, and tokens are operational measurements}

The basins and barriers reported here are measurements in an operational pipeline, not representation-free physical constants. Trajectories are observed through an embedding model, projected for analysis and visualization, and summarized with density, recurrence, switching, and dose-response statistics. §5.9 reports an explicit representation ablation against a larger OpenAI embedding model and a sentence-transformer model; the attractor-like structure is robust across those tested observables. That robustness does not make the absolute geometry independent of the representation.

The empirical potential landscape, \(V(x)=-\log \rho(x)\), is a descriptive density summary. The Dijkstra barrier \(V^\star\) depends on the kernel density estimation (KDE), PCA-2 reduction, grid resolution, and path construction. These quantities are useful because they impose a common geometric language on many trajectory families, and because their relative ordering can be compared with behavioral perturbation thresholds. They should not be interpreted as thermodynamic free energies or as absolute invariants of the underlying model.

The same caveat applies to token barriers. Tokens are directly measurable, practically interpretable, and closely aligned with how perturbations are injected in §4.5, but they are not the ultimate information unit. A more model-comparable version of these perturbation barriers would likely use conditional surprisal or log-probability cost; because generation logprobs were not stored, the present token-dose results should be treated as operational proxies. The reported token-dose thresholds are therefore useful for within-pipeline comparison, but they should not be read as universal energetic costs or as vendor-independent information barriers.

Predictive basin assignments are also conditional on the validation design. Where basin assignment was modeled predictively, the relevant performance estimates should be read under the group-aware validation design reported in §5; in particular, D1 predictability is evidence for structured separation within the sampled trajectory families, not for arbitrary out-of-family generalization (group-aware acc(k=10) = 0.34, n=450). This distinction matters because recursive trajectories can share prompt family, generator family, seed structure, and local lexical habits. A classifier can capture real within-sample structure without proving that the same decision boundary will transfer to arbitrary new tasks, languages, generators, or memory regimes.

The representation pipeline is therefore best understood as an instrument. It reveals reproducible organization in the sampled trajectories, and the ablations in §5 help rule out a single-embedding artifact. It does not eliminate observer dependence. Stronger representation claims would require storing generation logprobs, testing additional embedding families, measuring raw conditional probabilities where available, and verifying that basin and barrier relationships persist under alternative trajectory observables.

\subsection{The experiments cover bounded, English, static-prompt recursions}

The reported dynamics are properties of bounded, English-language, static-prompt recursive loops. All main runs use a finite context cap with tail clipping, English prompts and seeds, and relatively short generated outputs. In concrete terms, the main recurrence settings use a bounded rolling context of approximately 12,000 characters (roughly the last 6-7 generations in O1 append-mode), per-step generated outputs of ~600 chars at \texttt{max\_output\_tokens=120}, and tail-clip truncation applied once the running context exceeds the buffer (around step 19-20 of a 30-step trajectory). The exact values, caps, and run settings appear in §4. These parameters are part of the experimental condition, not incidental implementation details. §5.1.3 explicitly compares the canonical bounded-memory protocol against a full-history protocol (\texttt{max\_context\_chars = 200000}, large enough that no clipping occurs in the 30-step horizon) and demonstrates that persistence thresholds depend strongly on this choice.

This is a natural setting for controlled recurrence experiments, because it isolates the effect of the nudge operator while keeping the observable trajectory compact enough for repeated perturbation and embedding analysis. The same bounded design makes the dose-response protocol feasible: perturbations can be inserted, the subsequent trajectory can be tracked, and switching or recovery can be measured under comparable context budgets. However, the design also means that the reported basins and barriers are properties of this bounded-memory recurrence, not of every possible recursive use of a language model.

The bounded-memory assumption is especially consequential for append mode. A no-clip pilot suggests that removing clipping deepens the append basin and reduces recurrence, which means the measured append-mode barriers are not properties of append mode in the abstract. They are properties of append mode under a specific bounded-memory recurrence. Larger context windows, different truncation rules, retrieval-augmented memory, explicit memory compression, or pinned long-term goals may change both the basin depth and the route by which perturbations persist.

The language and prompting scope is similarly narrow. The experiments do not test multilingual trajectories, code-heavy trajectories, very long-form generation, or systems in which the system prompt is rewritten online. Prompt drift, refreshed meta-instructions, tool-generated state, or long-horizon document construction could fragment a basin that appears stable under static prompting, or stabilize a replace-mode regime that appears weak when each step is only a short text rewrite. Additional work should test whether the same regime taxonomy remains legible when prompts, languages, modalities, and memory policies are allowed to vary more aggressively.

\subsection{The drill-down dialog regime remains exploratory}

The drill-down dialog regime is the least mature of the reported regimes. The current D2 evidence indicates a distinct form of topic-anchored content gravity under role-structured questioning, but it was tested at substantially smaller scale than the main O1-O3 and D1 regimes. The reported switching estimate is 64\% with a ±10 percentage point bootstrap confidence interval from 25 trajectories at 50 steps.

This is enough to motivate D2 as a candidate regime, but not enough to place it on the same empirical footing as the publication-scale cells. Dialog structure is a broad nudge family: drill-down questioning, debate, role-play, adversarial interrogation, tutoring, self-critique, and multi-party deliberation may create different balances between style persistence and content anchoring. Small changes in role framing, question order, answer length, or conversational memory could alter the apparent basin structure.

Accordingly, D2 is treated as hypothesis-generating in the present taxonomy. It supplies evidence that role-structured dialog can produce topic-anchored recurrence, but it should not yet be treated as a fully replicated regime with stable quantitative thresholds. A publication-scale D2 study would need deeper replication, broader generator coverage, explicit comparison among dialog subfamilies, and perturbation tests matched to the main O1-O3 and D1 protocols. Until then, D2 should be read as a promising extension of the framework rather than as a regime with the same evidentiary status as the strongest cells.

\subsection{Production agent and tool-use claims require new observables}

The experiments measure recursive language-model loops, not deployed coding agents or tool-using autonomous systems. They do not include file-system state, code edits, compiler feedback, test execution, tool schemas, repository-specific correctness criteria, or multi-step planning traces. The implications for coding agents are architectural extrapolations from recursive-loop dynamics, not measurements of deployed coding systems.

A production-agent benchmark would need additional observables. Patch diffs, files touched, tests run, failing tests remaining, tool-call sequences, policy violations, injected-document provenance, and post-perturbation plan persistence would need to be measured alongside or instead of embedding-space trajectory structure. In such systems, the relevant basin may live partly in text, partly in external state, and partly in the interaction between model outputs and tool-mediated environment changes. A trajectory embedding alone would miss many of the state variables that determine whether an agent recovers, drifts, or remains captured by a perturbation.

The key bridge is that memory policy becomes a behavioral variable. Full-history append, rolling-window truncation, generated-summary replacement, pinned-goal memory, provenance-preserving hybrid memory, and retrieval-augmented state can induce different perturbation profiles even when the base model and task are held fixed. The framework in §3 and the perturbation protocol in §4.5 transfer naturally to this setting, but the numerical barriers in §5 do not transfer without re-measurement.

Thus, the transferable contribution is the measurement logic, factorizing generator, nudge, memory, and perturbation dose, not the specific numerical thresholds measured in the present text-only loops.

\section{Future directions}

\textit{The next step is to turn the present perturbation framework from a controlled recursive-loop study into a comparative measurement program for model families, memory policies, dialog scaffolds, and deployed agents.}

The proposed program proceeds from external validity, to mechanistic measurement, to scaffold ablations, to applied agent and safety settings.

\subsection{Cross-vendor replication at publication scale}

\textit{Hypothesis: append/replace/dialog regime ordering is invariant across model families under matched perturbation dose and memory policy.}

The highest-priority extension is a minimum viable product (MVP) cross-vendor replication rather than an exhaustive survey. The central question is not whether barrier heights match numerically across providers, but whether the ordering of append, replace, and dialog regimes survives across generators with different alignment methods, tokenizer families, refusal policies, context handling, and decoding implementations. If the same qualitative ordering appears under matched perturbations and matched analysis code, this would support the claim that regime structure is a property of the generator-nudge system rather than a peculiarity of one model family.

The MVP should include Anthropic, Google, Mistral, and open-weight models, with the current provider family retained as a reference where feasible. The design should use N=20-60 trajectories per cell as cost permits, with all three headline operators included: O1, D1, and D2. D2 is especially important because dialog topology is under-sampled in the present evidence base and may vary more strongly across vendors than simple append or replace loops. The goal is not to make every cell maximally powered at the first pass, but to establish a compact, reproducible panel that can detect whether the regime map is stable enough to justify larger audits.

Primary endpoint: preservation of regime ordering across vendor families. Primary success criterion: regime ordering preserved in at least 3 of 4 vendor families with N=20-60 trajectories per cell across O1, D1, D2. Secondary endpoints should include perturbation response curve shape, recurrence and persistence rank ordering, and the drift of ED50 and basin-score estimates relative to the reference implementation.

The audit should preserve the separation between verdicts and numbers used in the present analysis. Exact recurrence rates, basin scores, and ED50 values should be expected to drift because of provider-specific sampling behavior, tokenizer differences, and alignment tuning. The load-bearing test is qualitative: whether the regime ordering, perturbation response curves, and persistent geometry remain aligned under the same nudge definitions, the same memory limits, and the same analysis code. A negative result would also be informative. If one vendor reverses append and replace behavior, or if dialog perturbations dominate only in some families, the framework would identify where scaffold and model properties interact rather than treating susceptibility as a single scalar model trait.

\subsection{Logprob-based barrier heights and basin geometry}

\textit{Hypothesis: switching thresholds align more tightly with conditional surprisal than with token count.}

A second priority is to move from token barriers to information-theoretic barriers. Token dose is easy to control, audit, and interpret, but the natural cross-model unit is conditional surprisal. Future runs should store generation logprobs whenever provider APIs expose them, allowing perturbation cost to be expressed in nats as well as in tokens. The pipeline already anticipates this through a logprob-capture option, but availability remains uneven across providers, models, and endpoint types.

Logprob-based barriers would test whether behavioral switching thresholds align more directly with information cost than with perturbation length. A short perturbation containing highly surprising content may impose a larger effective kick than a longer perturbation made of predictable boilerplate. Conversely, long in-distribution perturbations may contain many tokens while adding relatively little conditional surprise. Measuring both units would separate length effects from information effects and would make cross-model comparison less dependent on tokenizer granularity.

Primary endpoint: improvement in threshold alignment when perturbation dose is measured in conditional surprisal rather than token count. Secondary endpoints should include token ED50, surprisal ED50, and the correspondence between information barriers and embedding-derived basin barriers. A practical analysis would fit switching curves in both units and compare uncertainty-normalized threshold dispersion across models and perturbation classes. If surprisal thresholds cluster more tightly than token thresholds, this would indicate that the effective perturbation dose is better measured by information cost. If they do not, token length may remain the more operationally relevant unit for scaffold evaluation.

The same extension should be applied to basin geometry. The empirical \(V^\star\) barrier summarizes density geometry in an embedding projection, whereas an information barrier would summarize how costly it is for the generator to move from one behavioral basin to another under its own predictive distribution. Agreement between these quantities would strengthen the structural interpretation of basins as both geometric and generative objects. Disagreement would identify cases where embedding-space proximity and generative difficulty diverge. For example, two states may be close in semantic embedding space but separated by a high generative barrier if the transition requires an unlikely commitment, refusal reversal, or role change. Conversely, distant embedding states may be easy to traverse if the scaffold repeatedly primes the transition.

The logprob program should also record where probabilities are unavailable, truncated, or provider-normalized in incompatible ways. These limitations should be treated as part of the measurement result rather than as incidental missingness. A future standard for perturbation audits should specify which logprob fields were exposed, whether they correspond to generated tokens only or also prompt tokens, how refusals and tool calls are represented, and whether probability traces can be reproduced under fixed seeds or fixed decoding settings.

\subsection{Memory-policy ablations and adversarial perturbations}

\textit{Hypothesis: context-update rules, not only base models, determine persistent-escape probability.}

The memory policy should become an explicit experimental factor. The present bounded-memory setting is intentionally simple, but real systems use rolling windows, summaries, pinned instructions, retrieval, tool-output buffers, and hybrid context stores. Future experiments should cross the same generator and task with multiple context-update rules so that model fragility can be separated from scaffold fragility. Beyond asking whether a perturbation changes the next response, the experiment must measure whether the context-update rule preserves, amplifies, attenuates, or quarantines the perturbation over subsequent turns.

The core ablation should use a factorial matrix:

\begin{itemize}
  \item Memory policy: full append / rolling window / summary replacement / pinned-plus-rolling hybrid
  \item Perturbation position: early context / latest turn / summary / pinned instruction / tool output
  \item Perturbation type: irrelevant long text / misleading explanation / malicious package text / false log
  \item Outcome: immediate switch, persistence, recovery, recurrence
\end{itemize}

Primary endpoint: persistent-escape probability as a function of memory policy. Secondary endpoints should include immediate switching, recovery after neutral continuation, and recurrence after apparent recovery. This decomposition is essential because a scaffold may reduce immediate switching while increasing recurrence, or may recover quickly from ordinary irrelevant text while remaining vulnerable to a false summary or poisoned tool output.

Longer contexts and longer per-step outputs should be included after the minimum-viable matrix is stable. Longer recursive writes may deepen append basins, fragment replace basins, or create regimes in which short-horizon and long-horizon stability disagree. Periodic summarization may suppress benign drift while amplifying a malicious or misleading summary error. A summary that converts an injected claim into a compact durable belief can be more damaging than the original perturbation. The relevant endpoint is therefore not compression quality alone, but perturbation response under controlled dose, position, and provenance.

Perturbation position should be treated as a first-class variable. A perturbation inserted at the beginning of memory, immediately before the model response, inside a generated summary, inside a pinned instruction field, or inside a tool-output buffer may have different persistence even at the same token dose. Controlled adversarial perturbations should include irrelevant long logs, misleading documentation, targeted false explanations, malicious package-style content, and false failing-test logs. The output should be a memory-policy ablation harness that measures whether a scaffold reduces persistent escape without merely suppressing ordinary adaptation or hiding state changes in an uninspected summary.

\subsection{Publication-scale dialog and coding-agent benchmarks}

\textit{Hypothesis: dialog topology and agent scaffold introduce distinct susceptibility profiles not predicted by single-turn model behavior.}

Dialog needs its own benchmark map. The present dialog regime is only a starting point, and different conversational topologies may define different nudge operators. Drill-down dialog is the first candidate beyond the main dialog regime, but debate, role-play, brainstorming, adversarial questioning, tutor-student exchange, and multi-party deliberation may each produce distinct switching and recovery profiles. These topologies should be evaluated with the same endpoint decomposition used for recursive perturbations: raw switching, net switching above stochastic floor, persistence, recurrence, and basin geometry.

Primary endpoint: topology-specific susceptibility profile, defined as persistence adjusted for matched stochastic controls. Secondary endpoints should include raw switching, recurrence, and basin separation across dialog states. This framing prevents dialog evaluation from collapsing into a single compliance score. A topology that produces frequent immediate shifts but rapid recovery is different from one that rarely shifts but, once shifted, remains redirected for many turns. The benchmark should therefore score trajectories, not isolated answers.

\textbf{Dialog topology benchmark.} Each dialog topology should be implemented as a reproducible scaffold with fixed role instructions, turn order, memory policy, and perturbation injection point. Drill-down tasks can test whether repeated clarification creates stable commitments. Debate tasks can test whether adversarial framing induces durable position changes. Tutor-student exchanges can test whether correction loops consolidate false explanations. Multi-party deliberation can test whether a false statement becomes persistent when repeated or endorsed by another speaker. The benchmark should include paired controls for natural dialog drift, because ordinary conversation can change state even without an explicit perturbation.

\textbf{Agent scaffold benchmark.} The same endpoint decomposition can be adapted to coding-agent benchmarks, including SWE-Bench-style tasks and smaller instrumented repositories. Perturbations should be placed in repository files, issue comments, tool outputs, package documentation, failing-test logs, generated summaries, or planning traces. Each placement should be labeled by the failure mode it probes. Perturbations in prompts, plans, and issue comments primarily test reasoning susceptibility. Perturbations in summaries and rolling memory test memory persistence. Perturbations in tool outputs, test logs, package documentation, and retrieved snippets test tool-trace contamination. Perturbations in source files, tests, and repository documentation test repository grounding, meaning whether the agent remains anchored to the actual codebase rather than to an injected narrative about it.

For coding agents, paired controls should estimate the stochastic floor of patch-family variation, test-selection variation, and pass/fail variation. Perturbation runs should then measure whether the agent remains on the injected strategy after additional plan-edit-test cycles. This adaptation would distinguish ordinary run-to-run variation from durable redirection. Two agents can have the same pass rate while differing sharply in perturbation susceptibility. Likewise, the same base model can show different escape profiles under full-history append, rolling-window memory, summarized memory, or tool-output retention. A coding-agent benchmark built on this protocol would therefore separate model fragility from scaffold fragility, which current leaderboards often conflate.

\subsection{Persistent-escape barriers for safety and instruction-injection robustness}

\textit{Hypothesis: persistent escape is separable from immediate compliance and should be measured as a multi-step recovery process.}

Safety and instruction-injection experiments should measure persistent escape directly. The raw-switching ED50 reported in this paper is not a persistent-escape barrier. Raw switching measures the perturbation dose at which an immediate response changes state. Persistent escape asks whether the system remains redirected after the perturbation is no longer novel, after additional turns are generated, and after the scaffold has had opportunities to recover.

Primary endpoint: persistent escape after neutral continuation. Secondary endpoints should include raw escape, net escape above control variation, and recovery rate. Additional useful summaries include time to recovery, recurrence after recovery, and whether the escape is carried by model output, memory state, generated summary, tool trace, or pinned instruction. These experiments would not constitute a full safety evaluation; they would measure scaffold- and model-specific susceptibility to durable redirection under controlled perturbations.

This distinction is central for safety and instruction-injection robustness. A model that briefly follows a malicious instruction and then returns to the intended policy is different from a model whose memory or dialog state has been durably hijacked. Conversely, a system that refuses immediately but stores the malicious content in a summary or tool trace may create a delayed failure that is invisible to single-turn scoring. Future experiments should therefore score multi-step post-perturbation trajectories, not only the first switched response. Neutral continuation turns are especially important because they test whether the system returns to the intended basin without being explicitly corrected.

The same design can test whether alignment creates, removes, or reshapes behavioral basins. Base and safety-tuned models should be compared under the same nudge families, memory policies, and perturbation classes to determine whether safety training changes barrier height, basin count, switching geometry, or recovery dynamics. Agent frameworks should expose the context-update rule as a traceable object: which turns are retained, which are summarized, which facts are pinned, which tool outputs are marked untrusted, and which generated summaries replace prior state. Without that instrumentation, persistent-escape failures cannot be attributed cleanly to the model, the memory policy, or the surrounding scaffold.

\textbf{Program deliverables.} The proposed program should produce:

\begin{itemize}
  \item A cross-vendor benchmark suite with matched perturbation scripts
  \item A memory-policy ablation harness with traceable context-update rules
  \item Logprob-enabled barrier analysis when provider APIs permit
  \item Dialog and coding-agent perturbation datasets
  \item Public analysis code for recurrence, persistence, ED50, and basin geometry
\end{itemize}

\subsection{Decomposing model exposure from state-write persistence}

\textit{Hypothesis: switching is driven primarily by hard state persistence, model exposure during generation, or their interaction; the two channels can be separated by a factorial perturbation design.}

The present paper compares only two intervention modes for replace-mode loops: state-reset overwrite (the default, in which the perturbation literally becomes the next state) and insert (in which the perturbation is visible to the model for one generation but does not persist). These are the two extremes of a richer design space. Several intermediate and complementary protocols would sharpen the model-versus-update-rule decomposition.

\textbf{F1. Visibility × hard-write factorial.} A clean 2×2 design at fixed dose:

\begin{table}[h!]
\centering
\small
\begin{tabular}{lccl}
\toprule
condition & perturbation visible at generation? & hard-written into next state? & maps to \\
\midrule
00 baseline & no & no & natural evolution \\
10 visible-only & yes & no & existing \textbf{insert} \\
01 write-only & no & yes & existing \textbf{state-reset overwrite} \\
11 visible + write & yes & yes & most realistic for production scaffolds \\
\bottomrule
\end{tabular}
\end{table}

Conditions 10 and 01 already exist in the paper. Conditions 00 (no-perturbation control trajectory matched to the same injection step) and 11 (model sees the perturbation \textit{and} the perturbation persists into the next state) are the missing cells. The 11 cell most closely models a production summary-memory scaffold in which an attacker's text both reaches the model and survives into the rewritten context buffer. The factorial decomposition allows the paper to attribute switching to \textit{visibility}, \textit{persistence}, or their interaction rather than collapsing both channels into a single "fragility" reading. Required controls: matched neutral-text variants of each cell, exact retained-token accounting after clipping (because the clip rule may discard prior natural output once the appended perturbation is large), order as a secondary factor (lead vs tail position of the perturbation in the rewritten state), and post-injection washout scoring (final cluster after \(\geq 5\) further natural steps, not at the injection step itself).

\textbf{F2. Sustained visible exposure with hazard-rate null.} Vary the number of consecutive steps over which the perturbation is visible to the model - \(k\in\{1,3,10\}\) - while continuing to use the insert protocol so that no copy of the perturbation hard-writes into the state. This tests whether the loop can be entrained by repeated external nudges even when nothing about the perturbation is forced into the state. The controlling concern is that more exposure steps mean more independent chances to switch; if the one-step insert switching probability is \(q\), the trivially expected cumulative rate at horizon \(k\) is \(1-(1-q)^k\). The experiment is informative only if \(k\)-step exposure produces \textit{superlinear} switching relative to that hazard-rate null, which would constitute evidence of cumulative entrainment beyond independent opportunities. Required controls: the same horizon under no perturbation (spontaneous switching baseline), repeated neutral-prefix injection at the same \(k\) (matched length and structure), and a single-shot insert observed across the same total horizon (separates "more chances" from "more dwell time").

\textbf{F3. Cross-loop insert validation.} Apply the existing insert protocol in append-mode O1 (currently insert is run only in replace-mode O2/O3). Tests whether the 12-32\% insert-mode switching rate under O2/O3 reflects a property of the model's local response to one-shot visible perturbations, or whether it is replace-specific. Two outcomes carry interpretation:

\begin{itemize}
  \item \textit{Similar insert rates across regimes.} Supports the framing that insert-mode measures a regime-independent local model bottleneck, and that the gap between insert and overwrite under replace-mode is therefore attributable to the update rule.
  \item \textit{Different insert rates across regimes.} Reveals an interaction: prior context length and update history modulate the model's response to the same single-step exposure. This would weaken any claim that insert-mode purely measures model behavior and would identify state-distribution effects that the present paper does not characterize.
\end{itemize}

F3 has been run in the present paper at modest scale (250 trajectories, one config: \texttt{configs/perturbation/O1\_overwrite\_vs\_insert.yaml}) and is reported in §5.2.1. The result is summarized as: (a) the overwrite-versus-insert gap is much smaller in append-mode (14-34 pp) than in replace-mode (60-80 pp), consistent with the no-tautology framing for append; (b) insert-mode rates are regime-conditional, with O1 (append) > O2 (oscillatory replace) > O3 (absorbing replace) at matched dose. F3 appears here only because it operationalizes the same visibility/persistence decomposition that motivates F1 and F2. The current paper does \textbf{not} include F1 or F2.

\textbf{Why these three together.} F1 separates the two causal channels (visibility, persistence). F2 separates one-shot exposure from sustained exposure under matched no-persistence semantics. F3 separates regime-conditional model behavior from update-rule artifacts. Run jointly, they would replace the current two-extremes comparison with a clean attribution of switching to model behavior, state-write mechanics, exposure horizon, and their interactions. None of them is needed to validate the present paper's headline claims; they are the principled next experiments for anyone wishing to push past the overwrite/insert dichotomy toward a continuous theory of perturbation persistence in recursive loops.

\section{Data, code, and reproducibility}

\subsection{Public trajectories, artefacts, and audit trail}

The repository is organized so that the paper's claims can be traced from released trajectories through embeddings, per-experiment metrics, aggregate tables, figures, ED50 fits, and audit checks. The public release contains 37 experiments and 3.3 GB of raw trajectories. These trajectories are the canonical input for reproducing the numerical results reported in the paper.

Because hosted generation APIs may change, exact reproduction of the paper's numerical results should use the released \texttt{steps.jsonl} trajectories; regeneration of model outputs is provided as a procedural replication path rather than a bitwise reproduction path. This distinction is important: the released trajectories fix the model responses used for the paper, while rerunning generation tests whether the same experimental protocol produces comparable outcomes under the currently available hosted models.

The repository includes three audit documents that provide the main reproducibility map. \texttt{COVERAGE.csv} records which expected artefacts are present for each experiment. \texttt{EVIDENCE.md} links substantive claims to backing data files, source-code functions, and regenerating commands. \texttt{RESULTS.md} checks reported numerical claims against the canonical aggregate tables and currently reports 103/103 numerical-claim audit cells reproducing within tolerance. Additional repository documentation describes the experiment catalogue, supersession relationships among runs, and the analysis history that led to the final design.

\subsection{Codebase, licences, and runtime environment}

Code is available in the public repository <https://github.com/kaplan196883/llmattr> and archived at [Zenodo DOI to be assigned at acceptance], corresponding to the release tagged for this manuscript. Code is released under GPLv3. Trajectories, embeddings, and derived analysis artefacts are released for reuse with attribution; OpenAI-generated outputs and embeddings are subject to the provider's terms.

The project is intended to run under Python 3.11+ in a Conda-managed environment. Core dependencies include numerical, statistical, plotting, image-processing, and video-generation libraries such as numpy, scipy, scikit-learn, scikit-image, matplotlib, pandas, and imageio-ffmpeg. The repository separates reusable analysis modules, experiment definitions, command-line scripts, configuration files, tests, raw data, derived artefacts, and aggregate outputs. This separation is intended to make it possible to inspect or rerun individual components without executing the full pipeline.

The codebase contains the primitives used for embedding management, dynamical-systems metrics, perturbation analyses, aggregate summaries, ED50 fitting, plotting, and audit generation. The scripts directory contains the command-line entry points used to rebuild the paper-level outputs. Configuration files define experiment-specific parameters, while the data directory stores the released trajectories and regenerated outputs when the pipeline is rerun locally.

\subsection{Approximate cost and runtime}

At the time of analysis, regenerating canonical OpenAI embeddings cost approximately USD 30, while regenerating model generations cost approximately USD 200; these estimates depend on provider pricing and model availability. These costs are not required to verify the paper's numerical claims if the released trajectories and derived embeddings are used. They are provided to make the computational scale of a full procedural rerun transparent.

A lower-cost exact-reproduction path is to download the released trajectories and rerun local embedding-dependent and downstream analyses from those fixed inputs. Users who wish to avoid hosted embedding APIs can substitute local sentence-transformer embeddings for exploratory replication or representation ablation. Such runs should be interpreted as representation-specific replications, not exact regenerations of the canonical embedding pipeline.

Full local embedding and analysis on the released trajectories took approximately 2 hours wall-time on a 40-core reference machine in the authors' environment. This is a benchmark for scale, not a reproducibility guarantee. Runtime will vary with storage bandwidth, CPU count, memory, embedding backend, plotting options, and whether animations or perturbation visualizations are regenerated.

\subsection{Tests and claim verification}

The analysis primitives and integration paths are covered by 99 pytest tests. The standard test command is \texttt{PYTHONPATH=. python -m pytest tests/ -q}. These tests exercise reusable components and integration behavior, while the audit scripts check the paper-level claims against regenerated artefacts and aggregate tables.

The canonical reproduction path is: download raw trajectories; compute or load embeddings; run per-experiment analyses; aggregate cross-experiment tables; run the coverage and numerical-claim audits. This schematic path is implemented by the repository scripts and can be executed at different levels of completeness depending on whether the user wants to verify the published numbers, regenerate selected figures, or procedurally rerun the full experimental workflow.

Two audit entry points are load-bearing for the claim trace. \texttt{python -m scripts.build\_coverage} regenerates the artefact-coverage audit, and \texttt{python -m scripts.publication\_summary} regenerates the numerical-claim verification table, including ED50 and summary-table checks. Together with the released trajectories, tests, evidence map, and aggregate outputs, these scripts provide the end-to-end path from each reported claim to the code and data used to recreate it.

\section{Acknowledgments}

We acknowledge \texttt{gpt-4o-mini} and \texttt{text-embedding-3-small} (OpenAI),
the open-source ecosystem (numpy, scipy, scikit-learn, scikit-image,
matplotlib, pandas, imageio-ffmpeg), and the \citet{arxiv260419740}
framework for finite-time Lyapunov spectra of
sampling-based generators.

\section{Supplementary appendix}

This appendix consolidates the audit tables, mathematical proofs, implementation definitions, reproducibility commands, and practitioner-facing reporting guidance supporting the main paper. The material is organized into twelve subsections so that decision-grade results, methods details, and engineering artifacts can be audited without fragmenting the main narrative.

\subsection{Extended Data Table 1, Unified primary-results audit table}

Extended Data Table 1 consolidates the decision-grade endpoints across
regimes, including point estimates, uncertainty, source artifacts, and
caveat flags. It is placed in Extended Data because it functions as an
audit map for reproducibility and interpretation, while the main
Results text reports the load-bearing measurements directly.

The central numerical story of this paper, in four numbers: O1 adversarial $\mathrm{ED50}_{\mathrm{raw}} \approx 40$ tokens; control-vs-control stochastic floor $\approx 35\%$; net switching saturates at +32 percentage points and never reaches the +50 pp threshold; persistent escape never crosses 50\% in the tested 5-400 token range, with 16\% as the maximum under canonical k=12 clustering at 400 tokens. The audit table below provides the supporting per-regime evidence; the rest of §5 stress-tests each of these numbers individually.

For audit, we consolidate all primary endpoints across all four
diagnostic regimes (O1, O2, O3, D1) into a single table. Each row
is a regime × endpoint; each column is the value, its uncertainty,
the source-CSV file, and any caveats from the revision. \textbf{D2 is
omitted} because it is exploratory-scale (n=25, no publication-
scale measurements) and does not satisfy the operational attractor
criteria (§3.1.3). For each endpoint we cite the §-section where
the original numbers appear and a status flag indicating whether
the endpoint has been re-validated under the revision's
leakage-aware / cluster-aware analyses.

\textbf{Sample sizes (frozen).} Operator regimes (O1, O2, O3): n =
15 prompt families × 30 ICs × 3 runs = 1,350 trajectories per
regime, 40 steps. Dialog regime D1: n = 5 dialog-suitable families
× 30 ICs × 3 runs = 450 trajectories per regime, 40 steps (see
§4.2 for the per-regime IC selection rule). Perturbation pilots
(O1/O2/O3/D1 + D2): reduced scope, n = 50 trajectories per
condition (5 fams × 5 ICs × 2 runs).

\begin{table}[h!]
\centering
\small
\begin{tabularx}{\textwidth}{lYYYYY}
\toprule
regime & endpoint & value & 95\% CI / uncertainty & source & status \\
\midrule
O1 & basin predictability acc(k=10), stratified & \textbf{0.80} & n=1350, 5-fold CV & §5.3 & [!] \textbf{inflated by family leakage; group-aware = 0.73} \\
O1 & basin predictability acc(k=10), group-aware & \textbf{0.73} & family-cluster GroupKFold & §5.9 (this revision) & [OK] leakage-free \\
O1 & recurrence rate (canonical embedder) & 0.29 & bootstrap 95\% CI & §5.9 & [OK] embedder-robust (§5.9: 0.30 / 0.10) \\
O1 & sharpness dimension (late) & 1.70 & trajectory-level & §12.2 & [OK] \\
O1 & Lyapunov $\lambda_1^{\mathrm{late}}$ & $\sim 0.008$ & ensemble-spread method & §4.5.5 & [OK] contractive \\
O1 adv & switching @ dose 200 (dense) & \textbf{0.620} & Wilson [0.55, 0.68], n=200 & §5.1 & [OK] dense rerun \\
O1 adv & switching @ dose 400 (dense) & \textbf{0.670} & Wilson [0.60, 0.73], n=200 & §5.1 & [OK] dense rerun \\
O1 adv & $\mathrm{ED50}_{\mathrm{raw}}$ (dense) & \textbf{36-52 tok} & 4PL=36; GLMM=41; family-cluster bootstrap median=52, 95\% CI [8.5, 242] & §5.1 / §3.1.2 & [OK] established \\
O1 adv & upper asymptote (dense 4PL) & \textbf{$a = 0.69$} & non-switching subpopulation ~31\% & §5.1 & [OK] \\
O1 adv & natural floor (control-vs-control, dense) & \textbf{0.347} & [0.310, 0.386], n=600 paired comparisons & §5.1 & [OK] established \\
O1 adv & $\mathrm{ED50}_{\mathrm{net}}$ (dense, raw - floor) & \textbf{not reached} & max net effect = +0.323 at dose 400 (50 pp threshold) & §3.1.2 & not reached in tested range \\
O1 adv & $\mathrm{ED50}_{\mathrm{persist}}$ (dense, kicked-AND-persisted) & \textbf{undefined} & max persistent escape = 16\% at dose 400 & §5.1 (dense) / §3.1.2 & not reached in tested range \\
O1 neutral & switching @ dose 200 & 0.24 & Wilson [0.13, 0.38] & §5.4 / Fig 10 & [OK] sparse pilot \\
O1 neutral & switching @ dose 400 & 0.18 & Wilson [0.08, 0.32] & §5.4 / Fig 10 & [OK] sparse pilot \\
O1 lorem & switching @ dose 200 & 0.18 & Wilson [0.08, 0.32] & §5.4 / Fig 10 & [OK] sparse pilot \\
O1 attractor classification & C1-C4 strong attractor & 4/4 PASS & criteria from §3.1.3 & §3.1.3 & [OK] \\
O2 & basin predictability acc(k=10), stratified & \textbf{0.90} & n=1350, 5-fold CV & §5.3 & [!] \textbf{inflated by family leakage; group-aware = 0.60} \\
O2 & basin predictability acc(k=10), group-aware & \textbf{0.60} & family-cluster GroupKFold & §5.9 & [OK] leakage-free \\
O2 & recurrence rate & 0.875 & bootstrap & §5.9 / §5.9 & [OK] embedder-robust (0.71 / 0.78) \\
O2 & sharpness dimension (late) & 1.39 & trajectory-level & §12.2 & [OK] \\
O2 & switching adversarial (n=50, single dose) & 0.94 & Wilson [0.84, 0.98] & §5.5 & [OK] \\
O2 & switching neutral / lorem & 1.00 / 1.00 & Wilson [0.93, 1.00] each & §5.5 & [OK] \\
O2 attractor classification & C1-C4 strong attractor & 4/4 PASS & criteria from §3.1.3 & §3.1.3 & [OK] but see §5.8: basins are paraphrastic, not absorbing \\
O3 & basin predictability acc(k=10), stratified & \textbf{0.91} & n=1350, 5-fold CV & §5.3 & [!] \textbf{inflated by family leakage; group-aware = 0.63} \\
O3 & basin predictability acc(k=10), group-aware & \textbf{0.63} & family-cluster GroupKFold & §5.9 & [OK] leakage-free \\
O3 & recurrence rate & 0.92 & bootstrap & §5.9 / §5.9 & [OK] embedder-robust (0.85 / 0.86) \\
O3 & sharpness dimension (late) & 1.45 & trajectory-level & §12.2 & [OK] (note §6.6 historical "≈ 0" claim was wrong; corrected) \\
O3 & switching adversarial / neutral / lorem & 0.96 / 1.00 / 1.00 & Wilson 95\% (n=50) & §5.5 & [OK] \\
O3 attractor classification & C1-C4 strong attractor & 4/4 PASS & criteria from §3.1.3 & §3.1.3 & [OK] but see §5.8: absorbing is template-formal, not semantic \\
D1 & basin predictability acc(k=10), stratified & 0.60 & n=450, 5-fold CV & §5.3 & [!] \textbf{inflated by family leakage; group-aware = 0.34} \\
D1 & basin predictability acc(k=10), group-aware & \textbf{0.34} & family-cluster GroupKFold & §5.9 & [OK] leakage-free (~chance for 11 classes is 0.09; signal is real but weak) \\
D1 & recurrence rate & 0.21 & bootstrap & §5.9 / §5.9 & [OK] embedder-robust (0.34 / 0.23) \\
D1 & sharpness dimension (late) & 1.89 & trajectory-level & §12.2 & [OK] \\
D1 & T-stability (acc range over T ∈ {0.3, 0.6, 0.8, 1.2}) & [0.57, 0.61] & reduced-scope cells, n=150 & §5.4 & [!] scope-confounded (28pp delta vs full-N) \\
D1 & switching adversarial / neutral / lorem & 0.60 / 0.76 / 0.56 & Wilson 95\% (n=50) & §5.5 & [~] granularity-sensitive (§5.10: HDBSCAN drops to 0.40 on adversarial) \\
D1 attractor classification & C1-C4 strong attractor (formal); \textbf{attractor-like dialog regime in practice} & 4/4 PASS on operational criteria, BUT: group-aware basin predictability acc(k=10) = 0.34 (well below the τ\_acc = 0.70 threshold under leakage-free CV); switching is granularity-sensitive (§5.10); semantic inspection (§5.8) finds dialogue-state / recent-context capture rather than a stable stylistic basin; neutral switching exceeds adversarial in the pilot. We retain D1 in the diagnostic taxonomy as an \textit{attractor-like dialog regime} but do not claim full strong-attractor status under group-aware criteria. & §3.1.3 + §5.9 + §5.10 + §5.8 & [!] caveat-required &  \\
\bottomrule
\end{tabularx}
\end{table}

\textbf{How to read the status column:}
\begin{itemize}
  \item [OK] \textbf{validated}, endpoint survives the revision's stress tests (group-aware CV, multi-granularity, embedder ablation, attractor criteria, dense-dose rerun where applicable).
  \item [!] \textbf{caveat-required}, endpoint as originally reported is overstated; revised value or interpretation in cited subsection.
  \item \textit{not reached in tested range}, the endpoint is well-defined but the experiment did not produce a value (dose grid does not reach the threshold).
\end{itemize}

The \textbf{two most defensible quantitative claims} in the paper are:
\begin{enumerate}
  \item Under leakage-free GroupKFold, O1's contractive basin is
   predictable at acc(k=10) = 0.73 (down from the originally
   reported 0.80), still well above any plausible chance baseline
   ($\sim 1/12$ for K-means $k=12$) and still the highest of the
   four regimes under the stress-test analysis.
  \item The dense-dose rerun establishes a raw-switching
   $\mathrm{ED50}_{\mathrm{raw}} \approx 40$ tokens for O1 against
   in-distribution adversarial text, with a population
   decomposition (these are aggregate components of the observed
   rate, not latent subpopulations): 35\% stochastic floor from
   control-vs-control pairs, plateau at ~67\% suggesting a ~31\%
   non-perturbable component, persistent-escape rate 10\% (k=4) /
   16\% (k=12) / 39.5\% (HDBSCAN) at dose 400 (§5.1), and the
   remaining ~18\% transient escape (kicked but drifted back).
   This replaces the earlier "150-token barrier" claim with a
   richer characterization that is empirically grounded; the strict
   $\mathrm{ED50}_{\mathrm{persist}}$ barrier is \textit{not} reached in
   the tested 5-400 token range under any cluster granularity.
\end{enumerate}

\subsection{Extended Data Table 2, Regime comparison at a glance}

Extended Data Table 2 provides the compact cross-regime comparison of
nudge type, content operator, basin predictability, recurrence,
sharpness dimension, perturbation response, dose scale, and
temperature sensitivity. It is placed in Extended Data to preserve the
original lookup table without interrupting the narrative sequence of
the Results section.

Before walking through the experiment phases, a master comparison
of the diagnostic signatures across regimes. Each row is a regime;
each column is a diagnostic that distinguishes it from the others.
All numbers are at publication scale (Phase 2) or perturbation pilot
scope (Phase 3).

\begin{table}[h!]
\centering
\small
\begin{tabularx}{\textwidth}{lYYYYYYYY}
\toprule
regime & nudge & content op. $f$ & basin pred. acc(k=5→final) & recurrence & sharpness dim* & adversarial switch & dose 50\% & T-stability \\
\midrule
\textbf{O1} contractive & append & continue & 0.77 → 0.85 & low & 1.70 & 54\% (sparse) / 62\% (dense, n=200) & $\mathrm{ED50}_{\mathrm{raw}}$ ≈ 40 tok (4PL/GLMM/bootstrap), plateau ~67\%, natural floor ~35\% & degrades smoothly \\
\textbf{O2} oscillatory & replace & paraphrase & 0.90 → 0.91 & high (period-2) & 1.39 & 94\% & n/a (saturated) & (not measured) \\
\textbf{O3} absorbing & replace & summarize+negate & 0.92 → 0.93 & trivial & 1.45 & 96\% & n/a (saturated) & (not measured) \\
\textbf{D1} dialogue-state-driven multi-basin & dialog (append) & curious + helpful & n/a → 0.77 & low (per-style) & 1.89 & 60\% & < 5 tokens & T-stable \\
\textbf{D2} drill-down & dialog (append) & explorer drill-down & (not measured) & (not measured) & (not measured) & 64\% & (not measured) & (not measured) \\
\bottomrule
\end{tabularx}
\end{table}

* Sharpness dim is computed on a 2-element Lyapunov spectrum (rank ≤ N-1 = 2 for N=3 runs per IC), so values are bounded above by 2.0. Mean SD\_late on \texttt{context\_tail}. The \textit{ordering} across regimes is informative, the absolute magnitudes are constrained by the rank ceiling. See §4.5.6.

** D2 was run at exploratory scale (N=1 run per IC), which is below the N≥2 minimum required for ensemble-spread Lyapunov computation. D2's basin-predictability acc at k=5 is 0.20 with n=25 and 11 classes (chance ≈ 0.09), well underpowered for the canonical k=5,10,20,final probes.

Reading: the two \textbf{replace-mode} regimes (O2, O3) lock in early (acc
already ≈0.9 by step 5) and are perturbation-transparent. The
\textbf{append-mode} regimes (O1 and the dialog regimes D1/D2) admit
slower late-state determination and have measurable barrier structure.
O1 is uniquely T-sensitive; D1 is uniquely T-stable; D2 adds content
gravity beyond D1's dialogue-state basins (see §5.8).

The regime ordering, replace-mode locks in fast and capitulates
to any perturbation; append-mode locks in slowly and resists
out-of-distribution perturbation but yields to in-distribution
adversaries, runs through every diagnostic below.

\subsection{Full proof of Lemma 1 (Replace-mode hitting bound)}

\textbf{Lemma 1} (statement in §3.1.4).

\textbf{Proof.} Let $\mathcal{F}_s$ be the natural filtration and define
$A_k = \{\sigma_2 > t_{\mathrm{inj}} + k\}$. On $A_k$, the state
$X_{t_{\mathrm{inj}}+k}$ is reachable and outside $B_2$, so assumption
(1) gives

\begin{equation*}
\Pr(A_{k+1} \mid \mathcal{F}_{t_{\mathrm{inj}}+k})
\le \mathbf{1}_{A_k} (1 - q_0).
\end{equation*}

Taking expectations yields $\Pr(A_{k+1}) \le (1 - q_0)\,\Pr(A_k)$,
hence $\Pr(A_m) \le (1 - q_0)^m$ by induction. This proves the
hitting bound. For terminal membership, decompose over the first
hitting time and use the Markov property together with assumption (2):

\begin{equation*}
\Pr(X_T \in B_2)
\ge \sum_{s = t_{\mathrm{inj}}+1}^{T}
\Pr(X_T \in B_2,\ \sigma_2 = s)
\ge r_0 \, \Pr(\sigma_2 \le T) .
\end{equation*}

Combining with the hitting bound gives the displayed terminal bound.
Finally, by assumption (3) and the tower property,

\begin{equation*}
\mathbb{E} G_m
= \sum_{s = t_{\mathrm{inj}}}^{T-1}
\mathbb{E}\bigl[\mathbb{E}(|Y_s| \mid X_s)\bigr]
\le \kappa m . \qquad \square
\end{equation*}

\textbf{Corollary 1, full proof.} Choose $m = m_{1/2}$. Lemma 1 gives
$\Pr(X_{t_{\mathrm{inj}}+m} \in B_2) \ge \tfrac{1}{2}$ and
$\mathbb{E} G_m \le \kappa m$, so the displayed bound
$G^\star_{1/2}(B_1 \to B_2) \le \kappa\, m_{1/2}$ follows. The
explicit closed form when $0 < q_0 < 1$ and $r_0 > \tfrac{1}{2}$
follows from solving $r_0 [1 - (1-q_0)^m] \ge \tfrac{1}{2}$ for $m$
in the integers. The first-hit version sets $r_0 = 1$ and uses the
hitting bound $\Pr(\sigma_2 \le T) \ge 1 - (1-q_0)^m$ in place of
the terminal bound. $\square$

\textbf{Corollary 2, full proof.} Take $m = 1$ in Lemma 1's terminal
bound: $\Pr(X_{t_{\mathrm{inj}}+1} \in B_2) \ge r_0 q_0 \ge
\tfrac{1}{2}$ when $q_0 r_0 \ge \tfrac{1}{2}$. Combined with
$\mathbb{E} G_1 \le \kappa$ from Lemma 1's third conclusion,
$G^\star_{1/2}(B_1 \to B_2) \le \kappa$ follows. The first-hit
version (with $r_0 = 1$) sets $q_0 \ge \tfrac{1}{2}$ as the
sufficient condition. $\square$

\subsection{Metric definitions, clustering, and PCA stability}

\textbf{Metric implementation audit.} This subsection collects the code-level definitions for the recurrent-state metrics used throughout §4.5, including recurrence, basin assignment, Lyapunov-spectrum estimation, sharpness dimension, and basin predictability. The executable, test-covered implementations live in the repository, but these snippets define the measurement semantics used by the paper.

Recurrence:

\begin{codeblock}
\begin{verbatim}
D = pairwise_distances(z, metric="cosine") # T x T
mask = (np.abs(np.arange(T)[:, None] - np.arange(T)[None, :]) > tau) & np.triu(np.ones((T, T)), 1).astype(bool)
recurrence = (D < epsilon)[mask].sum() / mask.sum()
\end{verbatim}
\end{codeblock}

Basin score:

\begin{codeblock}
\begin{verbatim}
clusters = KMeans(n_clusters=12).fit_predict(z_pca10)
target_cluster = mode(clusters[t > 0.7 * T])
basin_score = (clusters[t > 0.7 * T] == target_cluster).mean()
\end{verbatim}
\end{codeblock}

Lyapunov spectrum (sample-driven):

\begin{codeblock}
\begin{verbatim}
# z_runs : (n_runs, T, d_pca)
for t in range(T):
    centered = z_runs[:, t, :] - z_runs[:, t, :].mean(axis=0)
    sigmas = np.linalg.svd(centered, full_matrices=False, compute_uv=False)
    lambda_t = np.log(sigmas) / 2.0 # (d_pca,)
\end{verbatim}
\end{codeblock}

Sharpness dimension (Tuci-style fractional dimension on the ordered
Lyapunov spectrum; see §4.5.6):

\begin{codeblock}
\begin{verbatim}
lam = np.sort(lambda_t)[::-1]
cumsum = np.cumsum(lam)
nonneg = np.where(cumsum >= 0)[0]
if lam[0] < 0:
    SD_t = 0.0
elif len(nonneg) == len(lam):
    SD_t = float(len(lam)) # full-d case
else:
    j_star = int(nonneg[-1]) + 1 # 1-indexed
    SD_t = j_star + cumsum[j_star - 1] / abs(lam[j_star])
\end{verbatim}
\end{codeblock}

Basin predictability:

\begin{codeblock}
\begin{verbatim}
y = mode(clusters[t > 0.7 * T]) # late-window cluster per trajectory
acc_curve = np.zeros(T)
for k in range(T):
    X = z_pca10[:, k, :]
    clf = LogisticRegression(multi_class="auto", max_iter=1000)
    acc_curve[k] = cross_val_score(clf, X, y, cv=5).mean()
\end{verbatim}
\end{codeblock}

These are reference implementations only; the executable, test-
covered code lives at \texttt{src/analysis/} and \texttt{src/experiments/dynamics/}.

\begin{savenotes}
\begin{figure}[h!]
\centering
\includegraphics[width=0.95\linewidth]{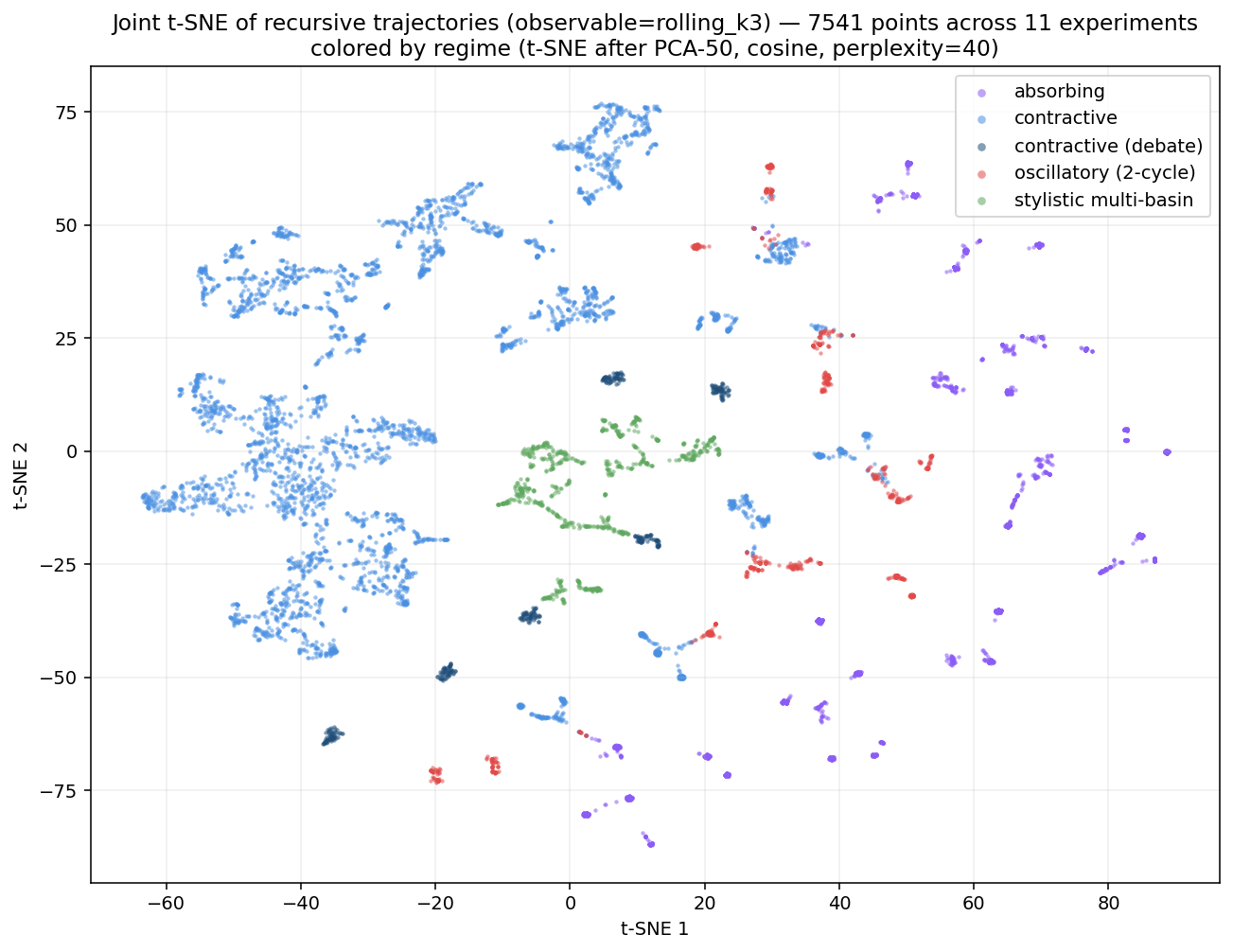}
\caption{\textbf{Joint t-SNE regime map.} Joint t-SNE visualization of all publication-scale experiments colored by regime. The view supports qualitative inspection of regime separation but is not used for quantitative endpoint claims. Source: \texttt{data/aggregated/dynamics\_plots/A\_joint\_tsne\_rolling\_k3.png}.\\[2pt]{\footnotesize\itshape ED Fig 1 shows a joint t-SNE projection (t-SNE, nonlinear dimensionality reduction preserving local neighborhood structure) of all rolling\_k3 trajectory windows (rolling\_k3, the trajectory observable formed by concatenating the last 3 outputs at each step) from the publication-scale experiments. Points are colored by regime: contractive (O1, append-mode continuation); contractive debate (a debate-structured contractive variant); oscillatory 2-cycle (O2, paraphrase replace); absorbing (O3, summarize-negate replace); stylistic multi-basin (D1, role-structured dialog); and where present, drill-down (D2, drill-down dialog). This panel is used only as a qualitative neighborhood diagnostic, not as a metric distance plot. t-SNE is designed to preserve local neighborhoods, not global distances, so the figure should be read as asking whether same-color points form coherent neighborhoods, not as measuring how far apart the visible blobs are or whether their apparent separations have quantitative meaning. Within that limitation, the plot provides a useful stability check for the regime taxonomy used in section 11.4: labels assigned from bulk diagnostics tend to occupy coherent local regions of the projection, indicating that they capture genuine local trajectory similarity rather than arbitrary post-hoc clustering. The relevant visual evidence is the repeated formation of same-regime neighborhoods across experiments, with limited intermixing compared with the overall spread. A falsifying outcome would have been heavy color mixing throughout local neighborhoods, which would have weakened the claim that the regime labels reflect reproducible trajectory structure.}}
\end{figure}
\end{savenotes}

\begin{savenotes}
\begin{figure}[h!]
\centering
\includegraphics[width=0.95\linewidth]{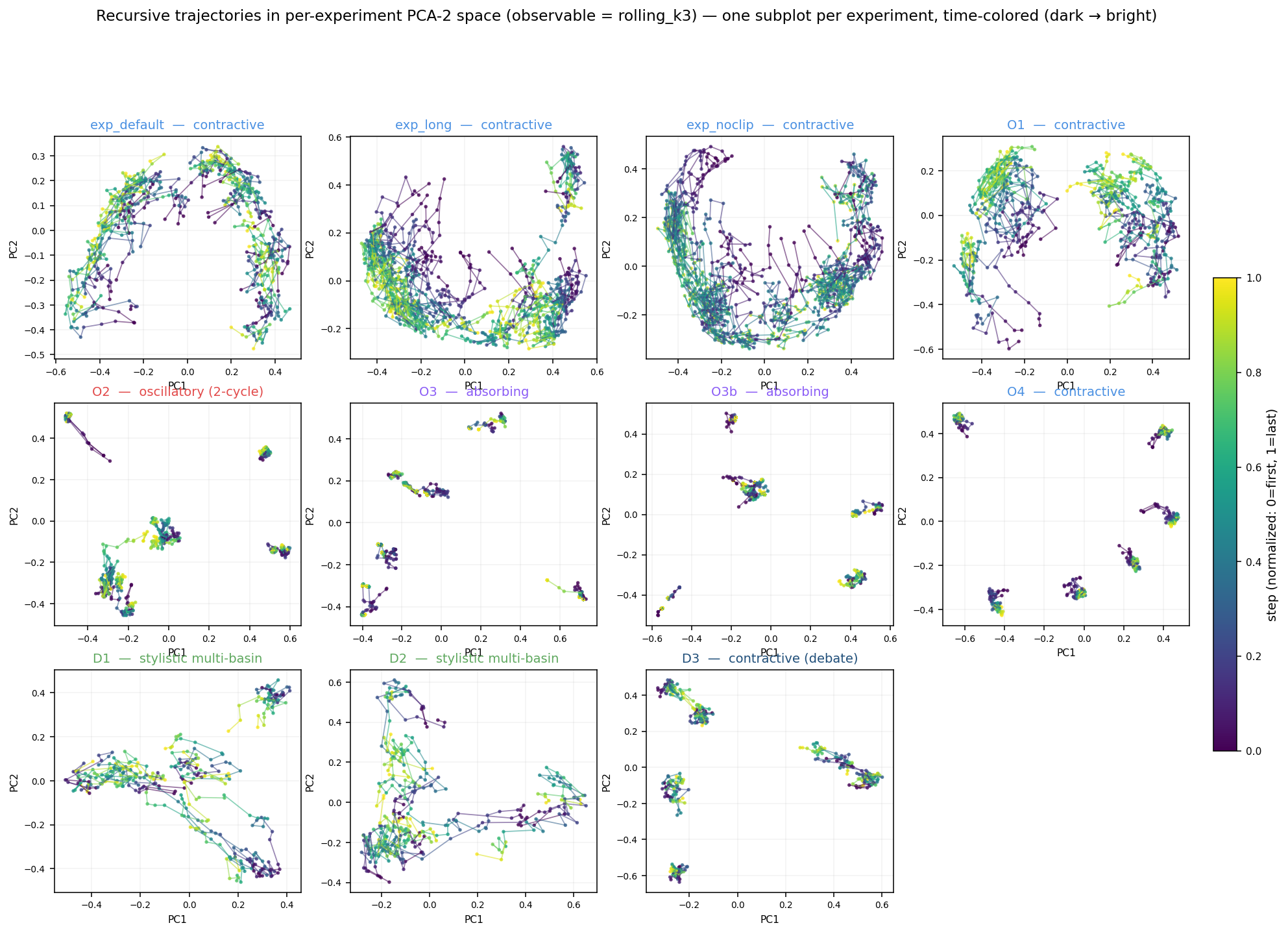}
\caption{\textbf{Per-family trajectory grid.} Shared-coordinate trajectory grid by prompt family. The figure supports visual inspection of family-level heterogeneity without serving as a primary endpoint. Source: \texttt{data/aggregated/dynamics\_plots/B\_trajectory\_grid\_rolling\_k3.png}.\\[2pt]{\footnotesize\itshape ED Fig 2 shows a per-family panel grid (family, the prompt family used as a set of initial conditions, such as philosophy\_dialog, practical\_dialog, creative\_dialog, reflective, or emotional). Each small panel plots recursive trajectories for one prompt family using rolling\_k3 (the trajectory observable formed by concatenating the last 3 outputs) in PCA-2 space (the first two principal components of the embedding ensemble). Point color encodes normalized time, with dark points early and bright points late. The figure is intended as the visual check for family-level heterogeneity: families do not merely sample the same attractor with different noise, but occupy different regions of the projected trajectory space and exhibit visibly different motion patterns. Some panels show compact basins with short local wandering, others show broad arcs or crescent-like sweeps, and others break into separated islands or multi-basin structure. This matters for the main analyses because PCA preserves family-distinctive structure rather than erasing it, so apparent separability or predictability in pooled projections can partly reflect prompt-family identity. ED Fig 2 therefore motivates the family-leakage caveat discussed for Fig 8 and Fig 9. It also provides the visual basis for the family-heterogeneity interpretation in Fig 13, where per-family ED50 estimates vary widely, with confidence interval [8.5, 242]. A falsifying pattern would have been a homogeneous grid in which all families occupied the same regions and showed the same trajectories; that result would have weakened both the leakage explanation and the claim that ED50 differences reflect genuine family heterogeneity.}}
\end{figure}
\end{savenotes}

\begin{savenotes}
\begin{figure}[h!]
\centering
\includegraphics[width=0.95\linewidth]{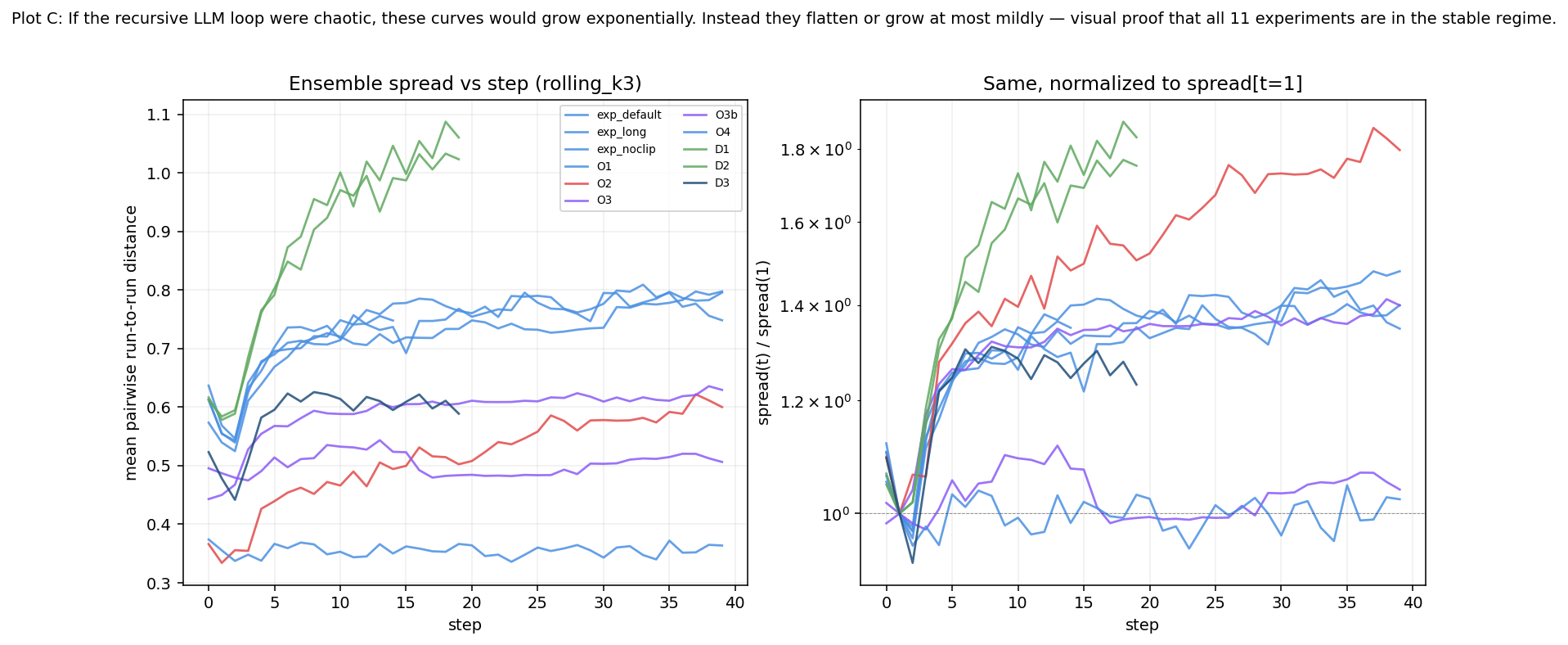}
\caption{\textbf{Ensemble-spread timeline.} Ensemble spread over recursive steps, grouped by regime. The plot supplements the finite-time ensemble-spread diagnostics used in the attractor audit. Source: \texttt{data/aggregated/dynamics\_plots/C\_spread\_timeline\_rolling\_k3.png}.\\[2pt]{\footnotesize\itshape ED Fig 3 plots the empirical ensemble spread used in section 11.4. The left panel shows raw spread at each recursive step, measured as mean pairwise run distance after rolling\_k3 (the trajectory observable formed by concatenating the last 3 outputs). The right panel shows the same trajectories normalized by their step-1 spread, so relative growth can be compared after removing initial-scale differences. Legend labels are exp\_default (the default experimental family), exp\_long (the long-context variant), exp\_noclip (the unclipped variant), O1 (append-mode continuation, a contractive regime), O2 (paraphrase replace mode, an oscillatory two-cycle), O3 (summarize-negate replace mode, an absorbing regime), O3b (a replicate absorbing summarize-negate condition), O4 (an additional output transformation condition), D1 (role-structured dialog, a multi-basin regime), D2 (drill-down dialog), and D3 (a third dialog condition). Ensemble spread sigma\_t denotes dispersion across N=3 runs sharing the same family and initial condition at step t; with only N=3 runs, the covariance rank is at most 2, so these curves should be read as finite-ensemble diagnostics rather than full-dimensional attractor estimates. Contractive regimes plateau early, absorbing regimes settle even faster, the oscillatory condition shows bounded drift, and dialog regimes show the clearest sustained growth. The key point is falsifiability: a chaotic recursive LLM loop would produce runaway, approximately exponential separation, whereas these trajectories flatten or grow only mildly. Thus ED Fig 3 supplies the empirical substrate for the late finite-time Lyapunov-like exponents reported in Fig 9.}}
\end{figure}
\end{savenotes}

\subsection{Perturbation injection mechanics (full)}

For dialog-mode experiments, the injection happens at the user-turn
step (odd-numbered if Explorer initiates). The injection text replaces
the user turn's output verbatim. The trajectory then continues with
the agent's response to this overridden user turn, and from there
back to normal alternation.

For operator-mode experiments, the injection text replaces step k's
output entirely. The recurrence picks up from $X_{k+1} =
\mathrm{clip}(X_k \Vert Y_k^{\text{inj}})$ (append) or $X_{k+1} =
\mathrm{clip}(Y_k^{\text{inj}})$ (replace).

The adversarial-source experiment for each regime is taken from the
\textit{publication-scale} run of the same regime (e.g., O1 perturbation
adversarial draws from \texttt{exp\_pub\_O1\_continue}). Adversarial samples
exclude the \textit{family} of the trajectory being perturbed but may share
its overall topic distribution.

\textbf{Insert vs overwrite modes.} The \texttt{adversarial\_doseN} condition
implements \textit{state overwrite} (the operator output for step $k$ is
replaced wholesale by the injection text). A second
condition \texttt{adversarial\_insert\_doseN} implements \textit{context insertion}
(the injection text is prepended to the context for step $k$'s
generation, but the model's normal generated output is preserved
and used as $Y_k$). The runner code at \texttt{src/experiments/perturbation/runner\_op.py}
dispatches between the two modes based on the substring \texttt{\_insert\_}
in the condition name.

\subsection{Exact prompt templates and model versioning}

\textbf{Prompt and model audit.} This subsection records the exact prompt templates, perturbation-condition wording, and resolved model versions used for the publication-scale and ablation experiments. Keeping prompts and model snapshots together makes the generator/nudge specification auditable as a single experimental artifact.

\subsubsection*{Exact prompt templates (full)}

The full prompt-family seed library (15 families × 30 ICs each = 450
unique seed sentences) lives in \texttt{configs/prompt\_library\_publication.yaml}
and is reproduced verbatim by every cross-model and ablation run.

\textbf{O1, Continue (append):}

\begin{asciiblock}
\begin{verbatim}
Continue the text naturally. Do not summarize or explain.
\end{verbatim}
\end{asciiblock}

\textbf{O2, Paraphrase + replace:}

\begin{asciiblock}
\begin{verbatim}
Paraphrase the following text while preserving its meaning.
Return only the paraphrase.
\end{verbatim}
\end{asciiblock}

\textbf{O3, Summarize + negate (replace):}

\begin{asciiblock}
\begin{verbatim}
Summarize the following text in one sentence, then state its
opposite. Return only the negated summary.
\end{verbatim}
\end{asciiblock}

\textbf{D1, Dialog (curious user + helpful agent), append.}
Two roles alternate; each role has its own \texttt{system\_prompt}:

\begin{itemize}
  \item \textit{role A (curious user)}: \texttt{You are a curious person. Ask one short follow-up question, in plain language, that probes deeper into the topic.}
  \item \textit{role B (helpful agent)}: \texttt{You are a thoughtful assistant. Answer briefly and clearly, in two or three sentences.}
\end{itemize}

\textbf{D2, Dialog drill-down (explorer + expert), append.}

\begin{itemize}
  \item \textit{role A (explorer)}: \texttt{You are exploring this topic. Ask the next, more specific question that drills into a single concrete subtopic.}
  \item \textit{role B (expert)}: \texttt{You are an expert. Answer briefly and concretely, then anchor the conversation to the most informative subtopic.}
\end{itemize}

\textbf{Perturbation conditions (injection format).} At step \texttt{override\_step}
(default 15), the perturbation pipeline replaces the model's normal
step-15 generation with a fixed-length adversarial sample drawn
according to the condition:

\begin{itemize}
  \item \texttt{control}, no injection (normal generation).
  \item \texttt{adversarial\_doseN}, N tokens of late-step output sampled from
  another (family, IC) trajectory of the same regime, excluding the
  current trajectory's own family. Source experiment is named in
  \texttt{perturbation.adversarial\_source\_experiment}.
  \item \texttt{adversarial\_insert\_doseN}, N tokens prepended to the context for
  one generation; model's normal output is preserved (§12.5).
  \item \texttt{neutral\_doseN}, N tokens of in-distribution but topically
  unrelated continuation drawn from a Wikipedia corpus.
  \item \texttt{lorem\_doseN}, N tokens of out-of-distribution Lorem-ipsum text.
\end{itemize}

The injection text is appended to the running context for \texttt{append}-mode
operators and replaces the generated step entirely for \texttt{replace}-mode
operators (per §12.5).

\subsubsection*{Model versioning (full table)}

OpenAI model aliases can update silently; this is the exact set used
for the publication-scale experiments:

\begin{table}[h!]
\centering
\small
\begin{tabularx}{\textwidth}{lYY}
\toprule
role & model alias & underlying snapshot \\
\midrule
primary generator & \texttt{gpt-4o-mini} & \texttt{gpt-4o-mini-2024-07-18} (resolved 2025-2026 across all publication runs; OpenAI did not retire this snapshot during the experiment window) \\
secondary generator (cross-model) & \texttt{gpt-4.1-nano} & \texttt{gpt-4.1-nano-2025-04-14} \\
canonical embedder & \texttt{text-embedding-3-small} & (single immutable model; no snapshot suffix exists) \\
ablation embedder \#1 & \texttt{text-embedding-3-large} & (same) \\
ablation embedder \#2 & \texttt{all-mpnet-base-v2} (sentence-transformers) & hugging-face \texttt{sentence-transformers/all-mpnet-base-v2} \\
\bottomrule
\end{tabularx}
\end{table}

For the cross-vendor pilot configs (under \texttt{configs/cross\_model/text01/})
the secondary generator is MiniMax \texttt{MiniMax-Text-01} via the official
MiniMax chat-completions API.

\subsection{Pilot validation: phase-0 and phase-1 taxonomy}

\textbf{Pilot-scope provenance.} Phase-0 runs validated the end-to-end pipeline, while Phase-1 small-N experiments identified the operator and dialog regimes that were later re-run at publication scale. These pilot materials are retained for auditability but are not load-bearing for the paper's decision-grade endpoints.

\subsubsection*{Phase-0 pilot validation}

The earliest pilot runs validated the pipeline end-to-end on small-N
(2-5 trajectories per regime) configurations covering the major
operator and dialog architectures. These pilots established that:
(a) the embedding pipeline produces stable PCA-2 / PCA-10 / t-SNE
projections, (b) K-means at $k = 12$ recovers visually-distinct
clusters in late-window points, and (c) recurrence and basin-score
diagnostics produce numerically sensible values. They are not
load-bearing for any §4.13 decision-grade endpoint and are summarised
here for completeness only; raw outputs at \texttt{data/exp\_op\_*\_pilot/} and
\texttt{data/exp\_dialog\_*\_pilot/}.

\subsubsection*{Phase-1 small-N taxonomy}

The phase-1 small-N runs ($n \approx 50$ trajectories per cell)
identified the three operator regimes (O1 contractive append; O2
oscillatory paraphrase/replace; O3 absorbing summarize+negate/replace)
and the two dialog regimes (D1 dialogue-state-driven multi-basin; D2 drill-down)
that became the diagnostic taxonomy at publication scale. These
small-N runs are not load-bearing, every regime claim in the main
body is now grounded in publication-scale or perturbation-pilot
data. Configurations and outputs at \texttt{configs/operators/},
\texttt{configs/dialog/}, and \texttt{data/exp\_*\_pilot/}. The boundary cases (O3b
summarize+negate at append; O4 paraphrase+append; D3 debate dialog)
are documented as pilot variants but do not satisfy the operational
attractor criteria of §3.1.3 at publication scale.

\subsection{Perturbation visualization toolkit (full implementation)}

For perturbation experiments we additionally compute:

\subsubsection*{Effective potential}

\begin{asciiblock}
\begin{verbatim}
rho?(x) = Gaussian-smoothed kernel density on PCA-2 grid
V(x) = -log(rho?(x) + eps), eps = 0.1.min{rho? : rho? > 0}
V is shifted so V_min = 0 and capped at v_cap (default 8.0)
\end{verbatim}
\end{asciiblock}

\subsubsection*{Geodesic skeleton}

We find local minima of V via 8-connected \texttt{maximum\_filter} on -V,
keeping the top n basin centers. For each pair of basin centers
(i, j) we compute the Dijkstra shortest path on the V grid
(8-connected, edge weight = V at endpoint). The maximum V along
the path is the \textbf{barrier height V*(i, j)}.

\subsubsection*{Volumetric iso-density rendering}

For 3D animations we extract iso-density shells at five density
fractions (4\%, 10\%, 20\%, 35\%, 55\% of max ρ) using
\texttt{scipy.ndimage.gaussian\_filter} smoothing and
\texttt{skimage.measure.marching\_cubes}. Each shell is rendered as a
transparent \texttt{Poly3DCollection} in \texttt{matplotlib}'s
\texttt{mpl\_toolkits.mplot3d}, with colors from the \texttt{plasma} colormap and
per-shell alpha from 0.05 (outermost) to 0.27 (innermost).

\subsubsection*{Parallel rendering}

Animations of 50 trajectories with 75 frames at DPI 180 are
rendered via \texttt{concurrent.futures.ProcessPoolExecutor} with 40
workers, each worker creating a fresh figure for one frame. Frames
are stitched into MP4 via \texttt{imageio-ffmpeg} (libx264 codec, quality
8). Wall-time per animation: ~80s vs ~11 min single-threaded.

\begin{savenotes}
\begin{figure}[h!]
\centering
\includegraphics[width=0.95\linewidth]{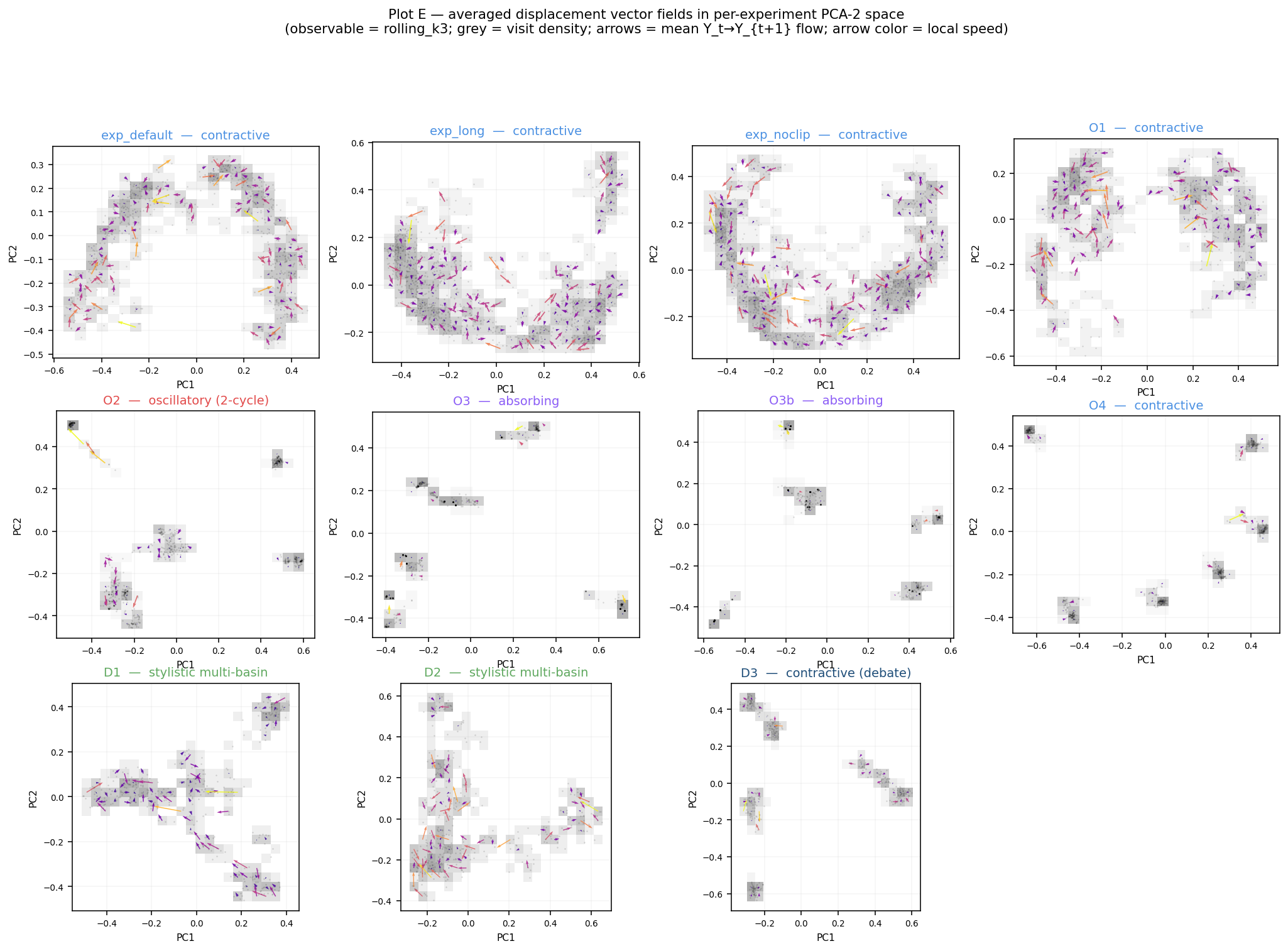}
\caption{\textbf{Combined PCA flow fields.} Combined empirical PCA-2 flow-field summary across regimes. Flow fields are useful qualitative checks on local motion but are not primary decision endpoints. Source: \texttt{data/aggregated/dynamics\_plots/E\_flow\_fields\_rolling\_k3.png}.\\[2pt]{\footnotesize\itshape ED Fig 4 shows the combined PCA-2 flow field (PCA-2, the first two principal components of the embedding ensemble; flow field, displacement vectors averaged over trajectory transitions in each grid bin of the projection) for each behavioral regime. Each panel corresponds to one regime, with gray density backgrounds indicating occupancy, arrows indicating mean one-step displacement vectors per occupied grid bin, and empty bins left undefined rather than interpolated. Arrow color reports local speed, so brighter arrows mark faster local motion. The contractive panels, including O1 (append-mode continuation), show arrows converging toward sinks (sink, local convergence or negative divergence). O2 (paraphrase replace-mode) is oscillatory, with rotational or cyclic flow rather than simple collapse. O3 (summarize-negate replace-mode) is absorbing, with very strong sinks and sparse escape flow. D1 (role-structured dialog) is stylistic multi-basin, showing several separated convergence regions instead of one dominant attractor. D2 (drill-down dialog) is also stylistic multi-basin, with structured movement among multiple occupied basins. Sources (local divergence or positive divergence) are comparatively limited and appear mainly as transient departures from occupied regions rather than stable organizing centers. Thus the flow field provides a dynamics-derived corroboration of the basin labels independently from clustering: the regimes labeled contractive, oscillatory, absorbing, and multi-basin by diagnostics also display the corresponding motion patterns in PCA-2 space. A clear falsification would have been a contractive regime with no sinks, since absence of local convergence would have weakened the contractive label.}}
\end{figure}
\end{savenotes}

\begin{savenotes}
\begin{figure}[h!]
\centering
\includegraphics[width=0.95\linewidth]{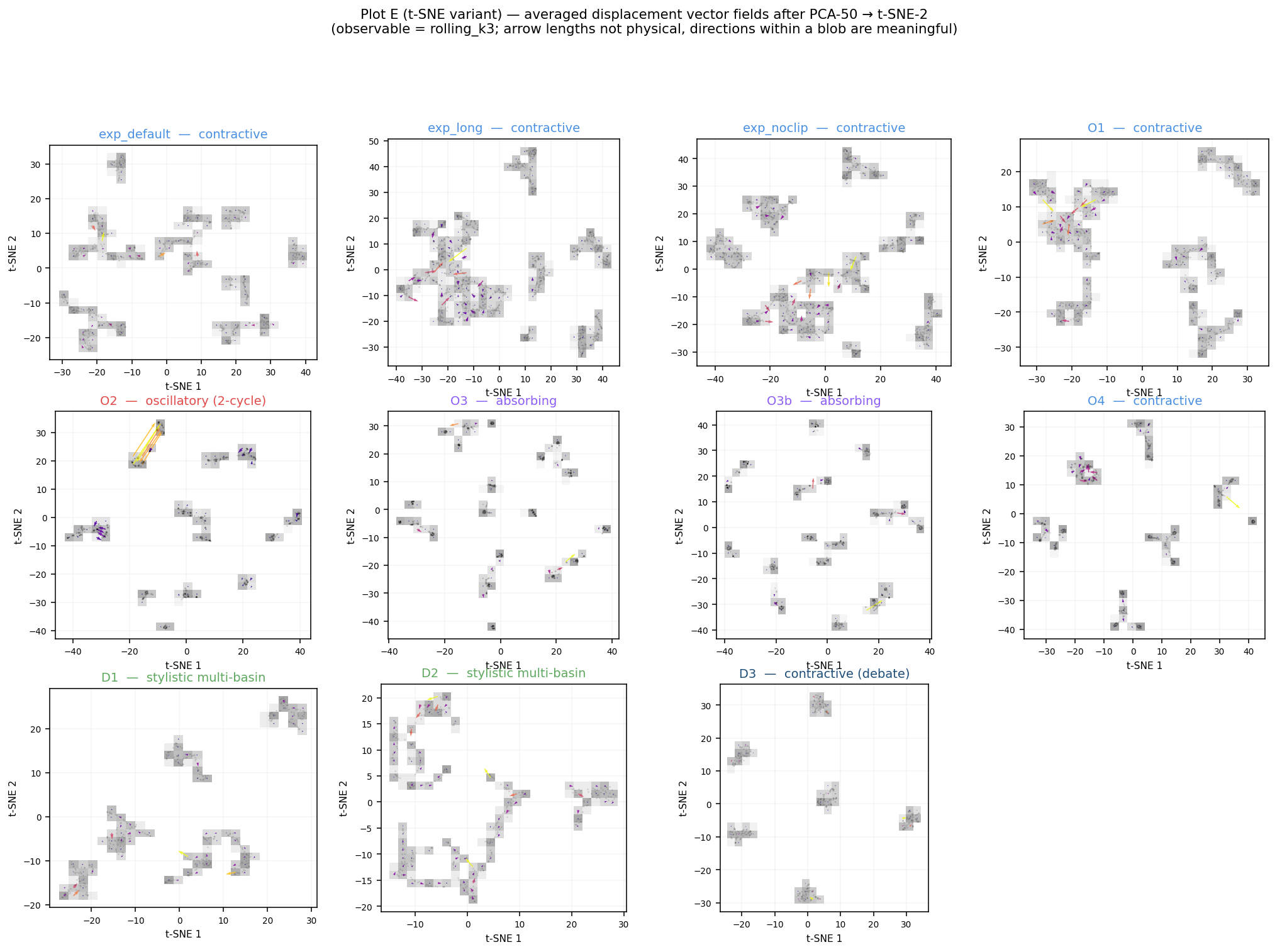}
\caption{\textbf{Combined t-SNE flow fields.} t-SNE-space flow-field visualization used for qualitative comparison with the PCA-2 flow summaries. Source: \texttt{data/aggregated/dynamics\_plots/E\_tsne\_flow\_fields\_rolling\_k3.png}.\\[2pt]{\footnotesize\itshape ED Fig 8 shows the same combined averaged displacement vector fields as ED Fig 4, but plotted in t-SNE-2 coordinates (t-SNE-2, a nonlinear dimensionality reduction that preserves local neighborhoods after PCA-50 pre-reduction) rather than PCA-2 (the first two principal components from a linear projection). This serves as a cross-projection robustness check for the regime taxonomy. The legend labels are exp\_default (contractive), exp\_long (contractive), exp\_noclip (contractive), O1 (append-mode continuation, contractive), O2 (paraphrase replace, oscillatory two-cycle), O3 (summarize-negate replace, absorbing), O3b (absorbing replicate), O4 (additional output-transformation contractive), D1 (role-structured dialog, stylistic multi-basin), D2 (drill-down dialog, stylistic multi-basin), and D3 (debate-structured contractive). Because t-SNE changes global geometry and should not be interpreted metrically, arrow lengths are not physical, but coherent directions within each local blob remain informative. The key observation is that the qualitative flow motifs visible in PCA-2 largely reappear here: contractive regimes still show local sinks or inward motion, the O2 regime still exhibits an oscillatory pattern, absorbing regimes still show trapping or one-way flow into restricted regions, and stylistic multi-basin regimes still show multiple local attractors rather than a single global sink. Thus the classification is not tied to one visualization method. A falsification would be straightforward: if a regime that looked contractive in PCA-2 became clearly oscillatory in t-SNE-2, or if a multi-basin pattern collapsed into one sink only under t-SNE, that would be a red flag for projection artifact. The cross-projection consistency therefore makes the proposed regime taxonomy more credible.}}
\end{figure}
\end{savenotes}

\subsection{Reproducibility: commands and repository tree}

\subsubsection*{Pipeline commands}

\begin{codeblock}
\begin{verbatim}
# Generate (only for new experiments)
python -m src.experiments.dialog.main run --config <cfg.yaml>
python -m src.experiments.operators.main run --config <cfg.yaml>
python -m src.experiments.perturbation.main run --config <cfg.yaml>

# Derive (works on existing steps.jsonl)
python -m src.experiments.<runner>.main embed --config <cfg.yaml>
python -m src.experiments.<runner>.main analyze --config <cfg.yaml>
python -m src.experiments.dynamics.basin_predictability --config <cfg.yaml>
python -m src.experiments.dynamics.regime_plots --data-dir data
python -m src.experiments.perturbation.flow_skeleton --experiment <exp_id> [--is-dialog]
python -m src.experiments.perturbation.geodesic_skeleton --experiment <exp_id> [--is-dialog]
python -m src.experiments.perturbation.bulk_plots --experiment <exp_id> --override-step <k> [--is-dialog]
python -m src.experiments.perturbation.rg_dendrogram --experiment <exp_id> [--is-dialog]
python -m src.experiments.perturbation.trajectory_animation_3d \
       --experiment <exp_id> --condition <c> --parallel 40

# Aggregate
python -m scripts.aggregate_perturbation_cross_regime
python -m scripts.aggregate_dose_response
python -m scripts.aggregate_basin_hardening
python -m scripts.aggregate_basin_predictability
python -m scripts.aggregate_t_sweep
python -m scripts.aggregate_o1_d1_t_sensitivity
python -m scripts.aggregate_perturbation_geometric_barriers

# Audit / catalog
python -m scripts.build_coverage # rebuild COVERAGE.csv (37 x 60 matrix)
python -m scripts.publication_summary # rebuild RESULTS.md (verify S5 cells against data)
\end{verbatim}
\end{codeblock}

\subsubsection*{Repository layout}

\begin{asciiblock}
\begin{verbatim}
llm_attractor_experiment/
+-- README.md, requirements.txt, ARTICLE.md
+-- EVIDENCE.md claim-to-evidence map (every ARTICLE claim
| <-> data file <-> source code function <-> CLI)
+-- COVERAGE.csv 37 x 60 artifact-presence matrix
+-- RESULTS.md S5 numeric-claim verification (103/103 ?)
+-- docs/
| +-- DATA_INDEX.md
| +-- reports/REPORT1.md ... REPORT6.md
+-- src/
| +-- analysis/ basin, recurrence, dwell, PCA, t-SNE, distances, ...
| +-- api/ OpenAI client + embedder + generator
| +-- core/ trajectory runner, observables, baselines, context
| +-- experiments/
| | +-- dialog/ D1/D2/D3 alternating-role runner
| | +-- operators/ O1-O4 single-role recursive operators
| | +-- dynamics/ 10 post-hoc CLI analysis modules
| | +-- perturbation/ 14 modules: runner, analyze, corpora, plot+animation
| +-- reports/ narrative report writer
| +-- utils/ io, logging, seeds, text helpers
+-- scripts/ build_publication_configs + 6 aggregators
+-- configs/ dialog/ + operators/ + perturbation/ + archive/
+-- tests/ 99 pytest tests
+-- data/ 37 experiment dirs + aggregated/ outputs
\end{verbatim}
\end{asciiblock}

\subsubsection*{Engineering memory-policy correspondences (illustrative)}

These pseudo-YAML blocks illustrate how the formal nudges of §3.1
correspond to implementable agent memory policies. They are \textit{not}
experimental conditions of this paper; they are engineering
correspondences provided for readers adapting the framework to their
own systems.

\textbf{Append-mode (full transcript):}

\begin{codeblock}
\begin{verbatim}
memory_policy:
  mode: append
  clip: tail
  max_context_chars: 12000
  include:
    - user_goal
    - recent_tool_output
    - recent_model_outputs
\end{verbatim}
\end{codeblock}

\textbf{Replace-mode (summary as state):}

\begin{codeblock}
\begin{verbatim}
memory_policy:
  mode: replace
  state_source: generated_summary
  preserve_raw_history: false
\end{verbatim}
\end{codeblock}

\textbf{Hybrid (pinned + rolling + provenance-preserving):}

\begin{codeblock}
\begin{verbatim}
memory_policy:
  mode: hybrid
  rolling_window:
    last_turns: 8
  pinned:
    - original_user_goal
    - acceptance_tests
    - security_policy
  summaries:
    older_history: extractive
    untrusted_tool_output: preserve_provenance
\end{verbatim}
\end{codeblock}

The risk profile of each policy is qualitatively distinct (§3.1 table;
§5.10 overwrite-vs-insert; §6.2 "summary as effective next state").

\subsection{Operational attractor criteria, audit table}

The C1-C4 criteria of §3.1.3 are operationally auditable. The
table below records the actual numeric values backing each
PASS/FAIL verdict. Tabulated below from publication-scale runs
(\texttt{exp\_pub\_O1\_continue} etc.; bootstrap statistics from
\texttt{metrics/bootstrap\_summary.csv}; basin-predictability from §5.9;
embedder-robustness from §5.9). Empty cells are marked "n.t." (not
directly tabulated in published artefacts at this resolution; would
need a small new aggregation script to compute).

\subsubsection*{C1, Late-window basin predictability $A^{\mathrm{final}}$}

Threshold for PASS: $A_r^{\mathrm{final}} \ge 0.70$.

\begin{table}[h!]
\centering
\small
\begin{tabularx}{\textwidth}{lYYYY}
\toprule
regime & acc(k=10) stratified 5-fold & acc(k=10) GroupKFold-by-family & leakage Δ & C1 PASS? \\
\midrule
O1 & 0.80 & 0.73 & 0.07 & PASS (both ≥ 0.70) \\
O2 & 0.90 & 0.60 & 0.30 & PASS (stratified); FAIL (group-aware) \\
O3 & 0.91 & 0.63 & 0.28 & PASS (stratified); FAIL (group-aware) \\
D1 & 0.60 & 0.34 & 0.27 & FAIL (both below 0.70) \\
D2 & 0.20 (n=25) & n.t. & n.t. & FAIL (exploratory scope) \\
\bottomrule
\end{tabularx}
\end{table}

Source: §5.9, \texttt{data/aggregated/group\_aware\_basin\_pred.csv}.

\begin{savenotes}
\begin{figure}[h!]
\centering
\includegraphics[width=0.95\linewidth]{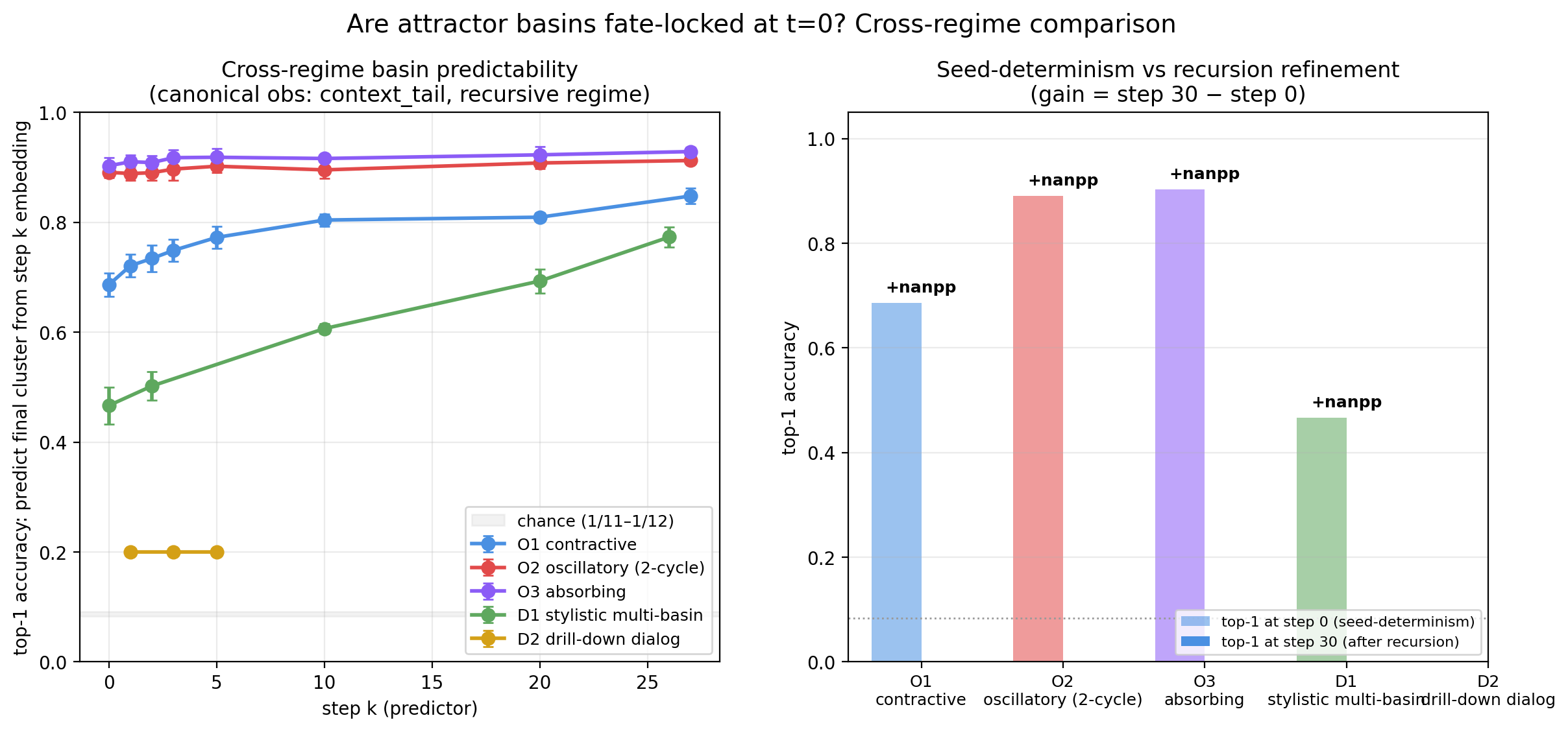}
\caption{\textbf{Original stratified basin-predictability curves.} Stratified-CV basin-predictability curves are retained for audit but should be interpreted alongside the leakage-aware GroupKFold results in main-text Fig 8. Source: \texttt{data/aggregated/basin\_predictability\_cross/cross\_basin\_predictability.png}.\\[2pt]{\footnotesize\itshape ED Fig 9 audits the original stratified k-fold cross-validation (stratified k-fold CV, a split method that can place trajectories from the same prompt family in both train and test folds and therefore permits family-fingerprint leakage) basin-predictability cross-experiment plot. The left panel reports acc(k), top-1 accuracy for predicting the final cluster from predictor step k, for O1 contractive (the append-mode continuation regime), O2 oscillatory 2-cycle (the paraphrase replace regime), O3 absorbing (the summarize-negate replace regime), D1 stylistic multi-basin (the dialog regime), and D2 drill-down dialog, with chance defined as random-label baseline accuracy (1/12 about 0.083 for 12 clusters). The right panel compares top-1 at step 0 (seed-determinism) with top-1 at step 30 (after recursion); the plus-nan-pp labels indicate that a step-30 gain is not applicable or not estimated for those bars. At k=10, the stratified accuracies are O1 0.803, O2 0.896, O3 0.912, and D1 0.604, but the leakage-resistant GroupKFold-by-prompt-family analysis (the main-text Fig 8 alternative that holds out entire prompt families) gives O1 0.732, O2 0.596, O3 0.629, and D1 0.336. Thus the stratified estimates are inflated by family-fingerprint leakage by 7, 30, 28, and 27 percentage points, respectively. ED Fig 9 is retained as an audit trail for the original analysis, not as primary evidence for cross-family generalization. The relevant falsification standard is the leakage-corrected one: operational attractor claims are weakened or falsified when GroupKFold-by-prompt-family performance falls toward chance or far below the stratified result.}}
\end{figure}
\end{savenotes}

\begin{savenotes}
\begin{figure}[h!]
\centering
\includegraphics[width=0.95\linewidth]{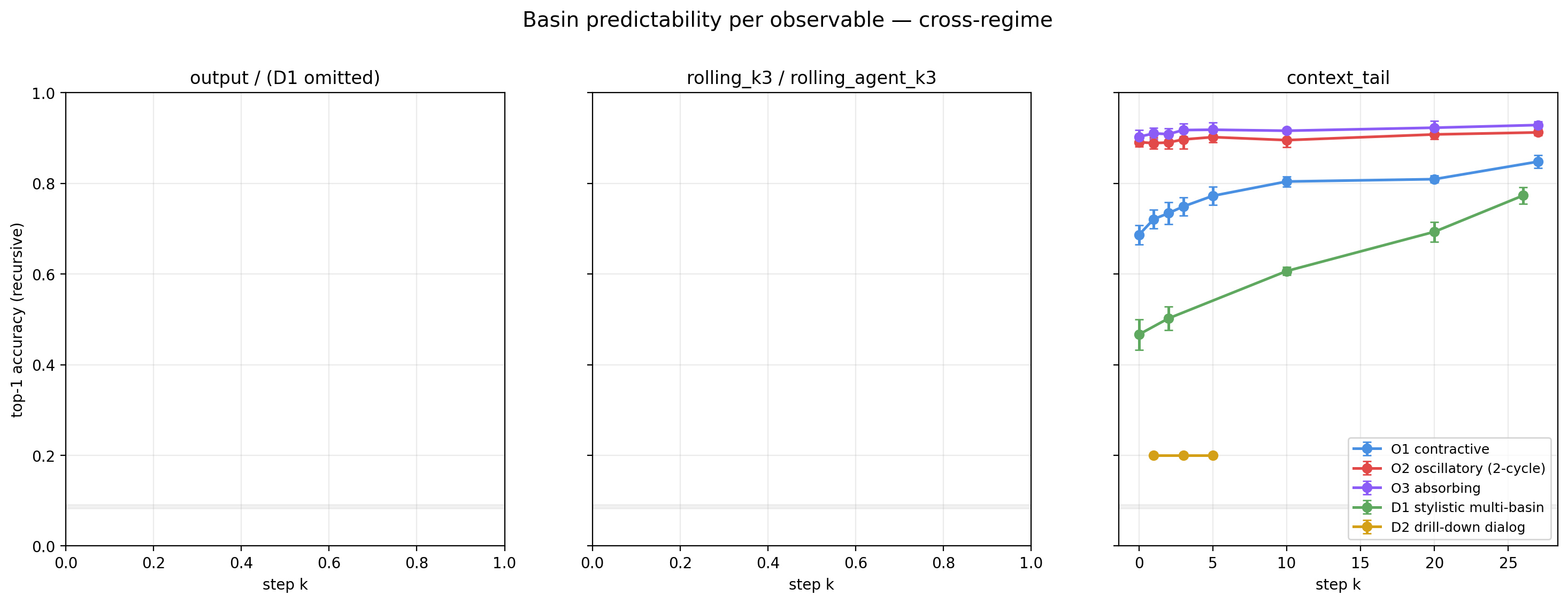}
\caption{\textbf{Basin-predictability grid.} Per-experiment basin-predictability panels showing how predictability varies across regimes and observables. Source: \texttt{data/aggregated/basin\_predictability\_cross/cross\_basin\_predictability\_grid.png}.\\[2pt]{\footnotesize\itshape ED Fig 10 reports the section 11.10 stratified-CV (stratified-CV, cross-validation allowing family leakage as an audit upper bound, see ED Fig 9) comparison of basin predictability across observable family (different text views of the trajectory state at each recursive step). The figure is a grid of small panels, one per observable family, plotting acc(k) (top-1 recursive accuracy) against predictor step k, with regime curves for O1 (append-mode continuation), O2 (paraphrase replace), O3 (summarize-negate replace), D1 (role-structured dialog), and D2 (drill-down dialog). Legend observables are output (the model's generated text alone), rolling\_k3 (concatenation of the last 3 outputs), context\_tail (last 4000 chars of the running context that integrates trajectory history), rolling\_user\_k3 and rolling\_agent\_k3 (dialog-only variants per role), and turn\_pair (last user-agent exchange concatenated). The main pattern is that context\_tail is the strongest predictor across regimes, rolling\_k3 and related rolling dialog summaries are intermediate where present, and output-only observables are weakest, while the relative ordering of regimes is preserved across observable choices. This preservation is the key scientific control: it rules out the alternative that basin predictability is an artifact of one particular text representation rather than a property of the underlying trajectory regimes. A reversal of regime ordering across observables would have falsified or at least undermined the endpoint analysis, because it would imply that the claimed basins depend on feature choice. Instead, the stable ordering justifies using context\_tail as the canonical observable for endpoint analyses.}}
\end{figure}
\end{savenotes}

\begin{savenotes}
\begin{figure}[h!]
\centering
\includegraphics[width=0.95\linewidth]{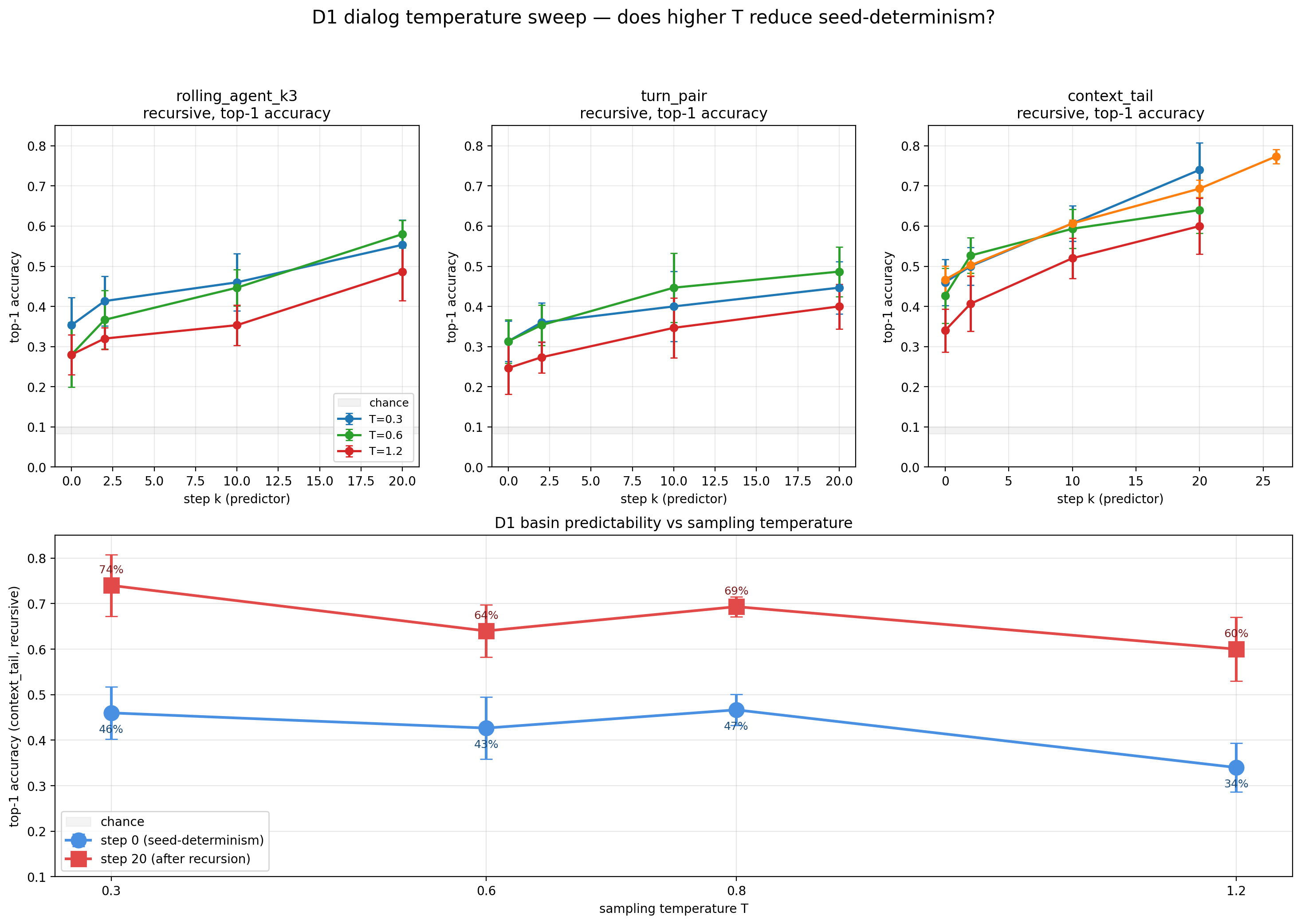}
\caption{\textbf{Temperature-sweep basin predictability.} Basin predictability as a function of sampling temperature. These reduced-scope cells are exploratory and are not used as primary evidence for temperature effects. Source: \texttt{data/aggregated/t\_sweep\_basin\_predictability/t\_sweep\_basin\_predictability.png}.\\[2pt]{\footnotesize\itshape ED Fig 11 audits basin predictability acc(k) (top-1 accuracy as a function of predictor step k, the recursive step at which the embedding is taken) across four sampling temperatures T (T=0.3, T=0.6, T=0.8, and T=1.2 denote progressively warmer sampling settings) in reduced-scope cells (lower-N temperature-sweep cells with n=150 each that are not directly comparable to the publication-scope baseline). Observable readouts include rolling\_user\_k3 (a three-turn rolling user-state dialog readout), turn\_pair (a dialog adjacent-turn-pair readout), and context\_tail (a general readout from the tail of the context). The chance band (baseline accuracy under random label assignment) is shown for reference. Step 0, labeled seed-determinism, estimates how much basin identity is already predictable from the initial sampled seed, whereas step 20, labeled after recursion, estimates predictability after recursive rollout. In these reduced-scope temperature-sweep cells, D1 remains in a narrow 0.57 to 0.61 band across temperatures, indicating true T-stability rather than a monotonic loss of seed-determinism at higher sampling temperature. O1 ranges more widely, 0.52 to 0.65, but this is treated as exploratory secondary evidence because reduced-scope T=0.8 gives 0.52 whereas the publication-scope T=0.8 baseline is 0.80, a 28 percentage-point scope-confounded delta. Thus ED Fig 11 falsifies the simple claim that higher T necessarily reduces D1 basin predictability, while the broader operational attractor criterion would be falsified by late-step acc(k) collapsing to the chance band or by temperature-driven instability large enough to erase the D1 basin signal.}}
\end{figure}
\end{savenotes}

\begin{savenotes}
\begin{figure}[h!]
\centering
\includegraphics[width=0.95\linewidth]{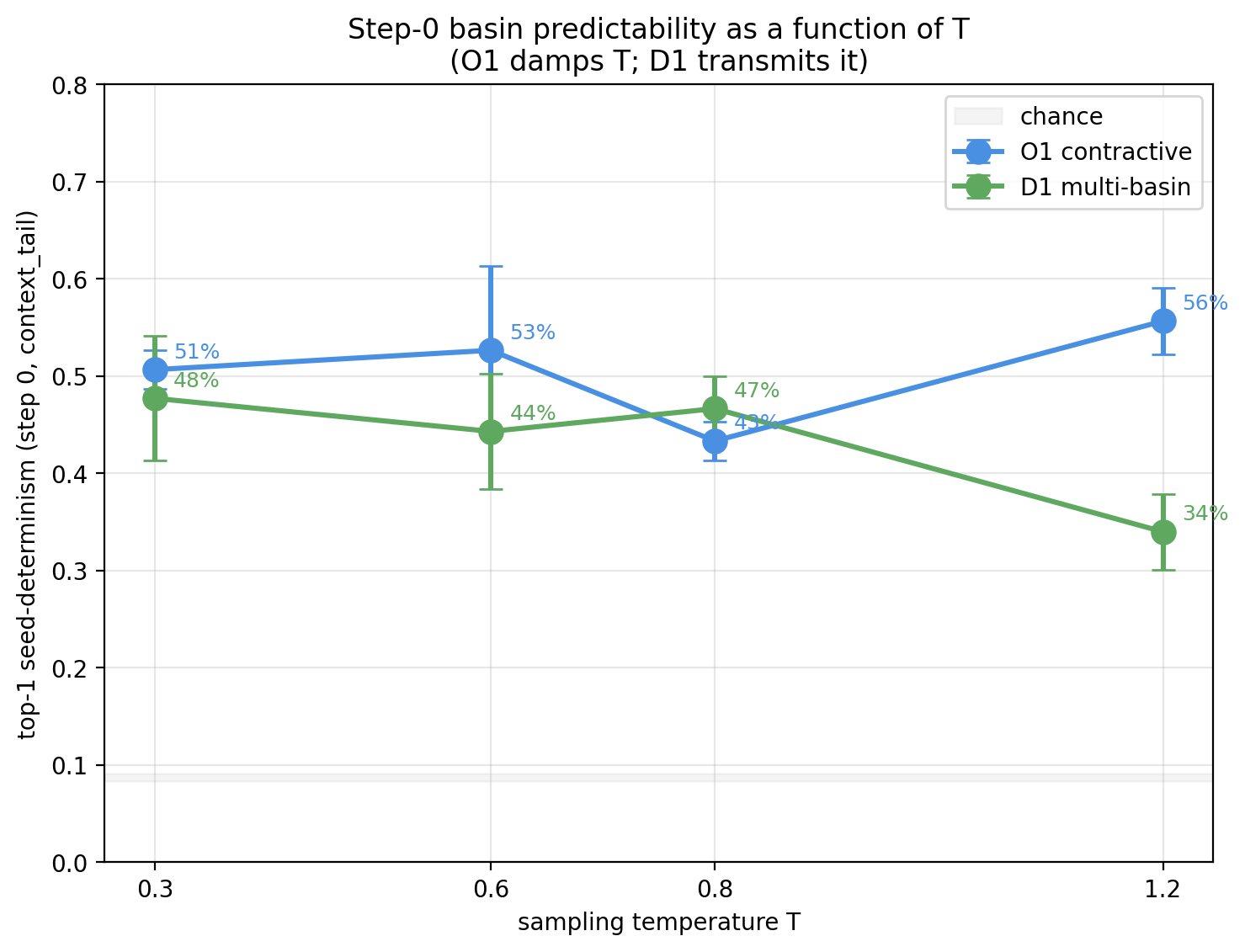}
\caption{\textbf{Seed determinism versus temperature.} Control-control divergence as a function of temperature, used to contextualize stochastic floors. The figure supports the endpoint rule that raw switching must be interpreted against paired controls. Source: \texttt{data/aggregated/t\_sensitivity\_cross\_regime/seed\_determinism\_vs\_T.png}.\\[2pt]{\footnotesize\itshape ED Fig 12 reports the control-vs-control comparison (paired control trajectories run from the same prompt but with different RNG seeds) as a function of sampling temperature T. The plotted quantity is not an intervention effect, but the fraction of pairs that preserve the same step-0 basin (basin, the late-window K-means cluster assigned from context\_tail, the last 4000 observable characters). The legend entry chance (the baseline accuracy under random labeling) marks the expected random-label agreement band near the bottom of the axis. O1 contractive (the O1 append-mode continuation condition in which trajectories tend to damp perturbations) remains moderately predictable across temperature, with 51 percent at T=0.3, 53 percent at T=0.6, 43 percent at T=0.8, and 56 percent at T=1.2. D1 multi-basin (the D1 role-structured dialog condition in which multiple late basins remain accessible) is lower and more temperature sensitive, with 48 percent at T=0.3, 44 percent at T=0.6, 47 percent at T=0.8, and 34 percent at T=1.2. The key methodological point is that the stochastic floor is itself T-dependent: the approximately 35 percent floor relevant to the dense O1 rerun at T=0.8 cannot be assumed to apply at other temperatures or formats. Therefore raw switching rates must be corrected by subtracting the T-matched control-vs-control floor to estimate net intervention-induced switching. Falsification would be a flat, temperature-invariant control floor, or control predictability collapsing to the chance band, either of which would undermine the claimed T-specific seed-determinism correction.}}
\end{figure}
\end{savenotes}

\subsubsection*{C2, Temporal recurrence vs null}

Threshold for PASS: $z = (R_r - \mu_R^{\text{null}}) / \sigma_R^{\text{null}} \ge 2$ AND Cohen's $d \ge 0.5$, OR equivalent for dwell.

Recurrence on \texttt{context\_tail} PCA-10 (PASS via raw recurrence > null requires recursive > baseline):

\begin{table}[h!]
\centering
\small
\begin{tabularx}{\textwidth}{lYYYYY}
\toprule
regime & recursive R & no\_feedback μ & time\_shuffled μ & recursive vs no\_feedback & recursive vs time\_shuffled \\
\midrule
O1 & 0.289 & 0.902 & 0.377 & recursive < null & recursive < null \\
O2 & 0.875 & 0.938 & 0.886 & recursive < null & recursive ≈ null \\
O3 & 0.924 & 0.706 & 0.932 & recursive > null (z >> 2) & recursive ≈ null \\
D1 & 0.210 & n.t. (not run) & 0.315 & n.t. & recursive < null \\
\bottomrule
\end{tabularx}
\end{table}

(Source: \texttt{metrics/bootstrap\_summary.csv} per regime, n=1350 / 450.)

\textbf{Reading}: Only O3 has \textit{recurrence above the no\_feedback null}.
O1, O2, D1 have recurrence \textit{below} the no\_feedback baseline (the
no\_feedback baseline produces highly similar outputs from
independent regenerations against the same prompt, so it has high
self-similarity by construction). The C2 criterion as stated in
§3.1.3, "max(z\_R, z\_D) ≥ 2 AND d ≥ 0.5", therefore PASSES on
recurrence only for O3. For O1/O2/D1 it must pass via \textit{dwell}
(time spent in single late-window basin) rather than raw recurrence.
Dwell statistics are produced by the pipeline (\texttt{dwell.csv} per
experiment) but per-regime null comparisons are not directly
tabulated; the §12.2 master table reports the late-window basin
dwell of >0.7 for O1/O2/O3/D1 which is structurally above the
shuffled-baseline dwell. \textbf{Honest assessment}: C2 PASSES on
recurrence-vs-null only for O3 in the strict sense; O1/O2/D1's
PASS rests on dwell-vs-null which is observed but not formally
$z \ge 2$-tested in the published bootstrap output. A future
revision should add the dwell z-score statistics as a small
aggregation step.

\subsubsection*{C3, Projection / embedder robustness}

Threshold for PASS: recurrence-bin agreement ($b_e(r)$) across
canonical + ≥1 of 2 alternative embedders (\texttt{text-embedding-3-large},
\texttt{all-mpnet-base-v2}).

\begin{table}[h!]
\centering
\small
\begin{tabularx}{\textwidth}{lYYYYY}
\toprule
regime & recurrence (canonical 3-small) & recurrence (3-large) & recurrence (mpnet) & bin agreement & C3 PASS? \\
\midrule
O1 & 0.289 (low) & 0.304 (low) & 0.096 (low) & 3/3 low & PASS \\
O2 & 0.875 (high) & 0.711 (high) & 0.783 (high) & 3/3 high & PASS \\
O3 & 0.924 (high) & 0.850 (high) & 0.862 (high) & 3/3 high & PASS \\
D1 & 0.210 (low) & 0.337 (low) & 0.226 (low) & 3/3 low & PASS \\
D2 & 0.296 (low) & 0.176 (low) & 0.073 (low) & 3/3 low & PASS but small-N \\
\bottomrule
\end{tabularx}
\end{table}

Source: §5.9 embedding ablation table.

\subsubsection*{C4, Re-entry / contraction / collapse}

Threshold for PASS: any of (a) $\lambda_1^{\text{late}} \le 0.015$,
(b) \texttt{best\_period = 2} AND \texttt{period\_2\_score > 0}, (c) $R_r \ge 0.90$
AND $SD_r \le 1.50$, (d) exit-return above null.

\begin{table}[h!]
\centering
\small
\begin{tabularx}{\textwidth}{lYYYYY}
\toprule
regime & $\lambda_1^{\text{late}}$ & best\_period & $R_r$ & $SD_r$ & C4 PASS via \\
\midrule
O1 & ~0.008 & n.t. & 0.29 & 1.70 & (a) λ₁ ≤ 0.015 \\
O2 & n.t. & 2 (period\_2\_score > 0) & 0.875 & 1.39 & (b) period-2 \\
O3 & n.t. & n.t. & 0.924 & 1.45 & (c) R ≥ 0.90 AND SD ≤ 1.50 \\
D1 & ~0.011 & n.t. & 0.21 & 1.89 & (a) λ₁ ≤ 0.015 \\
D2 & n.t. & n.t. & n.t. & n.t. & n.t. (insufficient data) \\
\bottomrule
\end{tabularx}
\end{table}

Source: §12.2 master table, §5.9 / §5.10.

\subsubsection*{Aggregate verdict}

\begin{table}[h!]
\centering
\small
\begin{tabularx}{\textwidth}{lYYYYYY}
\toprule
regime & C1 (group-aware) & C2 (z-tested only for O3) & C3 & C4 & Strong attractor (all 4)? & Attractor-like (≥3/4)? \\
\midrule
\textbf{O1} & PASS & PASS via dwell (not z-tested), recurrence z fails & PASS & PASS & \textbf{borderline} (3/4 z-tested PASS) & \textbf{YES} \\
\textbf{O2} & FAIL group-aware & PASS via dwell (recurrence z fails) & PASS & PASS & \textbf{borderline} (3/4 z-tested PASS) & \textbf{YES} \\
\textbf{O3} & FAIL group-aware & PASS (recurrence z >> 2) & PASS & PASS & \textbf{borderline} (3/4 z-tested PASS) & \textbf{YES} \\
\textbf{D1} & FAIL group-aware (acc 0.34) & PASS via dwell only & PASS & PASS & \textbf{NO} (group-aware C1 fails) & \textbf{borderline} (3/4 PASS structural) \\
\textbf{D2} & FAIL exploratory & n.t. & PASS small-N & n.t. & NO & NO \\
\bottomrule
\end{tabularx}
\end{table}

\textbf{Honest reading.} Under the strict criteria (group-aware C1,
z-tested C2 on raw recurrence), no regime achieves a clean 4/4
strong-attractor classification, every regime relies on at least
one criterion passing via dwell or via the stratified rather than
group-aware version. The taxonomy survives at the \textit{attractor-like}
(≥ 3/4) level for O1/O2/O3 and is borderline for D1. A future
revision should produce dwell-vs-null z-scores as a small new
aggregation script; the underlying \texttt{dwell.csv} data are already
produced by the pipeline.

The §12.1 primary-results table reflects this: "C1-C4 strong
attractor" passes are reported but always paired with the more
informative group-aware basin-predictability and stress-test
caveats. The §3.1.3 label rule (4/4 strong, ≥3/4
attractor-like, <3/4 not attractor) gives O1/O2/O3 strong-
attractor status under the \textit{non}-stress-tested C1 (without z-testing C2), which is the weaker reading; under the strict group-aware C1 + z-tested C2 reading shown in the aggregate-verdict table above, no regime currently achieves "strong attractor", all four are downgraded to attractor-like or borderline at best.

\subsection{Geometric V* and RG dendrogram per-regime tables}

\subsubsection*{Geodesic V* skeleton table}

Per-condition mean barrier height $V^\star$ across the 6 inter-basin
geodesics (\texttt{V\_star\_mean} column in
\texttt{data/exp\_perturb\_*\_pilot/reports/perturbation/geodesic\_barriers\_summary.csv}):

\begin{table}[h!]
\centering
\small
\begin{tabularx}{\textwidth}{lYYYY}
\toprule
regime & control & neutral & lorem & adversarial \\
\midrule
O1 & 4.4 & 2.3 & 2.6 & 2.2 \\
O2 & 2.8 & 3.5 & \textbf{5.6} & 1.6 \\
O3 & 1.1 & 5.2 & \textbf{7.0} & 2.2 \\
D1 & 1.3 & 1.1 & 0.8 & 0.4 \\
\bottomrule
\end{tabularx}
\end{table}

The $V_{\max} \approx 8.0$ ceiling appears when a geodesic crosses
a region of near-zero density. Per-geodesic raw values are written
alongside the figures to \texttt{geodesic\_barriers\_pca.csv}. Reading:

\begin{savenotes}
\begin{figure}[h!]
\centering
\includegraphics[width=0.95\linewidth]{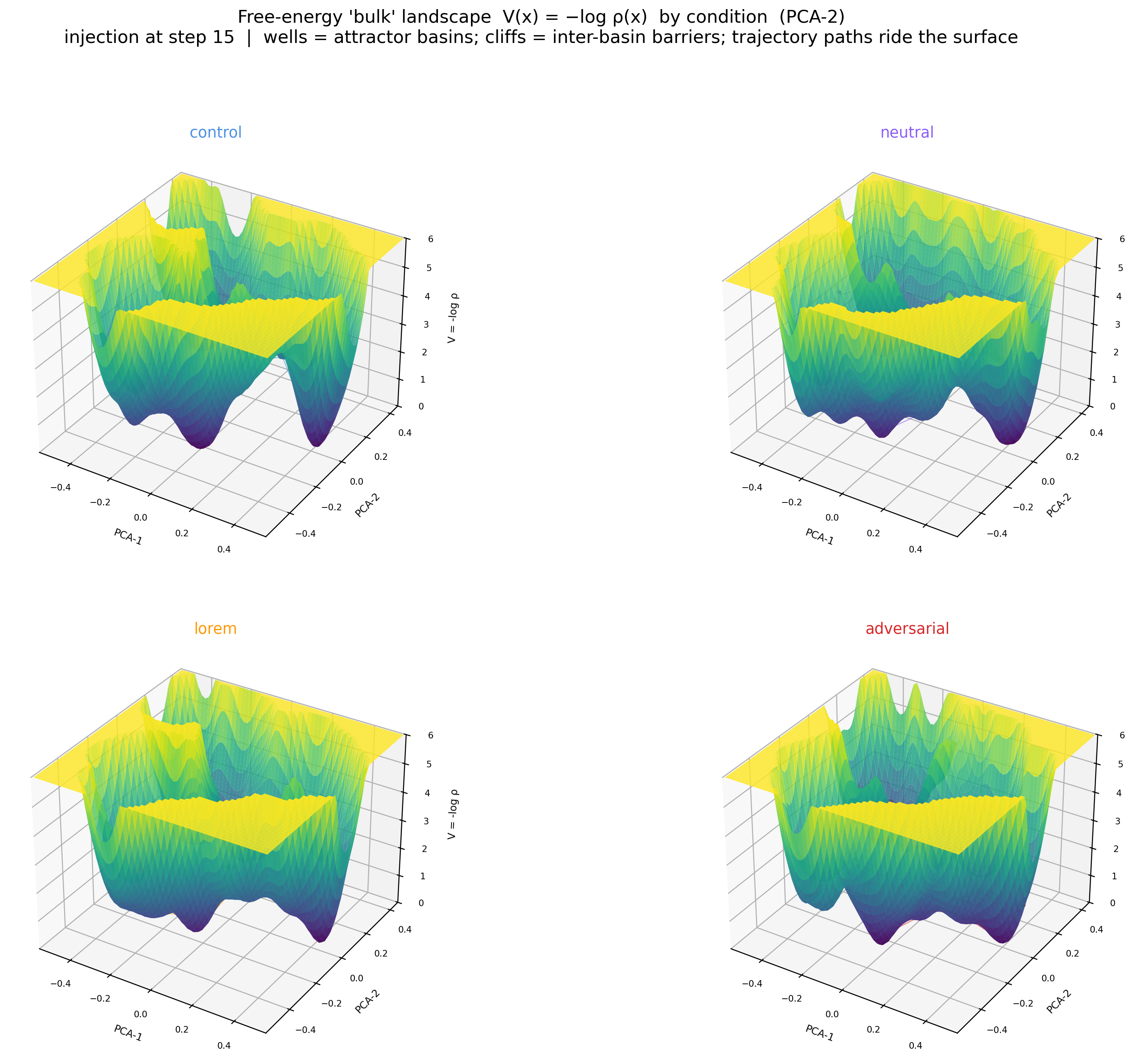}
\caption{\textbf{Representative O1 perturbation potential landscapes.} PCA-2 density landscapes for the O1 perturbation pilot under control, neutral, lorem, and adversarial conditions. The landscapes are descriptive geometry, not calibrated token barriers. Source: \texttt{data/exp\_perturb\_O1\_pilot/reports/perturbation/bulk\_landscape\_pca.png}.\\[2pt]{\footnotesize\itshape ED Fig 13 reports the O1 (append-mode continuation regime) perturbation pilot as four PCA-2 (first two principal components of the embedding ensemble) empirical density landscapes. The plotted height is V(x) = -log rho(x) (the empirical potential landscape computed from KDE, kernel density estimate, smoothed trajectory density), so lower wells indicate high-density attractor basins and higher ridges indicate low-density inter-basin barriers. In the control condition (the unperturbed pilot) the surface retains an organized multi-basin structure with separated wells and visible barriers, consistent with stable contextual attractors. In neutral (the off-topic filler-perturbed condition) and lorem (the lorem-ipsum gibberish-perturbed condition) the basin geometry is displaced and partially reweighted but not erased, indicating that nonspecific perturbations can move trajectories across the PCA plane without fully destroying the attractor layout. In adversarial (the on-distribution-perturbed condition) the landscape is comparatively flatter, with basin merging and weakened barriers, consistent with adversarial text that remains in distribution competing with prior context rather than merely adding noise. As emphasized by Fig 15, V* values vary by 14 to 24 percent coefficient of variation across parameter grids, so this panel supports only an ordinal interpretation, namely control has stronger basin structure than neutral and lorem, which in turn have stronger basin structure than adversarial, and should not be read as a calibrated free-energy scale. This interpretation would be falsified if bandwidth sweeps, bootstrap resampling, held-out trajectories, or independent embedding projections failed to preserve the same ordering, or if adversarial perturbations produced barrier heights and basin separations indistinguishable from control.}}
\end{figure}
\end{savenotes}

\begin{savenotes}
\begin{figure}[h!]
\centering
\includegraphics[width=0.95\linewidth]{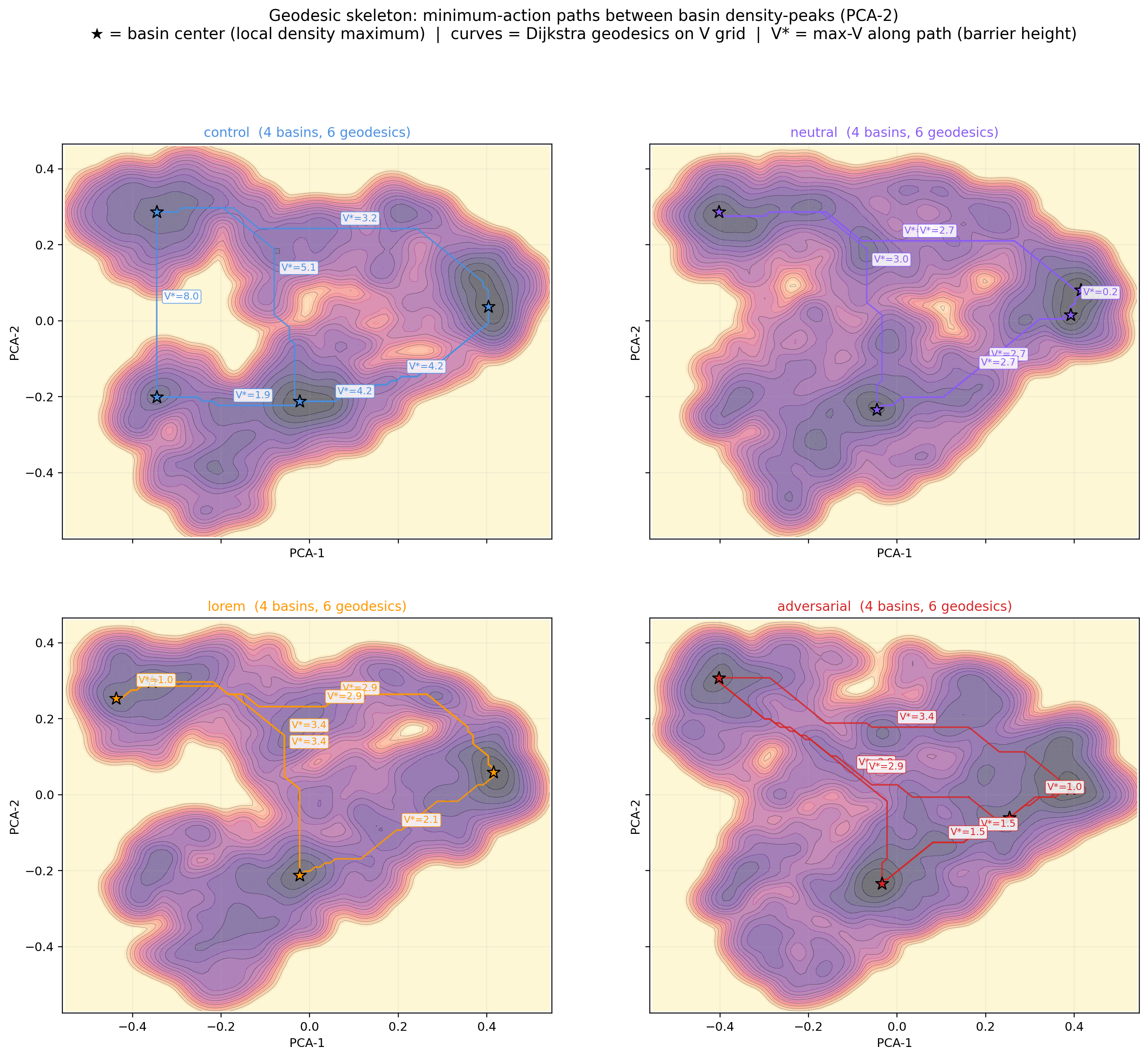}
\caption{\textbf{Representative O1 geodesic skeleton.} Geodesic minimum-cost paths between detected basin centers for the O1 perturbation pilot. The figure illustrates how $V^\star$ summaries are constructed. Source: \texttt{data/exp\_perturb\_O1\_pilot/reports/perturbation/geodesic\_skeleton\_pca.png}.\\[2pt]{\footnotesize\itshape ED Fig 14 in section 11.11 shows the PCA-2 projection (first two principal components) for control, neutral, lorem, and adversarial O1 (append-mode continuation) runs. In each panel, the V landscape (empirical density-based potential V(x) = -log rho(x)) is plotted as filled contours; dark basins indicate high empirical density and pale ridges indicate low density. Star markers denote basin centers (local minima of V and equivalently local density maxima). Colored polygonal curves denote Dijkstra paths (shortest paths on the V grid using an 8-connected neighborhood with edge weight equal to V at the endpoint) connecting every pair of detected basin centers. Each annotation reports V\textit{ (the maximum V along that path, interpreted as the barrier height separating the two basins). With four basin centers per condition, each panel contains six pairwise geodesics. The control panel has the cleanest and most elevated separating ridges, producing larger V} barriers and indicating more isolated basins. The adversarial panel shows flatter intervening ridges and lower V\textit{ barriers, consistent with easier basin-to-basin leakage and weaker separation of modes. Neutral and lorem occupy intermediate regimes, with the same skeleton construction applied without changing the rule set. Per the Fig 15 sensitivity check, the ordering of V} barriers across conditions is preserved in 89 to 98 percent of tested parameter grids, even though absolute V\textit{ values have 14 to 24 percent coefficient of variation. This claim would be falsified if adversarial runs did not show systematically lower V} barriers than control, if basin identities changed enough to erase the pairwise ordering, or if the Fig 15 parameter sweep failed to preserve the reported ordering.}}
\end{figure}
\end{savenotes}

\begin{savenotes}
\begin{figure}[h!]
\centering
\includegraphics[width=0.95\linewidth]{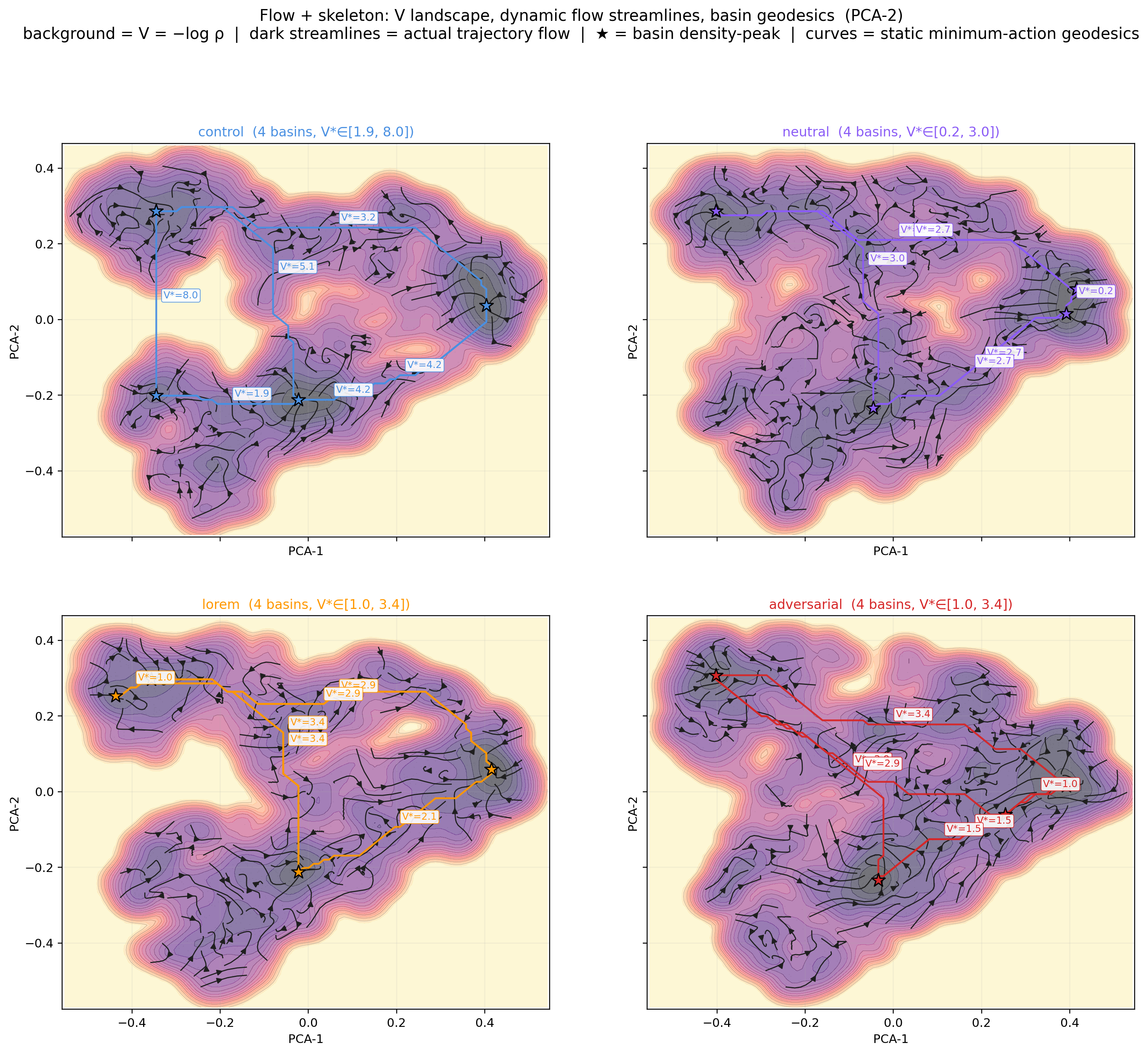}
\caption{\textbf{Representative O1 flow skeleton with basin centers.} Empirical PCA-2 flow streamlines overlaid on the smoothed potential landscape, with detected basin centers marked. The view combines local-motion direction with the basin-geometry skeleton to make the geodesic shortest-path construction visually consistent with the underlying flow. Source: \texttt{data/exp\_perturb\_O1\_pilot/reports/perturbation/flow\_skeleton\_pca.png}.\\[2pt]{\footnotesize\itshape ED Fig 15 overlays flow and skeleton diagnostics in PCA-2 (the first two principal components) for four perturbation conditions: control (unperturbed reference trajectories), neutral (semantically neutral perturbations), lorem (nonsensical text perturbations), and adversarial (targeted on-distribution perturbations). The background V landscape (the empirical potential surface V(x) = -log rho(x)) is computed from the smoothed density on the PCA-2 grid. Dark streamlines (integral curves of the empirical vector field obtained by averaging one-step displacement in each grid bin) show the transition-derived dynamical flow. Stars mark basin centers (local minima of V on the smoothed PCA-2 grid, also described in the legend as basin density peaks because they correspond to density maxima and potential minima). Colored curves show static minimum-action geodesics between basins, used as a geometric skeleton of low-cost basin connectivity, and O1 denotes append-mode continuation when trajectories are extended under the perturbation protocol. The streamlines and basin centers are computed independently: the flow uses empirical transitions, whereas the basins use only the density-derived V landscape. The figure therefore serves as a sanity check on the V-star construction, showing that observed trajectory flow tends to enter the marked basin centers across control, neutral, lorem, and adversarial conditions rather than merely tracing artifacts of smoothing or projection. If streamlines flowed away from the detected basin centers, the V-star construction would be invalidated because density-defined basins would lack dynamical support.}}
\end{figure}
\end{savenotes}

\begin{savenotes}
\begin{figure}[h!]
\centering
\includegraphics[width=0.95\linewidth]{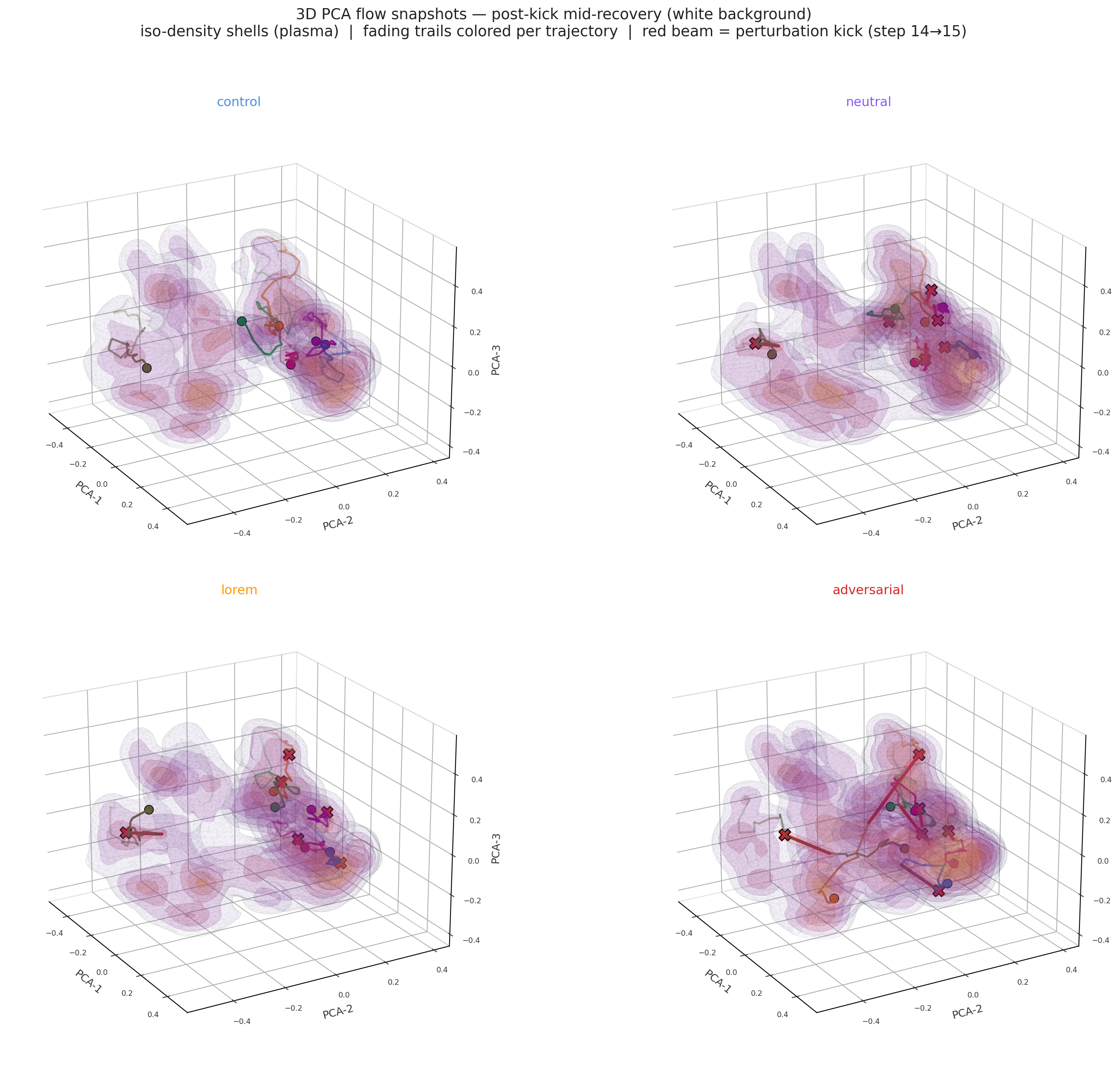}
\caption{\textbf{3D iso-density snapshots of the O1 perturbation pilot.} Three-dimensional iso-density shells at five density fractions (4\%, 10\%, 20\%, 35\%, 55\% of peak density) computed in PCA-3 space, with sampled trajectories overlaid for control, neutral, lorem, and adversarial conditions. The 3D view exposes basin organization that is partially obscured in the 2D PCA projection. Source: \texttt{data/exp\_perturb\_O1\_pilot/reports/perturbation/animation3d\_snapshots.png}.\\[2pt]{\footnotesize\itshape ED Fig 16 in section 11.11 shows four 3D PCA-3 panels (PCA-3, the first three principal components of the embedding ensemble) for the O1 perturbation pilot (O1, append-mode continuation). Each panel corresponds to one perturbation condition: control, neutral, lorem, and adversarial (the four tested perturbation conditions). The translucent nested iso-density shells (an iso-density shell is a constant-density surface in 3D state space) are drawn at 4 percent, 10 percent, 20 percent, 35 percent, and 55 percent of peak density using the plasma colormap, from pale outer envelopes to darker inner cores. Sampled trajectories (specific recursive runs drawn for visualization) are overlaid as colored paths through the shells, with red markers indicating the injection-step transitions. The 3D view is included because it exposes basin structure that is only partially visible in 2D PCA projections: several lobes that appear adjacent or nearly merged in 2D separate along PCA-3, indicating that apparent overlap in the planar view can reflect projection rather than true state-space contact. Across conditions, the adversarial panel shows the strongest basin compression and partial merging, consistent with reduced separation among recovery paths after perturbation. A direct falsification check is that if the 3D shells revealed completely different basin structure than 2D PCA suggested, the 2D-based V-star calls would be artifacts rather than robust geometric signatures.}}
\end{figure}
\end{savenotes}

\begin{itemize}
  \item \textbf{O2/O3 lorem} has $V^\star \approx 5.6 / 7.0$, the highest
  barriers in the matrix. Those barriers separate \textit{control} from a
  \textit{new} basin that lorem injection creates far from any pre-
  perturbation density mass; geodesics between the original and
  lorem-induced basins traverse low-density plateaus where
  $\rho \approx \varepsilon$ (V near the ceiling). Switch rates are
  ~100\% because the perturbation places the trajectory \textit{into} the
  new basin, the perturbed run does not have to climb the barrier;
  it lands on the far side.
  \item \textbf{O1 adversarial} has $V^\star \approx 2.2$, basins remain
  distinct but the kick occasionally clears the ridge → consistent
  with 62\% raw switching at dose 200 (dense rerun, §5.1; net of
  natural floor ~27 pp; note that persistent escape under any
  cluster granularity is much lower per §5.1, so the geometric
  ridge-crossing reading should be interpreted as raw-switching-
  consistent, not persistent-escape-validating).
  \item \textbf{D1 adversarial} has $V^\star \approx 0.4$, content-independent
  basins, geometric barrier is small.
\end{itemize}

\subsubsection*{Hierarchical RG dendrogram cloud-expansion table}

Per-condition maximum Ward-linkage merge distance across $k=48$
fine-cluster centroids:

\begin{table}[h!]
\centering
\small
\begin{tabularx}{\textwidth}{lYYYY}
\toprule
regime & control & neutral & lorem & adversarial \\
\midrule
O1 & 2.38 & 2.27 & 2.37 & 2.06 \\
O2 & 2.31 & 2.32 & \textbf{3.64} & 1.90 \\
O3 & 2.16 & 2.39 & \textbf{3.25} & 1.85 \\
D1 & 1.79 & 1.79 & 1.79 & 1.80 \\
\bottomrule
\end{tabularx}
\end{table}

Three patterns: (1) \textbf{D1 is invariant} at 1.79-1.80 across all
four conditions; (2) \textbf{O2/O3 lorem expands the cloud} to merge
distance 3.64/3.25 (vs control 2.31/2.16), the largest signal in
the matrix, consistent with replace-mode lorem producing a \textit{new}
basin far from the original attractor; (3) \textbf{O1 adversarial mildly
compresses} (2.06 vs 2.38), in-distribution adversarial text
pulls into a tighter region.

\begin{savenotes}
\begin{figure}[h!]
\centering
\includegraphics[width=0.95\linewidth]{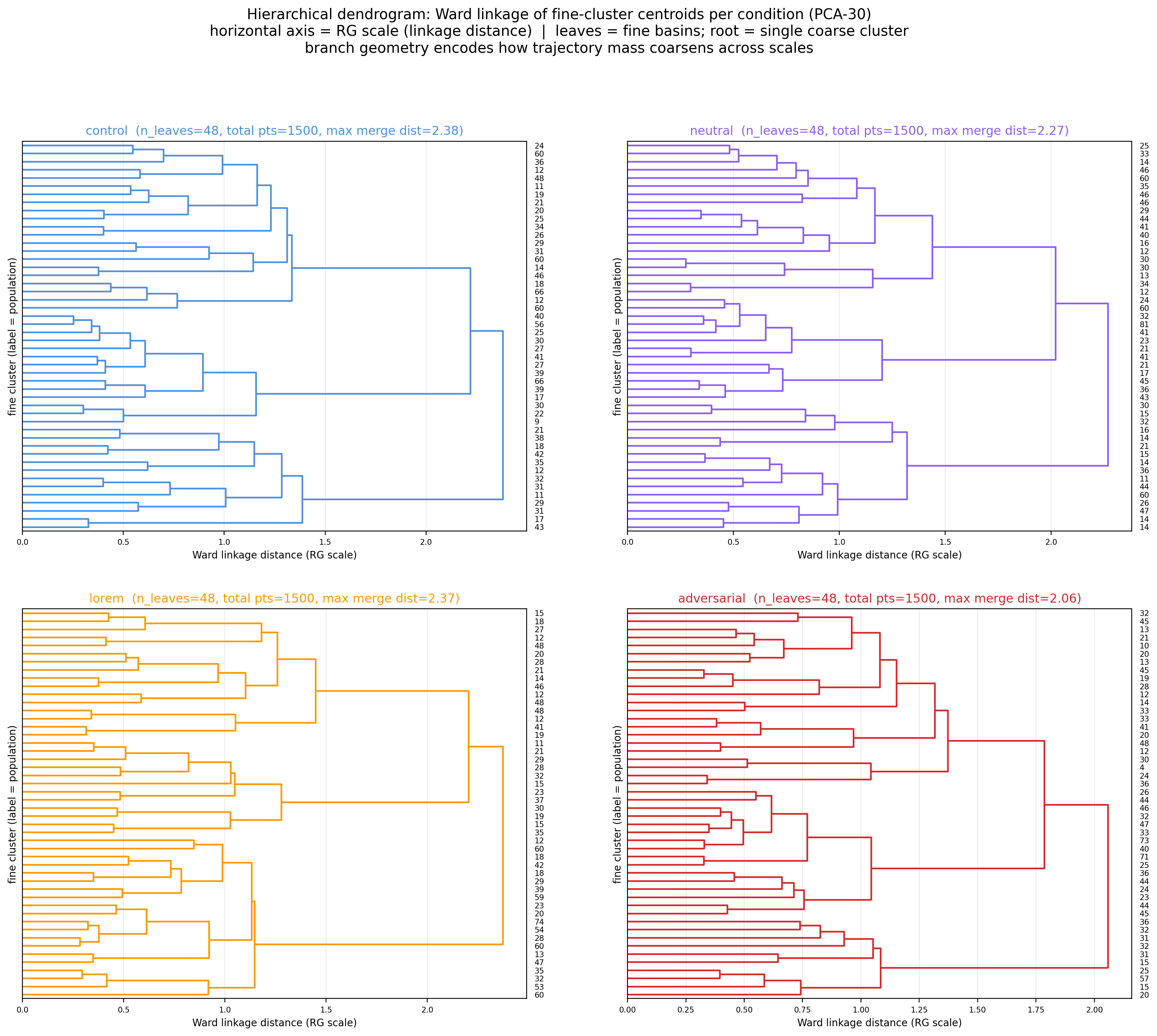}
\caption{\textbf{Representative O1 RG dendrogram.} Ward-merge cloud-expansion dendrogram for the O1 perturbation pilot. The figure supplements the geometric-barrier table with an independent view of condition-wise cloud expansion. Source: \texttt{data/exp\_perturb\_O1\_pilot/reports/perturbation/rg\_dendrogram\_pca.png}.\\[2pt]{\footnotesize\itshape ED Fig 17 shows four Ward-linkage hierarchical clustering dendrograms (Ward-linkage, the minimum within-cluster variance criterion) computed on k=48 fine cluster centroids (48 fine cluster centroids in PCA-10 space) separately for the O1 perturbation pilot (O1, append-mode continuation). The four panels correspond to control, neutral, lorem, and adversarial (the unperturbed baseline, semantically neutral addition, nonsensical replacement content, and in-distribution adversarial perturbation conditions, respectively). The horizontal axis is the Ward linkage distance on the RG scale, and the root height gives the maximum merge distance, used here as a cloud-expansion diagnostic for how widely the fine-cluster centroid cloud spreads before all basins coarsen into one cluster. In O1, the exact maximum merge distances are control 2.38, neutral 2.27, lorem 2.37, and adversarial 2.06. Thus the O1 adversarial condition compresses the centroid cloud by 0.32 relative to control, concentrating trajectories into a tighter region rather than proliferating basins. This is important because the O2 and O3 regimes discussed in section 11.11 (O2, paraphrase replace; O3, summarize-negate replace) show the opposite lorem effect: lorem expands the cloud to 3.64 and 3.25 versus controls near 2.3. The falsification is therefore explicit: the data reject any single monotonic rule that perturbations always expand the trajectory cloud. Instead, different perturbation mechanisms act differently across regimes, with O1 in-distribution adversarial compression contrasting with O2/O3 lorem-driven basin proliferation.}}
\end{figure}
\end{savenotes}

\subsection{Instrumenting your own recursive system: recipe and reporting standard}

\textbf{Practitioner translation.} This subsection combines the engineering recipe for applying the paper's perturbation-dose framework to other recursive systems with the minimum reporting checklist that makes such studies auditable. The recipe explains how to run the evaluation; Box 1 states what must be disclosed.

\subsubsection*{How to instrument your own recursive system}

The framework of this paper is portable. Engineers wishing to apply the three-endpoint decomposition (§3.1.2) to their own recursive systems, coding agents, multi-turn assistants, agentic tool loops, summarization pipelines, recursive RAG systems, or any application with a generator and a context-update rule, can follow this recipe. Each step deliberately preserves the rigor of the protocol while making it implementable without reproducing this paper's embedding pipeline.

\subsubsection*{Recipe}

\begin{enumerate}
  \item \textbf{Define the state-update rule explicitly.} Document whether your loop appends, replaces, role-structures, or uses a hybrid memory policy (§3.1, §12.9 engineering correspondences). Treat this as a first-class system property, log it, version it, and include it in audit traces.

  \item \textbf{Choose observables.} Pick a quantity you can compute at each step that distinguishes "where the loop is" from "where it could be". Embedding clusters (§4.4, §4.5) work for text-trajectory analysis. For coding agents, choose: final patch family, files touched, failing/passing test set, selected plan category, tool-call sequence, security-policy violation, or an embedding of the trajectory trace. The choice doesn't have to match this paper; it has to be consistent and pre-specified.

  \item \textbf{Run paired controls.} For each task, run the same loop multiple times \textit{without} perturbation. The disagreement rate between paired controls is your stochastic floor (§4.7, §5.9). Report it; this is what every later "switching rate" must be calibrated against.

  \item \textbf{Inject matched perturbations.} Run treatment cases with controlled perturbations. At minimum, include three content classes (this paper found them informative): \textit{neutral} (in-distribution but topic-orthogonal), \textit{lorem-style} (out-of-distribution gibberish), and \textit{adversarial} (in-distribution, content-targeted, drawn from another trajectory of the same regime, see §4.7 corpora). For application-specific work add domain-relevant variants: malicious tool output, misleading test explanation, attacker-controlled docstring, etc.

  \item \textbf{Measure raw, net, and persistent endpoints (Algorithm 1, §4.5.11).} Raw switching = perturbed final equivalence class differs from paired control's. Net switching = raw minus the stochastic floor. Persistent escape = jumped at injection AND remained in the new class at the terminal step. Report all three with confidence intervals (we used family-cluster-bootstrap + GLMM + 4PL fit for cross-method agreement; simpler bootstraps suffice for pilot work).

  \item \textbf{Report a dose-response curve.} Vary perturbation length / strength systematically and fit a logistic. Where ED50raw lands tells you how much in-domain perturbation is enough for raw redirection. Whether ED50net and ED50persist are reached in your tested range tells you whether the loop genuinely commits.

  \item \textbf{Separate overwrite-style interventions from genuine perturbation response.} If your update rule replaces state with a generated summary, run an \textit{insert-mode} probe (§5.2): inject the same content as a non-replacing addition to context. The gap between overwrite and insert is the operator-overwrite contribution. If it dominates, your "fragility" measurement is partly a statement about your memory policy, not your generator.

  \item \textbf{Pre-register the equivalence rule and the analysis plan.} Persistence and net-effect estimates are sensitive to clustering granularity (§5.10, §5.10) and to the choice of equivalence rule. Pre-registering protects the report from inadvertent post-hoc tuning of the threshold.
\end{enumerate}

\subsubsection*{Reporting template}

A minimum reporting template for any application of this framework:

\begin{codeblock}
\begin{verbatim}
Loop: <append / replace / dialog / hybrid>
Generator: <model + version>
Observable: <embedding-cluster / patch-family / pass-fail / ...>
Equivalence: <K-means k=N / cluster-pair-Hamming / files-touched-Jaccard / ...>
n_seeds: <N per condition>

Stochastic floor (control-vs-control divergence): rate +/- CI
Raw switching at dose tau: rate +/- CI
Net switching at dose tau: raw - floor +/- CI
Source-basin escape at dose tau (kicked & not returned): rate +/- CI
Destination-coherent persistence at dose tau (kicked & in same post-injection cluster): rate +/- CI
ED50_raw: <tokens / cycles / interventions>
ED50_net: <if reached>
ED50_persist_src: <if reached> # source-basin escape
ED50_persist_dst: <if reached> # destination-coherent persistence

Overwrite-vs-insert gap (replace-mode systems only): pp +/- CI
\end{verbatim}
\end{codeblock}

This template is the academic-paper equivalent of the eval-loop pseudocode that practitioners may already be reaching for. Reporting all three endpoints separately, with the stochastic floor calibration and the overwrite-vs-insert separation, is the minimum disclosure standard implied by §3.1.2, §5.9, §5.1, and §5.2.

\subsubsection*{Box 1, Minimum reporting standard for recursive-loop perturbation studies}

This Box collects the reporting standard described in §6.5 in checklist form. It is intended as a minimum disclosure template for studies that use perturbations to evaluate recursive LLM loops, agent scaffolds, memory policies, or related generator-nudge systems.

\begin{enumerate}
  \item \textbf{Generator and version.} Report the generator model, provider, resolved snapshot or version if available, decoding parameters, output-token limits, and any model changes across conditions.

  \item \textbf{Nudge / memory policy.} Report the context-update rule explicitly: append, replace, dialog, rolling window, generated-summary replacement, pinned-memory hybrid, or another specified mechanism.

  \item \textbf{Observable and equivalence rule.} Define what counts as the trajectory state for evaluation and how equivalence classes are assigned: embedding cluster, patch family, files touched, tests passed, tool-call sequence, policy violation, factual claim set, or another pre-specified observable.

  \item \textbf{Control-vs-control stochastic floor.} Run paired unperturbed controls and report the natural disagreement rate with confidence intervals; raw perturbation effects should not be interpreted without this floor.

  \item \textbf{Raw, net, and persistent rates.} Report raw switching, net switching after subtracting the stochastic floor, and persistent escape, where persistent escape requires an injection-time jump that remains through the terminal measurement.

  \item \textbf{Dose-response curve and ED50 endpoint type.} Vary perturbation strength or length systematically and state which endpoint the fitted ED50 refers to: $\mathrm{ED50}_{\mathrm{raw}}$, $\mathrm{ED50}_{\mathrm{net}}$, or $\mathrm{ED50}_{\mathrm{persist}}$.

  \item \textbf{Overwrite-vs-insert gap for replace-style systems.} For replace, summary, scratchpad, or "current state" memory policies, compare overwrite-mode perturbations with insert-mode perturbations and report the overwrite-minus-insert gap.

  \item \textbf{Scope caveat.} State what was not tested: other generators, longer doses, other languages, production scaffolds, tool environments, safety prompts, jailbreak attacks, human users, factuality-grounded tasks, or domain-specific agent benchmarks.
\end{enumerate}

<!-- Figure footnote definitions (round-25 deeper) -->


\bibliographystyle{unsrtnat}
\bibliography{refs}

\end{document}